\pdfoutput=1
\documentclass[11pt]{article}
\usepackage{fullpage}
\usepackage[hyphens]{url}
\usepackage{hyperref}
\hypersetup{
breaklinks=true,
    colorlinks=true, 
    linkcolor=black, 
    citecolor=black, 
    filecolor=black,
    urlcolor=black 
}
\usepackage{caption}
\usepackage{makecell}
\usepackage{afterpage}

\usepackage{algorithm}
\usepackage{algorithmic}

\usepackage{amssymb}
\usepackage{mathtools}
\usepackage{amsthm}
\usepackage{mathrsfs}
\usepackage{subcaption}

\usepackage[capitalize,noabbrev]{cleveref}
\usepackage{minitoc}
\usepackage[utf8]{inputenc} 
\usepackage[T1]{fontenc}    
\usepackage{booktabs}       
\usepackage{amsfonts}       
\usepackage{nicefrac}       
\usepackage{microtype}      
\usepackage{diagbox}
\usepackage{enumerate}
\usepackage[shortlabels]{enumitem}
\usepackage{tabularx}
\usepackage{verbatim}
\usepackage{afterpage}
\usepackage{float}
\usepackage{tcolorbox} 
\usepackage{listings}  

\usepackage{multirow}
\usepackage{color}
\usepackage{xcolor}
\usepackage[numbers]{natbib}

\allowdisplaybreaks[4]

\theoremstyle{plain}
\newtheorem{theorem}{Theorem}[section]

\newtheorem{asmp}{Assumption}[section]
\newtheorem{prop}{Proposition}[section]

\newtheorem{lem}{Lemma}[section]
\newtheorem{remark}[theorem]{Remark}
\newtheorem{cor}{Corollary}
\newtheorem{model}{Model}

\newtheoremstyle{remarkstyle}
  {}                    
  {}                    
  {\normalfont}         
  {}                    
  {\itshape}            
  {.}                   
  { }                   
  {}                    
\theoremstyle{remarkstyle}

\definecolor{wjs}{RGB}{200,0,50}

\definecolor{hyw}{RGB}{153,000,000}

\newcommand{\new}[1]{#1}

\hypersetup{
colorlinks=true,
filecolor=black,
citecolor=blue,
urlcolor=black,
}

\usepackage{amsmath,amsfonts,bm}

\def\1{\bm{1}}

\def\eps{{\varepsilon}}

\def\rd{{\textnormal{d}}}

\def\vh{{\bm{h}}}

\DeclareMathAlphabet{\mathsfit}{\encodingdefault}{\sfdefault}{m}{sl}
\SetMathAlphabet{\mathsfit}{bold}{\encodingdefault}{\sfdefault}{bx}{n}

\def\epsinital{{\widehat{\eps}_{\mathrm{ini}} }}

\def\epsrefine{{\widehat{\eps}_{\mathrm{rfn}} }}
\def\epsopt{{\widehat{\eps}_{\mathrm{opt}} }}

\renewcommand{\xi}{\zeta}

\def\0{{\bf 0}}
\def\1{{\bf 1}}

\def\MM{{\mathcal M}}
\def\AM{{\mathcal A}}
\def\BM{{\mathcal B}}

\def\GM{{\mathcal G}}
\def\FM{{\mathcal F}}

\def\HM{{\mathcal H}}

\def\LM{{\mathcal L}}

\def\OM{{\mathcal O}}
\def\PM{{\mathcal P}}

\def\SM{{\mathcal S}}
\def\TM{{\mathcal T}}

\def\UM{{U}}
\def\VM{{\mathcal V}}
\def\WM{{\mathcal W}}

\def\RB{{\mathbb R}}

\def\EB{{\mathbb E}}
\newcommand{\EBpr}{\mathbb{E}^{\mathrm{pr}}}

\newcommand{\bFpr}{\Bar{F}^{\mathrm{pr}}}
\newcommand{\bFprP}{\Bar{F}_{\bP}^{\mathrm{pr}}}
\newcommand{\bgP}{\Bar{g}_{\bP}}

\def\PB{{\mathbb P}}

\newcommand{\KL}{D_{\mathrm{KL}}}
\newcommand{\TV}{D_{\mathrm{TV}}}
\newcommand{\Chi}{D_{\chi^2}}

\newcommand{\Var}{\mathrm{Var}}

\def\rd{{\mathrm{d}}}

\def\vopt{v_{\mathrm{opt}}}
\def\hvopt{\widehat{v}_{\mathrm{opt}}}
\def\vind{v_{\mathrm{ind}}}
\def\YWM{Y^{\mathrm{wm}}}

\def\Key{{\mathtt{Key}}}

\def\token{{w}}

\def\Voca{{\WM}}
\newcommand\bP{\bm{P}}

\title{Optimal Estimation of Watermark Proportions in Hybrid AI-Human Texts}
  
\author{
{Xiang Li\thanks{University of Pennsylvania; Email: \texttt{lx10077@upenn.edu}. } } 
\and
{Garrett G.\ Wen\thanks{Yale University; Email: \texttt{gang.wen@yale.edu}. }} 
\and
{Weiqing He\thanks{University of Pennsylvania; Email: \texttt{weiqingh@sas.upenn.edu}. }} 
\and
{Jiayuan Wu\thanks{University of Pennsylvania; Email: \texttt{jyuanw@wharton.upenn.edu}. }}  
\and
{Qi Long\thanks{University of Pennsylvania; Email: \texttt{qlong@upenn.edu}. Joint corresponding author.}}
\and
{Weijie J.\ Su\thanks{University of Pennsylvania; Email: \texttt{suw@wharton.upenn.edu}. Joint corresponding author.}}
}

\date{June 27, 2025}

\begin{document}

\maketitle

\begin{abstract}

Text watermarks in large language models (LLMs) are an increasingly important tool for detecting synthetic text and distinguishing human-written content from LLM-generated text. While most existing studies focus on determining whether entire texts are watermarked, many real-world scenarios involve mixed-source texts, which blend human-written and watermarked content. In this paper, we address the problem of optimally estimating the watermark proportion in mixed-source texts. We cast this problem as estimating the proportion parameter in a mixture model based on \emph{pivotal statistics}. First, we show that this parameter is not even identifiable in certain watermarking schemes, let alone consistently estimable. In stark contrast, for watermarking methods that employ continuous pivotal statistics for detection, we demonstrate that the proportion parameter is identifiable under mild conditions. We propose efficient estimators for this class of methods, which include several popular unbiased watermarks as examples, and derive minimax lower bounds for any measurable estimator based on pivotal statistics, showing that our estimators achieve these lower bounds. Through evaluations on both synthetic data and mixed-source text generated by open-source models, we demonstrate that our proposed estimators consistently achieve high estimation accuracy.

\end{abstract}

\section{Introduction}
\label{sec:intro}

Over the past two years, the rapid advancement of large language models (LLMs) has profoundly transformed content generation, with diverse applications spanning creative writing, code generation, text summarization, and educational assistance \citep{openai2023}. Their rapid adoption, however, has raised pressing concerns regarding content authenticity in critical domains such as scientific peer review \citep{liang2024monitoring}, education~\citep{milano2023large}, journalism~\citep{starbird2019disinformation}, and data authenticity~\citep{radford2023robust,das2024under}.

To mitigate potential misuse of these powerful generative tools, researchers have developed \emph{text watermarking} methods that embed imperceptible yet statistically verifiable signals into LLM-generated output~\citep{kirchenbauer2023watermark,scott2023watermarking,kuditipudi2023robust}. These watermarking techniques facilitate reliable detection of synthetic text, enabling one to distinguish fully watermarked content from human-written text~\citep{piet2023mark}.

However, real-world scenarios frequently involve mixed-source texts, where human-written and LLM-generated content are commonly interwoven \citep{zhang2024llm,dugan2023real,li2024optimal}. This is particularly evident in academic peer reviews, as demonstrated by \citet{liang2024monitoring}, who reported a growing prevalence of LLM-modified reviews in machine learning conferences. Beyond academia, evidence suggests that human-AI collaboration is increasingly reshaping content creation practices across various domains \citep{bommasani2021opportunities}. In this context, quantifying the extent of human versus LLM contribution becomes crucial \citep{xie2024measuring}. Indeed, accurate estimates of human-written proportions have significant implications for determining content ownership and authenticity in academic writing, education, and legal documents \citep{wu2021ai}. For instance, a homework assignment essay with only 10\% student contribution might be deemed unacceptable, while one with 80\% or more student input might be considered permissible.

While prior studies have primarily focused on binary detection of fully watermarked versus fully human-written texts \citep{li2024statistical}, recent work has begun to address the challenges posed by mixed-source content. From a statistical standpoint, however, the problem of estimating the proportion of LLM-generated content in mixed-source texts remains insufficiently explored, despite its critical role in use cases such as authorship attribution. Notably, \citet{li2024optimal} formulated watermark detection in mixed-source texts as a mixture detection problem and identified optimal detection methods through the lens of statistical robustness. Other studies have examined the more granular problem of identifying watermarked segments within mixed-source texts. Examples include \citet{scott2023watermarking,kirchenbauer2023reliability}'s continuous span search, \citet{zhao2024efficiently}'s online learning approach, and \citet{li2024segmenting}'s application of change point detection to identify watermarked segments.

While identifying watermarked segments can, in principle, suffice for our purpose, it often yields an unnecessarily fine level of granularity and, more critically, may be statistically infeasible for pinpointing precisely which segments are human-authored. In many scenarios, a single numerical estimate of the proportion of LLM-generated content is more convenient and interpretable. For instance, it is practical to require AI-generated content to not exceed a 10\% threshold in critical documents such as legal briefs and academic papers, while a higher limit (e.g., 50\%) may be tolerable in less critical contexts such as homework assignments. Moreover, precisely localizing human-written text is arguably more challenging than simply estimating the overall proportion. This parallels findings from the broader mixture detection literature, wherein identifying individual signal instances is inherently more difficult than detecting their aggregate presence. Indeed, signals can be statistically detectable even when it is information-theoretically impossible to locate each instance accurately \citep{donoho2004higher,tony2017optimal}.

\subsection{Our Contributions}
In this paper, we address the problem of estimating the proportion of watermarked contents in mixed-source texts under general watermarking schemes. Our main contributions are:

\noindent\textbf{Problem formulation and identifiability}: We formulate the estimation problem as a mixture model parameter estimation task using pivotal statistics \citep{li2024statistical}. 
Roughly speaking, a pivotal statistic quantifies the evidence that a given word (or token) is watermarked. Through rigorous analysis, we show that the proportion parameter in green-red list watermarking schemes \citep{kirchenbauer2023watermark} is not identifiable, which leads to the inherent bias in traditional approaches like maximum likelihood estimation (MLE). However, for watermarks with continuous pivotal statistics, we establish conditions under which identifiability can be achieved.

\noindent\textbf{New estimators with theoretical guarantees}: We develop computationally efficient estimators for identifiable cases and provide theoretical guarantees for their estimation accuracy. Our comprehensive experimental validation, conducted on both synthetic data and mixed-source texts from open-source LLMs, shows the consistent and superior performance of our proposed estimators.

\noindent\textbf{Matched minimax lower bounds}: We establish minimax lower bounds for broad classes of measurable estimators and prove that our proposed methods generally achieve these bounds, thereby establishing their theoretical optimality.

\subsection{Related Work}

Most watermarking studies focus on the binary detection problem, which aims to determine whether a text is fully watermarked or entirely human-written. However, this narrow scope fails to address real-world scenarios where texts often originate from mixed sources. While recent work has begun exploring the detection of watermarked segments in mixed-source texts, their approaches differ significantly from ours.
\citet{li2024optimal} investigated the optimal detection of watermarked segments under human editing, assuming that the text is initially fully generated by a watermarked LLM and later modified by a human editor. Using the Gumbel-max watermark as an example, they demonstrate that a family of truncated goodness-of-fit tests achieves optimal detection, where optimality is defined by the phase transition boundary and feast-favorable efficiency notion introduced in \citep{li2024statistical}. However, their work focuses exclusively on detection and does not address estimation.

Other researchers attempt to locate watermarked segments by assigning scores to individual tokens or intervals of tokens and identifying watermarked (or LLM-generated) content based on extreme values. For example, \citet{kirchenbauer2023reliability} searched for the largest intervals or continuous token spans with the highest $z$-scores by testing all possible window sizes and traversing the entire text for each size, resulting in a computational complexity of $\widetilde{O}(n^2)$. \citet{scott2023watermarking} improved on this by using dynamic programming to identify the largest continuous token span, achieving a reduced complexity of $O(n^{1.5})$. \citet{zhao2024efficiently} defined token scores based on the expectations of pivotal statistics and employs an online learning algorithm for estimation, achieving a complexity of $O(n\log n)$ for texts of length $n$. Inspired by change point detection, \citet{li2024segmenting} used bootstrap-based $p$-values as token scores, but the method incurs a higher computational cost of approximately $\widetilde{O}(n^2)$ due to its analysis of dependencies across all sliding token windows and pseudorandomness.

Despite these advancements, existing methods neither address the problem of watermark proportion estimation nor provide a systematic analysis of their performance or hyperparameter tuning (e.g., threshold selection). In contrast, our work presents the first comprehensive framework for watermark proportion estimation. We address key questions such as what to estimate (estimand and estimability), how to estimate (estimation methods), how accurately it can be estimated (estimation accuracy), and the theoretical limits of estimation (minimax lower bounds). Furthermore, our proposed estimator achieves ${O}(n)$ complexity by requiring only a logarithmic number of passes through the samples.

\vspace{-0.1in}
\section{Preliminaries}\label{sec:pre}

LLMs, such as the GPT series \citep{radford2019language, brown2020language}, generate text by autoregressively sampling discrete units called tokens. Let $\Voca$ denote the vocabulary (token universe), and let $\token \in \Voca$ be a single token. At each step $t$, given the prefix $\token_{1:(t-1)} := \token_1 \token_2 \cdots \token_{t-1}$, the model computes a multinomial distribution $\bP_t := (P_{t,w})_{w \in \Voca}$ over the vocabulary, and samples the next token $\token_t \sim \bP_t$. This distribution $\bP_t$, called the next-token prediction (NTP) distribution, is unknown to the verifier, as it depends on the user-supplied prompt, and potentially hidden system prompts \citep{vaswani2017attention}.

\paragraph{A general framework for watermarking.}
We adopt a general three-party framework that reflects real-world applications of watermark-based verification \citep{kuditipudi2023robust,xie2024debiasing}.  
The language model provider (e.g., OpenAI) embeds statistical watermark signals into the generated text using a designated watermarking scheme. Under this setup, every token generated by the model is watermarked by design.  
The user (e.g., a student) interacts with the model, receives the watermarked output, and may optionally modify the text before sharing or submitting it.  
The verifier (e.g., a teacher) is tasked with estimating how many tokens in a given document originated from the watermarked LLM.

Watermarking schemes operate by modifying the standard sampling process to embed statistical signals that can later be recovered for verification. This is typically achieved by introducing a pseudorandom variable $\xi_t$ at each step $t$, which serves similarly as a random seed that guides token selection. 
Instead of sampling $\token_t$ directly from $\bP_t$, the model applies a decoding rule $\token_t := \SM(\bP_t, \xi_t)$, where $\SM$ is a (possibly stochastic) decoding function. The pseudorandom variable $\xi_t$ is generated using cryptographic techniques and can be deterministically reconstructed by the verifier if the construction is known.
In this work, $\xi_t = \AM(\token_{(t-m):(t-1)}, \Key)$ for some hash function $\AM$, context window size $m$, and secret key $\Key$. Given the full text $\token_{1:n}$, the hash function, and the key, one can reconstruct $\xi_{1:n}$. In this work, we assume $\xi_t$ are i.i.d. samples from a known distribution, consistent with the perfect pseudorandomness assumption used in prior work \citep{wu2023dipmark, zhao2024permute}. A watermark is called unbiased if the resulting token still marginally follows the original NTP distribution $\bP_t$; otherwise, it is biased \citep{li2024statistical}.

In contrast, human-written tokens are also drawn from $\bP_t$ but do not depend on $\xi_t$, leading to a different joint distribution over $(\token_t, \xi_t)$.
Detection thus reduces to testing whether $\token_t$ and $\xi_t$ are statistically dependent. For human-written text, they are independent, since human users know nothing about the pseudorandomness; for watermarked text, dependence is introduced through the decoding rule $\SM$.
To detect this dependence, \citet{li2024statistical} proposed computing a pivotal statistic $Y_t = Y(\token_t, \xi_t)$, designed to control Type I error and boost power.
These statistics are scalars that summarize the degree of dependence between $\token_t$ and $\xi_t$, which follow a known distribution under the null (i.e., no watermark). Specifically, $Y_t \sim \mu_0$ i.i.d. when no watermark is present, and $Y_t \mid \bP_t \sim \mu_{1, \bP_t}$ under the alternative due to watermark signals.
Many watermarking schemes can be cast in this form, and detection is based solely on the sequence $Y_{1:n}$. This approach enables the verifier to statistically infer the presence and extent of watermarked content, even without access to the original prompt or NTP distributions.

\begin{table}[!t]
\vspace{-0.1in}
\caption{Summary of three watermarks. }
\label{tab:comparison}
\vspace{-0.1in}
\centering
\resizebox{\linewidth}{!}{
\begin{tabular}{c|c|c|c|c|c}
\toprule
\textbf{Watermark} & \textbf{Pseudorandom $\xi$} & \textbf{Decoder $\SM(\bP, \xi)$} & 
\parbox{3.2cm}{\centering \textbf{Pivotal statistic} \\ $Y(\token, \xi)$ }
 & \textbf{$\mu_0$} & \textbf{$\mu_{1,\bP}$} \\
\midrule
\multirow{2}{*}{\makecell{Gumbel-max \\ \citep{scott2023watermarking}}} & 
\parbox{3.2cm}{\centering $\xi = (U_1, \ldots, U_{|\Voca|})$ with each $U_{\token}$ i.i.d. from $\UM(0, 1)$} 
& \parbox{3.1cm}{\centering Gumbel-max trick: \\
$\arg\max_{\token} \frac{\log \xi_{\token}}{P_\token}$ (which follows $\bP$)} 
& \parbox{2.8cm}{\centering $U_{w}$}  
& $\UM(0, 1)$  
& \parbox{3.2cm}{\centering The CDF is: $\sum\limits_{\token} P_{\token} r^{1/P_{\token}}$ with $r \in [0, 1]$
\citep{li2024statistical}} \\
\midrule
\multirow{2}{*}{\makecell{Inverse transform \\\citep{kuditipudi2023robust}}} &
\parbox{3.2cm}{\centering $\xi= (U, \pi)$ with $U \sim \UM(0, 1)$ and $\pi$ a permutation on $\Voca$}  
& \parbox{3.1cm}{\centering $\pi^{-1} \circ F_{\pi}^{-1}(U)$ where $F_\pi^{-1}$ is the generalized inverse of $F_\pi$} 
& \parbox{3.1cm}{\centering $1-|U - \eta_{\pi}(\token)|$ where $\eta_{\pi}(\token) = \frac{\pi(\token)-1}{|\Voca|-1}$}  
& \parbox{2.8cm}{\centering Asymp. CDF: $r^2$ with $r \in [0, 1]$} 
& \parbox{3.2cm}{\centering
Asymp. CDF is $(1-\frac{1-r}{P_{(1)}})^2$ with $r \in [1-P_{(1)}, 1]$ \citep{li2024statistical}} \\
\midrule
\makecell{Green-red list \\
\citep{kirchenbauer2023watermark}} & 
\parbox{3.2cm}{\centering $\xi = \GM$ which is a random set of $\Voca$ with $|\GM| = |\Voca| \cdot \gamma $}  
& \parbox{3.1cm}{\centering sampled from $\bP_{gr}$
where $(P_{gr})_{\token} \propto P_\token \cdot \exp(\delta \1_{\token \in \GM}) $
} 
& \parbox{3.1cm}{\centering $\1_{\token \in \GM}$, whether the token $\token$ is in the green set $\GM$.}  
& \parbox{2.8cm}{\centering $\mathrm{Ber(\gamma)}$} 
& \parbox{3.2cm}{\centering $\mathrm{Ber(\mu)}$ where $\mu = \frac{\sum\limits_{\token \in \GM} P_{\token} \exp(\delta) }{\sum\limits_{\token \in \GM} P_{\token} \exp(\delta) + \sum\limits_{\token \notin \GM } P_{\token} }$ \citep{kirchenbauer2023watermark}} \\
\bottomrule
\end{tabular}}
\vspace{-0.1in}
\end{table}

\paragraph{Three representative watermarks.}

We study three representative watermarking schemes under the general decoding framework described above. These schemes differ primarily in the choice of decoder $\SM$ and the construction of the pseudorandom number $\zeta_t$. Once these components are specified, the watermarking scheme is fully determined. For detection purposes, once the pivotal statistics are defined, the null distribution $\mu_0$ and the alternative distributions $\mu_{1, \bP}$ are also fixed. Therefore, it suffices to highlight these structural differences. Table~\ref{tab:comparison} summarizes the key components of each scheme.

We briefly introduce each scheme below.
The Gumbel-max watermark \citep{scott2023watermarking}, recognized as the first unbiased watermark, has been implemented internally at OpenAI. It uses the Gumbel-max trick \citep{gumbel1948statistical} to sample tokens, which by construction ensures unbiasedness.
The inverse transform watermark \citep{kuditipudi2023robust} applies inverse transform sampling to the CDF $F_\pi(x) = \sum_{\token} P_{\token} \1_{\pi(\token) \le x}$, which defines a distribution over the vocabulary $\Voca$ based on a permutation $\pi$.
The green-red list watermark, though biased, is widely used due to its simplicity. It partitions the vocabulary into green and red token sets, secretly boosts the probability of green tokens, and signals watermarking if the observed frequency of green tokens exceeds an expected threshold. Follow-up work has improved this scheme through parameter tuning \citep{wouters2024optimizing} and by developing unbiased \citep{hu2023unbiased, wu2023dipmark, xie2024debiasing} or robust variants \citep{zhao2024provable, zhu2024duwak}.

\vspace{-0.1in}
\section{Proportion Estimation Methods}
\label{sec:main-results}

In this section, we study how to estimate the proportion of watermarked content in mixed-source text. Section \ref{sec:estimand} defines the problem, Section \ref{sec:estimability} analyzes its estimability, and Section \ref{sec:method} presents our estimators. 

\subsection{Target Estimand}
\label{sec:estimand}
The first step is to identify the quantity we aim to estimate and the specific model where it arises.
Most existing watermark detection methods rely on pivotal statistics \citep{li2024statistical}, which provide quantitative evidence of watermark signals at each token–pseudorandom pair $(\token_t, \xi_t)$.
A key property of these statistics is the following structural dichotomy. 
When the watermark signal in $(\token_t, \xi_t)$ is disrupted---such as by modifying $\token_t$ or altering its preceding tokens so that $\xi_t$ no longer generates $\token_t$---$\token_t$ becomes statistically independent of $\xi_t$ due to the statistical sensitivity of hash functions. 
In such cases, the pivotal function $Y$ ensures that $Y_t = Y(\token_t, \xi_t)$ always follows a known distribution $\mu_0$, regardless of the distribution of $\token_t$. Conversely, if the watermark signal in $(\token_t, \xi_t)$ remains intact, $\token_t$ continues to depend on $\xi_t$ via the decoder $\SM$ so that $Y_t \sim \mu_{1, \bP_t}$ with $\bP_t$ the NTP distribution for $\token_t$. 
So, given $\bP_t$, $Y_t$ follows either the null $\mu_0$ or the alternative $\mu_{1, \bP_t}$.

However, the verifier does not observe which pivotal statistics retain watermark dependence, and our objective is not to localize them individually. 
Instead, we estimate the aggregate proportion of watermark-preserving pivotal statistics in a given text, denoted by $\eps$. 
This leads to the following mixture formulation.

\begin{model}[\citet{li2024optimal}]
\label{model:previous}
The pivotal statistic $Y_t$ satisfies
\begin{equation}
\label{eq:hypothesis}
Y_t \mid \bP_t \sim (1-\eps) \mu_0 + \eps \mu_{1, \bP_t}, \quad \forall t \in [n],
\end{equation}
where $\eps$ represents the proportion of watermark signals, our estimand of interest. Rigorously, there exists a $\sigma$-field $\{\FM_t\}_{t \ge 1}$ so that $\bP_t \in \FM_{t-1}$ and $Y_t \in \FM_t$ with
\[
\EB[\mathbf{1}\{Y_t \le x\} | \FM_{t-1}] = (1-\eps) F_0(x) + \eps F_{\bP_t}(x).
\]
Here, $\bP_t$ depends on the history of tokens $\token_{1:(t-1)}$, so we view it as $\FM_{t-1}$-measurable.
\end{model}

In this model, $\eps$ represents the marginal proportion of pivotal statistics that preserve watermark dependence, rather than the modification probability of individual tokens. Thus, it captures the ground-truth watermark-retention rate under the verifier's information constraints. The model is most appropriate for approximately position-homogeneous editing patterns, as this retention rate is assumed to be constant across $t$.\footnote{This modeling may fail, for example, if human editors systematically revise some parts of a text, such as the beginning of a paragraph, more heavily than others.}

\begin{remark}
The $\sigma$-field $\FM_t$ formalizes the information flow from the verifier’s perspective up to token $t$. 
It includes the variables that determine the conditional law of $Y_t$, such as the token history $\token_{1:t}$, the (latent) NTP distributions $\bP_{1:(t+1)}$, and the (latent) editing indicators.
Although $\FM_t$ is used to specify conditional distributions, some of its components---such as $\bP_t$ and the editing status---are not observable to the verifier. 
The model therefore represents the verifier’s uncertainty about watermark integrity through a conditional mixture structure, without imposing structural assumptions on the editing mechanism.
\end{remark}

\begin{remark}
It is worth noting that the watermark proportion $\eps$ is defined in terms of pivotal statistics rather than individual tokens. As introduced in Section~\ref{sec:pre}, each pivotal statistic is a deterministic function of $(m+1)$ tokens. As a result, modifying a single token may affect $(m+1)$ pivotal statistics, leading to a discrepancy between the proportion of watermarked tokens and that of watermarked pivotal statistics. We adopt this definition for its simplicity and because nearly all existing detection rules rely on pivotal statistics.
\end{remark}

\subsection{Estimability}
\label{sec:estimability}

The next question is whether the proportion $\eps$ is well-defined---or, more precisely, identifiable---when the NTP distributions $\bP_{1:n}$ are unknown and potentially vary, as is the case for real-world verifiers. Identifiability, in this context, means that $\eps$ can be uniquely determined from the observed pivotal statistics alone, even when $\bP_{1:n}$ are unknown and arbitrary.
If $\eps$ is not identifiable, multiple values of $\eps$ could yield the same data distribution, making it impossible to obtain a unique and accurate estimate. This ambiguity challenges the interpretability of any downstream use of the estimate, as its real-world meaning is unclear.

\paragraph{Impossibility for the green-red list watermark.}
Unfortunately, for the green-red list watermark, the proportion $\eps$ is not identifiable. To illustrate this, consider a straightforward example of proportion estimation in an i.i.d.\ binary mixture model.
\begin{lem}
\label{lem:non-identifiable}
If data points are i.i.d.\ from the binary mixture $(1-\eps) \mathrm{Ber}(\gamma) + \eps \mathrm{Ber}(\mu)$ where both $\eps$ and $\mu$ are unknown while $\gamma$ is known, then $\eps$ is not identifiable.
\end{lem}


The real-world use of green-red list watermarks is more complex than the simplified setting in Lemma~\ref{lem:non-identifiable}. In practice, each $Y_t \mid \bP_t \sim (1 - \eps) \mathrm{Ber}(\gamma) + \eps \mathrm{Ber}(\mu_t)$, where $\mu_t$ depends on the unknown NTP distribution $\bP_t$. The expression for $\mu_t$ is given in Table~\ref{tab:comparison}. As a result, we arrive at the following corollary. The non-identifiability in this case arises because each $Y_t$ is binary and thus carries limited information about the underlying parameters.
\begin{cor}
\label{cor:non-identify-green-red-list}
Under Model~\ref{model:previous}, $\eps$ is non-identifiable for the green-red list watermark. 
\end{cor}

\paragraph{Identifiability for continuous pivotal statistics.}
Interestingly, the estimation problem becomes fundamentally different when $Y_t$ is continuous. Specifically, by assuming \(\mu_{1, \bP}\) differs uniformly from \(\mu_0\) as in Assumption \ref{asmp:main}, \(\eps\) becomes identifiable (Lemma \ref{lem:identifiable}).

\begin{asmp}
\label{asmp:main}
Let $F_0$ and $F_{\bP}$ denote the CDFs of $\mu_0$ and $\mu_{1,\bP}$, with densities $f_0$ and $f_{\bP}$, respectively. We assume that (i) both CDFs are absolutely continuous and vanish at zero, (ii) $F_{\bP}(x)/F_0(x)\to0$ as $x\to0$ whenever $\bP$ is non-singular (no entry equals $1$), (iii) $f_{\bP}(x)/f_0(x)$ is uniformly bounded over $x$ and $\bP$, $F_0$-almost everywhere, and (iv) all NTP distributions are non-singular almost surely.
\end{asmp}

\begin{lem}
\label{lem:identifiable}
Under Assumption \ref{asmp:main} and Model \ref{model:previous}, $\eps$ is identifiable. 
\end{lem}

This result applies to both the Gumbel-max and inverse transform watermark schemes, as their pivotal statistics are continuous and satisfy Assumption~\ref{asmp:main}. As a result, $\eps$ is identifiable for these two watermarks. 

\begin{lem}\label{lem:identifiable-two-watermarks}
Assumption \ref{asmp:main} holds for the Gumbel-max and inverse transform watermark.
\end{lem}

\begin{remark}
For the inverse transform watermark, \citet{li2024statistical} obtains the simple limiting CDFs in Table~\ref{tab:comparison} under the condition $P_{(2)} \cdot \log |\Voca| \to 0$ as $|\Voca| \to \infty$. This condition is used only to justify the limiting approximation. The exact finite-vocabulary CDFs satisfy Assumption~\ref{asmp:main} without this condition; see the proof of Lemma~\ref{lem:identifiable-two-watermarks}.
\end{remark}

\subsection{Estimation Methods}
\label{sec:method}

The proportion $\eps$ satisfies the equation
\begin{equation}
\label{eq:eps-equation}
\Bar{F}(x) = (1 - \eps) F_0(x) + \eps \Bar{F}_{\bP}(x)~\text{for any}~x,
\end{equation}
where $\Bar{F}(x) = \frac{1}{n} \sum_{t=1}^n \PB(Y_t \le x)$ is the averaged marginal CDF of $Y_t$, $F_0(x)$ is the null CDF, and 
$\Bar{F}_{\bP}(x) = \frac{1}{n} \sum_{t=1}^n \EB[F_{\bP_t}](x)$ is the expected averaged CDF under the watermarked alternative.
Equation \eqref{eq:eps-equation} implies that for any weight function $v:\RB \mapsto \RB$, it must satify
\[
\EB_{\bar{F}}[v] = (1-\eps) \EB_{{F}_0}[v] + \eps \EB_{\Bar{F}_{\bP}}[v],
\]
where expectations $\EB_{\bar{F}}[v]$, $\EB_{F_0}[v]$, and $\EB_{\Bar{F}_{\bP}}[v]$ represent integrals with respect to $\Bar{F}$, $F_0$, and $\Bar{F}_{\bP}$, respectively. Rearranging this expression gives
\begin{equation}
\label{eq:general-eps-equation}
\eps = \frac{\EB_{{F}_0}[v] - \EB_{\bar{F}}[v]}{\EB_{{F}_0}[v] - \EB_{\Bar{F}_{\bP}}[v]}.
\end{equation}
Equation \eqref{eq:general-eps-equation} forms the basis for estimators inspired by moment methods \citep{blischke1962moment}, which match observed sample moments with theoretical ones. Here, $\EB_{F_0}[v]$ is computable because $F_0$ is known and can be approximated through Monte Carlo simulation. For the observed mixture samples $Y_1, \ldots, Y_n$, $\EB_{\bar{F}}[v]$ is estimated by the sample average $\frac{1}{n} \sum_{t=1}^n v(Y_t)$. However, the term $\EB_{\Bar{F}_{\bP}}[v]$ is more challenging because $\Bar{F}_{\bP}$ is unknown and depends on the unknown NTP distributions $\bP_1, \ldots, \bP_n$, which also vary across tokens. Estimating these NTP distributions directly is impractical due to their high dimensionality and variability. For example, in models like OPT-1.3B, the vocabulary size $|\Voca| = 50,272$ \citep{zhang2022opt}, meaning each $\bP_t$ is a distribution over tens of thousands of tokens. With text lengths $n$ often around 100, the number of parameters far exceeds the number of samples, making estimation infeasible.

\paragraph{Initial attempt by ignoring $\Bar{F}_{\bP}$.}  
To simplify, we initially ignore $\Bar{F}_{\bP}$ by choosing the weight function $\vind(x) = \1\{x \le \delta\}$, where $\delta > 0$ is a small constant specified by the verifier. For small $\delta$, the unknown term $\Bar{F}_{\bP}(\delta)$ becomes negligible compared to $F_0(\delta)$ because Assumption \ref{asmp:main} implies that $\lim_{\delta \to 0} \frac{\Bar{F}_{\bP}(\delta)}{F_0(\delta)} = 0$.
This leads to the following estimator:
\begin{equation}
\label{eq:fraction}
\epsinital(\delta) = 1 - \frac{\widehat{F}(\delta)}{F_0(\delta)},
\end{equation}
where $\widehat{F}(x) = \frac{1}{n} \sum_{t=1}^n \1\{Y_t \le x\}$ is the empirical CDF of the observed pivotal statistics. While this estimator is simple, it has two significant limitations. First, relying on a small $\delta$ means only a few samples are used which reduces statistical efficiency. For example, if $\delta$ is chosen too small, most observed values $Y_t$ may exceed $\delta$, leading to unstable estimates due to limited sample sizes. Second, the choice of $\vind(x)$ is heuristic, and better weight functions may improve performance.

\paragraph{Refinement by estimating $\Bar{F}_{\bP}$.}  
To overcome these limitations, we refine the estimator by approximating $\Bar{F}_{\bP}$ instead of assuming it is negligible. Instead of estimating each $\bP_t$ individually, we estimate the averaged alternative CDF $\Bar{F}_{\bP}$ using pivotal statistics collected from a comparable open-source model. Specifically, let $\YWM_1, \ldots, \YWM_N$ denote pivotal statistics derived from purely watermarked texts, and define their empirical CDF as $\widehat{F}_{\bP}(x) = \frac{1}{N} \sum_{t=1}^N \1\{\YWM_t \le x\}$. Since this estimation process can be performed prior to observing our samples, $\widehat{F}_{\bP}$ is treated as a non-random CDF. Incorporating this additional information, the refined estimator is given by
\begin{equation}
\label{eq:refined-estimator}
\epsrefine(\delta) = \frac{F_0(\delta) - \widehat{F}(\delta)}{F_0(\delta)-\widehat{F}_{\bP}(\delta)}.
\end{equation}

\paragraph{Optimal weight function.}  
Finally, we address the choice of weight function $v$. To gain some insights, let's consider an ideal (albeit impractical) case where $\Bar{F}_{\bP}$ is perfectly known such that $\widehat{F}_{\bP} = \Bar{F}_{\bP}$. In this case, for a given weight function $v$, the resulting estimator $\frac{\EB_{{F}_0}[v] - \EB_{\widehat{F}}[v]}{\EB_{{F}_0}[v] - \EB_{\Bar{F}_{\bP}}[v]}$ is unbiased for $\eps$.
To quantify its fluctuations, we bound the deviation of $\EB_{\widehat{F}}[v]$ using martingale concentration inequalities, such as Freedman’s inequality. These bounds are governed by the predictable quadratic variation, which is defined by $\left\langle \EB_{\widehat{F}}[v] \right\rangle := \frac{1}{n^2}\sum_{t=1}^n \Var\!\left(v(Y_t)\mid \FM_{t-1}\right)$ for the random variable $\EB_{\widehat{F}}[v] := \frac{1}{n}\sum_{t=1}^n v(Y_t)$.
Therefore, the expected deviation of the above ideal estimator is controlled by this quantity, which leads to the upper bound in \eqref{eq:upper-bound-variance} (see Appendix \ref{proof:upper-bound-variance} for the proof).

\begin{equation}\label{eq:upper-bound-variance}
\EB\left\langle\frac{\EB_{{F}_0}[v] - \EB_{\widehat{F}}[v]}{\EB_{{F}_0}[v] - \EB_{\Bar{F}_{\bP}}[v]}\right\rangle \le \frac{\Var_{\bar{F}}(v)}{n (\EB_{{F}_0}[v] - \EB_{\Bar{F}_{\bP}}[v])^2}.
\end{equation}

\begin{remark}
\label{rmk:tightness}
One can verify that \eqref{eq:upper-bound-variance} becomes an equality when all NTP distributions are identical and deterministic. In other words, the bound is attained in the extreme case where $Y_1, \ldots, Y_n$ are iid. Accordingly, the minimax lower bound is tight in this regime.
\end{remark}

We then minimize this variance upper bound to determine the optimal weight function.
\begin{lem}[Optimal weight function] 
\label{lem:optimal-weight}
For Inequality \eqref{eq:upper-bound-variance}, the optimal value over all weight functions satisfies
\[
\min_{v} \frac{\Var_{\bar{F}}(v)}{[\EB_{{F}_0}[v] - \EB_{\Bar{F}_{\bP}} [v]]^2} 
= \left[ \int \frac{[1-g(x)]^2}{(1-\eps)+\eps g(x)}\rd F_0(x)\right]^{-1}
\]  
and the optimal solution (up to constant factors) is
\begin{equation}
\label{eq:optimal-weight-true}
\vopt(x) = \frac{1 - g(x)}{(1-\eps) + \eps g(x)}, 
~\text{with}~
g(x) = \frac{\rd \Bar{F}_{\bP}(x)}{\rd F_0(x)}.
\end{equation}
\end{lem}

However, directly using this optimal weight function $\vopt$ presents a challenge: it depends on both $\Bar{F}_{\bP}$ and $\eps$, which are unknown. While $\Bar{F}_{\bP}$ can be estimated empirically by $\widehat{F}_{\bP}$, substituting $\vopt$ into \eqref{eq:general-eps-equation} introduces $\eps$ into both sides of the equation, creating a self-referential problem. To resolve this, we define the refined estimator $\epsopt$ as the fixed point of the following operator, that is, $\epsopt = \widehat{\TM}(\epsopt)$ and
\begin{equation}
\label{eq:emprical-operator}
\widehat{\TM}(\eps) = \PM\left(\frac{\int \hvopt(\eps, x) \left[\rd F_0(x) - \rd \widehat{F}(x)\right]}{\int \hvopt(\eps, x) \left[\rd F_0(x) - \rd \widehat{F}_{\bP}(x)\right]}\right),
\end{equation}
where $\PM(x)$ projects $\eps$ onto the interval $[\eps_{\min}, 1-\eps_{\min}]$. Here, $\eps_{\min}$ is a sufficiently small number (e.g., $10^{-3}$) to ensure numerical stability and $\hvopt$ is the empirical version of the optimal weight function, defined as:
\begin{equation}
\label{eq:optimal-weight}
\hvopt(\eps, x) = \frac{1 - \widehat{g}(x)}{(1-\eps) + \eps \widehat{g}(x)}, \quad \widehat{g}(x) = \frac{\rd \widehat{F}_{\bP}(x)}{\rd F_0(x)}.
\end{equation}
Here, $\widehat{g}(x)$ represents an empirical estimate of the density ratio function ${g}(x) = \frac{\rd \Bar{F}_{\bP}(x)}{\rd F_0(x)}$, with its specific form detailed in the experiments section. Since $\widehat{g}(x)$ can be computed prior to the proportion estimation, we treat it as a non-random function in the theoretical analysis.

The intuition behind $\epsopt$ is that it balances two key factors: the weight function $\hvopt$ leverages information about $\widehat{F}_{\bP}$ to improve accuracy, while the fixed-point formulation ensures that the unknown $\eps$ is consistently estimated. This approach effectively incorporates both theoretical insights and empirical data to refine the estimation process.
As Lemma \ref{lem:contraction-mapping} implies, $\epsopt$ generally exists and is unique because the operator is a contraction mapping when $\widehat{g}$ is sufficiently accurate.
\begin{lem}
\label{lem:contraction-mapping}
Assume that $0 < \TV( F_0, \widehat{F}_{\bP})$, $ \sup_{x}|\widehat{g}(x)| < \infty$, and $\EB_{F_0}[\widehat{g}]^{-2} < \infty$.
Define $\bgP = \frac{1}{n} \sum_{t=1}^n \frac{\rd F_{\bP_t}}{\rd F_0}$ and $\Delta_g = \EB_{\bP} \left[\sup_{x \in [0, 1]}\frac{|\widehat{g}(x)-\bgP(x)|}{\min\{1, \widehat{g}(x) \}} \right]$.
If $\Delta_g$ is sufficiently small and $n$ is sufficiently large, then with high probability, $\widehat{\mathcal{\TM}}$ is a contraction mapping on the interval $[\eps_{\min}, 1 - \eps_{\min}]$ for small values of $\eps_{\min}$.
\end{lem}

\section{Theoretical Analysis}
In this section, we present a theoretical analysis of the estimators introduced earlier. Formal theories and detailed proofs for all results are provided in Appendix \ref{app:proofs}.

\subsection{Accuracy for Initial Attempts}
\label{sec:accuray-init}

\begin{theorem}
\label{thm:upper-bound-eps-inital}
Let $\eps \in [0, 1]$ and $\delta \in (0, 1]$ be fixed.
Under Model \ref{model:previous}, it follows that 
\begin{equation*}
\EB \left|\epsinital(\delta) - \eps  \right|
\le  \frac{\sigma_n}{\sqrt{n}} + \eps  \frac{\bar{F}_{\bP}(\delta)}{F_0(\delta)}
\quad \text{where} \quad
\sigma_n^2 = \frac{\bar{F}(\delta)(1-\bar{F}(\delta))}{[F_0(\delta)]^2},
\end{equation*}
\end{theorem}

\begin{cor}
\label{cor:no-bias}
Following Theorem \ref{thm:upper-bound-eps-inital}, we have
\[
\EB \left| \epsinital(\delta_n) -\eps\right| \le 
\frac{1}{\sqrt{n F_0(\delta_n)}} \cdot \sqrt{(1-\eps)+ \eps\frac{\Bar{F}_{\bP}(\delta_n)}{F_0(\delta_n)}}
+ \eps \frac{\Bar{F}_{\bP}(\delta_n)}{F_0(\delta_n)}.
\]
Under Assumption \ref{asmp:main}, the right hand side $\to 0$ if $\delta_n \to 0$ and $ n F_0(\delta_n) \to \infty$ as $n \to \infty$.
\end{cor}

Theorem \ref{thm:upper-bound-eps-inital} shows that the mean absolute error (MAE) of $\epsinital(\delta)$, i.e., $\EB |\epsinital(\delta) - \eps|$, is of the order $\frac{\sigma_n}{\sqrt{n}} + \text{bias}$. The noise term, $\frac{\sigma_n}{\sqrt{n}}$, arises from the randomness in the collected data. 
The bias term, $\eps \frac{\Bar{F}_{\bP}(\delta)}{F_0(\delta)}$, is due to neglecting the unknown term $\Bar{F}_{\bP}$ in the estimator $\epsinital$. In general, this bias cannot be eliminated, as we will show in Section \ref{sec:inherent-bias} using the green-red list watermark as an example.  
However, for watermarks satisfying Assumption \ref{asmp:main}, this bias can be mitigated by setting $\delta$ sufficiently small, leveraging the condition $\lim\limits_{\delta \to 0} \frac{\Bar{F}_{\bP}(\delta)}{F_0(\delta)} = 0$, as show in Corollary \ref{cor:no-bias}.
In summary, adaptively tuning $\delta$ can reduce the bias and make the estimator asymptotically unbiased.

\subsection{Accuracy for Refinements}
\label{sec:accuray-refinement}

\begin{theorem}
\label{thm:accuracyRfn}
Let $\Delta_g$ be defined in Lemma \ref{lem:contraction-mapping}.
Let $o_g(1)$ (or $o(1)$) denote a quantity that converges to zero when $\Delta_g$ (or $\frac{1}{n}$) converges to zero. 
Under Model \ref{model:previous}, it follows that 
\[
\EB |\epsrefine(\delta) - \eps| \le   \frac{{\sigma}_n^{\star}}{\sqrt{n}} + o_g(1).
\]
If $0 < \TV( F_0, \Bar{F}_{\bP})$, $\sup\limits_{x}|{g}(x)| < \infty$ and $\EB_{F_0}[g]^{-2} < \infty$, then either setting $\eps_{\min}$ as a small constant or choosing $\eps_{\min} = \frac{\log n}{\sqrt{n}}$, we have that
\begin{align}
\label{eq:error-refine}
  \EB |\epsopt - \eps| &\lesssim   \frac{{\tau}_n^{\star} + o(1)}{\sqrt{n}} + o_g(1)+\eps_{\min}\1\{\eps \notin [\eps_{\min}, 1-\eps_{\min}]\}.
\end{align}
Here, $\sigma_n^\star \geq \tau_n^\star$ are two variance terms defined by
\begin{equation}
\label{eq:two-variance}
[\sigma_n^\star]^2 = \frac{\bar{F}(\delta)(1-\bar{F}(\delta))}{[F_0(\delta) - \Bar{F}_{\bP}(\delta)]^2}
\quad \text{and} \quad
[\tau_n^\star]^2  = \left[ \int \frac{[1-g(x)]^2}{(1-\eps)+\eps g(x)}\rd F_0( x)\right]^{-1}. 
\end{equation}
\end{theorem}
\begin{remark}
The error bound \eqref{eq:error-refine} remains valid when $ \EB_{F_0}[g]^{-2} = \infty$ and $\eps_{\min} = \frac{\log n}{n^{1/4}}$, but achieving it requires a stronger condition: $ n^{1/4} \cdot \Delta_g \to 0 $. See Appendix \ref{proof:accuracyRfn-final} for details. The same argument also holds for Lemma \ref{lem:contraction-mapping}. See Appendix \ref{proof:accuracyRfn} for details.
\end{remark}

\begin{remark}
We derive the optimal estimator $\epsopt$ under the idealized condition that $\Bar{F}_{\bP}$ is perfectly known, that is, $\widehat{F}_{\bP} = \Bar{F}_{\bP}$. 
However, Theorem~\ref{thm:accuracyRfn} does not rely on this assumption.
In Theorem~\ref{thm:accuracyRfn}, the quantity $\Delta_g$ measures the accuracy of $\widehat g$. 
The term $o_g(1)$ captures the effect of the deviation of $\widehat g$ and vanishes as $\Delta_g \to 0$.
\end{remark}

Theorem \ref{thm:accuracyRfn} provides an analysis of the MAEs for the refined estimators $\epsrefine(\delta)$ and $\epsopt$. The analysis assumes that the density ratio estimate $\widehat{g}$ is sufficiently accurate, satisfying the condition $\Delta_g$ being small. 
For $\epsrefine(\delta)$, the MAE is given by $\frac{\sigma_n^\star}{\sqrt{n}}$. In contrast, analyzing the MAE for $\epsopt$ requires additional moment conditions, which are necessary to study the fixed-point solutions of the operator $\widehat{\TM}$ in \eqref{eq:emprical-operator}. Under these conditions, $\epsopt$ achieves an MAE of $\frac{\tau_n^\star}{\sqrt{n}}$, if we ignore higher-order terms and assume the stability constant $\eps_{\min}$ serves as a valid lower bound for the true proportion $\eps$.
Note that $[\tau_n^\star]^2$ corresponds to the optimal value of the minimization problem in Lemma \ref{lem:optimal-weight}, as a result of which, $\sigma_n^\star \geq \tau_n^\star$. This guarantees that $\epsopt$ consistently achieves a smaller MAE than $\epsrefine(\delta)$, highlighting the advantage of using the optimal weight function. 
As will be shown in the experiments, $\epsopt$ not only achieves better accuracy but also exhibits higher stability.

\subsection{Minimax Lower Bounds}

\begin{theorem}
\label{thm:minimax}
For a given $n$, let $Y_1, \ldots, Y_n$ be generated under Model \ref{model:previous}, where $\eps_{0}$ represents the true proportion.
For a fixed pair $(\eps,\bar F_{\bP})$, define the model class
\[
\MM(\eps,\bar F_{\bP}) := \Bigl\{
\text{LLMs with true proportion $\eps$ that generate } \{\bP_t\}_{t=1}^n
\ \text{satisfying} \
\bar F_{\bP} = \frac{1}{n}\sum_{t=1}^n \EB[F_{\bP_t}] \Bigr\}.
\]
Denote $\BM(\eps)$ by a small open interval containing $\eps$. 
Let $[\sigma_n^\star]^2$ and $[\tau_n^\star]^2$ be the variance term defined in \eqref{eq:two-variance} and $\sup_{\MM(\eps_0,\bar F_{\bP})}$ corresponds to the worst LLM within that class.
\begin{enumerate}[label=$(\alph*)$]
    \item If $\GM$ is the set of measurable functions of the indicators $\mathbf{1}\{Y_1 \leq \delta\}, \ldots, \mathbf{1}\{Y_n \leq \delta\}$, 
    \[
    \inf_{\widehat{\eps} \in \GM}
    \sup_{\eps_0 \in \BM(\eps)}
    \sup_{\MM(\eps_0,\bar F_{\bP})}
    \EB \bigl|\widehat{\eps} - \eps_0\bigr|
    \;\ge\;
    \frac{\sigma_n^\star}{\sqrt{2n}}.
    \]
    \item If $\GM$ is the set of measurable functions of the random vector $(Y_1, \ldots, Y_n)$, 
    \[
    \inf_{\widehat{\eps} \in \GM} 
    \sup_{\eps_0 \in \BM(\eps)}
    \sup_{\MM(\eps_0,\bar F_{\bP})}
    \EB\bigl|\widehat{\eps} - \eps_{0}\bigr|  \;\ge\; \frac{\tau_n^\star}{\sqrt{2n}}
    \]
\end{enumerate}
\end{theorem}

Theorem~\ref{thm:minimax} provides minimax lower bounds for the MAEs under Model~\ref{model:previous}. 
To obtain instance-dependent bounds, we take the worst case over all LLMs whose NTP sequences satisfy the aggregate condition $\bar F_{\bP} = \frac{1}{n}\sum_{t=1}^n \EB[F_{\bP_t}].$
In the proof, the worst case occurs when the $\bP_t$'s are drawn i.i.d.\ from a fixed distribution, as discussed in Remark~\ref{rmk:tightness}.
In our result, the lower bounds, $\sigma_n^\star/\sqrt{n}$ and $\tau_n^\star/\sqrt{n}$, depend only on the averaged CDF $\Bar{F}_{\bP}$. 
This quantity does not depend on the order of the pivotal statistics. 
It summarizes the overall behavior of the alternative distributions without requiring access to each individual $\bP_t$. 
In practice, such aggregate information can often be estimated using similar LLMs or related data.
In contrast, sharper bounds would require knowing the full sequence $\{\bP_t\}_{t=1}^n$, that is, which token corresponds to which NTP distribution. 
This information is typically unavailable in real verification settings.

The implications are twofold. First, these lower bounds are tight and achieved by our estimators, up to the estimation error $o_g(1)$ in estimating $\Bar{F}_{\bP}$: $\epsrefine(\delta)$ matches the bound $\frac{\sigma_n^\star}{\sqrt{n}}$ among indicator-based estimators, while $\epsopt$ attains the smaller bound $\frac{\tau_n^\star}{\sqrt{n}}$ among all measurable estimators. Second, achieving these bounds critically depends on incorporating the correction term $\Bar{F}_{\bP}$; without it, the initial estimator fails to reach the optimal rate, as detailed in Appendix~\ref{proof:inital-estimator}.

\subsection{Inherent Bias Caused by Non-identifiability}
\label{sec:inherent-bias}

The green-red list watermark does not satisfy Assumption \ref{asmp:main}, so we cannot expect the aforementioned estimators to obtain high accuracy for this watermark. One piece of evidence is that, although Theorem \ref{thm:upper-bound-eps-inital} still holds for the green-red list watermark, the bias term $\eps \frac{\Bar{F}_{\bP}(\delta)}{F_0(\delta)} = \eps \left(1 - \eps + \eps \frac{1 - \Bar{\mu}_t}{1 - \gamma}\right)$ does not depend on $\delta$ and will never vanish as $\delta \to 0$. 
This is consistent with the non-identifiability issue: $\bar{\mu}_t$ is unobservable since each $\bP_{t}$ is unobservable (see Table \ref{tab:comparison} for their relation), and the binary $Y_t$ signals provide negligible information about $\bar{\mu}_t$. Even attempting to estimate $\bar{\mu}_t$ fails to resolve this issue. For instance, applying MLE to the i.i.d. binary mixture model in Lemma \ref{lem:non-identifiable} to estimate unknown parameters $\mu$ and $\eps$ introduces inherent bias, as shown in Theorem \ref{thm:failure-inherent-bias}.


\begin{theorem}
\label{thm:failure-inherent-bias}
Let $Y_1, \ldots, Y_n \overset{\text{i.i.d.}}{\sim} (1 - \eps) \mathrm{Ber}(\gamma) + \eps \mathrm{Ber}(\mu)$, where only $\eps$ and $\mu$ ($1 > \mu > \gamma > 0$) are unknown. Let $\widehat{e}$ denote the average of these $n$ samples. Consider MLE that minimizes the $L_2$-regularized negative log-likelihood loss:
\begin{equation}
\label{eq:regularized-solutions}
(\widehat{\eps}_{\lambda}, \widehat{\mu}_{\lambda}) = \arg\min \left\{ - \log L(\eps, \mu) + \lambda (\mu^2 + \eps^2) \right\},
\end{equation}
where $L(\eps, \mu) = [(1 - \eps) \gamma + \eps \mu]^{\widehat{e}} \left[1 - (1 - \eps) \gamma - \eps \mu\right]^{1 - \widehat{e}}$. We then have the following results
\begin{equation}
\label{eq:limit-solutions}
\lim_{\lambda \to 0} \widehat{\eps}_{\lambda} = x \sqrt{x^2 + \gamma}, \quad \lim_{\lambda \to 0} \widehat{\mu}_{\lambda} = x^2 + \gamma,
\end{equation}
 for $n$ sufficiently large where $x > 0$ is the solution to $x^3 \sqrt{x^2 + \gamma} = \widehat{e} - \gamma$.
\end{theorem}

As we will see in Figure \ref{fig:simulation-bias}, the estimate $\widehat{\eps}$ almost never equals the ground truth $\eps$ even when $n$ is sufficiently large and $\lambda$ is very small. This case is very different from standard linear regression, where regularized MLE 
(almost surely) converges to the underlying true parameter as regularization diminishes and sample size diverges \citep{shao2003mathematical}. This discrepancy arises because non-identifiability is an inherent problem of the (watermarking) model, not a flaw of the estimation methods.

\section{Experiments}
\label{sec:LLM-experiments}

We first evaluate estimation accuracy and inherent bias on simulation datasets and then check performance on open-source LLMs. Additional details and results are in the appendix. The code is available at: \url{https://github.com/lx10077/WatermarkProportion}.

\subsection{Simulation Study}


Following \citep{li2024optimal}, we use a vocabulary of size $|\Voca| = 10^3$ and generate 2,500 sequences of 400 tokens each, yielding $n = 10^6$ pivotal statistics as watermarked samples. The generation process mimics LLM behavior with a control parameter $\Delta \in (0, 1)$. For each token $\token_t$, we construct its distribution $\bP_t$ by uniformly sampling the largest probability from $\UM(0, 1-\Delta)$ and interpolating between Zipf's law \citep{zipf2016human} and uniform distributions to ensure valid probabilities over $\Voca$ (Algorithm \ref{alg:dominate-Ps}). We then sample $\token_t \sim \bP_t$.
To simulate mixed-source data, we generate $n = 10^6$ null samples from $F_0$. Across 200 uniformly spaced values $\eps \in [0, 1]$, we create mixture datasets combining $n(1-\eps)$ null samples with $n\eps$ watermarked samples to estimate each $\eps$. We aggregate all watermarked statistics into a single dataset for estimating $\widehat{g}$ and $\widehat{F}_{\bP}$, though some pivotal statistics in this aggregated dataset may be entirely irrelevant to the current task of estimating a specific $\eps$.

\paragraph{An equivalence for implementation.}
Since $F_0$ is known, the probability integral transform \citep{dodge2003oxford} ensures that $F_0(Y_1), \ldots, F_0(Y_n)$ are i.i.d. $\UM(0, 1)$ samples. We can therefore assume without loss of generality that $\mu_0 = \UM(0, 1)$, with $\mu_{1, \bP} = F_{\bP} \circ F_0^{-1}$ where $F_0^{-1}$ is the generalized inverse. Assumption \ref{asmp:main} remains valid under this transformation.
For all experiments, we apply this transform to ensure $\mu_0 = \UM(0, 1)$, so our pivotal statistics are basically $F_0(Y_1), \ldots, F_0(Y_n)$. For simplicity, we continue denoting them as $Y_1, \ldots, Y_n$.


\paragraph{How we obtain $\widehat{g}$.}
The approach to estimating the density ratio $g$ is highly flexible, with many methods available in the statistical literature. 
We estimate the density ratio $g$ using a simple histogram-based approach. The data in $[0, 1]$ is divided into $N = 500$ bins, with bin heights normalized to form probability densities that integrate to 1, yielding a piecewise constant estimator. While more sophisticated methods like kernel density estimation could provide smoother estimates, they are computationally prohibitive for the large datasets ($n = 10^6$). 
The histogram approach offers better computational efficiency and scalability.


\begin{figure*}[t!]
\caption{
Figure \ref{fig:main-CDF} shows the CDFs of mixture sample distributions for the Gumbel-max watermark with $\Delta=0.1$ and varying $\eps$.
Figures \ref{fig:simulation-MAE-gumbel}, \ref{fig:simulation-MAE-inverse} and \ref{fig:simulation-MAE-greenredlist} show MAEs of three estimators across various ground truth $\eps$ values when $\Delta = 0.1$. 
Figure \ref{fig:simulation-bias} shows the MLE results for estimating both $\eps$ and $\mu$ in the case of i.i.d. non-identifiable mixture binary data.
}
\label{fig:simulation}
\centering
\begin{subfigure}{0.32\textwidth}
\includegraphics[width=\textwidth]{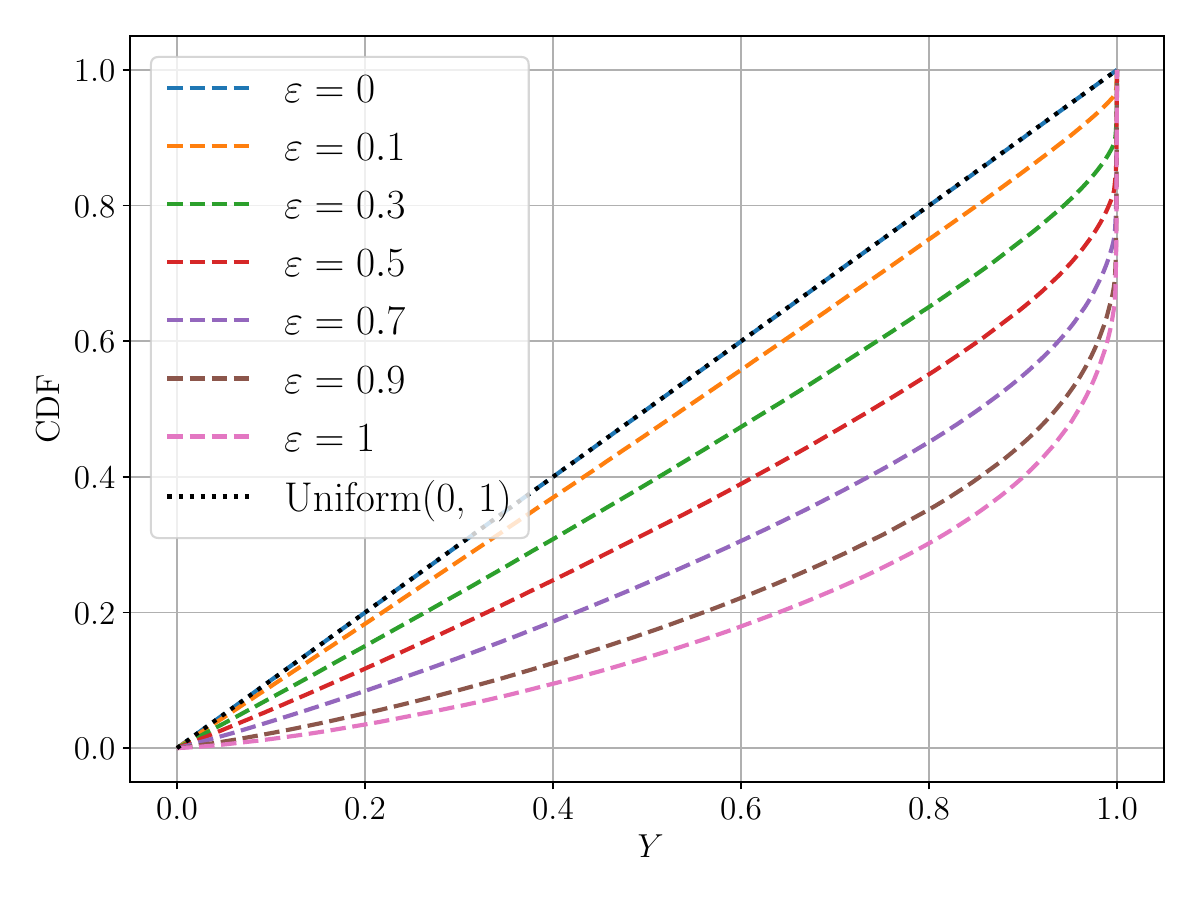}
\caption{CDFs of mixture dist.}
\label{fig:main-CDF}
\end{subfigure}
\begin{subfigure}{0.32\textwidth}
\includegraphics[width=\textwidth]{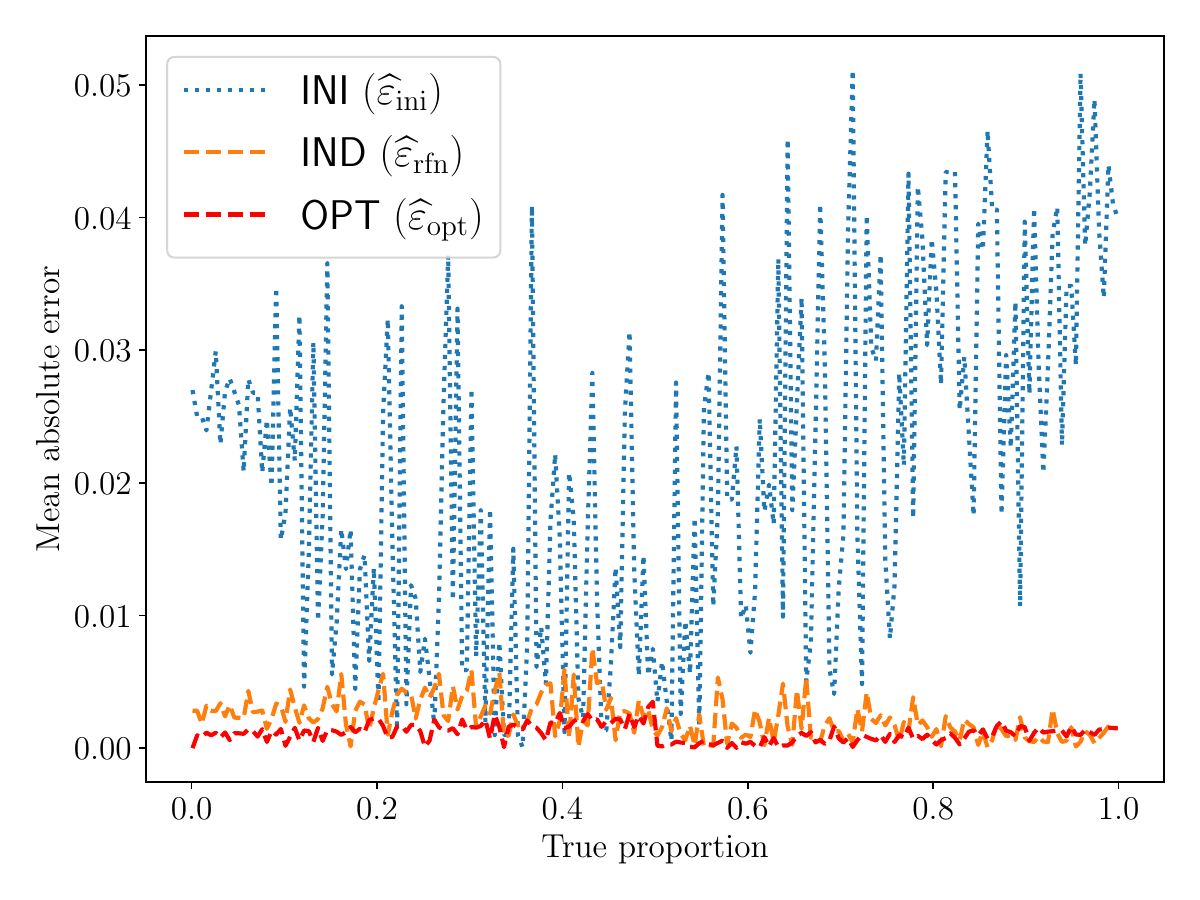}
\caption{MAEs for Gumbel-max}
\label{fig:simulation-MAE-gumbel}
\end{subfigure}
\begin{subfigure}{0.32\textwidth}
\includegraphics[width=\textwidth]{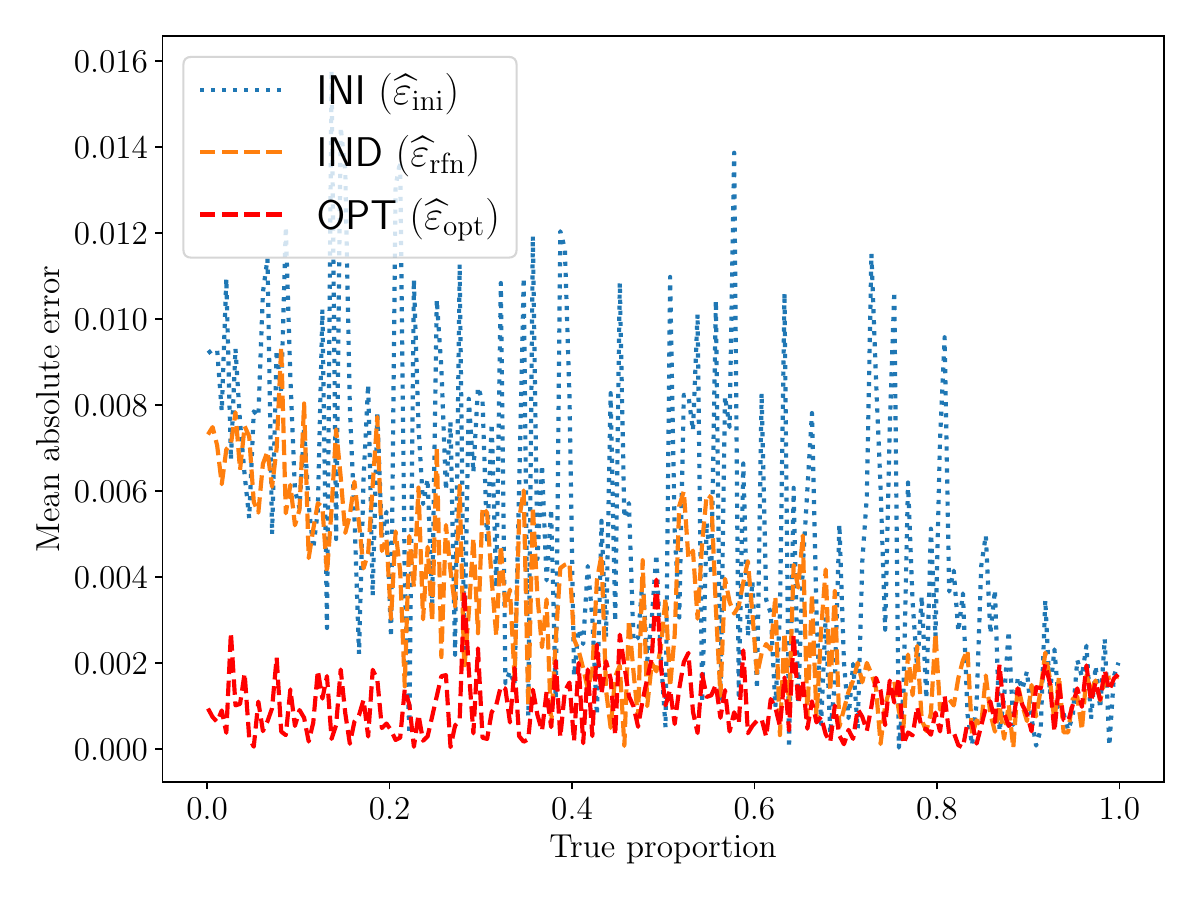}
\caption{MAEs for inverse transform}
\label{fig:simulation-MAE-inverse}
\end{subfigure} 
\begin{subfigure}{0.32\textwidth}
\includegraphics[width=\textwidth]{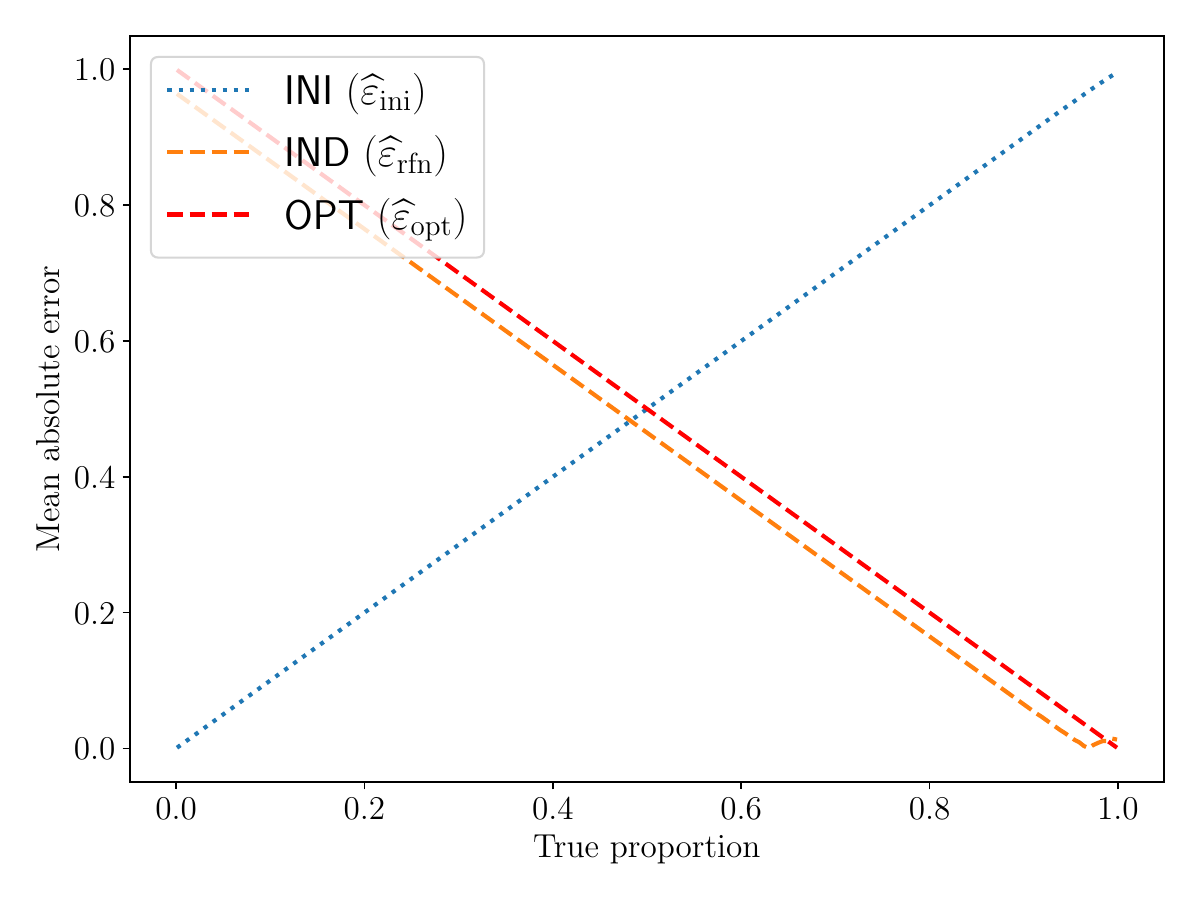}
\caption{MAEs for green-red list}
\label{fig:simulation-MAE-greenredlist}
\end{subfigure}
\begin{subfigure}{0.32\textwidth}
\includegraphics[width=\textwidth]{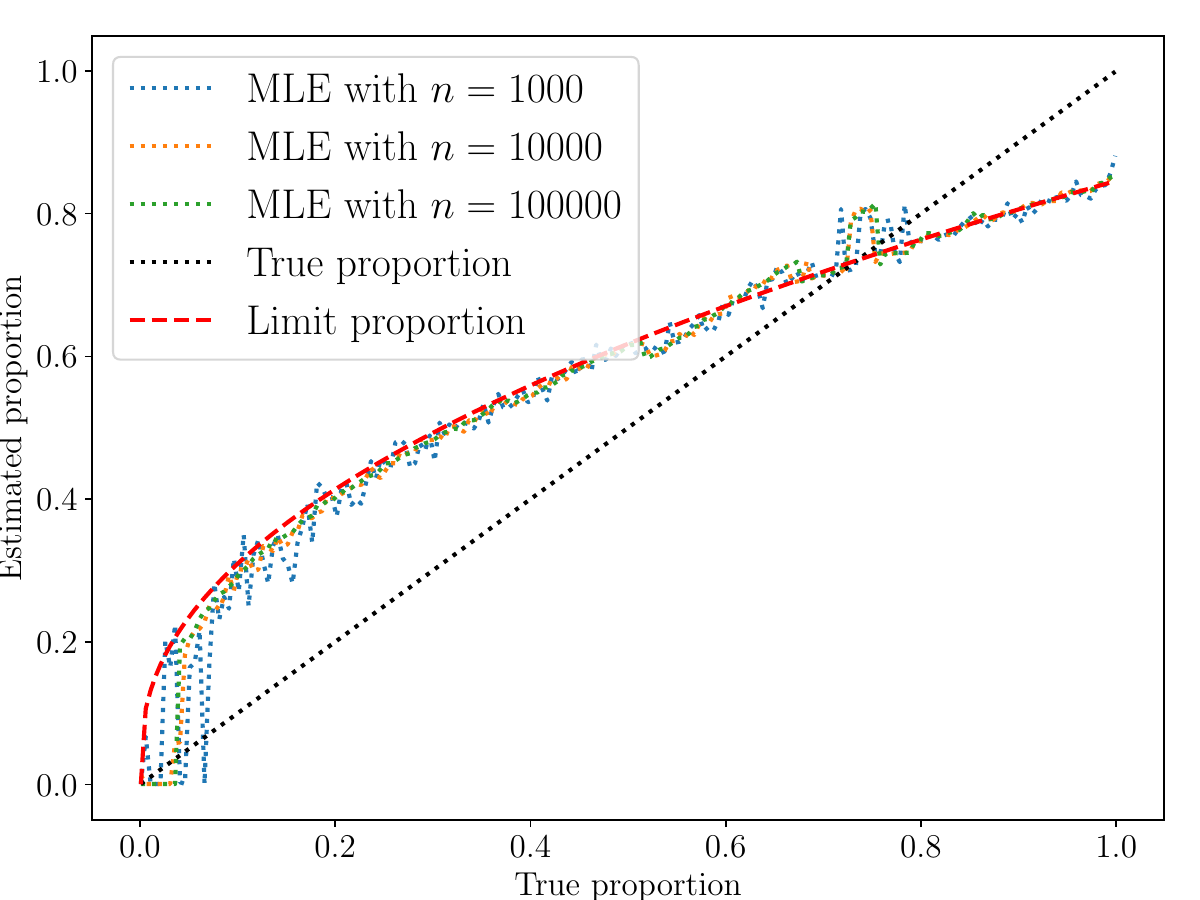}
\caption{Inherent bias (green-red list)}
\label{fig:simulation-bias}
\end{subfigure}
\end{figure*}

\paragraph{CDFs of mixture distributions.}
Figure \ref{fig:main-CDF} shows the CDFs of mixture sample distributions for the Gumbel-max watermark when $\Delta = 0.1$. As $\eps$ increases, the CDFs diverge further from the uniform distribution, which corresponds to $\eps = 0$ (pure null samples). When $\eps = 1$, the mixture consists entirely of watermarked samples, resulting in the greatest deviation from uniformity. 
Additional results for other values of $\Delta$ and the inverse transform watermark are provided in Figures \ref{fig:gumbel_cdf} and \ref{fig:inverse_cdf} in the appendix.

\paragraph{Estimation accuracy.} 
We now examine the estimation accuracy, measured by the mean absolute errors (MAEs) on the $n = 10^6$ mixed-source samples. Three estimators are compared: (i) \textsf{INI}, the initial estimator $\epsinital(\delta)$ tuned with the best $\delta$ from $\{10^{-1}, 10^{-2}, 10^{-3}\}$, (ii) \textsf{IND}, the refined estimator $\epsrefine(\delta)$ with $\delta$ similarly tuned, and (iii) \textsf{OPT}, the refined estimator $\epsopt$ employing the optimal weight function $\hvopt$ defined in \eqref{eq:optimal-weight}. 

Figures \ref{fig:simulation-MAE-gumbel} and \ref{fig:simulation-MAE-inverse} visualize the MAEs of these estimators at various ground truth values $\eps$ when $\Delta = 0.1$.
The results reveal that \textsf{OPT} consistently outperforms both \textsf{INI} and \textsf{IND} across all $\eps$ values, particularly at higher $\eps$, where watermarked samples dominate. This shows the benefits of the optimal weight function $\hvopt$ in improving estimation accuracy. While less accurate than \textsf{OPT}, \textsf{IND} still shows improvements over \textsf{INI}.

\begin{table}[t!]
\vspace{-0.1in}
\caption{
Averaged MAEs calculated over 200 ground truth $\eps$ values which are uniformly distributed on $[0, 1]$. Standard deviations are provided in parentheses, and all values are reported in units of $10^{-4}$. Bold numbers denote the best performance.
}
\label{tab:MAE}
\vspace{-0.2in}
\begin{center}
\begin{small}
\begin{tabular}{c|c|cccccc}
\toprule
Watermarks  & Methods  &$\Delta= 0.1$ & $\Delta= 0.2$ & $\Delta= 0.3$ & $\Delta= 0.4$ & $\Delta= 0.5$ & $\Delta= 0.6$ \\
\midrule
\multirow{3}{*}{Gumbel-max} 
& \textsf{INI} & 210(132) & 140(99) & 59(40) & 95(60) & 33(23) & 26(13) \\
& \textsf{IND} & 23(15) & 22(14) & 31(17) & 13(9) & 23(13) & 20(13) \\
& \textsf{OPT} & \textbf{11(7)} & \textbf{13(8)} & \textbf{14(7)} & \textbf{9(5)} & \textbf{11(7)} & \textbf{13(7)} \\
\midrule
\multirow{3}{*}{Inverse}
& \textsf{INI} & 51(36) & 47(35) & 16(11) & 16(10) & 13(10) & 33(18) \\
& \textsf{IND} & 32(22) & 11(8) & \textbf{14(10)} & 15(10) & 13(10) & 33(18) \\
& \textsf{OPT} & \textbf{10(7)} & \textbf{10(7)} & 16(\textbf{10}) & \textbf{14(8)} & \textbf{11(7)} & \textbf{13(8)} \\
\bottomrule
\end{tabular}
\end{small}
\end{center}
\end{table}

As further evidence, Table \ref{tab:MAE} presents the averaged MAEs across different $\Delta$ values. For both watermark types, the \textsf{OPT} estimator consistently achieves the lowest average MAEs with the smallest standard deviations, as highlighted by the bold numbers. This shows not only the accuracy of \textsf{OPT} but also its stability across varying conditions. 
For example, in the Gumbel-max watermark setting with $\Delta = 0.1$, \textsf{OPT} records an MAE of $11 \times 10^{-4}$, a significant improvement over \textsf{INI} ($210 \times 10^{-4}$) and \textsf{IND} ($23 \times 10^{-4}$). 
This performance pattern persists across other $\Delta$ values, showing the consistent superiority of \textsf{OPT}.
The improvement of \textsf{OPT} over \textsf{IND} further implies the critical role of the optimal weight function in enhancing estimation accuracy. This is expected, as the weight function is derived by minimizing the variance of the estimator (see Lemma \ref{lem:optimal-weight}), thus inherently providing robustness over diverse settings.


\paragraph{Non-identifiability.}
Our estimators are expected to fail on the green-red list watermark because $\eps$ is not identifiable in this case, given that NTP distributions $\bP_{1:n}$ are inaccessible. 
A further issue is that binary $Y^{\mathrm{wm}}_{1:n} \equiv 1$, which results in $\widehat{F}_{\bP} \equiv 1$ and provides no information about $\bar{F}_{\bP}$, leading to abnormal estimation behavior. Figure \ref{fig:simulation-MAE-greenredlist} confirms this theoretical prediction: \textsf{INI} often gives zero estimates while \textsf{IND} and \textsf{OPT} give estimates near one.

\paragraph{Inherent bias.}
We then empirically validate the inherent bias predicted by Theorem \ref{thm:failure-inherent-bias} by generating $n$ i.i.d. samples from $(1 - \eps) \text{Ber}(\gamma) + \eps \text{Ber}(\mu)$, computing the $L_2$-regularized MLE via \eqref{eq:regularized-solutions}, and plotting $\widehat{\eps}_{\lambda}$ against $\eps$. Figure \ref{fig:simulation-bias} shows results for $\lambda = 10^{-2}$ and $(\gamma, \mu) = (0.3, 0.9)$, where $\widehat{\eps}_{\lambda}$ consistently deviates from true $\eps$, confirming the predicted bias. The red dashed line representing the theoretical limit from \eqref{eq:limit-solutions} closely matches empirical estimates. Similar patterns across different parameter settings (Figure \ref{fig:inherent-bias} in the appendix) show that non-identifiability renders standard estimation methods ineffective.

\subsection{Open-source Model Experiments}
We evaluate watermark proportion estimation using open-source LLMs under a controlled setup. We begin by sampling 500 documents from two corpora: the news-like C4 dataset \citep{raffel2020exploring} and the scientific arXiv dataset \citep{cohan2018discourse}. For each document, we extract the last 50 tokens as a prompt and generate 500 continuation tokens using one of three models: OPT-1.3B, OPT-13B \citep{zhang2022opt}, or LLaMA3.1-8B \citep{dubey2024llama}.
To introduce mixture data, at each step $t$, the LLM generates $\token_t$ using a specified watermarking scheme with probability $\eps$; with the remaining probability, it samples $\token_t$ directly from $\bP_t$ without watermarking.
We further apply 1-sequence repeated context masking \citep{dathathri2024scalable}, which watermarks $\token_t$ only if the current context window $\token_{(t-m):(t-1)}$ is unique in the generation history. This technique improves text quality but lowers the realized watermark proportion below the nominal $\eps$, yielding a more realistic mix of watermarked and unwatermarked tokens.
All experiments are conducted under two temperature settings $\mathrm{T} \in \{0.7, 1\}$.
We adopt \textsf{WPL} \citep{zhao2024efficiently} as our baseline since no existing method directly estimates watermark proportions.\footnote{We exclude another token-level localization method, \textsf{SeedBS} \citep{li2024segmenting}, due to its poor accuracy and prohibitive runtime.}
\textsf{WPL} is a token-level localization method that assigns a binary label to each token indicating whether it is watermarked, and the overall proportion can be estimated by averaging these labels.

\begin{figure*}[t!]
\vspace{-0.1in}
\caption{
MAEs of estimators across various true proportions $\eps$ and temperature parameters on the OPT-1.3B model and C4 dataset. Additional results for other models and datasets are provided in Appendix \ref{appen:LLM-experiment}.
}
\label{fig:1.3B-LLM}
\centering
\begin{subfigure}{0.4\textwidth}
\includegraphics[width=\textwidth]{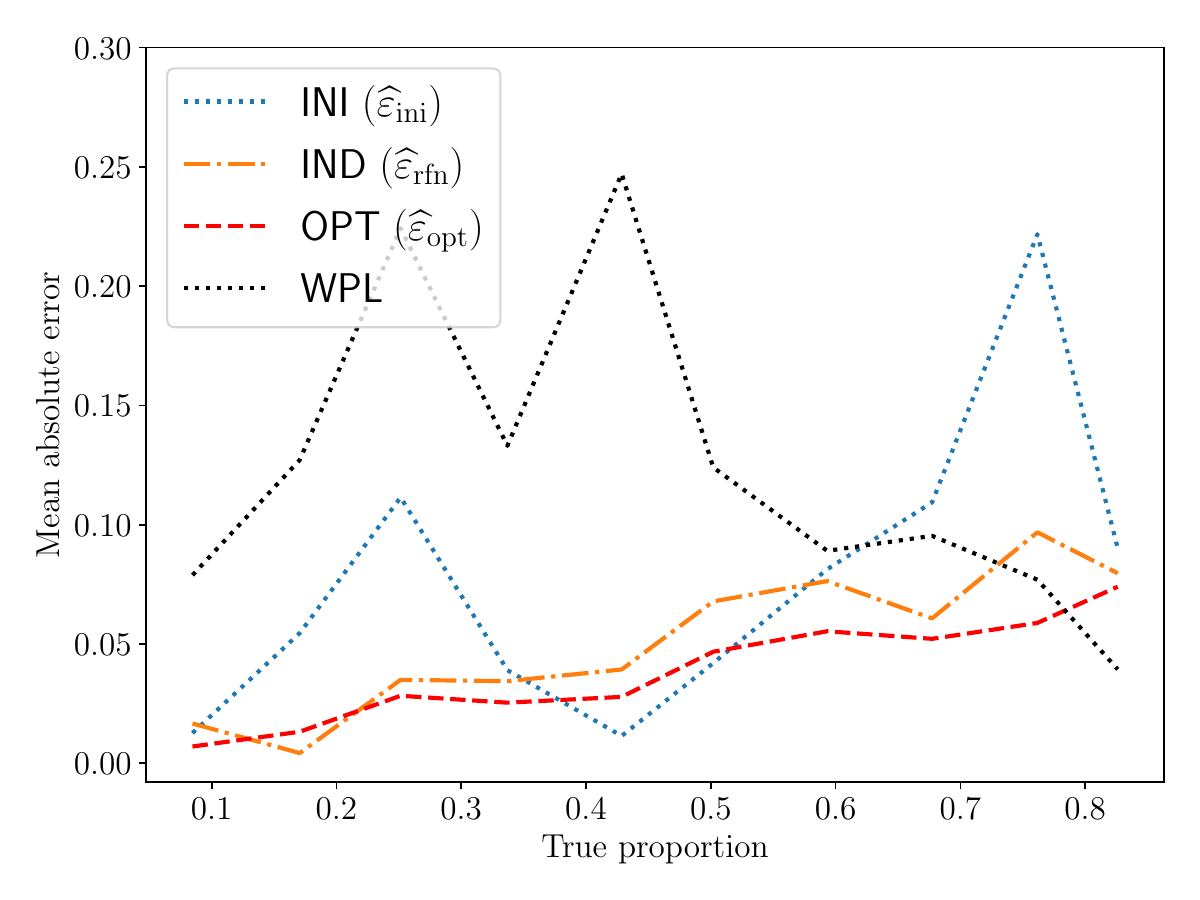}
\caption{Gumbel-max with $\mathrm{T}=0.7$}
\label{fig:1.3B-gumbel-0.7}
\vspace{-0.1in}
\end{subfigure} 
\begin{subfigure}{0.4\textwidth}
\includegraphics[width=\textwidth]{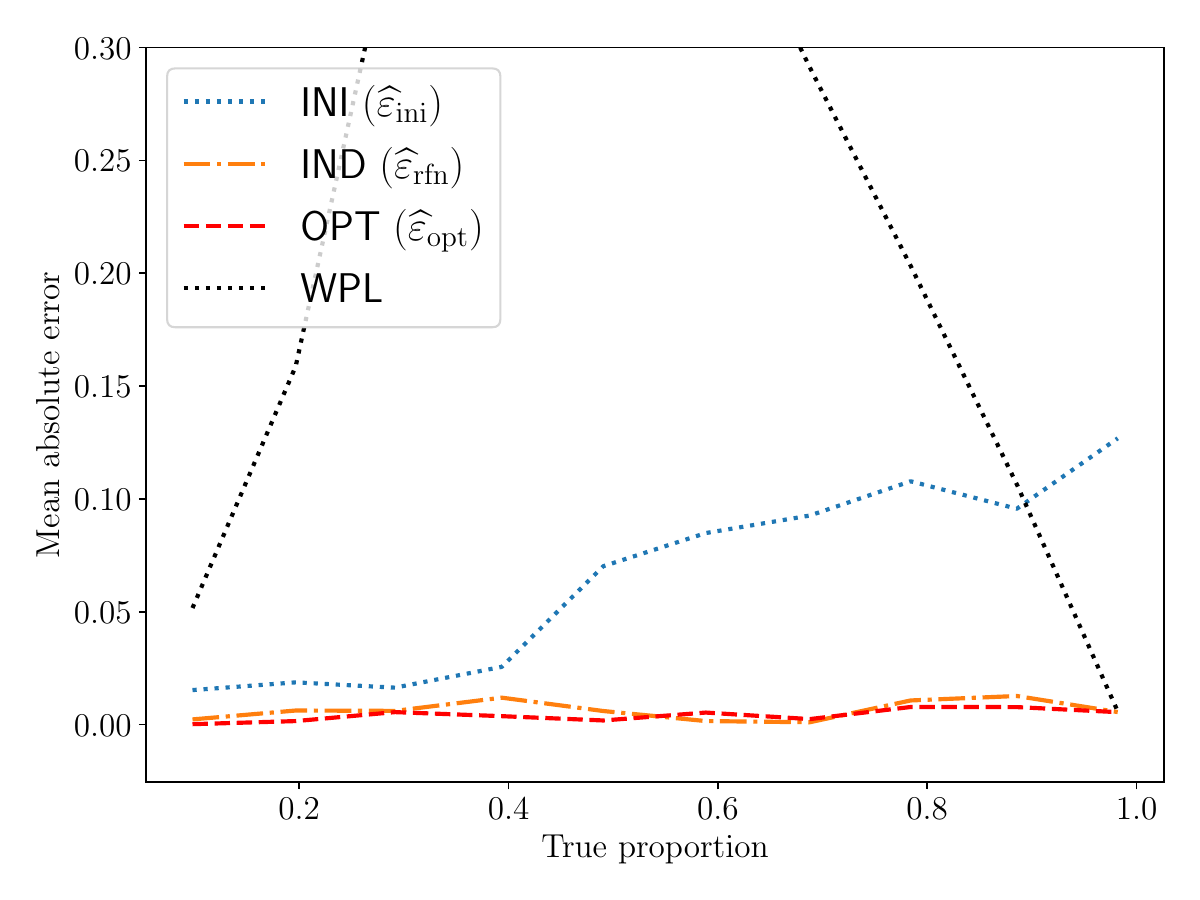}
\caption{Gumbel-max with $\mathrm{T}=1$}
\label{fig:1.3B-gumbel-1}
\vspace{-0.1in}
\end{subfigure} 
 
\begin{subfigure}{0.4\textwidth}
\includegraphics[width=\textwidth]{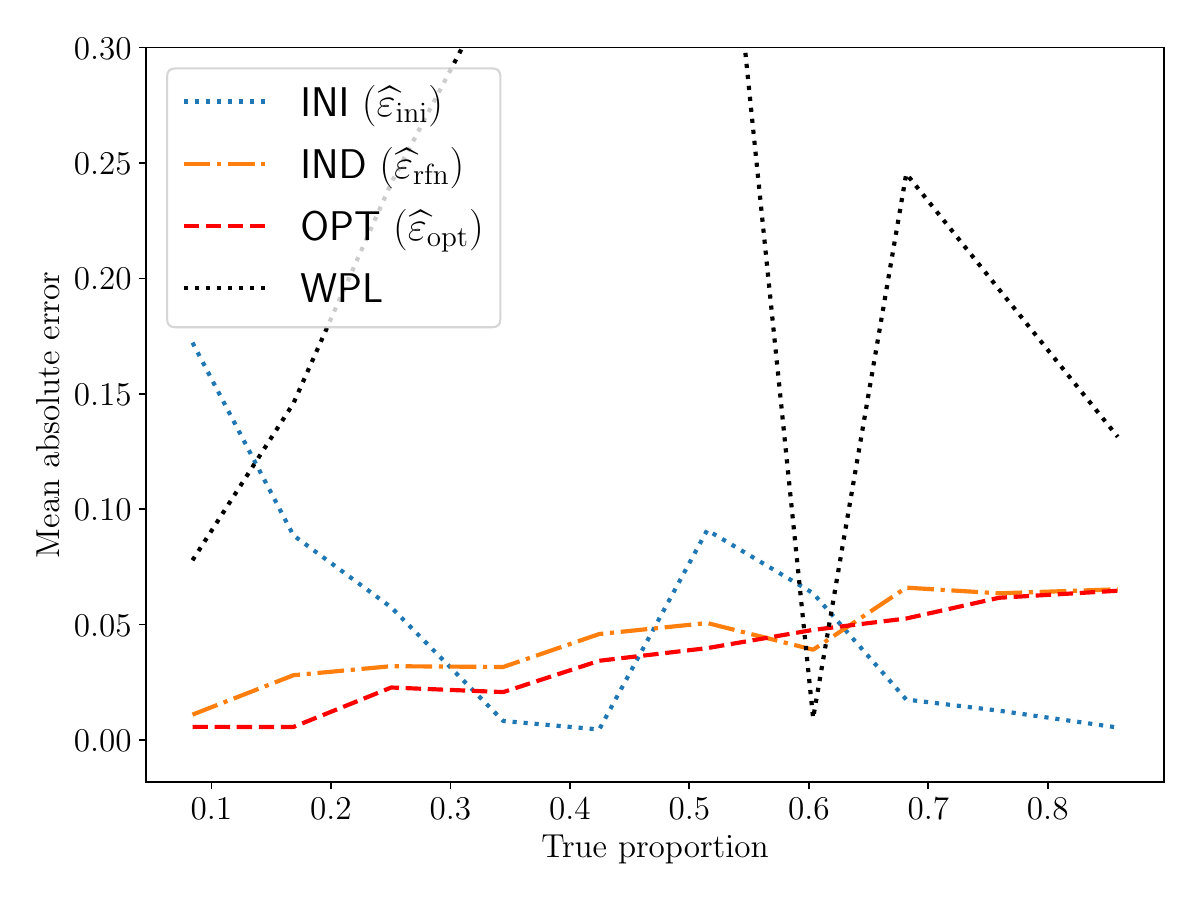}
\caption{Inverse with $\mathrm{T}=0.7$}
\label{fig:1.3B-inverse-0.7}
\vspace{-0.1in} 
\end{subfigure}
\begin{subfigure}{0.4\textwidth}
\includegraphics[width=\textwidth]{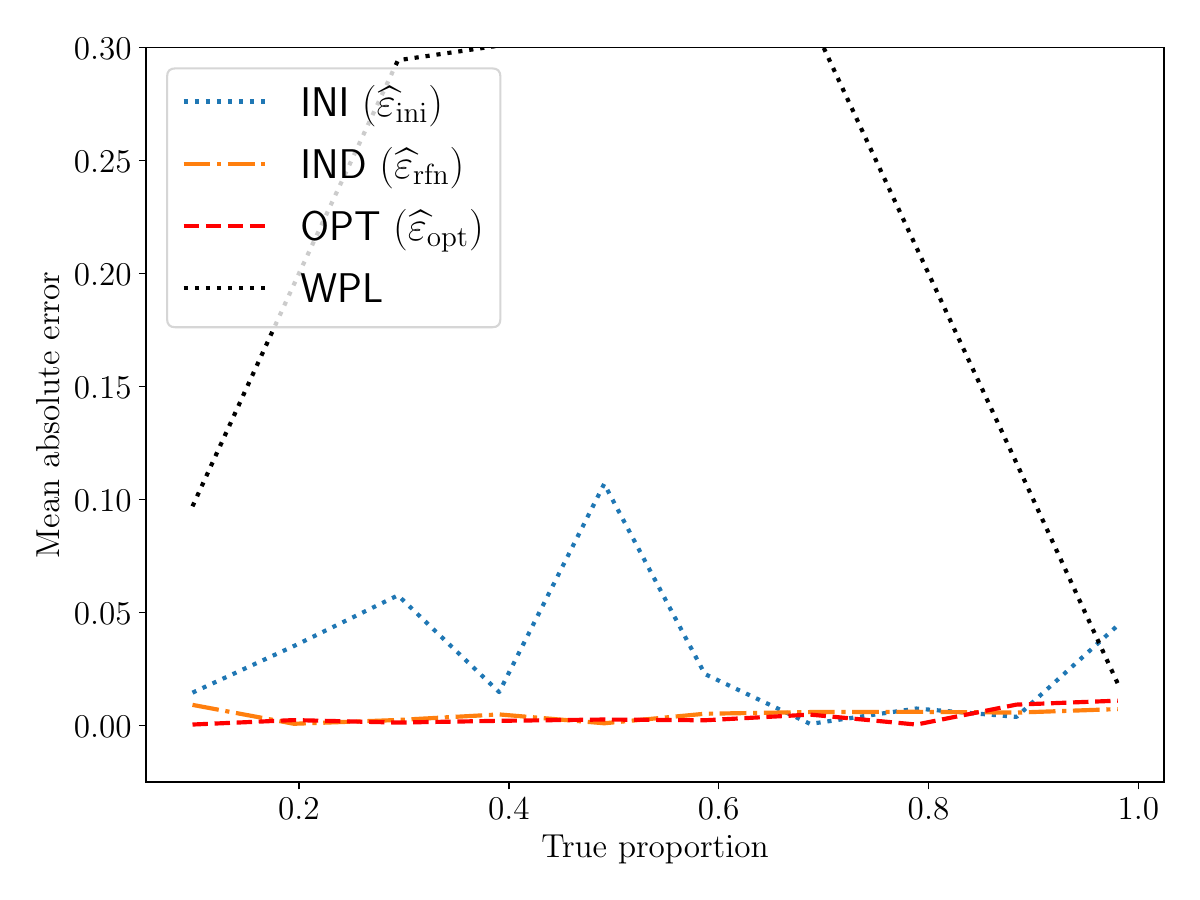}
\caption{Inverse with $\mathrm{T}=1$}
\label{fig:1.3B-inverse-1}
\vspace{-0.1in}
\end{subfigure}
\end{figure*}

\paragraph{Estimation accuracy.}
MAE results for the 1.3B-OPT model and C4 dataset are shown in Figure \ref{fig:1.3B-LLM}, while results for other LLMs and datasets, with similar performance patterns, are provided in the appendix. Table \ref{tab:LLM-MAE} summarizes the mean and standard deviation of MAEs, offering an overview of estimation performance across models, temperatures, and watermarks. In nearly all cases, the optimal estimator \textsf{OPT} achieves the smallest error with the least variation, demonstrating strong empirical performance. In contrast, \textsf{WPL} performs much worse, with MAEs higher than any of our estimators. This result is expected, as token-level localization is inherently more challenging than estimating a global proportion, making \textsf{WPL} suboptimal for our task.

\begin{table}[t!]
\vspace{-0.1in}
\caption{
Averaged MAEs calculated over 10 ground truth $\eps$ values on open-sources model experiments. Standard deviations are provided in parentheses, and all values are reported in units of $10^{-3}$. Bold numbers denote the best performance.
}
\label{tab:LLM-MAE}
\begin{center}
\vspace{-0.3in}
\begin{small}
\resizebox{\textwidth}{!}{%
\begin{tabular}{ccc|c|ccc|c|ccc}
\toprule
\multirow{2}{*}{Models} & \multirow{2}{*}{Datasets} & \multirow{2}{*}{$\mathrm{T}$} & \multicolumn{4}{c|}{Gumbel-max} & \multicolumn{4}{c}{Inverse transform} \\
\cmidrule{4-11}
& & & 
\textsf{WPL} & \textsf{INI} & \textsf{IND} & \textsf{OPT} & 
\textsf{WPL} & \textsf{INI} & \textsf{IND} & \textsf{OPT} \\
\midrule
\multirow{3}{*}{OPT-1.3B} 
  & C4 &0.7 
  & 123(62) &  77(59) & 51(28) &\textbf{39(20)} 
  & 214(125) & 52(52) & 43(\textbf{17}) &\textbf{35}(21) \\
 & C4 & 1 
 & 254(167)   &  65(40) & 6(4) &\textbf{4(3)} 
 & 247(142) &  31(31) & 5(\textbf{3}) &\textbf{4(3)} \\
 & Arxiv & 1 
 &  275(184)   &  70(105) & 19(8) &\textbf{11(6)} 
 &  286(174) &  17(14) & 18(\textbf{8}) &\textbf{12(8)} \\
\midrule
\multirow{3}{*}{OPT-13B} 
& C4 &0.7 
&  119(90)  &   122(94) & 34(19) &\textbf{26(15)} 
& 212(135)  & 49(28) & 25(\textbf{12}) &\textbf{20(12)} \\
&C4& 1 
& 195(156) & 56(40) &8(5) &\textbf{5({3})}
& 250(143) &  94(43) & 70(48) &\textbf{68(40)} \\
& Arxiv& 1 
& 253(162) &  51(27) & 26(\textbf{9}) &\textbf{17}(10)
& 262(140) &  27(15) & 21(12) &\textbf{16(8)} \\
\midrule
\multirow{3}{*}{LLaMA-8B}&C4& 0.7 
&  82(71) & 60(34) & 90(42) &\textbf{75(37)}
& 160(113)&  60(34) & 90(42) &\textbf{75(37)} \\
&C4& 1 
&  263(178) & 45(22) & 6(3) &\textbf{4(2)} 
& 148(77)  &  30(32) & \textbf{2(1)} &{5(4)} \\
&Arxiv& 1
&  236(201)  &  44(18) & 18(9) &\textbf{15(7)} 
& 291(176) &  32(32) & 19(11) &\textbf{16(6)} \\
\bottomrule
\end{tabular}}
\end{small}
\end{center}
\vspace{-0.2in}
\end{table}

\paragraph{Performance after modification.}
We then evaluate the performance of proportion estimators under post-processing modifications. Starting from a fully watermarked dataset, we modify a given proportion of tokens using random substitution, deletion, or insertion, and estimate the resulting watermark proportion. The true proportion is defined as the fraction of pivotal statistics following the alternative distribution. Specifically, if a token $w_t$ or any of its $m$ preceding tokens is modified, we treat $Y_t$ as null. Thus, modifying an $\eps$-fraction of tokens may affect up to an $m\eps$-fraction of pivotal statistics, resulting in a watermark proportion of at least $1 - m\eps$. Results on the C4 dataset are shown in Table~\ref{tab:LLM-MAE-robust}, with similar findings on the arXiv dataset in Table~\ref{tab:LLM-MAE-arxiv} in the appendix.



We have two observations.
First, our estimators remain robust under post-processing modifications, with \textsf{OPT} achieving the lowest MAE and variance in most cases, though \textsf{INI} or \textsf{IND} may occasionally perform better.
Second, all estimators perform slightly better than in the mixture setting (Table~\ref{tab:LLM-MAE}). This improvement arises because, in the post-processing setup, only a subset of tokens is modified, while the rest remain fully watermarked. This results in longer uninterrupted spans of watermarked text, making the underlying signal easier to detect than in the mixture setting, where watermarked and non-watermarked tokens are interleaved throughout.
This improvement also underscores the broad applicability of our formulation in Section~\ref{sec:estimand}.

\begin{table}[t!]
\vspace{-0.1in}
\caption{
Average MAEs under common modifications, computed over 11 ground-truth $\eps$ values on the C4 dataset with temperature 1. Standard deviations are shown in parentheses, and all values are scaled by $10^{-3}$. Boldface denotes the best performance.
}
\label{tab:LLM-MAE-robust}
\begin{center}
\vspace{-0.1in}
\begin{small}
\resizebox{\textwidth}{!}{%
\begin{tabular}{cc|c|ccc|c|ccc}
\toprule
\multirow{2}{*}{Models} & \multirow{2}{*}{Edit types} & \multicolumn{4}{c|}{Gumbel-max} & \multicolumn{4}{c}{Inverse transform} \\
\cmidrule{3-10}
& & 
\textsf{WPL} & \textsf{INI} & \textsf{IND} & \textsf{OPT} & 
\textsf{WPL} & \textsf{INI} & \textsf{IND} & \textsf{OPT} \\
\midrule
\multirow{3}{*}{OPT-1.3B} 
  & Substitution
  & 103(63) & 56(24) & 3(2) &\textbf{1(1)}
  & 275(109) & 46(35) & 5(3) &\textbf{2(1)} \\
 & Insertion 
 & 105(66) & 70(35) & \textbf{8(5)} &\textbf{8(5)} 
 & 282(84) & 38(23) & 9(8) &\textbf{9(5)}\\
 & Deletion 
 & 170(88) & \textbf{38}(27) & 71(19) &66\textbf{(18)} 
 &   268(108) & 35(27) & 68(18) &\textbf{65(19)} \\
\midrule
\multirow{3}{*}{OPT-13B} 
& Substitution 
&  64(33) & 78(55) & 5(5) &\textbf{1(1)}
&  262(85) & 52(28) & 4(2) &\textbf{1(1)} \\
&Insertion
&  60(33) & 61(32) & 12\textbf{(4)} &8\textbf{(5)} 
& 244(95) & 50(40) & \textbf{8}(7) &\textbf{8(5)} \\
& Deletion
& 100(56) & \textbf{49}(38) & 63\textbf{(15)} & {66(18)} 
& 259(89) & \text{26}(23) & 72\textbf{(17)} &68\textbf{(17)} \\
\midrule
\multirow{3}{*}{LLaMA-8B}
&Substitution
&  127(70) & 54(28) & 6(5) &\textbf{1(1)} 
& 236(127) & 50(37) & 3(4) &\textbf{2(1)} \\
&Insertion
&  126(73) & 46(22) & 7(4) &\textbf{2(1)} 
& 243(121) & 18(15) & 8(6) &\textbf{3(2)} \\
&Deletion
& 201(60) & \textbf{34}(22) & 56(24) &38\textbf{(20)} 
& 270(109) & 54(47) & 54(21) &\textbf{36(19)} \\
\bottomrule
\end{tabular}}
\end{small}
\end{center}
\vspace{-0.2in}
\end{table}


\section{Discussion}
In this paper, we address the problem of estimating the proportion of watermarked pivotal statistics in mixed-source texts. We formulate this task as a coefficient estimation problem in a time-varying mixture model and establish conditions under which the proportion is identifiable. For the green-red list watermark, the proportion is not identifiable due to the binary nature of its pivotal statistic, which provides limited information. As a result, common estimation methods, such as maximum likelihood estimation, suffer from inherent and unavoidable bias. In contrast, for watermarks with continuous pivotal statistics, such as the Gumbel-max and inverse transform watermarks, we demonstrate that the proportion is identifiable. We propose several efficient estimators and show that they achieve the minimax optimal lower bounds within their respective classes. Our methods exhibit strong performance on both synthetic datasets and LLM-generated data.

Several open directions remain for future research. This work measures the extent of AI-generated content through pivotal statistics but does not account for its importance within mixed-source texts. For instance, 10\% AI-generated content might carry critical significance in some contexts but be less relevant in others. Incorporating additional information, such as token semantics or entropies, could enable more nuanced proportion estimation. Exploring how to integrate these features is a promising avenue for future study.
\new{Our estimator is order-agnostic and relies only on the empirical distribution of pivotal statistics. This ensures robustness under limited verifier information but does not exploit potential sequential structure. Developing order-sensitive methods that leverage temporal dependence remains an important direction. Moreover, our work operates at the token level and models watermark integrity through a marginal mixture formulation of pivotal statistics. This abstraction reflects the verifier’s limited information, as the specific editing pattern is unobserved. Developing models that explicitly incorporate structured editing patterns or semantic-level dependencies remains an open direction.}
Another direction is to investigate the theoretical limits of identifying AI-generated segments, including possible phase transitions, similar to classic work \citep{donoho2004higher,tony2017optimal}. Developing efficient and adaptive methods for segment identification could further advance the field. 
\new{Finally, our theoretical guarantees require the density ratio estimation error $\Delta_g$ to be sufficiently small. In practice, accurately estimating $g$ can be challenging, especially under distributional shifts. Developing more robust procedures that relax this requirement is an important and practically relevant open question.}


\section*{Acknowledgments}
We would like to thank two anonymous referees for their constructive comments that helped improve the presentation of the paper. This work was supported in part by NIH grants, RF1AG063481 and U01CA274576, NSF DMS-2310679, a Meta Faculty Research Award, and Wharton AI for Business. The content is solely the responsibility of the authors and does not necessarily represent the official views of the NIH.

\bibliographystyle{abbrvnat}
\bibliography{bib/chatgpt,bib/privacy,bib/stat}

@inproceedings{brown2020language,
title={Language models are few-shot learners},
author={Brown, Tom and Mann, Benjamin and Ryder, Nick and Subbiah, Melanie and Kaplan, Jared D and Dhariwal, Prafulla and Neelakantan, Arvind and Shyam, Pranav and Sastry, Girish and Askell, Amanda and others},
booktitle={Advances in neural information processing systems},
volume={33},
pages={1877--1901},
year={2020}
}

@article{das2024under,
title={Under the Surface: {T}racking the Artifactuality of {LLM}-Generated Data},
author={Das, Debarati and De Langis, Karin and Martin, Anna and Kim, Jaehyung and Lee, Minhwa and Kim, Zae Myung and Hayati, Shirley and Owan, Risako and Hu, Bin and Parkar, Ritik and others},
journal={arXiv preprint arXiv:2401.14698},
year={2024}
}

@article{milano2023large,
title={Large language models challenge the future of higher education},
author={Milano, Silvia and McGrane, Joshua A and Leonelli, Sabina},
journal={Nature Machine Intelligence},
volume={5},
number={4},
pages={333--334},
year={2023},
publisher={Nature Publishing Group UK London}
}

@article{starbird2019disinformation,
title={Disinformation's spread: {B}ots, trolls and all of us},
author={Starbird, Kate},
journal={Nature},
volume={571},
number={7766},
pages={449--450},
year={2019},
publisher={Nature Publishing Group}
}

@article{radford2019language,
title={Language models are unsupervised multitask learners},
author={Radford, Alec and Wu, Jeffrey and Child, Rewon and Luan, David and Amodei, Dario and Sutskever, Ilya and others},
journal={OpenAI blog},
volume={1},
number={8},
pages={9},
year={2019}
}

@article{zhang2022opt,
title={{OPT}: {O}pen pre-trained transformer language models},
author={Zhang, Susan and Roller, Stephen and Goyal, Naman and Artetxe, Mikel and Chen, Moya and Chen, Shuohui and Dewan, Christopher and Diab, Mona and Li, Xian and Lin, Xi Victoria and others},
journal={arXiv preprint arXiv:2205.01068},
year={2022}
}

@article{
kuditipudi2023robust,
title={Robust Distortion-free Watermarks for Language Models},
author={Rohith Kuditipudi and John Thickstun and Tatsunori Hashimoto and Percy Liang},
journal={Transactions on Machine Learning Research},
issn={2835-8856},
year={2024},
url={https://openreview.net/forum?id=FpaCL1MO2C}
}

@article{raffel2020exploring,
title={Exploring the limits of transfer learning with a unified text-to-text transformer},
author={Raffel, Colin and Shazeer, Noam and Roberts, Adam and Lee, Katherine and Narang, Sharan and Matena, Michael and Zhou, Yanqi and Li, Wei and Liu, Peter J},
journal={Journal of Machine Learning Research},
volume={21},
number={1},
pages={5485--5551},
year={2020},
publisher={JMLRORG}
}

@inproceedings{
kirchenbauer2023reliability,
title={On the Reliability of Watermarks for Large Language Models},
author={John Kirchenbauer and Jonas Geiping and Yuxin Wen and Manli Shu and Khalid Saifullah and Kezhi Kong and Kasun Fernando and Aniruddha Saha and Micah Goldblum and Tom Goldstein},
booktitle={International Conference on Learning Representations},
year={2024},
url={https://openreview.net/forum?id=DEJIDCmWOz}
}

@misc{scott2023watermarking,
title = {Watermarking of Large Language Models},
organization = {Simons Institute for the Theory of Computing},
author = {Aaronson, Scott},
howpublished = {\url{https://simons.berkeley.edu/talks/scott-aaronson-ut-austin-openai-2023-08-17}},
year={2023},
month = {August},
}

@inproceedings{piet2023mark,
  title={{MARKMyWORDS}: {A}nalyzing and Evaluating Language Model Watermarks},
  author={Piet, Julien and Sitawarin, Chawin and Fang, Vivian and Mu, Norman and Wagner, David},
  booktitle={IEEE Conference on Secure and Trustworthy Machine Learning},
  pages={68--91},
  year={2025},
  organization={IEEE}
}

@inproceedings{vaswani2017attention,
title={Attention is all you need},
author={Vaswani, Ashish and Shazeer, Noam and Parmar, Niki and Uszkoreit, Jakob and Jones, Llion and Gomez, Aidan N and Kaiser, {\L}ukasz and Polosukhin, Illia},
booktitle={Advances in neural information processing systems},
volume={30},
year={2017}
}

@inproceedings{zhao2024efficiently,
title={Efficiently Identifying Watermarked Segments in Mixed-Source Texts},
author={Zhao, Xuandong and Liao, Chenwen and Wang, Yu-Xiang and Li, Lei},
booktitle = {Proceedings of the 63rd Annual Meeting of the Association for Computational Linguistics (Volume 1: Long Papers)},
year={2025},
pages = {6304--6316}
}

@inproceedings{li2024segmenting,
  title={Segmenting Watermarked Texts From Language Models},
  author={Li, Xingchi and Li, Guanxun and Zhang, Xianyang},
  booktitle={Neural Information Processing Systems},
year={2024}
}

@inproceedings{zhu2024duwak,
title={{D}uwak: {D}ual Watermarks in Large Language Models},
author={Zhu, Chaoyi and Galjaard, Jeroen and Chen, Pin-Yu and Chen, Lydia Y},
booktitle={Findings of the Association for Computational Linguistics},
year={2024},
page={11416--11436}
}

@article{bommasani2021opportunities,
  title={On the opportunities and risks of foundation models},
  author={Bommasani, Rishi and Hudson, Drew A and Adeli, Ehsan and Altman, Russ and Arora, Simran and von Arx, Sydney and Bernstein, Michael S and Bohg, Jeannette and Bosselut, Antoine and Brunskill, Emma and others},
  journal={arXiv preprint arXiv:2108.07258},
  year={2021}
}

@article{dathathri2024scalable,
  title={Scalable watermarking for identifying large language model outputs},
  author={Dathathri, Sumanth and See, Abigail and Ghaisas, Sumedh and Huang, Po-Sen and McAdam, Rob and Welbl, Johannes and Bachani, Vandana and Kaskasoli, Alex and Stanforth, Robert and Matejovicova, Tatiana and others},
  journal={Nature},
  volume={634},
  number={8035},
  pages={818--823},
  year={2024},
  publisher={Nature Publishing Group UK London}
}

@inproceedings{dugan2023real,
  title={Real or fake text?: {I}nvestigating human ability to detect boundaries between human-written and machine-generated text},
  author={Dugan, Liam and Ippolito, Daphne and Kirubarajan, Arun and Shi, Sherry and Callison-Burch, Chris},
  booktitle={AAAI Conference on Artificial Intelligence},
  volume={37},
  pages={12763--12771},
  year={2023}
}

@inproceedings{xie2024measuring,
  title={Measuring Human Contribution in {AI}-Assisted Content Generation},
  author={Xie, Yueqi and Qi, Tao and Yi, Jingwei and Yang, Xiyuan and Whalen, Ryan and Huang, Junming and Ding, Qian and Xie, Yu and Xie, Xing and Wu, Fangzhao},
  booktitle={Proceedings of the 64th Annual Meeting of the Association for Computational Linguistics (Volume 1: Long Papers)},
  pages={6168--6190},
  year={2026},
  publisher={Association for Computational Linguistics},
  doi={10.18653/v1/2026.acl-long.279},
  url={https://aclanthology.org/2026.acl-long.279/}
}

@inproceedings{zhang2024llm,
  title={{LLM}-as-a-{C}oauthor: {C}an Mixed Human-Written and Machine-Generated Text Be Detected?},
  author={Zhang, Qihui and Gao, Chujie and Chen, Dongping and Huang, Yue and Huang, Yixin and Sun, Zhenyang and Zhang, Shilin and Li, Weiye and Fu, Zhengyan and Wan, Yao and others},
  booktitle={Findings of the Association for Computational Linguistics: NAACL 2024},
  pages={409--436},
  year={2024}
}

@inproceedings{
wouters2024optimizing,
title={Optimizing Watermarks for Large Language Models},
author={Bram Wouters},
booktitle={International Conference on Machine Learning},
year={2024}
}

@inproceedings{wu2021ai,
  title={{AI} creativity and the human-{AI} co-creation model},
  author={Wu, Zhuohao and Ji, Danwen and Yu, Kaiwen and Zeng, Xianxu and Wu, Dingming and Shidujaman, Mohammad},
  booktitle={Human-Computer Interaction International Conference},
  pages={171--190},
  year={2021},
  organization={Springer}
}

@inproceedings{
zhao2024provable,
title={Provable Robust Watermarking for {AI}-Generated Text},
author={Xuandong Zhao and Prabhanjan Vijendra Ananth and Lei Li and Yu-Xiang Wang},
booktitle={International Conference on Learning Representations},
year={2024},
url={https://openreview.net/forum?id=SsmT8aO45L}
}

@inproceedings{
zhao2024permute,
title={Permute-and-{F}lip: {A}n optimally robust and watermarkable decoder for {LLMs}},
author={Xuandong Zhao and Lei Li and Yu-Xiang Wang},
booktitle={The Thirteenth International Conference on Learning Representations},
year={2025},
url={https://openreview.net/forum?id=YyVVicZ32M}
}

@inproceedings{kirchenbauer2023watermark,
title = 	 {A Watermark for Large Language Models},
author =       {Kirchenbauer, John and Geiping, Jonas and Wen, Yuxin and Katz, Jonathan and Miers, Ian and Goldstein, Tom},
booktitle = 	 {International Conference on Machine Learning},
pages = 	 {17061--17084},
year = 	 {2023},
volume = 	 {202}
}

@inproceedings{wu2023dipmark,
title={A Resilient and Accessible Distribution-Preserving Watermark for Large Language Models},
author={Wu, Yihan and Hu, Zhengmian and Guo, Junfeng and Zhang, Hongyang and Huang, Heng},
booktitle={Proceedings of the 41st International Conference on Machine Learning},
series={Proceedings of Machine Learning Research},
volume={235},
pages={53443--53470},
year={2024},
publisher={PMLR},
url={https://proceedings.mlr.press/v235/wu24h.html}
}

@inproceedings{hu2023unbiased,
title={Unbiased Watermark for Large Language Models},
author={Hu, Zhengmian and Chen, Lichang and Wu, Xidong and Wu, Yihan and Zhang, Hongyang and Huang, Heng},
booktitle={International Conference on Learning Representations},	year={2024},
url={https://openreview.net/forum?id=uWVC5FVidc}
}

@misc{openai2023, 
title={{ChatGPT}: {O}ptimizing language models for dialogue}, 
howpublished = {\url{http://web.archive.org/web/20230109000707/https://openai.com/blog/chatgpt/}},
journal={OpenAI}, publisher={OpenAI}, author={OpenAI}, year={2023}, month={Jan}}

@inproceedings{cohan2018discourse,
  title={A Discourse-Aware Attention Model for Abstractive Summarization of Long Documents},
  author={Cohan, Arman and Dernoncourt, Franck and Kim, Doo Soon and Bui, Trung and Kim, Seokhwan and Chang, Walter and Goharian, Nazli},
  booktitle={Conference of the North American Chapter of the Association for Computational Linguistics: Human Language Technologies},
  pages={615--621},
volume={2},
  year={2018}
}

@article{dubey2024llama,
  title={The {L}lama 3 herd of models},
  author={Dubey, Abhimanyu and Jauhri, Abhinav and Pandey, Abhinav and Kadian, Abhishek and Al-Dahle, Ahmad and Letman, Aiesha and Mathur, Akhil and Schelten, Alan and Yang, Amy and Fan, Angela and others},
  journal={arXiv preprint arXiv:2407.21783},
  year={2024}
}

@inproceedings{liang2024monitoring,
title={Monitoring {AI}-modified content at scale: {A} case study on the impact of {C}hat{GPT} on {AI} conference peer reviews},
author={Liang, Weixin and Izzo, Zachary and Zhang, Yaohui and Lepp, Haley and Cao, Hancheng and Zhao, Xuandong and Chen, Lingjiao and Ye, Haotian and Liu, Sheng and Huang, Zhi and others},
booktitle = 	 {International Conference on Machine Learning},
year={2024}
}

@article{tony2017optimal,
	title={Optimal screening and discovery of sparse signals with applications to multistage high throughput studies},
	author={Cai, T Tony and Sun, Wenguang},
	journal={Journal of the Royal Statistical Society Series B: Statistical Methodology},
	volume={79},
	number={1},
	pages={197--223},
	year={2017},
	publisher={Oxford University Press}
}

@article{blischke1962moment,
 author = {W. R. Blischke},
 journal = {The Annals of Mathematical Statistics},
 number = {2},
 pages = {444--454},
 publisher = {Institute of Mathematical Statistics},
 title = {Moment Estimators for the Parameters of a Mixture of Two Binomial Distributions},
 volume = {33},
 year = {1962}
}

@article{li2024statistical,
  title={A statistical framework of watermarks for large language models: {P}ivot, detection efficiency and optimal rules},
  author={Li, Xiang and Ruan, Feng and Wang, Huiyuan and Long, Qi and Su, Weijie J},
  journal={The Annals of Statistics},
  volume={53},
  number={1},
  pages={322--351},
  year={2025},
  publisher={Institute of Mathematical Statistics},
  doi={10.1214/24-AOS2468}
}

@article{li2024optimal,
  title={Robust detection of watermarks for large language models under human edits},
  author={Li, Xiang and Ruan, Feng and Wang, Huiyuan and Long, Qi and Su, Weijie J},
  journal={Journal of the Royal Statistical Society Series B: Statistical Methodology},
  volume={88},
  number={2},
  pages={491--515},
  year={2026},
  publisher={Oxford University Press},
  doi={10.1093/jrsssb/qkaf056}
}

@book{dodge2003oxford,
  title={The {O}xford dictionary of statistical terms},
  author={Dodge, Yadolah},
  year={2003},
  publisher={Oxford University Press, USA}
}

@book{zipf2016human,
  title={Human behavior and the principle of least effort: {A}n introduction to human ecology},
  author={Zipf, George Kingsley},
  year={2016},
  publisher={Ravenio books}
}

@article{virtanen2020scipy,
  title={{SciPy} 1.0: {F}undamental algorithms for scientific computing in {P}ython},
  author={Virtanen, Pauli and Gommers, Ralf and Oliphant, Travis E and Haberland, Matt and Reddy, Tyler and Cournapeau, David and Burovski, Evgeni and Peterson, Pearu and Weckesser, Warren and Bright, Jonathan and others},
  journal={Nature methods},
  volume={17},
  number={3},
  pages={261--272},
  year={2020},
  publisher={Nature Publishing Group}
}

@book{thomas2006elements,
  title={Elements of information theory},
  author={Thomas, MTCAJ and Joy, A Thomas},
  year={2006},
  publisher={Wiley-Interscience}
}

@book{wainwright2019high,
  title={High-dimensional statistics: {A} non-asymptotic viewpoint},
  author={Wainwright, Martin J},
  volume={48},
  year={2019},
  publisher={Cambridge university press}
}

@book{shao2003mathematical,
	title={Mathematical statistics},
	author={Shao, Jun},
	year={2003},
	publisher={Springer Science \& Business Media}
}

@book{gumbel1948statistical,
	title={Statistical theory of extreme values and some practical applications: {A} series of lectures},
	author={Gumbel, Emil Julius},
	volume={33},
	year={1948},
	publisher={US Government Printing Office}
}

@article{donoho2004higher,
	title={HIGHER CRITICISM FOR DETECTING SPARSE HETEROGENEOUS MIXTURES},
	author={Donoho, David and Jin, Jiashun},
	journal={The Annals of Statistics},
	volume={32},
	number={3},
	pages={962--994},
	year={2004},
	publisher={Citeseer}
}

@article{xie2024debiasing,
  title={Debiasing watermarks for large language models via maximal coupling},
  author={Xie, Yangxinyu and Li, Xiang and Mallick, Tanwi and Su, Weijie and Zhang, Ruixun},
  journal={Journal of the American Statistical Association},
  volume={120},
  number={551},
  pages={1424--1436},
  year={2025},
  publisher={Taylor \& Francis},
  doi={10.1080/01621459.2025.2520455}
}

@inproceedings{radford2023robust,
  title={Robust speech recognition via large-scale weak supervision},
  author={Radford, Alec and Kim, Jong Wook and Xu, Tao and Brockman, Greg and McLeavey, Christine and Sutskever, Ilya},
  booktitle={International Conference on Machine Learning},
  pages={28492--28518},
  year={2023},
  organization={PMLR}
}

\appendix
\newpage
\begin{appendix}
\onecolumn

\part*{Appendices} 

\section{Proofs of Theoretical Analysis}
\label{app:proofs}

\paragraph{Notations.}
We begin by introducing several notations that will be used throughout the proof.

Recall that $\widehat{F}_{\bP}$ denotes the empirical CDF of the pivotal statistics obtained from the surrogate model and the associated datasets. Since this quantity can be computed prior to the estimation procedure, we treat it as deterministic in the proof.

For a sequence of generated tokens $w_{1:n}=w_1\ldots w_n$, let $\bP_1, \ldots, \bP_n$ denote their underlying NTP distributions. Due to the autoregressive generation of LLMs, these distributions are random.
Given the corresponding sequence of pivotal statistics $Y_1, \ldots, Y_n$, define
\begin{equation}
\label{eq:bFpr}
\bFpr(x) 
= \frac{1}{n} \sum_{t=1}^n 
\EB\!\left[\1\{Y_t \le x\} \mid \FM_{t-1}\right],
\end{equation}
which represents the average conditional CDF of $Y_t$ given the past filtration $\FM_{t-1}$. Similarly, define
\begin{equation}
\label{eq:bFprP}
\bFprP(x) 
= \frac{1}{n} \sum_{t=1}^n F_{\bP_t}(x),
\end{equation}
the empirical average of the conditional alternative CDFs. Here ``pr'' stands for \emph{predictive}, emphasizing that these quantities are defined through conditional (predictive) distributions given the past filtration.
As a reminder, the NTP distributions $\bP_1, \ldots, \bP_n$ are random and unknown. Consequently, both $\bFpr$ and $\bFprP$ are random functions. We therefore introduce their expectation counterparts
\[
\Bar{F} = \EB[\bFpr],
\qquad
\Bar{F}_{\bP} = \EB[\bFprP],
\]
which are deterministic. In particular,
\[
\Bar{F}(x) 
= \frac{1}{n} \sum_{t=1}^n \PB(Y_t \le x),
\qquad
\Bar{F}_{\bP}(x) 
= \frac{1}{n} \sum_{t=1}^n \EB\!\left[F_{\bP_t}(x)\right].
\]

\subsection{Proof of Results in Section \ref{sec:estimability}}
We provide the proofs of Lemmas \ref{lem:non-identifiable}, \ref{lem:identifiable}, and \ref{lem:identifiable-two-watermarks} in this subsection.

\begin{proof}[Proof of Lemma \ref{lem:non-identifiable}]
Each data point is equivalently i.i.d. from $\mathrm{Ber}((1-\eps) \gamma + \eps \mu)$. Since both $\eps$ and $\mu$ are unknown, there are multiple combinations of $(\eps, \mu)$ that yield the same value for $(1-\eps) \gamma + \eps \mu$, resulting in the same data distribution.
\end{proof}

\begin{proof}[Proof of Lemma~\ref{lem:identifiable}]
By Assumption~\ref{asmp:main}, the following hold:
(i) $F_0$ is known and deterministic,
(ii) $\lim_{x\to0} \Bar F(x)/F_0(x)$ is determined by the marginal (population) distribution of the observed data (because $\Bar F(x) =  \frac{1}{n} \sum_{t=1}^n \PB(Y_t \le x)$ depends only on the marginal law of $Y$), and
(iii) $\lim_{x \to 0}\bFprP(x)/F_0(x) = 0$ almost surely with respect to the randomness of the generated NTP distributions, since such distributions are non-singular with probability one.

From \eqref{eq:eps-equation}, for any $x$,
\[
\Bar F(x) 
= (1-\eps)F_0(x) + \eps \Bar F_{\bP}(x).
\]
Dividing both sides by $F_0(x)$, letting $x \to 0$, and taking expectation over the randomness in $\bP$, we obtain
\[
\lim_{x \to 0} \frac{\Bar F(x)}{F_0(x)}
= (1-\eps) + \eps \lim_{x \to 0} \frac{\bar{F}_{\bP}(x)}{F_0(x)}
= (1-\eps) + \eps \lim_{x \to 0} \frac{\EB[\bFprP](x)}{F_0(x)}.
\]
By Assumption~\ref{asmp:main}(iii), there is a constant $C$ such that $\rd F_{\bP}/\rd F_0\le C$ almost everywhere, uniformly over the NTP distribution $\bP$. Hence, $F_{\bP}(x)\le C F_0(x)$, and $\bFprP(x)/F_0(x)$ is uniformly bounded for any sequence of $\bP_1, \ldots, \bP_n$.
By (iii), the second term on the right-hand side vanishes by the dominated convergence theorem: the integrand is dominated by a constant (since $\bFprP(x)/F_0(x)$ is uniformly bounded in $x$) and satisfies $\bFprP(x)/F_0(x)\to 0$ as $x\to 0$ almost surely from Assumption \ref{asmp:main}.
Therefore,
\[
\lim_{x \to 0} \frac{\Bar F(x)}{F_0(x)}
= 1-\eps,
\]
which implies that
\[
\eps = 1 - \lim_{x \to 0} \frac{\Bar F(x)}{F_0(x)}.
\]
Since the right-hand side is uniquely determined by the marginal distribution of the observed data and $F_0$ is known, $\eps$ is identifiable.
\end{proof}

\begin{proof}[Proof of Lemma \ref{lem:identifiable-two-watermarks}]
It suffices to prove that the null and alternative CDFs of the considered watermarks satisfy Assumption \ref{asmp:main}. We examine each watermark separately.
\\
\textbf{Gumbel-max watermark:}
The CDFs are given by
\begin{align*}
F_0(x) = x,~~\text{and}~~F_{\bP}(x) &= \sum_{\token} P_{\token} x^{1/P_{\token}}.
\end{align*}
Both CDFs are absolutely continuous and vanish at zero.
Since $\bP = (P_1, \ldots, P_{|\Voca|})$ is non-singular, we have $1/P_{\token} > 1$ for all $\token \in \Voca$. Therefore,
\[
F_0'(0) = 1 > F_{\bP}'(0) = 0
~~\implies~~
\lim_{x \to 0} \frac{F_{\bP}(x)}{F_0(x)} = 0.
\]
Finally, we show that $\operatorname*{ess\,sup}_{x} \left| \frac{\rd F_{\bP}(x)}{\rd F_0(x)} \right|$ is uniformly bounded over all $\bP$.
A direct calculation gives
\[
\frac{\rd F_{\bP}(x)}{\rd F_0(x)}
=
\sum_{w} x^{1/P_w - 1}.
\]
Since $x \in [0,1]$ and $P_w \in (0,1]$, we have 
$x^{1/P_w - 1} \le 1$ for each $w$. 
Therefore,
\[
\frac{\rd F_{\bP}(x)}{\rd F_0(x)}
\le
|\Voca|.
\]
\textbf{Inverse transform watermark:}
Let $\pi$ range over all permutations from $\Voca$ to $[|\Voca|]$, and write
\[
w_{\pi,i}=\pi^{-1}(i),\qquad
\eta_i=\frac{i-1}{|\Voca|-1},\qquad
a_{\pi,i}=\sum_{j=1}^i P_{w_{\pi,j}},\qquad a_{\pi,0}=0.
\]
For $x\in[0,1]$, define
\[
A_i(x)=\left\{u\in[0,1]:1-|u-\eta_i|\le x\right\}.
\]
The exact finite-vocabulary characterization in \citet{li2024statistical} gives the null and alternative CDFs, respectively, as
\begin{align*}
F_0(x)
&=\frac{1}{|\Voca|}\sum_{i=1}^{|\Voca|} \operatorname{Leb}\bigl(A_i(x)\bigr),\\
F_{\bP}(x)
&=\frac{1}{|\Voca|!}\sum_{\pi}\sum_{i=1}^{|\Voca|}
\operatorname{Leb}\Bigl((a_{\pi,i-1},a_{\pi,i}]\cap A_i(x)\Bigr),
\end{align*}
where $\operatorname{Leb}$ denotes interval length. Indeed, under the null, the permuted rank of the observed token is uniform on $[|\Voca|]$ and independent of $U$. Under the alternative, conditional on $\pi$, the inverse-transform decoder selects $w_{\pi,i}$ precisely when $U\in(a_{\pi,i-1},a_{\pi,i}]$.

Each term above is a piecewise-linear Lipschitz function of $x$. Hence, $F_0$ and $F_{\bP}$ are absolutely continuous and satisfy $F_0(0)=F_{\bP}(0)=0$, verifying part (i).

We next verify part (ii). Suppose that $\bP$ is non-singular, so $P_{(1)}<1$. If the selected token has rank $i=1$, then $U\le P_{w_{\pi,1}}\le P_{(1)}$ and $Y=1-U\ge1-P_{(1)}$. If $i=|\Voca|$, then $U>1-P_{w_{\pi,|\Voca|}}\ge1-P_{(1)}$ and $Y=U>1-P_{(1)}$. For any interior rank $2\le i\le |\Voca|-1$, the fact that $U\in[0,1]$ gives
\[
Y=1-|U-\eta_i|\ge\min\{\eta_i,1-\eta_i\}\ge\frac{1}{|\Voca|-1}.
\]
Therefore, with
\[
\delta_{\bP}:=\min\left\{1-P_{(1)},\frac{1}{|\Voca|-1}\right\}>0,
\]
the alternative statistic satisfies $Y\ge\delta_{\bP}$ almost surely. If $0<x<\delta_{\bP}$, the event $\{Y\le x\}$ is therefore empty, and hence $F_{\bP}(x)=0$.

It remains to evaluate the null CDF near zero. Under the null, the rank $i$ is uniform on $[|\Voca|]$ and independent of $U$. For the two endpoint ranks,
\[
\PB(Y\le x\mid i=1)=\PB(U\ge1-x)=x,
\qquad
\PB(Y\le x\mid i=|\Voca|)=\PB(U\le x)=x.
\]
If $2\le i\le |\Voca|-1$ and $0<x<1/(|\Voca|-1)$, the preceding interior-rank bound gives $Y\ge1/(|\Voca|-1)>x$, so the corresponding conditional probability is zero. Averaging over the uniformly distributed rank gives $F_0(x)=2x/|\Voca|$. Consequently,
\[
\lim_{x\to0}\frac{F_{\bP}(x)}{F_0(x)}=0,
\]
which proves part (ii).

Finally, let $f_0$ and $f_{\bP}$ denote the almost-everywhere derivatives of the two exact CDFs. Since $\operatorname{Leb}(A_1(x))=\operatorname{Leb}(A_{|\Voca|}(x))=x$, the two endpoint terms imply
\[
f_0(x)\ge\frac{2}{|\Voca|}\qquad\text{for almost every }x\in(0,1).
\]
For each fixed $\pi$ and $i$, the corresponding interval-intersection term in $F_{\bP}$ is $2$-Lipschitz in $x$. Summing over the $|\Voca|$ ranks and averaging over $\pi$ therefore gives
\[
f_{\bP}(x)\le2|\Voca|\qquad\text{for almost every }x\in(0,1).
\]
Thus,
\[
\frac{\rd F_{\bP}}{\rd F_0}(x)
=\frac{f_{\bP}(x)}{f_0(x)}
\le |\Voca|^2
\qquad F_0\text{-almost everywhere},
\]
uniformly over $\bP$, which verifies part (iii). Part (iv) is precisely the maintained non-singularity condition on the NTP distributions. Thus, both watermarks satisfy Assumption~\ref{asmp:main}, and the inverse-transform verification does not require an asymptotic condition on $P_{(2)}$.
\end{proof}

\subsection{Proof of Lemma \ref{lem:optimal-weight}}
\begin{proof}[Proof of Lemma \ref{lem:optimal-weight}]

We aim to maximize the following functional
\[
\LM(v) = \frac{\left( \mathbb{E}_0[v] - \mathbb{E}_{\bar{F}_{\bP}}[v] \right)^2}{\text{Var}_{\bar{F}}(v)},
\]
where \( v(x) \) is the function to optimize and \( \vopt(\eps, x) \) is defined as:
\[
\vopt(\eps, x) = \frac{\rd F_0}{\rd \bar{F}}(x) - \frac{\rd \bar{F}_{\bP}}{\rd \bar{F}}(x).
\]
In the main text, we use the equivalent form of $\vopt(\eps, x)$:
\[
\vopt(\eps, x) = \frac{1-g(x)}{(1-\eps)+\eps g(x)}
~~\text{where}~~
g(x) = \frac{\rd \bar{F}_{\bP}}{\rd F_0}(x).
\]

The first step is to simplify the expression for $\LM(v)$. 
Using the definition of \( \vopt(\eps, x) \), its numerator becomes:
\[
\mathbb{E}_0[v] - \mathbb{E}_{\bar{F}_{\bP}}[v] 
= \int v(x) \left( \rd F_0(x) - \rd \bar{F}_{\bP}(x) \right)
= \int v(x) \vopt(\eps, x) \rd \bar{F}(x).
\]
Next, we analyze the denominator. Assume, without loss of generality, that \( v(x) \) is centered under \( \bar{F} \), i.e., \( \int v(x) \rd \bar{F}(x) = 0 \). This is because adding a constant does not affect the value of $\LM(v)$; that is, $\LM(v) = \LM(v + C)$ for any constant $C$.
Then:
\[
\Var_{\bar{F}}(v) = \int v^2(x) \rd \bar{F}(x) - \left( \int v(x) \rd \bar{F}(x) \right)^2 = \int v^2(x) \rd \bar{F}(x).
\]
Thus, the objective simplifies to:
\[
\LM(v) = \frac{\left( \int v(x) \vopt(\eps, x) \rd \bar{F}(x) \right)^2}{\int v^2(x) \rd \bar{F}(x)}.
\]

By the Cauchy-Schwarz inequality, for two functions \( v(x) \) and \( \vopt(\eps, x) \) under the measure \( \bar{F} \), we have:
\[
\left( \int v(x) \vopt(\eps, x) \rd \bar{F}(x) \right)^2 \leq \left( \int v^2(x) \rd \bar{F}(x) \right) \left( \int \vopt^2(\eps, x) \rd \bar{F}(x) \right).
\]
Equality holds if and only if \( v(x) \) is proportional to \( \vopt(\eps, x) \), i.e. for some constant \( c \),
\[
v(x) = c \cdot \vopt(\eps, x).
\]

Finally, plugging this optimal weight function, we have that
\[
\LM(\vopt)
=\frac{\left( \int \vopt^2(\eps, x) \rd \bar{F}(x) \right)^2}{\int \vopt^2(\eps, x) \rd \bar{F}(x)} 
= \int \frac{(\rd F_0(x) - \rd \bar{F}_{\bP}(x))^2}{\rd \bar{F}(x)}
= \int \frac{(1-g(x))^2}{(1-\eps)+\eps g(x)}\rd F_0(x).
\]

\end{proof}

\subsection{Proof of Inequality (\ref{eq:upper-bound-variance})}
\label{proof:upper-bound-variance}

\begin{proof}[Proof of Inequality \eqref{eq:upper-bound-variance}]
In the following, we prove that
\begin{equation}\tag{\ref{eq:upper-bound-variance}}
\EB\left\langle\frac{\EB_{{F}_0}[v] - \EB_{\widehat{F}}[v]}{\EB_{{F}_0}[v] - \EB_{\Bar{F}_{\bP}}[v]}\right\rangle \le \frac{\Var_{\bar{F}}(v)}{n (\EB_{{F}_0}[v] - \EB_{\Bar{F}_{\bP}}[v])^2}.
\end{equation}
Recall that under Model \ref{model:previous}, $\widehat{F}$ is the empirical CDF of $n$ dependent pivotal statistics $Y_1, \ldots, Y_n$ where each $Y_t|\FM_{t-1} \sim (1-\eps) F_0 + \eps F_{\bP_t}$ with $\bP_{t}, Y_{t-1} \in \FM_{t-1}$.

As a reminder, for the random variable $\EB_{\widehat F}[v]=\frac{1}{n}\sum_{t=1}^n v(Y_t)$, its predictable quadratic variation is $\left\langle \EB_{\widehat F}[v]\right\rangle = \frac{1}{n^2}\sum_{t=1}^n \Var\!\left(v(Y_t)\mid \FM_{t-1}\right).$
Taking expectations and using the law of total variance,
\begin{align*}
\EB\langle \EB_{\widehat{F}}[v] \rangle
&= \frac{1}{n^2} \sum_{t=1}^n \EB [\Var(v(Y_t)|\FM_{t-1})]\\
&= \frac{1}{n^2} \sum_{t=1}^n  \left[  \EB[v^2(Y_t)] - \EB[\EB[v(Y_t)|\FM_{t-1}]^2]
\right]\\
&\overset{(a)}{\le} \frac{1}{n^2} \sum_{t=1}^n \EB[v^2(Y_t)]
- \frac{1}{n}\EB \left( \frac{1}{n} \sum_{t=1}^n \EB[v(Y_t)|\FM_{t-1}] \right)^2\\
&\overset{(b)}{\le} \frac{1}{n^2} \sum_{t=1}^n \EB[v^2(Y_t)]
- \frac{1}{n} \left( \frac{1}{n} \sum_{t=1}^n \EB[v(Y_t)] \right)^2\\
&= \frac{1}{n} \left( \EB_{\Bar{F}}[v^2] - \EB_{\Bar{F}}[v]^2 \right) = \frac{1}{n} \Var_{\Bar F}(v)
\end{align*}
where (a) follows from the inequality $\sum_{t=1}^n a_n^2 \ge \frac{1}{n}(\sum_{t=1}^n a_n)^2$ applied to $a_t=\EB[v(Y_t)\mid\FM_{t-1}]$ and (b) uses Jensen's inequality such that $\EB[X^2] \ge \EB[X]^2$.

Using the last inequality, it then follows that
\begin{align*}
\EB\left\langle\frac{\EB_{{F}_0}[v] - \EB_{\widehat{F}}[v]}{\EB_{{F}_0}[v] - \EB_{\Bar{F}_{\bP}}[v]}\right\rangle 
&= \frac{\EB\langle \EB_{\widehat{F}}[v] \rangle}{ (\EB_{{F}_0}[v] - \EB_{\Bar{F}_{\bP}}[v])^2} 
\le \frac{\Var_{\bar{F}}(v)}{n (\EB_{{F}_0}[v] - \EB_{\Bar{F}_{\bP}}[v])^2}
\end{align*}

Finally, we characterize when the inequality $\EB\langle \EB_{\widehat{F}}[v] \rangle  \le \frac{1}{n} \Var_{\Bar F}(v)$ becomes an equality. This occurs if and only if both steps (a) and (b) hold with equality.
Step (a) becomes an equality if and only if $\EB[v(Y_t)\mid \FM_{t-1}]
$ is the same for all $t$ almost surely. This requires that the conditional distributions of $Y_t$ given $\FM_{t-1}$ coincide, i.e., $\bP_t = \bP_0$ for all $t$.
Step (b) becomes an equality if and only if $\frac{1}{n}\sum_{t=1}^n \EB[v(Y_t)\mid \FM_{t-1}]$ is almost surely constant, which further requires that $\bP_0$ is deterministic.
In summary, equality holds when $Y_1,\ldots,Y_n$ are independent and identically distributed.

\end{proof}

\subsection{Analysis for the Initial Estimator}
\label{proof:inital-estimator}

We first present the proof of Theorem \ref{thm:upper-bound-eps-inital} and then discuss the MAE associated with adaptively tuning the parameter $\delta$.

\begin{proof}[Proof of Theorem \ref{thm:upper-bound-eps-inital}]
Note that $\bFpr = (1-\eps) F_0+ \eps \bFprP$. We have the following decomposition:
\begin{align*}
\epsinital(\delta) = 1- \frac{\widehat{F}(\delta)}{F_0(\delta)} = \frac{\bFpr(\delta)-\widehat{F}(\delta)}{F_0(\delta)}- \eps \frac{\bFpr_{\bP}(\delta)}{F_0(\delta)}+ \eps.
\end{align*}
We denote by $p_t(\eps) :=\EB[\1\{Y_t\le \delta\}|\FM_{t-1}] =(1-\eps) F_0(\delta) + \eps F_{\bP_t}(\delta)$ for simplicity.
Due to the relation \eqref{eq:eps-equation}, we have $\bFpr(\delta) = \frac{1}{n}\sum_{t=1}^n p_t(\eps)$.
It then follows that
\begin{align*}
\EB \left|\epsinital(\delta) - \eps + \eps \frac{\bFprP(\delta)}{F_0(\delta)} \right|
&\overset{(a)}{\le} \sqrt{\EB \left|\epsinital(\delta) - \eps + \eps \frac{\bFprP(\delta)}{F_0(\delta)} \right|^2}\\
&= \sqrt{\EB \left| \frac{\bFpr(\delta)-\widehat{F}(\delta)}{F_0(\delta)}\right|^2}\\
&\overset{(b)}{=}  \frac{1}{F_0(\delta)n} \sqrt{\EB\sum_{t=1}^n p_t(\eps)(1-p_t(\eps))}\\
&\overset{(c)}{\le} \frac{1}{F_0(\delta)\sqrt{n}} \cdot \sqrt{ \left[\frac{1}{n}\sum_{t=1}^n \EB[p_t(\eps)] \right] 
\left(1 -\left[\frac{1}{n}\sum_{t=1}^n \EB[p_t(\eps)] \right] \right) } \\
&\overset{(d)}{=} \frac{\sigma_n}{\sqrt{n}},
\end{align*}
where $(a)$ and $(c)$ follow from Jensen's inequality, $(b)$ uses the Doob’s decomposition for square integrable martingales $\{X_t\}_{t=1}^n$ such that $\EB[\langle X \rangle]=\EB[\sum_{t=1}^n[X_t - \EB[X_t|\FM_{t-1}]]]^2 = \EB \sum_{t=1}^n\EB[X_t - \EB[X_t|\FM_{t-1}]]^2$, and $(d)$ utilizes the definition of $\sigma_n$ that is $\sigma_n^2 = \frac{\bar{F}(\delta)(1-\bar{F}(\delta))}{[F_0(\delta)]^2}.$

Finally, it follows from the triangle inequality that
\begin{align*}
\EB \left|\epsinital(\delta) - \eps  \right|
\le \frac{\sigma_n}{\sqrt{n}} + \eps \EB \left| \frac{\bFprP(\delta)}{F_0(\delta)} \right| =  \frac{\sigma_n}{\sqrt{n}} + \eps  \frac{\bar{F}_{\bP}(\delta)}{F_0(\delta)}.
\end{align*}
\end{proof}

Theorem \ref{thm:upper-bound-eps-inital} analyzes the mean absolute error (MAE) of $\epsinital(\delta)$ and shows it is of the order $\frac{\sigma_n}{\sqrt{n}} + \text{bias}$, where the bias term is given by $\eps \frac{\bar{F}_{\bP}(\delta)}{F_0(\delta)}$.
Theorem \ref{thm:minimax} establishes that the minimax optimal MAE within the class of all measurable functions of the indicator variables $\1\{Y_1 \leq \delta\}, \ldots, \1\{Y_n \leq \delta\}$ is $\frac{\sigma_n^\star}{\sqrt{n}}$. By Theorem \ref{thm:upper-bound-eps-inital}, the variance term $\sigma_n$ for $\epsinital(\delta)$ is given by $\frac{\bar{F}(\delta)(1-\bar{F}(\delta))}{[F_0(\delta)]^2}$. As shown in Proposition \ref{prop:inital-optimal}, this variance is smaller (up to constant factors) than the optimal variance $\sigma_n^\star$. However, the bias term $\eps \frac{\bar{F}_{\bP}(\delta)}{F_0(\delta)}$ generally increases the MAE of $\epsinital(\delta)$, making its overall MAE larger than $\frac{\sigma_n^\star}{\sqrt{n}}$. This is further detailed in Corollary \ref{cor:less-MAE-inital-estimation}.

\begin{prop}
\label{prop:inital-optimal}
Under Assumption \ref{asmp:main}, we have \( c \cdot \sigma_n \leq \sigma_n^\star \) for some \( c > 0 \), where 
\[
\sigma_n^2 = \frac{\bar{F}(\delta)(1-\bar{F}(\delta))}{[F_0(\delta)]^2}.
\]
\end{prop}

\begin{proof}[Proof of Proposition \ref{prop:inital-optimal}]
    We observe that under Assumption \ref{asmp:main}, $\lim_{\delta \to 0} \frac{F_{\bP}(\delta)}{F_0(\delta)}=0$ ensuring that there exists a positive constant $c_1 > 0$ such that 
    $    |F_0(\delta) - F_{\bP}(\delta)| \leq c F_0(\delta)$ for any $\bP$, due to the continuity of $F_{\bP}$ in $\bP$.
    It implies that we have
    \[
    \sigma_n^2 = \frac{\bar{F}(\delta)(1-\bar{F}(\delta))}{[F_0(\delta)]^2}  \le c^2 \frac{\bar{F}(\delta)(1-\bar{F}(\delta))}{[F_0(\delta)-\bar{F}_{\bP}(\delta)]^2} = c^2 \cdot [\sigma_n^\star]^2.
    \]

\end{proof}

\begin{cor}
\label{cor:less-MAE-inital-estimation}
Under the same setting as Theorem \ref{thm:upper-bound-eps-inital}, assume that $\bar{F}_{\bP}(x) \sim x^p$ and $F_0(x) \sim x$ as $x \to 0$, where \( p > 1 \). Then, if $\eps = \Theta(1)$, we have:
\[
\EB\left|\epsinital(\delta_n) - \eps  \right|
\leq \frac{\sigma_n}{\sqrt{n}} + \eps \frac{\bar{F}_{\bP}(\delta_n)}{F_0(\delta_n)} \sim \frac{1}{\sqrt{n \delta_n}} + \eps \delta_n^{p-1}.
\]
By selecting the optimal value $\delta_n = [2(p-1)\eps\sqrt{n}]^{-\frac{1}{p-0.5}}$, it follows that:
\[
\EB\left|\epsinital(\delta_n) - \eps  \right| \precsim \frac{\eps^{\frac{1}{p-0.5}}}{n^{\frac{1}{2}\left(1-\frac{1}{p-0.5}\right)}}.
\]
\end{cor}

This result highlights that the bias term $\eps \frac{\bar{F}_{\bP}(\delta_n)}{F_0(\delta_n)}$ limits the performance of $\epsinital(\delta_n)$. Even with adaptive tuning of $\delta$, its dependence on $n$ is strictly slower than $\frac{1}{\sqrt{n}}$, illustrating why $\epsinital(\delta)$ cannot achieve minimax optimality.

\subsection{Analysis for the Refined Estimator}

\begin{theorem}[Full version of the first part in Theorem \ref{thm:accuracyRfn}]
\label{thm:MAE-eps-refine}
Let $\eps \in [0, 1]$ and $\delta \in (0, 1]$ be fixed. Then,
\begin{equation*}
\EB \left|\epsrefine(\delta) - \eps + \eps \frac{\bar{F}_{\bP}(\delta)-\widehat{F}_{\bP}(\delta)}{F_0(\delta)-\widehat{F}_{\bP}(\delta)}\right| \leq \frac{\widehat{\sigma}_n}{\sqrt{n}},
~~\text{where}~~\widehat{\sigma}_n^2 = \frac{\bar{F}(\delta)(1-\bar{F}(\delta))}{[F_0(\delta)-\widehat{F}_{\bP}(\delta)]^2}.
\end{equation*}
\end{theorem}

\begin{cor}[Simplified MAE]
When $\EB_g\left[\sup_{x} \frac{|\widehat{g}(x) - g(x)|}{\min\{1, \widehat{g}(x)\}} \right] \to 0$, it follows that  $\widehat{\sigma}_n \to \sigma_n^\star$, implying $\sigma_n^\star = \widehat{\sigma}_n + o_g(1)$. Under this condition, the bias term $\eps \frac{\bar{F}_{\bP}(\delta) - \widehat{F}_{\bP}(\delta)}{F_0(\delta) - \widehat{F}_{\bP}(\delta)}$ also becomes $o_g(1)$. Consequently, we have
\[
\EB|\epsrefine(\delta) - \eps| \le \frac{\sigma_n^\star+o_g(1)}{\sqrt{n}}.
\]
\end{cor}

\begin{proof}[Proof of Theorem \ref{thm:MAE-eps-refine}]
Note that $\bFpr = (1-\eps) F_0+ \eps \bFprP$. We have the following decomposition:
\begin{align*}
\epsrefine(\delta) = \frac{F_0(\delta) - \widehat{F}(\delta) }{F_0(\delta) - \widehat{F}_{\bP}(\delta)} = \frac{\bFpr(\delta)-\widehat{F}(\delta)}{F_0(\delta)-\widehat{F}_{\bP}(\delta)}- \eps \frac{\bFprP(\delta)-\widehat{F}_{\bP}(\delta)}{F_0(\delta)-\widehat{F}_{\bP}(\delta)}+ \eps.
\end{align*}
We denote by $p_t(\eps) :=\EB[\1\{Y_t\le \delta\}|\FM_{t-1}] =(1-\eps) F_0(\delta) + \eps F_{\bP_t}(\delta)$ for simplicity.
Due to the relation \eqref{eq:eps-equation}, we have $\bFpr(\delta) = \frac{1}{n}\sum_{t=1}^n p_t(\eps)$.
It then follows that
\begin{align*}
\EB \left|\epsrefine(\delta) - \eps + \eps \frac{\bFpr_{\bP}(\delta)-\widehat{F}_{\bP}(\delta)}{F_0(\delta)-\widehat{F}_{\bP}(\delta)}\right|
&\overset{(a)}{\le} \sqrt{\EB \left|\epsrefine(\delta) - \eps + \eps \frac{\bFpr_{\bP}(\delta)-\widehat{F}_{\bP}(\delta)}{F_0(\delta)-\widehat{F}_{\bP}(\delta)}\right|^2}\\
&= \sqrt{\EB \left| \frac{\bFpr(\delta)-\widehat{F}(\delta)}{F_0(\delta)-\widehat{F}_{\bP}(\delta)}\right|^2}\\
&= \frac{\sqrt{\sum_{t=1}^n \EB[p_t(\eps)(1-p_t(\eps))]}}{n[F_0(\delta)-\widehat{F}_{\bP}(\delta)]} \\
&\overset{(b)}{\le} \frac{\sqrt{ 
 \left[\frac{1}{n}\sum_{t=1}^n \EB[p_t(\eps)] \right] 
\left(1 -\left[\frac{1}{n}\sum_{t=1}^n \EB[p_t(\eps)] \right] \right) }}{\sqrt{n}[F_0(\delta)-\widehat{F}_{\bP}(\delta)]}  \\
&\overset{(c)}{=} \frac{\widehat{\sigma}_n}{\sqrt{n}},
\end{align*}
where $(a)$ and $(b)$ follow from Jensen's inequality, and $(c)$ utilizes the definition of $\widehat{\sigma}_n$  that is $\widehat{\sigma}_n^2 = \frac{\bar{F}(\delta)(1-\bar{F}(\delta))}{[F_0(\delta)-\widehat{F}_{\bP}(\delta)]^2}.$
\end{proof}

\subsection{Analysis for the Optimal Estimator}
\label{proof:accuracyRfn}

\subsubsection{Proof of Lemma \ref{lem:contraction-mapping}}
Recall that we define an operator $\widehat{\TM}: [\eps_{\min}, 1-\eps_{\min}] \to [\eps_{\min}, 1-\eps_{\min}]$ in defining the optimal estimator $\epsopt$:

\begin{equation}
\label{eq:new-emprical-operator}
\widehat{\TM}(\eps) 
=  \PM \circ \widetilde{\TM}(\eps)~~\text{with}~~
\widetilde{\TM}(\eps) := \frac{\int \hvopt(\eps, x) \left[\rd {F}_0(x) - \rd \widehat{F}(x)\right]}{\int \hvopt(\eps, x) \left[\rd {F}_0(x) - \rd \widehat{F}_{\bP}(x)\right]},
\end{equation}
where $\PM(x) := \max\{\min\{x, 1-\eps_{\min}\}, \eps_{\min}\}$ is the projection operator, and $\eps_{\min}>0$ is a small constant to avoid numerical issues.

To establish Lemma \ref{lem:contraction-mapping}, which asserts that \( \widehat{\TM} \) is a contraction mapping with high probability, we first derive a supporting result, stated as Lemma \ref{lem:formal-contraction-mapping}. The proof of Lemma \ref{lem:formal-contraction-mapping} is deferred to Appendix \ref{proof:formal-contraction-mapping}.

\begin{lem}
\label{lem:formal-contraction-mapping}
Define the total variation distance between $F_0$ and $\widehat{F}_{\bP}$ as
$\TV(F_0, \widehat{F}_{\bP}) := \frac{1}{2}\int |1-\widehat{g}(x)| \rd F_0(x)$ and $C_g := \max\{1, \sup\limits_{x \in [0, 1]}|\widehat{g}(x)|\}$. 
Assume that $\TV(F_0, \widehat{F}_{\bP}) > 0$ and $C_g  < \infty$.
Define 
\begin{gather*}
\Delta_g = \EB_{\bP} \left[\sup\limits_{x \in [0, 1]}\frac{|\widehat{g}(x)-\bgP(x)|}{\min\{1, \widehat{g}(x) \}}\right]
\quad \text{and} \quad C_0 = \frac{C_g^2}{2\TV^3(F_0, \widehat{F}_{\bP})}.
\end{gather*}
Then, there exists a random variable $\EBpr \left[\frac{\rd \widetilde{\TM}(\eps)}{\rd \eps}\right]$ such that the following results hold:
\begin{subequations}
\begin{gather}
\PB\left(
\sup_{\eps \in [\eps_{\min}, 1-\eps_{\min}]}\left|
\frac{\rd \widetilde{\TM}(\eps)}{\rd \eps} - \EBpr \left[\frac{\rd \widetilde{\TM}(\eps)}{\rd \eps}\right] \right| \geq 
\frac{C_0}{\eps_{\min}^2} \sqrt{\frac{C\log n}{n}}\right) \le \frac{8}{\eps_{\min}^3} \cdot \frac{1}{n^{C-1}},
\label{eq:T-highprob-convergence} \\
\EB \left[\sup_{\eps \in [\eps_{\min}, 1-\eps_{\min}]}\left|\EBpr \left[\frac{\rd \widetilde{\TM}(\eps)}{\rd \eps}\right] \right|\right]
\leq \Delta_g \left[1 + C_0
\min \left\{  
\frac{1}{\eps_{\min}}, \EB_{F_0}[\widehat{g}]^{3}+\EB_{F_0}[\widehat{g}]^{-2}
\right\}  \right]. \label{eq:ET-bound}
\end{gather}
\end{subequations}
If we further have $\EB_{F_0}[\widehat{g}]^{-2} < \infty$, then \eqref{eq:T-highprob-convergence} can be refined to
\begin{equation}
\tag{15c}
\label{eq:T-highprob-convergence-moments}
\PB\left(
\sup_{\eps \in [\eps_{\min}, 1-\eps_{\min}]}\left|
\frac{\rd \widetilde{\TM}(\eps)}{\rd \eps} - \EBpr \left[\frac{\rd \widetilde{\TM}(\eps)}{\rd \eps}\right] \right| \geq 
C_0' \left[ 
\frac{3C\log n}{n \eps_{\min}^2} + \sqrt{\frac{3CM\log n}{n\eps_{\min}^2}}\right]
\right) \le \frac{8}{\eps_{\min}^3} \cdot \frac{1}{n^{C-1}}.
\end{equation}
where
\[
C_0' := \frac{C_g^3 \cdot \EB_{F_0} \left[\widehat{g} - \widehat{g}^{-1} \right]^2}{\TV^4(F_0, \widehat{F}_{\bP})}
\quad\text{and}\quad
M := \max_{\eps \in [0,1]} \EB_{(1-\eps)F_0 + \eps F_{\bP}} \left[\widehat{g} - \widehat{g}^{-1} \right]^2.
\]
\end{lem}

Lemma~\ref{lem:contraction-mapping} is a direct corollary of Lemma~\ref{lem:formal-contraction-mapping}. Another useful consequence of Lemma~\ref{lem:formal-contraction-mapping} is that it guarantees, with high probability, the quantity $\sup\limits_{\eps \in [\eps_{\min}, 1 - \eps_{\min}]} \left| \frac{\rd \widehat{\TM}(\eps)}{\rd \eps} \right|$ is bounded by a small value $A$ (see Corollary \ref{cor:contraction-mapping}).

\begin{cor}[Full version of Lemma \ref{lem:contraction-mapping}]
\label{cor:contraction-mapping}
Under the same assumptions and notations in Lemma \ref{lem:formal-contraction-mapping}, for any $C>1$, with probability at least $1-\frac{8}{\eps_{\min}^3} \cdot \frac{1}{n^{C-1}}$, 
\[
\sup_{\eps \in [\eps_{\min}, 1-\eps_{\min}]} \left|
\frac{\rd \widehat{\TM}(\eps)}{\rd \eps}\right|  \le A,
\]
where the bound $A$ takes the following form depending on whether a moment condition holds:
\begin{itemize}
\item If $\EB_{F_0}[\widehat{g}]^{-2} < \infty$, the bound $A$ is given by
\begin{align*}
A =  \Delta_g \left[1 + C_0 \left(
\EB_{F_0}[\widehat{g}]^3 +\EB_{F_0}[\widehat{g}]^{-2}
\right)\right]
+ C_0' \left[ 
\frac{3C\log n}{n \eps_{\min}^2} + \sqrt{\frac{3CM\log n}{n\eps_{\min}^2}}\right].
\end{align*}
\item If the moment condition $\EB_{F_0}[\widehat{g}]^{-2} < \infty$ fails, the bound $A$ becomes
\begin{align*}
A = \Delta_g \left[1 + \frac{C_0}{\eps_{\min}} \right]
+ \frac{C_0}{\eps_{\min}^2} \sqrt{\frac{3C\log n}{n}}.
\end{align*}
\end{itemize}
\end{cor}



\begin{proof}[Proof of Corollary \ref{cor:contraction-mapping}]
Note that the projection operator enforces the derivative to be zero only when its output lies outside the domain $[\eps_{\min}, 1-\eps_{\min}]$. Therefore, we have
\[
\sup_{\eps \in [\eps_{\min}, 1-\eps_{\min}]} \left|
\frac{\rd \widehat{\TM}(\eps)}{\rd \eps}\right| \leq
\sup_{\eps \in [\eps_{\min}, 1-\eps_{\min}]} \left|
\frac{\rd \widetilde{\TM}(\eps)}{\rd \eps}\right|.
\]  
 By \eqref{eq:ET-bound} in Lemma \ref{lem:formal-contraction-mapping}, it follows that or any $x >0$, 
\begin{align*}
\PB&\left(\sup_{\eps \in [\eps_{\min}, 1-\eps_{\min} ]} \left|
\frac{\rd \widetilde{\TM}(\eps)}{\rd \eps}\right|  \ge x + \Delta_g \left[1 + \frac{C_0}{4}
\min \left\{  
\frac{1}{\eps_{\min}^2}, \EB_{F_0}[\widehat{g}]^{3}+\EB_{F_0}[\widehat{g}]^{-2}
\right\}  \right]
\right) \\
&\le \PB\left(  \sup_{\eps \in [\eps_{\min}, 1-\eps_{\min} ]} \left|
\frac{\rd \widetilde{\TM}(\eps)}{\rd \eps}\right| 
-\sup_{\eps \in [\eps_{\min}, 1-\eps_{\min} ]}\left|\EBpr \left[\frac{\rd \widetilde{\TM}(\eps)}{\rd \eps}\right]\right| \ge x \right)\\
&\le \PB\left(  \sup_{\eps \in [\eps_{\min}, 1-\eps_{\min} ]} \left|
\frac{\rd \widetilde{\TM}(\eps)}{\rd \eps}-\EBpr \left[\frac{\rd \widetilde{\TM}(\eps)}{\rd \eps}\right] \right| \ge x \right).
\end{align*}
By setting $x = \frac{C_0}{\eps_{\min}^2} \sqrt{\frac{3C \log n}{n}}$ and applying \eqref{eq:T-highprob-convergence}, we complete the proof of the second bullet point. The proof of the first bullet point is similar and is therefore omitted.
\end{proof}

\subsubsection{Proof of Theorem \ref{thm:accuracyRfn}}
\label{proof:accuracyRfn-final}

The first part of Theorem \ref{thm:accuracyRfn} concerns the accuracy of the refined estimator $\epsrefine$, with its analysis provided in Theorem \ref{thm:MAE-eps-refine}. The second part addresses the accuracy of the optimal estimator $\epsopt$, which is analyzed in Theorem \ref{thm:formal-accuracyRfn-part2} whose proof is in Appendix~\ref{proof:formal-accuracyRfn-part2}. The inequality that $\sigma_n^\star \geq \tau_n^\star$ is established at the end of Appendix \ref{proof:minimax}.

\begin{theorem}[Full version of the second part of Theorem \ref{thm:accuracyRfn}]
\label{thm:formal-accuracyRfn-part2}
Assume $0 < \TV(\widehat{F}_{\bP}, F_0)$ and $\sup\limits_{x}|\widehat{g}(x)| < \infty$.
Let 
$C_g := \max\{1, \sup_{x \in [0, 1]}\widehat{g}(x)\} $.
Define
\begin{equation}
\label{eq:some-constants}
\bgP = \frac{1}{n} \sum_{t=1}^n \frac{\rd F_{\bP_t}}{\rd F_0},~~
\Delta_g = \EB_{\bP} \left[ \sup\limits_{x \in [0, 1]}\frac{|\widehat{g}(x)-\bgP(x)|}{\min\{1, \widehat{g}(x) \}} \right]
~~\text{and}~~C_0 = \frac{C_g^2}{\TV^3(F_0, \widehat{F}_{\bP})}.
\end{equation}
For a fixed true proportion \( \eps \in [0, 1] \), it follows that
\begin{align*}
\EB |\epsopt - \eps| 
&\leq \left(1 - A\right)^{-1} \left(\VM + \BM + \eps_{\min}\1\{\eps \notin [\eps_{\min}, 1-\eps_{\min}]\} \right) + \frac{8}{\eps_{\min}^3} \cdot \frac{1}{n^{C-1}}.
\end{align*}
where $C > 1$ is any given positive constant, the expression of $A$ is given in Corollary \ref{cor:contraction-mapping} (not restated here for brevity),
\(C_0\) are problem-dependent constants defined in Lemma \ref{lem:formal-contraction-mapping}, and \( \VM \) and \( \BM \) are the variance and bias terms, respectively, defined as
\[
\VM = \frac{\sqrt{\frac{1}{n} \int \left[\frac{1 - \widehat{g}(x)}{(1-\eps) + \eps \widehat{g}(x)}\right]^2 \rd \bar{F}(x)}}{\int \frac{[1 - \widehat{g}(x)]^2}{(1-\eps) + \eps \widehat{g}(x)} \rd F_0(x)},
~~\text{and}~~ 
\BM = \eps  
\EB \left|\frac{\int \hvopt(\eps, x) \left[\rd \widehat{F}_{\bP}(x) - \rd \bFprP(x)\right]}{\int \hvopt(\eps, x) \left[\rd {F}_0(x) - \rd \widehat{F}_{\bP}(x)\right]} \right| .
\]
\end{theorem}

\begin{cor}[Simplified MAE]
\label{cor:simplify-MAE-opt-eps}
Assume \( 0 < \TV(\bar{F}_{\bP}, F_0)\), \( \sup\limits_{x}|{g}(x)| < \infty \), and $\EB_{F_0}[g]^{-2} < \infty$. Let \( o_g(1) \) denote a term that converges to zero as \( \Delta_g \to 0 \) which is defined in \eqref{eq:some-constants}, and \( o(1) \) a term that converges to zero as \( n \to \infty \). 
If either setting $\eps_{\min}$ as a small constant or choosing $\eps_{\min} = \frac{\log n}{\sqrt{n}}$, we then have
\[
A =o_g(1) +  o(1), \quad \BM = o_g(1), \quad \VM = \frac{\tau_n^\star+ o_g(1)}{\sqrt{n}} .
\]
As a result, it follows that
\[
\EB |\epsopt - \eps| \precsim \frac{\tau_n^\star+ o(1)}{\sqrt{n}}  + o_g(1)+\eps_{\min}\1\{\eps \notin [\eps_{\min}, 1-\eps_{\min}]\}.
\]
The last inequality also holds without the condition $\EB_{F_0}[g]^{-2} < \infty$ given that we set $\eps_{\min} = \frac{\log n}{n^{1/4}}$ and ensure that $ n^{1/4} \cdot \Delta_g = o(1) $.
\end{cor}

\begin{proof}[Proof of Corollary \ref{cor:simplify-MAE-opt-eps}]
It suffices to simplify the variance term $\VM$ and the bias term $\BM$ defined in Theorem \ref{thm:formal-accuracyRfn-part2}.
Let $\Delta_g = \EB_{\bP} \left[ \sup_{x}\frac{|\widehat{g}(x)-\bgP(x)|}{\min\{1, \widehat{g}(x) \}} \right]$ denote the expected uniform estiamtion error between $\widehat{g}$ and $g$.
Note that $\VM$ is a continuous functional of the density ratio $\widehat{g}$.
Hence, as $\Delta_g \to 0$, it follows that $\sup_{x} \frac{|\widehat{g}(x)-g(x)|}{\min\{1, \widehat{g}(x)} \to 0$ (since the supremum of a family of convex functions is still convex) and 
\begin{equation*}
\VM \to \left[ \frac{1}{n} \int \frac{[1-g(x)]^2}{(1-\eps)+\eps g(x)} \rd F_0(x) \right]^{-\frac{1}{2}} = \frac{\tau_n^\star}{\sqrt{n}}
~~\implies~~ \VM = \frac{\tau_n^\star+o_g(1)}{\sqrt{n}}
\end{equation*}
where $o_g(1)$ denote a quantity that converges to zero when $\Delta_g$ converges to zero.

We then analyze the bias term $\BM$.
The numerator of the term $\BM$ is bounded by 
\begin{align*}
\left|\int \hvopt(\eps, x) \left[\rd \widehat{F}_{\bP}(x) - \rd \bFprP(x)\right]\right|
&= \left|\int  \frac{(1-\widehat{g}(x))(\widehat{g}(x)-\bgP(x))}{(1-\eps)+\eps \widehat{g}(x)} \rd F_0(x)\right|\\
&\le \Delta_g \cdot \int |1-\widehat{g}(x)| \rd F_0(x) 
= 2\Delta_g \cdot \TV(F_0, \widehat{F}_{\bP}).
\end{align*}
For the denominator, by \eqref{CgTVdistance}, we have
\begin{align*}
\left|\int \hvopt(\eps, x) \left[\rd F_0(x) - \rd \widehat{F}_{\bP}(x)\right]\right|&=\int  \frac{[1-\widehat{g}(x)]^2}{(1-\eps)+\eps \widehat{g}(x)}\rd {F}_0(x)\\
&\ge  \frac{\Chi(\widehat{F}_{\bP}, F_0)}{C_g} 
\ge \frac{4}{C_g} \TV^2(F_0, \widehat{F}_{\bP}).
\end{align*}
Here, \( C_g \geq 1 \) represents the upper bound of \( |\widehat{g}(x)| \). Since \( \sup_x |g(x)| = \sup_x |\rd \bar{F}_{\bP}(x)/\rd F_0(x)| < \infty \), we have \( C_g < \infty \) as \( \Delta_g \to 0 \). This ensures that the bounds involving \( C_g \) remain finite under the given assumptions. 
Combining these bounds, the bias term is bounded by 
\begin{equation*}
\BM \le \frac{C_g \Delta_g}{2\TV(F_0, \widehat{F}_{\bP})} = o_g(1).
\end{equation*}
The last equation uses the fact that as \( \Delta_g \to 0 \), the total variation distance \( \TV(F_0, \widehat{F}_{\bP}) \to \TV(F_0, \bar{F}_{\bP}) > 0 \).

\paragraph{With the moment condition $\EB_{F_0}[g]^{-2} < \infty$.}
Now suppose the moment condition $\EB_{F_0}[g]^{-2} < \infty$ holds, which implies that the moment condition in Corollary~\ref{cor:contraction-mapping} holds.
If $\eps_{\min}$ is set as a small constant, then 
\[
A = \OM\left(\Delta_g + \sqrt{\frac{\log n}{n}}\right) = o_g(1) + o(1).
\]  
If we set $\eps_{\min} = \log n / \sqrt{n}$, then we obtain 
\[
A = \OM\left( \Delta_g + \frac{1}{\log n} + \frac{1}{\sqrt{\log n}} \right) = o_g(1) + o(1).
\] 

\paragraph{Without the moment condition $\EB_{F_0}[g]^{-2} < \infty$.}
Finally, if the moment condition $ \EB_{F_0}[g]^{-2} < \infty $ does not hold, when we set $\eps_{\min} = \frac{\log n}{n^{1/4}}$, we only have
\[
A = \OM\left( \frac{\Delta_g }{\eps_{\min}} + \frac{1}{\sqrt{\log^3 n}} \right) = \OM\left( n^{1/4} \cdot \Delta_g \right) + o(1).
\] 
To ensure that the above $ A = o(1) $, we must impose the stronger requirement $ n^{1/4} \cdot \Delta_g = o(1) $, which means the estimation error in $ \widehat{g} $ must be significantly smaller than in the previous two cases.
\end{proof}

\subsubsection{Proof of Lemma \ref{lem:formal-contraction-mapping}}
\label{proof:formal-contraction-mapping}
\begin{proof}[Proof of Lemma \ref{lem:formal-contraction-mapping}]
In the following, all integrations are taken with respect to $x$. For simplicity, we omit explicit notation for the dependence on $x$.
By the definition of $\hvopt(\eps, x)$, and noting that $\widehat{g}(x) = \frac{\rd \widehat{F}_{\bP}(x)}{\rd F_0(x)}$, it follows that for any $\eps \in [0, 1]$,
\begin{align*}
\widetilde{\TM}(\eps) 
= \frac{\int \hvopt(\eps, x) \left[\rd {F}_0(x) - \rd \widehat{F}(x)\right]}{\int \hvopt(\eps, x) \left[\rd {F}_0(x) - \rd \widehat{F}_{\bP}(x)\right]}
= \frac{\int \frac{1-\widehat{g} }{(1-\eps)+\eps \widehat{g} } \left[\rd F_0  - \rd \widehat{F}  \right] }{\int  \frac{[1-\widehat{g} ]^2}{(1-\eps)+\eps \widehat{g} }\rd {F}_0 }.
\end{align*}
Direct calculation yields that
\begin{align*}
\frac{\rd \widetilde{\TM}(\eps) }{\rd \eps} 
= &\frac{\int \frac{[1-\widehat{g} ]^2}{[(1-\eps)+\eps \widehat{g} ]^2} \left[\rd F_0  - \rd \widehat{F}  \right] }{ \int\frac{[1-\widehat{g} ]^2}{(1-\eps)+\eps \widehat{g} }\rd {F}_0 } - \frac{ \int\frac{[1-\widehat{g} ]^3}{[(1-\eps)+\eps \widehat{g} ]^2}\rd {F}_0 }{\left[  \int\frac{[1-\widehat{g} ]^2}{(1-\eps)+\eps \widehat{g} }\rd {F}_0  \right]^2} \cdot \int 
 \frac{1-\widehat{g} }{(1-\eps)+\eps \widehat{g} } \left[\rd F_0  - \rd \widehat{F}  \right].
\end{align*}

\paragraph{Step 1: Expression for the predictive mean 
$\EB^{\mathrm{pr}}\!\left[\frac{\rd \widetilde{\TM}(\eps)}{\rd \eps}\right]$.}
Observe that $\frac{\rd \widetilde{\TM}(\eps)}{\rd \eps}$ depends on 
$\widehat{F}$ only through terms of the form 
$\int \varphi \, \rd \widehat{F}$, while all denominators depend only on $F_0$.
Hence, the derivative is linear in the empirical CDF $\widehat{F}$.
We thus introduce the predictive expectation operator 
\begin{equation}
\label{eq:EBpr}
\EB^{\mathrm{pr}}[\widehat{F}] = \bFpr
\end{equation}
where the expectation is taken with respect to the predictive randomness in $\widehat{F}$ while treating $F_0$ and $\widehat{g}$ as fixed.
By linearity, taking predictive expectation amounts to replacing 
$\widehat{F}$ by $\bFpr$ in the above expression. Consequently,
\begin{align*}
\EBpr \left[ \frac{\rd \widetilde{\TM}(\eps) }{\rd \eps} \right]
= &\frac{\int \frac{[1-\widehat{g} ]^2}{[(1-\eps)+\eps \widehat{g} ]^2} \left[\rd F_0  - \rd \bFpr  \right] }{ \int\frac{[1-\widehat{g} ]^2}{(1-\eps)+\eps \widehat{g} }\rd {F}_0 }   - \frac{ \int\frac{[1-\widehat{g} ]^3}{[(1-\eps)+\eps \widehat{g} ]^2}\rd {F}_0 }{\left[  \int\frac{[1-\widehat{g} ]^2}{(1-\eps)+\eps \widehat{g} }\rd {F}_0  \right]^2} \cdot \int 
 \frac{1-\widehat{g} }{(1-\eps)+\eps \widehat{g} } \left[\rd F_0  - \rd  \bFpr  \right].
\end{align*}
To distinguish, we use $\eps_0$ to denote the true proportion that generates $\widehat{F}$.
Observe the equation that
\[
\rd \bFpr
= (1-\eps_0) \rd F_0  + \eps_0 \rd  \bFprP
= \left[ (1-\eps_0) + \eps_0 \bgP  \right] \rd F_0
\quad \text{where} \quad
\bgP = \frac{1}{n} \sum_{t=1}^n \frac{\rd F_{\bP_t}}{\rd F_0}.
\]
Rearranging terms yields that
\begin{align}
\EBpr  & \left[ \frac{\rd \widetilde{\TM}(\eps) }{\rd \eps} \right] = \eps_0 \frac{\int \frac{[1-\widehat{g} ]^2[1-\bgP ]}{[(1-\eps)+\eps \widehat{g} ]^2}\rd F_0 }{ \int\frac{[1-\widehat{g} ]^2}{(1-\eps)+\eps \widehat{g} }\rd {F}_0 } 
- \eps_0 \frac{ \int\frac{[1-\widehat{g} ]^3}{[(1-\eps)+\eps \widehat{g} ]^2}\rd {F}_0 }{\left[  \int\frac{[1-\widehat{g} ]^2}{(1-\eps)+\eps \widehat{g} }\rd {F}_0  \right]^2} \cdot \int 
 \frac{[1-\widehat{g} ][1-\bgP]}{(1-\eps)+\eps \widehat{g} } \rd F_0  \nonumber \\
 &= \eps_0\left[   \frac{\int \frac{[1-\widehat{g} ]^2[1-\bgP]}{[(1-\eps)+\eps \widehat{g} ]^2}\rd F_0 }{ \int\frac{[1-\widehat{g} ]^2}{(1-\eps)+\eps \widehat{g} }\rd {F}_0 } - \frac{\int \frac{[1-\widehat{g} ]^3}{[(1-\eps)+\eps \widehat{g} ]^2}\rd F_0 }{ \int\frac{[1-\widehat{g} ]^2}{(1-\eps)+\eps \widehat{g} }\rd {F}_0 } \right]\nonumber\\
& \quad + \eps_0 \frac{ \int\frac{[1-\widehat{g} ]^3}{[(1-\eps)+\eps \widehat{g} ]^2}\rd {F}_0 }{\left[ \int\frac{[1-\widehat{g} ]^2}{(1-\eps)+\eps \widehat{g} }\rd {F}_0  \right]^2} \cdot 
\left[ \int 
 \frac{[1-\widehat{g} ]^2}{(1-\eps)+\eps \widehat{g} } \rd F_0 - \int 
 \frac{[1-\widehat{g} ][1-\bgP ]}{(1-\eps)+\eps \widehat{g} } \rd F_0 \right] \nonumber
\\
&=   \frac{\eps_0\int \frac{[1-\widehat{g} ]^2[\widehat{g} -\bgP ]}{[(1-\eps)+\eps \widehat{g} ]^2}\rd F_0 }{ \int\frac{[1-\widehat{g} ]^2}{(1-\eps)+\eps \widehat{g} }\rd {F}_0 } + \frac{ \eps_0\int\frac{[1-\widehat{g} ]^3}{[(1-\eps)+\eps \widehat{g} ]^2}\rd {F}_0 }{\left[ \int\frac{[1-\widehat{g} ]^2}{(1-\eps)+\eps \widehat{g} }\rd {F}_0  \right]^2} 
\int 
 \frac{[1-\widehat{g} ][\bgP -\widehat{g} ]}{(1-\eps)+\eps \widehat{g} } \rd F_0 . \label{eq:expectation-dT}
\end{align}


\paragraph{Step 2: Uniform upper bound for 
$\EB\!\left[\sup_{\eps \in [\eps_{\min}, 1-\eps_{\min}]}
\left|
\EB^{\mathrm{pr}}\!\left[\frac{\rd \widetilde{\TM}(\eps)}{\rd \eps}\right]
\right|\right]$.}
To proceed with the proof, we first derive individual bounds for each term in \eqref{eq:expectation-dT}.
Let $C_g = \max\{1, \sup\limits_{x\in[0, 1]}|\widehat{g}(x)|\}$.
We also recall the definition of the total variation (TV) distance:
\begin{equation}
\label{eq:TV}
\TV(F_0, \widehat{F}_{\bP}) := \frac{1}{2}\int |1-\widehat{g}(x)| \rd F_0(x).
\end{equation}
Since $0 \le \widehat{g}(x) < \infty$ and $\eps \in [\eps_{\min}, 1-\eps_{\min}]$, the following uniform bound holds:
\begin{equation}
\label{eq:fraction-bound}
-\frac{1}{\eps_{\min}} \le
-\frac{1}{\eps}\le \frac{1-\widehat{g}(x)}{(1-\eps) + \eps \widehat{g}(x)} \le \frac{1}{1-\eps} \le \frac{1}{\eps_{\min}}.
\end{equation}

\begin{enumerate}[label=$(\alph*)$]
\item Note that $\EB\left[\frac{|\widehat{g}(x)-\bgP(x)|}{\min\{1, \widehat{g}(x) \}} \right] \le \Delta_g$ for all $x \in [0, 1]$.
As a result, we have that
$\EB[|\widehat{g}(x)-\bgP(x)|] \le \Delta_g \cdot \min\{1, \widehat{g}(x) \} \le \Delta_g [(1-\eps)+ \eps \widehat{g}(x)]$ for any $\eps, x \in [0, 1]$.
Hence, the expectation of the absolute value of the first term in 
\eqref{eq:expectation-dT} is bounded by $\eps_0 \Delta_g$ for any $\eps \in [\eps_{\min},1- \eps_{\min}]$.

\item By the bound in \eqref{eq:fraction-bound} and the definition of TV distance in \eqref{eq:TV}, it follows that
\[
 \int\frac{|1-\widehat{g}(x)|^3}{[(1-\eps)+\eps \widehat{g}(x)]^2}\rd {F}_0(x)
 \le  \frac{1}{\eps_{\min}} \cdot \int\frac{[1-\widehat{g}(x)]^2}{(1-\eps)+\eps \widehat{g}(x)}\rd {F}_0(x).
\]
On the other hand, it follows that
\begin{align*}
 \int\frac{[1-\widehat{g}(x)]^3}{[(1-\eps)+\eps \widehat{g}(x)]^2}\rd {F}_0(x)
 &
 \le \int_{x: \widehat{g}(x) \le 1}\frac{[1-\widehat{g}(x)]^3}{\min\{1, \widehat{g}(x)\}^2}\rd {F}_0(x)\\
 &= \int_{x: \widehat{g}(x) \le 1}\frac{1}{[\widehat{g}(x)]^2}\rd {F}_0(x)
 \le \int \frac{1}{[\widehat{g}(x)]^2}\rd {F}_0(x).
\end{align*}
For the other direction, it follows that
\begin{align*}
 \int\frac{[\widehat{g}(x)-1]^3}{[(1-\eps)+\eps \widehat{g}(x)]^2}\rd {F}_0(x)
 &
 \le \int_{x: \widehat{g}(x) \ge 1}\frac{[\widehat{g}(x)-1]^3}{\min\{1, \widehat{g}(x)\}^2}\rd {F}_0(x)\\
 &= \int_{x: \widehat{g}(x) \ge 1} [\widehat{g}(x)]^3 \rd {F}_0(x)
 \le \int [\widehat{g}(x)]^3  \rd {F}_0(x).
\end{align*}

Combining the above bounds, we have
\[
 \left|\int\frac{[1-\widehat{g}(x)]^3 \rd {F}_0(x)}{[(1-\eps)+\eps \widehat{g}(x)]^2}\right|
 \le \min \left\{ 
\frac{1}{\eps_{\min}} \int\frac{[1-\widehat{g}(x)]^2}{(1-\eps)+\eps \widehat{g}(x)}\rd {F}_0(x),
 \EB_{F_0}[\widehat{g}]^3 + \EB_{F_0}[\widehat{g}]^{-2}
 \right\}.
\]


\item 
By the definition of $C_g$, we have that
\begin{align}
\int\frac{[1-\widehat{g}(x)]^2}{(1-\eps)+\eps \widehat{g}(x)}\rd {F}_0(x) 
&\ge \frac{1}{C_g} \int [1-\widehat{g}(x)]^2\rd {F}_0(x)  = \frac{\Chi(\widehat{F}_{\bP}, F_0)}{C_g}
\label{CgTVdistance}
\end{align}
where $\Chi(\widehat{F}_{\bP}, F_0):= \int [1-\widehat{g}(x)]^2\rd {F}_0(x)$ is the $\chi^2$-divergence between $\widehat{F}_{\bP}$ and $F_0$.

\item The definition of $\TV(F_0, \widehat{F}_{\bP})$ implies that
\begin{align*}
\EB_{\bP} \left|\int 
 \frac{[1-\widehat{g}(x)][\bgP(x)-\widehat{g}(x)]}{(1-\eps)+\eps \widehat{g}(x)} \rd F_0(x)\right|
 & \le \EB_{\bP}\left[ \sup_{x \in [0, 1]} \frac{|\bgP(x)-\widehat{g}(x)|}{\min\{1, \widehat{g}(x)\}} \right] \cdot \int |1-\widehat{g}(x)| \rd F_0(x) \\
 &\le \Delta_g \cdot \int |1-\widehat{g}(x)| \rd F_0(x) \\
 & = 2\Delta_g\TV(F_0, \widehat{F}_{\bP}).
\end{align*}

\end{enumerate}

To establish \eqref{eq:ET-bound}, we combine the bounds on the above four individual terms and apply them to the expression in \eqref{eq:expectation-dT}.
Note that $4 \TV^2(F_0, \widehat{F}_{\bP}) \le \Chi(\widehat{F}_{\bP}, F_0)$, $C_g \ge 1$, and $\TV(F_0, \widehat{F}_{\bP}) \le 1$.
We have that
\begin{align*}
\EB \left[\sup_{\eps \in [\eps_{\min}, 1-\eps_{\min}]}\left|\EBpr \left[\frac{\rd \widetilde{\TM}(\eps)}{\rd \eps}\right] \right|\right]
&\leq \eps_0 \Delta_g \left[1 + \frac{C_g^2}{\Chi(\widehat{F}_{\bP}, F_0)}
\min \left\{  
\frac{1}{\eps_{\min}}, \frac{
\EB_{F_0}[\widehat{g}]^{3}+\EB_{F_0}[\widehat{g}]^{-2}}{\sqrt{\Chi(\widehat{F}_{\bP}, F_0)}}
\right\}  \right]\\
&\leq \eps_0 \Delta_g \left[1 + \frac{C_g^2}{4\TV^2(F_0, \widehat{F}_{\bP})}
\min \left\{  
\frac{1}{\eps_{\min}}, \frac{
\EB_{F_0}[\widehat{g}]^{3}+\EB_{F_0}[\widehat{g}]^{-2}
}{2\TV(F_0, \widehat{F}_{\bP})}
\right\}  \right]\\
&\le \eps_0 \Delta_g \left[1 + \frac{C_g^2}{4\TV^3(F_0, \widehat{F}_{\bP})}
\min \left\{  
\frac{1}{\eps_{\min}}, \EB_{F_0}[\widehat{g}]^{3}+\EB_{F_0}[\widehat{g}]^{-2}
\right\}  \right].
\end{align*}

\paragraph{Step 3: Uniform concentration of 
$\frac{\rd \widetilde{\TM}(\eps)}{\rd \eps}$ around its predictive mean.}
Finally, we establish that the derivative 
$\frac{\rd \widetilde{\TM}(\eps)}{\rd \eps}$ concentrates uniformly around its predictive mean $\EB^{\mathrm{pr}}\!\left[ \frac{\rd \widetilde{\TM}(\eps)}{\rd \eps} \right].$ To this end, we invoke a uniform concentration result stated below, whose proof is provided in Appendix~\ref{proof:high-prob-concentration}.
\begin{lem}[High probability concentration]
\label{lem:high-prob-concentration}
Define two classes of functions as
\begin{equation}
\label{equ:H1H2}
\begin{aligned}
\HM_1 &= \left\{ \frac{1-\widehat{g}(x)}{(1-\eps)+\eps \widehat{g}(x)} : \eps \in [\eps_{\min}, 1-\eps_{\min}] \right\}, \\
\HM_2 &= \left\{ \frac{[1-\widehat{g}(x)]^2}{[(1-\eps)+\eps \widehat{g}(x)]^2} : \eps \in [\eps_{\min}, 1-\eps_{\min}] \right\}.
\end{aligned}
\end{equation}
For the pivotal statistics $Y_1, \ldots, Y_n$ generated under 
Model~\ref{model:previous}, for any $C > 1$,
\begin{gather}
\label{eq:concentration-no-moment}
\begin{split}
  \PB\left(\sup_{h \in \mathcal{H}_1} \left|\frac{1}{n}\sum_{t=1}^n [h(Y_t) -\EB[h(Y_t)|\FM_{t-1}]]\right| \geq \frac{1}{\eps_{\min}} \sqrt{\frac{3C\log n}{n}} \right) 
\leq \frac{4}{\eps_{\min}^3} \cdot \frac{1}{n^{C-1}},\\
\PB\left(\sup_{h \in \mathcal{H}_2} \left|\frac{1}{n}\sum_{t=1}^n [h(Y_t) - \EB[h(Y_t)|\FM_{t-1}] ]\right| \geq 
\frac{1}{\eps_{\min}^2} \sqrt{\frac{3C\log n}{n}}
\right) 
\leq \frac{4}{\eps_{\min}^3} \cdot \frac{1}{n^{C-1}}.  
\end{split}
\end{gather}
If we further have $\EB_{F_0}[\widehat{g}]^{-2} < \infty$, we then have
\[
M := \max_{\eps \in [0,1]} \EB_{(1-\eps)F_0 + \eps F_{\bP}} \left[ \left(\widehat{g} - \frac{1}{\widehat{g}}\right)^2 \right] < \infty.
\]
For every $C > 1$, 
\begin{gather}
\label{eq:concentration-with-moment}
\begin{split}
\PB\left(\sup_{h \in \mathcal{H}_1} \left|\frac{1}{n}\sum_{t=1}^n [h(Y_t) - \EB[h(Y_t)|\FM_{t-1}]]\right| \geq \frac{3C\log n}{n \eps_{\min}} + \sqrt{\frac{3CM\log n}{n}}\right) 
\leq \frac{4}{\eps_{\min}^3} \cdot \frac{1}{n^{C-1}},\\
\PB\left(\sup_{h \in \mathcal{H}_2} \left|\frac{1}{n}\sum_{t=1}^n [h(Y_t) -  \EB[h(Y_t)|\FM_{t-1}] ]\right| \geq \frac{3C\log n}{n \eps_{\min}^2} + \sqrt{\frac{3CM\log n}{n\eps_{\min}^2}}\right) 
\leq \frac{4}{\eps_{\min}^3} \cdot \frac{1}{n^{C-1}}.
\end{split}
\end{gather}
\end{lem}


We further consider two cases, depending on whether the moment condition $\EB_{F_0}[\widehat{g}]^{-2} < \infty$ holds. 

\paragraph{Without the moment condition $\EB_{F_0}[\widehat{g}]^{-2} < \infty$.}
Similar to the way we bounded the individual terms in \eqref{eq:expectation-dT}, we have:
\[
\frac{ \int\frac{|1-\widehat{g}|^3}{[(1-\eps)+\eps \widehat{g} ]^2}\rd {F}_0 }{\left[  \int\frac{[1-\widehat{g} ]^2}{(1-\eps)+\eps \widehat{g} }\rd {F}_0  \right]^2} 
\overset{\eqref{eq:fraction-bound}}{\le}
\frac{1}{\eps_{\min}} \cdot \frac{1}{\int\frac{[1-\widehat{g} ]^2}{(1-\eps)+\eps \widehat{g} }\rd {F}_0} 
\le C_1,
\quad \text{and} \quad
\frac{1}{\int\frac{[1-\widehat{g} ]^2}{(1-\eps)+\eps \widehat{g} }\rd {F}_0} \le C_2,
\]
where
\[
C_1 := \frac{C_g}{4\eps_{\min}\TV^2(F_0, \widehat{F}_{\bP})} = \frac{C_2}{\eps_{\min}}, \quad \text{and} \quad
C_2 := \frac{C_g}{4\TV^2(F_0, \widehat{F}_{\bP})}.
\]
With the above inequalities in place, we are now ready to prove \eqref{eq:T-highprob-convergence}:
\begin{align*}
\PB&\left(
\sup_{\eps \in [\eps_{\min}, 1-\eps_{\min}]}\left|
\frac{\rd \widetilde{\TM}(\eps)}{\rd \eps} - \EBpr \left[\frac{\rd \widetilde{\TM}(\eps)}{\rd \eps}\right] \right| \geq 
\left( \frac{C_1}{\eps_{\min}}+ \frac{C_2}{\eps_{\min}^2} \right) \sqrt{\frac{3C\log n}{n}}\right) \\
 &\le \PB\left( \frac{ \int\frac{|1-\widehat{g}|^3}{[(1-\eps)+\eps \widehat{g} ]^2}\rd {F}_0 }{\left[  \int\frac{[1-\widehat{g} ]^2}{(1-\eps)+\eps \widehat{g} }\rd {F}_0  \right]^2}
 \sup_{h \in \mathcal{H}_1}  \left|\frac{1}{n}\sum_{t=1}^n [h(Y_t) - \EB[h(Y_t)|\FM_{t-1}]]\right| \ge \frac{C_1}{\eps_{\min}} \sqrt{\frac{3C\log n}{n}} \right)\\
 & \qquad + 
 \PB\left( \frac{1}{\int\frac{[1-\widehat{g} ]^2}{(1-\eps)+\eps \widehat{g} }\rd {F}_0}
 \sup_{h \in \mathcal{H}_2}  \left|\frac{1}{n}\sum_{t=1}^n [h(Y_t) - \EB[h(Y_t)|\FM_{t-1}]]\right| \ge \frac{C_2}{\eps_{\min}^2} \sqrt{\frac{3C\log n}{n}} \right)\\
 &\le \PB\left(
 \sup_{h \in \mathcal{H}_1}  \left|\frac{1}{n}\sum_{t=1}^n [h(Y_t) - \EB[h(Y_t)|\FM_{t-1}]]\right| \ge \frac{1}{\eps_{\min}} \sqrt{\frac{3C\log n}{n}} \right)\\
 &\qquad  + \PB\left( 
 \sup_{h \in \mathcal{H}_2}  \left|\frac{1}{n}\sum_{t=1}^n[h(Y_t) - \EB[h(Y_t)|\FM_{t-1}]] \right| \ge \frac{1}{\eps_{\min}^2} \sqrt{\frac{3C\log n}{n}} \right)\\
 &\overset{\eqref{eq:concentration-no-moment}}{\le} \frac{8}{\eps_{\min}^3} \cdot \frac{1}{n^{C-1}}.
\end{align*}
 We can then rearrange the last inequality as follows:
\[
\PB\left(
\sup_{\eps \in [\eps_{\min}, 1-\eps_{\min}]}\left|
\frac{\rd \widetilde{\TM}(\eps)}{\rd \eps} -\EBpr \left[\frac{\rd \widetilde{\TM}(\eps)}{\rd \eps}\right]\right| \geq 
\frac{C_g}{2\eps_{\min}^2\TV^2(F_0, \widehat{F}_{\bP})} \sqrt{\frac{3C\log n}{n}}\right) \le \frac{8}{\eps_{\min}^3} \cdot \frac{1}{n^{C-1}}.
\]
We complete the proof by noting that $C_g \ge 1$ and $\TV(F_0, \widehat{F}_{\bP}) \le 1$.

\paragraph{With the moment condition $\EB_{F_0}[\widehat{g}]^{-2} < \infty$.}
By direct calculation, it follows that
\[
\frac{\left| \int\frac{[1-\widehat{g}]^3}{[(1-\eps)+\eps \widehat{g} ]^2}\rd {F}_0\right|}{\left[  \int\frac{[1-\widehat{g} ]^2}{(1-\eps)+\eps \widehat{g} }\rd {F}_0  \right]^2} 
\le  \frac{ 
\EB_{F_0}[\widehat{g}]^{3}+\EB_{F_0}[\widehat{g}]^{-2}
}{\left[\int\frac{[1-\widehat{g} ]^2}{(1-\eps)+\eps \widehat{g} }\rd {F}_0\right]^2} 
\le C_1',
\quad \text{and} \quad
\frac{1}{\int\frac{[1-\widehat{g} ]^2}{(1-\eps)+\eps \widehat{g} }\rd {F}_0} \le C_2,
\]
where the constants $C_1'$ and $C_2$ are defined as
\[
C_1' = \frac{C_g^2 \left[\EB_{F_0}[\widehat{g}]^{3}+\EB_{F_0}[\widehat{g}]^{-2}\right]}{16\TV^4(F_0, \widehat{F}_{\bP})}, 
\quad \text{and} \quad
C_2 = \frac{C_g}{4\TV^2(F_0, \widehat{F}_{\bP})}.
\]
With the concentration result in \eqref{eq:concentration-with-moment} in mind, we define the following deviation thresholds:
\[
t_1 = \frac{3C\log n}{n \eps_{\min}} + \sqrt{\frac{3CM\log n}{n}},
\quad \text{and} \quad
t_2 = \frac{3C\log n}{n \eps_{\min}^2} + \sqrt{\frac{3CM\log n}{n\eps_{\min}^2}}.
\]
With the above bounds and notation in place, we are now ready to prove \eqref{eq:T-highprob-convergence}:
\begin{align*}
\PB&\left(
\sup_{\eps \in [\eps_{\min}, 1-\eps_{\min}]}\left|
\frac{\rd \widetilde{\TM}(\eps)}{\rd \eps} - \EBpr \left[ \frac{\rd \widetilde{\TM}(\eps)}{\rd \eps}\right] \right| \geq 
C_1't_1 + C_2 t_2
\right) \\
 &\le \PB\left( \frac{ \left| \int\frac{[1-\widehat{g}]^3}{[(1-\eps)+\eps \widehat{g} ]^2}\rd {F}_0 \right| }{\left[  \int\frac{[1-\widehat{g} ]^2}{(1-\eps)+\eps \widehat{g} }\rd {F}_0  \right]^2}
 \sup_{h \in \mathcal{H}_1}  \left|\frac{1}{n}\sum_{t=1}^n [h(Y_t) - \EB[h(Y_t)|\FM_{t-1}]]\right| \ge  C_1' t_1 \right)\\
 & \qquad + 
 \PB\left( \frac{1}{\int\frac{[1-\widehat{g} ]^2}{(1-\eps)+\eps \widehat{g} }\rd {F}_0}
 \sup_{h \in \mathcal{H}_2}  \left|\frac{1}{n}\sum_{t=1}^n [h(Y_t) - \EB[h(Y_t)|\FM_{t-1}]]\right| \ge C_2 t_2\right)\\
 &\le \PB\left(
 \sup_{h \in \mathcal{H}_1}  \left|\frac{1}{n}\sum_{t=1}^n [h(Y_t) - \EB[h(Y_t)|\FM_{t-1}]]\right| \ge t_1 \right)  + \PB\left( 
 \sup_{h \in \mathcal{H}_2}  \left|\frac{1}{n}\sum_{t=1}^n [h(Y_t) - \EB[h(Y_t)|\FM_{t-1}]]\right| \ge t_2 \right)\\
 &\overset{\eqref{eq:concentration-with-moment}}{\le} \frac{8}{\eps_{\min}^3} \cdot \frac{1}{n^{C-1}}.
\end{align*}
Using the fact that $t_1 \le t_2$ and defining $C_0' := C_1' + C_2$, we can now rearrange the last inequality as follows:
\[
\PB\left(
\sup_{\eps \in [\eps_{\min}, 1-\eps_{\min}]}\left|
\frac{\rd \widetilde{\TM}(\eps)}{\rd \eps} - \EBpr \left[\frac{\rd \widetilde{\TM}(\eps)}{\rd \eps}\right] \right| \geq 
C_0' \left[ 
\frac{3C\log n}{n \eps_{\min}^2} + \sqrt{\frac{3CM\log n}{n\eps_{\min}^2}}\right]
\right) \le \frac{8}{\eps_{\min}^3} \cdot \frac{1}{n^{C-1}}.
\]
For notation simplicity, noting that $C_g \ge 1$ and $\TV(F_0, \widehat{F}_{\bP}) \le 1$, we introduce a slightly larger but more convenient upper bound $C_0'$ given by
\[
C_0' = \frac{C_g^3 \cdot \EB_{F_0}[\widehat{g} - \widehat{g}^{-1}]^{2} }{\TV^4(F_0, \widehat{F}_{\bP})}.
\]
It is straightforward to verify that $C_0' \ge C_1' + C_2$ still holds.
\end{proof}

\subsubsection{Proof of Theorem \ref{thm:formal-accuracyRfn-part2}}
\label{proof:formal-accuracyRfn-part2}
\begin{proof}[Proof of Theorem \ref{thm:formal-accuracyRfn-part2}]
By the moment conditions, \( 0 < \TV(F_0, \widehat{F}_{\bP}) \). 
By definition, $\epsopt$ is the fixed point of the equation $\eps = \widehat{\TM}(\eps) = \PM \circ \widetilde{\TM}(\eps)$, where the operators $\widehat{\TM}, \PM$ and $\widetilde{\TM}$ are defined in \eqref{eq:new-emprical-operator}. Let the true proportion that generates the observed samples $Y_1, \ldots, Y_n$ be denoted by $\eps_0$. Then, $\eps_0$ satisfies the fixed-point equation $\eps_0 = \TM(\eps_0)$, where $\TM$ is essentially an identity mapping for any NTP distributions:
\[
{\TM}(\eps) = \frac{\int \vopt(\eps, x) \left[\rd {F}_0(x) - \rd \bar{F}(x)\right]}{\int \vopt(\eps, x) \left[\rd {F}_0(x) - \rd \bar{F}_{\bP}(x)\right]},
\]
and $\vopt(\eps, x)$ is the optimal weight function. 
To see this, we substitute the definition of $\vopt$ and get
\[
{\TM}(\eps) = \frac{\int \frac{1-{g}(x)}{(1-\eps)+\eps {g}(x)} \left[\rd F_0(x) - \rd \bar{F}(x)\right]}{\int \frac{[1-{g}(x)]^2}{(1-\eps)+\eps {g}(x)} \rd {F}_0(x)} = \eps.
\]

As $\epsopt \in [\eps_{\min}, 1-\eps_{\min}]$ and $\eps_0 \in [0, 1]$ are solutions to their respective fixed-point equations, we have
\begin{align*}
| \epsopt - \eps_0|
&\le| \epsopt - \PM(\eps_0)| + |\eps_0 - \PM(\eps_0)|\\
&\overset{(a)}{\le} |\widehat{\TM}(\epsopt) - {\TM} \circ \PM(\eps_0)| + \eps_{\min}\1\{\eps \notin [\eps_{\min}, 1-\eps_{\min}]\}\\
&\le |\widehat{\TM}(\epsopt) -\widehat{\TM}\circ \PM(\eps_0)| + |\widehat{\TM}\circ \PM(\eps_0) - {\TM}\circ \PM(\eps_0)  | + \eps_{\min}\1\{\eps \notin [\eps_{\min}, 1-\eps_{\min}]\}\\
&\overset{(b)}{\le} |\widetilde{\TM}(\epsopt) -\widetilde{\TM}\circ \PM(\eps_0)| + |\widetilde{\TM}\circ \PM(\eps_0) - {\TM}\circ \PM(\eps_0)  | + \eps_{\min}\1\{\eps \notin [\eps_{\min}, 1-\eps_{\min}]\}\\
&\overset{(c)}{\le} L \cdot |\epsopt -\PM(\eps_0)| + |\widetilde{\TM}\circ \PM(\eps_0) - {\TM}\circ \PM(\eps_0)  | + \eps_{\min}\1\{\eps \notin [\eps_{\min}, 1-\eps_{\min}]\},
\end{align*}
where $(a)$ uses the relation that $|\eps_0 - \PM(\eps_0)| \le \eps_{\min}\1\{\eps \notin [\eps_{\min}, 1-\eps_{\min}]\}$.
This is because we have $|\eps_0 - \PM(\eps_0)| = 0$ if $\eps_0$ is within the interval $[\eps_{\min},1-\eps_{\min}]$; otherwise, $|\eps_0 - \PM(\eps_0)| \leq \eps_{\min}$. 
Here $(b)$ uses the relation $\PM \circ \TM \circ \PM = \TM \circ \PM$, and $(c)$ uses the fact that $\widetilde{\TM}$ is $L$-Lipschitz continuous, where  
\[
L =\sup_{\eps \in [\eps_{\min},1-\eps_{\min}]}\left|\frac{\rd \widetilde{\TM}(\eps) }{\rd \eps}\right|. 
\]
Rearranging the resulting inequality yields
\begin{equation}
\label{eq:intermedia-1}
| \epsopt - \eps_0| \le \frac{|\widetilde{\TM}\circ \PM(\eps_0) - {\TM}\circ \PM(\eps_0)| + 2\eps_{\min}\1\{\eps \notin [\eps_{\min}, 1-\eps_{\min}]\}}{1-L}.
\end{equation}

Without loss of generality, we assume $\eps_0 \in [\eps_{\min}, 1-\eps_{\min}]$. If $\eps_0$ lies outside this interval, we instead consider $\PM(\eps_0)$.
Let $\bFprP$ be defined in \eqref{eq:bFprP} and the operator $\EBpr(\cdot)$ be defined in \eqref{eq:EBpr}.
Now, let us derive a bound for \( \EB|\widetilde{\TM}(\eps_0) - {\TM}(\eps_0)| \).
It follows that
\begin{align}
\label{eq:epsopt-general-upper-bound}
   \EB &|\widetilde{\TM}(\eps_0) - {\TM}(\eps_0)  | \nonumber \\
   &= \EB|\widetilde{\TM}(\eps_0) - \EBpr [\widetilde{\TM}(\eps_0)] + \EBpr [\widetilde{\TM}(\eps_0)] - {\TM}(\eps_0)  | \nonumber \\  
    & \leq \EB|\widetilde{\TM}(\eps_0) - \EBpr [\widetilde{\TM}(\eps_0)] | + \EB|\EBpr [\widetilde{\TM}(\eps_0)] - {\TM}(\eps_0)  |\nonumber \\ 
    & \leq \sqrt{ \EB |\widetilde{\TM}(\eps_0) - \EBpr [\widetilde{\TM}(\eps_0)]|^2} +  \EB \left|\frac{\int \hvopt(\eps_0, x) \left[\rd {F}_0(x) - \rd \bFpr(x)\right]}{\int \hvopt(\eps_0, x) \left[\rd {F}_0(x) - \rd \widehat{F}_{\bP}(x)\right]} - \eps_0\right| \nonumber \\ 
    &\le \frac{ \sqrt{ \frac{1}{n} \int \left[ \frac{1-\widehat{g}(x)}{(1-\eps)+\eps \widehat{g}(x)}  \right]^2 \rd \bar{F}(x)  }}{ \int \frac{ [1-\widehat{g}(x)]^2 }{ (1-\eps)+\eps \widehat{g}(x) } \rd F_0(x) } + \eps_0  
    \EB \left|\frac{\int \hvopt(\eps_0, x) \left[\rd \widehat{F}_{\bP}(x) - \rd \bFprP(x)\right]}{\int \hvopt(\eps_0, x) \left[\rd {F}_0(x) - \rd \widehat{F}_{\bP}(x)\right]} \right| \nonumber \\
    &=: \VM + \BM.
\end{align}

Finally, we analyze the Lipschitz constant $L$.
By Lemma \ref{lem:formal-contraction-mapping}, we have 
\begin{equation}
\label{eq:bound-for-A}
\PB(L \ge A)  \le \frac{8}{\eps_{\min}^3} \frac{1}{n^{C-1}}.
\end{equation}
We then complete the proof by combining \eqref{eq:intermedia-1}, \eqref{eq:epsopt-general-upper-bound} and \eqref{eq:bound-for-A}.
\end{proof}

\subsection{Proof of Lemma \ref{lem:high-prob-concentration}}
\label{proof:high-prob-concentration}

\begin{proof}[Proof of Lemma \ref{lem:high-prob-concentration}]

Since the underlying NTP distributions $P_1,\dots,P_n$ are generated autoregressively and are therefore random, the observed pivotal statistics $Y_1,\dots,Y_n$ are generally not independent. 
To address this dependence, we cast the concentration problem in a martingale framework. 
Let $\FM_t=\sigma(Y_1,\dots,Y_t,\,P_1,\dots,P_{t+1})$ denote the natural filtration as specified in Model~\ref{model:previous}.
We treat the two function classes $\HM_1$ and $\HM_2$ separately, which are defined in \eqref{equ:H1H2}. 
Fix $k\in\{1,2\}$ and $\eps\in\Theta:=[\eps_{\min},1-\eps_{\min}]$. 
Let $h^{(k)}_{\eps}\in\HM_k$ denote the function associated with parameter $\eps$. 
Define the martingale difference sequence
\[
D^{(k)}_{t}(\eps)
:= h^{(k)}_{\eps}(Y_t)
- \EB\!\left[h^{(k)}_{\eps}(Y_t)\mid \FM_{t-1}\right],
\qquad t=1,\dots,n.
\]
By construction, $\EB[D^{(k)}_{t}(\eps)\mid \FM_{t-1}]=0$, and hence $\{D^{(k)}_{t}(\eps)\}_{t=1}^n$ forms a martingale difference sequence adapted to $\{\FM_t\}_{t=1}^n$. 
Our objective is to control the uniform deviation
\[
\sup_{\eps\in\Theta}
\left|
\frac{1}{n}\sum_{t=1}^n D^{(k)}_{t}(\eps)
\right|.
\]

\subsubsection{\texorpdfstring{$\eta$}{eta}-net Discretization}
Fix $k\in\{1,2\}$ and consider the function class $\HM_k$. 
Since both $\HM_1$ and $\HM_2$ are parameterized by the scalar 
$\eps \in \Theta := [\eps_{\min}, 1-\eps_{\min}]$, 
the following discretization argument applies uniformly to both classes.

To obtain uniform control over $\eps \in \Theta$, 
we discretize the parameter space via an $\eta$-net. 
Let $h^{(k)}_{\eps}$ denote the function in $\HM_k$ associated with parameter $\eps$. 
From the definitions of $\HM_1$ and $\HM_2$, we can bound the derivatives 
with respect to $\eps$ uniformly in $x$:
\[
\text{For } \mathcal{H}_1, \quad \left| \frac{\rd}{\rd \eps} \left(\frac{1-\widehat{g}(x)}{(1-\eps) + \eps \widehat{g}(x)}\right) \right| = \frac{(1-\widehat{g}(x))^2}{((1-\eps) + \eps \widehat{g}(x))^2} \leq \frac{1}{\eps_{\min}^2},
\]
\[
\text{For } \mathcal{H}_2, \quad \left| \frac{\rd}{\rd \eps} \left(\frac{[1-\widehat{g}(x)]^2}{[(1-\eps)+\eps \widehat{g}(x)]^2}\right) \right| = \frac{2|1-\widehat{g}(x)|^3}{((1-\eps) + \eps \widehat{g}(x))^3} \leq \frac{2}{\eps_{\min}^3}.
\]
For simplicity, we use the uniform Lipschitz constant $L_{\max} := \frac{2}{\eps_{\min}^3}$ for both classes.
We now construct a uniform $\eta$-net $\mathcal{N}$ over the interval $\Theta = [\eps_{\min}, 1-\eps_{\min}]$ with mesh size $\eta = \frac{\eps_{\min}^3}{2n}$. The total number of grid points is bounded by:
\[
|\mathcal{N}| \leq \frac{1}{\eta} = \frac{2n}{\eps_{\min}^3}.
\]
By definition, for any $\eps \in \Theta$, there exists $\eps_j \in \mathcal{N}$ such that 
$|\eps - \eps_j| \le \eta$. 
Using the Lipschitz property and the contraction property of conditional expectation, 
we obtain for each $t$,
\[
\bigl| D^{(k)}_{t}(\eps) - D^{(k)}_{t}(\eps_j) \bigr|
\le 2 L_{\max} \cdot |\eps - \eps_j|
\le 2 \cdot \frac{2}{\eps_{\min}^3}
\cdot \frac{\eps_{\min}^3}{2n}
= \frac{2}{n}.
\]
Summing over $t=1,\dots,n$ and dividing by $n$ yields the deterministic bound
\[
\sup_{\eps \in \Theta}
\left|
\frac{1}{n}\sum_{t=1}^n D^{(k)}_{t}(\eps)
\right|
\le
\max_{\eps_j \in \mathcal{N}}
\left|
\frac{1}{n}\sum_{t=1}^n D^{(k)}_{t}(\eps_j)
\right|
+ \frac{2}{n}.
\]
Applying a union bound over the finite set $\mathcal{N}$, we reduce the uniform deviation to a finite collection of pointwise bounds:
\begin{equation}
\label{eq:union-bound-grid}
\PB\!\left(
\sup_{\eps \in \Theta}
\left|
\frac{1}{n}\sum_{t=1}^n D^{(k)}_{t}(\eps)
\right|
\ge t
\right)
\le
|\mathcal{N}|
\max_{\eps_j \in \mathcal{N}}
\PB\!\left(
\left|
\sum_{t=1}^n D^{(k)}_{t}(\eps_j)
\right|
\ge
n\!\left(t - \frac{2}{n}\right)
\right).
\end{equation}
The above reduction holds for both $k=1$ and $k=2$. 
In the next step, we bound the pointwise tail probability 
for a fixed $\eps_j \in \mathcal{N}$ separately for the two function classes.

\subsubsection{Without Moment Conditions}

In the absence of additional moment assumptions, we apply the Azuma–Hoeffding inequality (in Lemma \ref{lem:azuma}) to control the right-hand side of \eqref{eq:union-bound-grid} for the classes $\HM_1$ and $\HM_2$, respectively. 
As a consequence, the resulting concentration bounds depend explicitly on $1/\eps_{\min}^2$.

\begin{lem}[Azuma–Hoeffding inequality]
\label{lem:azuma}
Let $\{D_t\}_{t=1}^n$ be a martingale difference sequence adapted to 
$\{\FM_t\}_{t=1}^n$. Suppose that $|D_t| \le c_t$ almost surely for all $t$. 
Then for any $x>0$,
\[
\PB\!\left(
\left|\sum_{t=1}^n D_t\right| \ge x
\right)
\le
2 \exp\!\left(
-\frac{x^2}{2\sum_{t=1}^n c_t^2}
\right).
\]
\end{lem}

\paragraph{Case $k=1$ (the class $\HM_1$).}

Since $\eps \in [\eps_{\min}, 1-\eps_{\min}]$ and $\widehat{g}(x) \ge 0$, we have 
$\sup_x |h^{(1)}_{\eps}(x)| \le 1/\eps_{\min}$. 
Consequently, the martingale difference satisfies 
$|D^{(1)}_t(\eps)| \le 1/\eps_{\min} =: c_t$. 
Applying Lemma~\ref{lem:azuma} with 
$x = n\!\left(t - 2/n\right)$ yields
\[
\PB\!\left(\left|\sum_{t=1}^n D^{(1)}_t(\eps_j)\right| 
\ge n\!\left(t - \frac{2}{n}\right)\right)
\le 
2 \exp\!\left(-\frac{n \eps_{\min}^2 (t - 2/n)^2}{2}\right).
\]
Let $t \ge \frac{1}{\eps_{\min}}\sqrt{\frac{3C\log n}{n}} \ge \frac{2}{n} + \frac{1}{\eps_{\min}}\sqrt{\frac{2C\log n}{n}}$ for any $C \ge 1$. 
Substituting this into \eqref{eq:union-bound-grid}, we obtain
\[
\PB\!\left(
\sup_{\eps \in \Theta}
\left|
\frac{1}{n}\sum_{t=1}^n D^{(1)}_{t}(\eps)
\right|
\ge 
\frac{1}{\eps_{\min}}\sqrt{\frac{3C\log n}{n}}
\right)
\le 
\frac{4}{\eps_{\min}^3}\,\frac{1}{n^{C-1}}.
\]

\paragraph{Case $k=2$ (the class $\HM_2$).}
Similarly, $\sup_x |h^{(2)}_{\eps}(x)| \le 1/\eps_{\min}^2$, implying 
$|D^{(2)}_t(\eps)| \le 1/\eps_{\min}^2$. 
Applying Lemma~\ref{lem:azuma} gives
\[
\PB\!\left(\left|\sum_{t=1}^n D^{(2)}_t(\eps_j)\right|
\ge n\!\left(t - \frac{2}{n}\right)\right)
\le 
2 \exp\!\left(-\frac{n \eps_{\min}^4 (t - 2/n)^2}{2}\right).
\]
Choosing 
$t \ge \frac{1}{\eps_{\min}^2}\sqrt{\frac{3C\log n}{n}} \ge \frac{2}{n} + \frac{1}{\eps_{\min}^2}\sqrt{\frac{2C\log n}{n}}$
yields the corresponding bound:
\[
\PB\!\left(
\sup_{\eps \in \Theta}
\left|
\frac{1}{n}\sum_{t=1}^n D^{(2)}_{t}(\eps)
\right|
\ge 
\frac{1}{\eps_{\min}^2}\sqrt{\frac{3C\log n}{n}}
\right)
\le 
\frac{4}{\eps_{\min}^3}\,\frac{1}{n^{C-1}}.
\]



\subsubsection{With Moment Conditions}
We now consider the setting in which a bound on the predictable quadratic variation is available. 
In this case, we exploit the variance structure of the martingale sequence via Freedman's inequality.

\begin{lem}[Freedman's inequality]
\label{lem:freedman}
Let $\{D_t\}_{t=1}^n$ be a martingale difference sequence adapted to 
$\{\FM_t\}_{t=1}^n$. Suppose that $|D_t| \le U$ almost surely for some deterministic constant $U > 0$.
Let 
$V_n = \sum_{t=1}^n \EB[D_t^2 \mid \FM_{t-1}]$ 
be the predictable quadratic variation. 
Then for any $x, V>0$,
\[
\PB\!\left(
\left|\sum_{t=1}^n D_t\right| \ge x
\ \text{and}\ 
V_n \le V
\right)
\le
2 \exp\!\left(
-\frac{x^2}{2(V + Ux/3)}
\right).
\]
\end{lem}

\paragraph{A moment condition.}
Unlike the i.i.d. setting, the quadratic variation in our case, i.e., $V_n$, is random. 
To apply Lemma~\ref{lem:freedman} directly, we need to bound it by a deterministic constant.
In the following, we show that if $\EB_{F_0}[\widehat g^{-2}] < \infty$, this is doable.
More specifically, we define the deterministic constant
\[
M 
:= 
\max_{\eps \in [0,1]}
\EB_{X \sim (1-\eps)F_0 + \eps F_{\bP}}
\!\left[
\left(\widehat g(X) - \frac{1}{\widehat g(X)}\right)^2
\right],
\]
and claim that $M < \infty$ whenever $\EB_{F_0}[\widehat g^{-2}] < \infty$.
Since $\widehat g$ is already uniformly bounded, it suffices to show that 
$\max_{\eps \in [0,1]}
\EB_{X \sim (1-\eps)F_0 + \eps F_{\bP}}[\widehat g(X)^{-2}]$ is finite.
By Assumption~\ref{asmp:main}, with probability one, all generated NTP distributions $\bP_t$ are non-singular and satisfy $\sup_x \sup_{\bP} \left| \rd F_{\bP}(x) / \rd F_0(x) \right| \le C$ for some constant $C \ge 1$.  
As a result, for any non-negative measurable function $\phi$, we have
$\EB_{X \sim (1-\eps)F_0 + \eps F_{\bP}}[\phi(X)]
= (1-\eps)\EB_{F_0}[\phi(X)]
+ \eps \EB_{F_{\bP}}[\phi(X)]
\le C \, \EB_{F_0}[\phi(X)]$.
Applying this with $\phi(x) = \widehat g(x)^{-2}$ yields the desired finite bound.

\paragraph{Case $k=1$ (the class $\HM_1$).}

Recall the algebraic inequality 
\[
|h_{\eps}^{(1)}(x)| = \frac{|1-\widehat{g}(x)|}{(1-\eps) + \eps \widehat{g}(x)}
\le \frac{|1-\widehat{g}(x)|}{\min\{1, \widehat{g}(x)\}}
=\begin{cases}
\widehat{g}(x) - 1 &~\text{if}~\widehat{g}(x) \ge 1\\
  \frac{1}{\widehat{g}(x)}-1  &~\text{if}~\widehat{g}(x) \le 1
\end{cases}
\le \left|\widehat g(x) - \frac{1}{\widehat{g}(x)} \right|.
\]
Therefore, for any $\eps \in \Theta$,
\[
\EB[(D_{t}^{(1)}(\eps))^2 \mid \FM_{t-1}]
\le
\EB[h_{1,\eps}^2(Y_t) \mid \FM_{t-1}]
\le
\EB\left[(\widehat g(x) - {\widehat{g}(x)}^{-1})^2 | \FM_{t-1} \right]
\le M
\quad \text{a.s.}
\]
This implies the predictable quadratic variation is deterministically bounded:
\[
V_n = \sum_{t=1}^n \EB[(D_{t}^{(1)}(\eps))^2 \mid \FM_{t-1}] \leq nM =: V \quad \text{a.s.}
\]
Applying Lemma~\ref{lem:freedman} with $V = nM, U = 2/\eps_{\min}$, and $x = n(t - 2/n)$ implies that for any $\eps_j \in \mathcal{N}$, 
\[
\PB\left(\left|\sum_{t=1}^n D_{t}^{(1)}(\eps_j)\right| \geq n\left(t - \frac{2}{n}\right)\right) \leq 2 \exp\left( - \frac{n^2(t - 2/n)^2}{2\left(nM + \frac{2n(t - 2/n)}{3\eps_{\min}}\right)} \right) = 2 \exp\left( - \frac{n (t - 2/n)^2}{2M + \frac{4(t - 2/n)}{3\eps_{\min}}} \right).
\]
For any $C \ge 1$, by setting
\[
t\ge \frac{3C \log n}{n \eps_{\min}} + \sqrt{\frac{3CM \log n}{n}} 
\geq \frac{2}{n} + \frac{4C \log n}{3 n \eps_{\min}} + \sqrt{\frac{2CM \log n}{n}},
\]
we have $\exp\left( - \frac{n (t - 2/n)^2}{2M + \frac{4(t - 2/n)}{3\eps_{\min}}} \right) \le n^{-C}.$
Substituting the last inequality into \eqref{eq:union-bound-grid}, we obtain
\[
\PB\!\left(
\sup_{\eps \in \Theta}
\left|
\frac{1}{n}\sum_{t=1}^n D^{(1)}_{t}(\eps)
\right|
\ge 
\frac{3C \log n}{n \eps_{\min}} + \sqrt{\frac{3C M \log n}{n}}
\right)
\le 
\frac{4}{\eps_{\min}^3}\,\frac{1}{n^{C-1}}.
\]

\paragraph{Case $k=2$ (the class $\HM_2$).}

The argument is analogous. Since $\sup_x |h_{\eps}^{(2)}(x)| \le 1/\eps_{\min}^2$, Freedman's inequality in Lemma~\ref{lem:freedman} applies with $U = 2/\eps_{\min}^2$ and $V = nM/\eps_{\min}^2$. Therefore,
\[
\PB\!\left(
\sup_{\eps \in \Theta}
\left|
\frac{1}{n}\sum_{t=1}^n D^{(2)}_{t}(\eps)
\right|
\ge 
\frac{3C \log n}{n \eps_{\min}^2} + \sqrt{\frac{3C M \log n}{n \eps_{\min}^2}}
\right)
\le 
\frac{4}{\eps_{\min}^3}\,\frac{1}{n^{C-1}}.
\]

\end{proof}

\subsection{Proof of Theorem \ref{thm:minimax}}
\label{proof:minimax}


\begin{proof}[Proof of Theorem \ref{thm:minimax}]

To establish the minimax lower bound, it suffices to consider a subclass 
of data-generating mechanisms under which $Y_1,\ldots,Y_n$ are i.i.d. 
Indeed, since the minimax risk over a larger model class is lower bounded 
by the minimax risk over any subclass, we may restrict attention to an 
order-invariant mechanism.
Specifically, we construct the following i.i.d.\ model. 
At each step $t$, the NTP distribution $\bP_t$ is sampled independently 
from a fixed distribution, and the pivotal statistic $Y_t$ is generated according to 
$(1-\eps_i)F_0 + \eps_i F_{\bP_t}$ independently across $t$. 
Under this construction, $Y_1,\ldots,Y_n$ are independent and identically distributed.

We then apply Le Cam’s two-point method to derive the minimax lower bound 
\citep[Chapter~15.2]{wainwright2019high}. From Proposition 15.1 and (15.14) in \citep{wainwright2019high}, it follows that
\begin{align}
 \inf_{\widehat{\eps} \in \GM} \sup_{\eps_0 \in \BM(\eps)} \EB|\widehat{\eps} - \eps_0| 
 & \ge \inf_{\widehat{\eps} \in \GM} \sup_{\eps_0 \in \{\eps_1, \eps_2\}} \EB|\widehat{\eps} - \eps_0| 
 \ge \frac{|\eps_1 - \eps_2|}{2} \left(1 - \TV(\rho_1, \rho_2)\right).  \label{eq:lower-bound-1} 
\end{align}
Here, $\eps_i$ for $i=1,2$ are carefully chosen proportions within $\BM(\eps)$, and $\rho_i$ denotes the joint probability distribution of the considered functional of pivotal statistics in $\GM$ when the underlying proportion is $\eps_i$. Recall that each $Y_t \mid \bP_t$ is generated according to $(1 - \eps_i) \mu_0 + \eps_i \mu_{1, \bP_t}$, where $\bP_t$ is independently and uniformly sampled from the set $\{\bP_1, \ldots, \bP_n\}$.
We explain the choice of $\eps_i$ as follows. Since $\BM(\eps)$ is an open interval containing $\eps$, we must have $\eps \in \BM(\eps)$ and $\eps + \Delta \in \BM(\eps)$ for a sufficiently small $\Delta$. We then select $\eps_1 = \eps + \Delta$ and $\eps_2 = \eps$.

In the following, we discuss two classes of estimators and identify the minimax lower bound for their mean absolute error. The proof ideas are essentially the same.

\paragraph{(a) Indicator functions.}
Let $\GM$ collect all measurable functions of the $n$-dimensional indicator functions $(\mathbf{1}\{Y_1 \leq \delta\}, \ldots, \mathbf{1}\{Y_n \leq \delta\})$.
We first establish the lower bound for $\widehat{\eps} \in \GM$.
Given that the $Y_t$'s are generated independently and identically, the joint distribution $\rho_i$ is the product of $n$ independent Bernoulli distributions:
\[
\rho_i = \mathrm{Ber}(p(\eps_i))^{\otimes n},
\]
where
\[
p(\eps_i) = (1 - \eps_i) F_0(\delta) + \eps_i \bar{F}_{\bP}(\delta).
\]
By the relation \eqref{eq:eps-equation} and the choice that $\eps_2 = \eps$, it follows that
\begin{equation}
\label{eq:average}
\bar{F}(\delta) = (1-\eps) F_0(\delta) + \eps  \bar{F}_{\bP}(\delta) = p(\eps) = p(\eps_2).
\end{equation}

Next, we further upper bound the total variation distance $\TV(\rho_1, \rho_2)$ in \eqref{eq:lower-bound-1} to simplify the expression.
\begin{align}
\TV(\rho_1, \rho_2)
&\overset{(a)}{\le} \sqrt{\frac{1}{2} \KL(\rho_1, \rho_2)} \nonumber \\
&\overset{(b)}{\le}\sqrt{\frac{n}{2} \KL(\mathrm{Ber}(p(\eps_1)), \mathrm{Ber}(p(\eps_2))) } \nonumber\\
&\overset{(c)}{\le}\sqrt{\frac{n|\eps_1-\eps_2|^2}{2}  \frac{[F_0(\delta)-\bar{F}_{\bP}(\delta)]^2}{p(\eps_2)(1-p(\eps_2))} } \nonumber\\
&\overset{(d)}{\le} \frac{1}{\sqrt{2}} |F_0(\delta)-\bar{F}_{\bP}(\delta)| \Delta \cdot \sqrt{ \frac{n}{\bar{F}(\delta)(1-\bar{F}(\delta))}} \nonumber\\
&\overset{(e)}{\le} \sqrt{\frac{n}{2}} \frac{\Delta}{\sigma_n^\star}, \label{eq:bound-TV}
\end{align}
where
\begin{enumerate}[label=$(\alph*)$]
    \item uses the Pinsker inequality, which allows us to upper bound the total variation distance by the square root of half the Kullback-Leibler (KL) divergence.
    \item employs the decoupling property of KL divergences and the fact that $\rho_i = \mathrm{Ber}(p(\eps_i))^{\otimes n}$, leading to
    \[
    \KL(\rho_1 \| \rho_2) = n \KL(\mathrm{Ber}(p(\eps_1)) \| \mathrm{Ber}(p(\eps_2))).
    \]
    \item utilizes the numerical inequality that
    \[
    \KL(\mathrm{Ber}(p) \| \mathrm{Ber}(q)) = p \log \frac{p}{q} + (1 - p) \log \frac{1 - p}{1 - q} \leq \frac{(p - q)^2}{q (1 - q)}.
    \]
    \item uses the notation $\Delta = |\eps_1 - \eps_2|$.
    \item uses the definition of $\sigma_n^\star$ that is $[\sigma_n^\star]^2 = \frac{\bar{F}(\eps)(1-\bar{F}(\eps))}{[F_0(\delta)-\bar{F}_{\bP}(\delta)]^2}.$
\end{enumerate}

Combining \eqref{eq:lower-bound-1} and \eqref{eq:bound-TV}, we obtain
\[
\inf_{\widehat{\eps} \in \GM} \sup_{\eps_0 \in \BM(\eps)} 
\EB|\widehat{\eps} - \eps_0| \ge 
\frac{\Delta}{2} \left(1 - \frac{1}{\sqrt{2}} \frac{\sqrt{n} \Delta}{\sigma_n^\star} \right).
\]
By setting $\Delta = \frac{\sigma_n^\star}{\sqrt{2n}}$ and assuming that $n$ is sufficiently large such that $\eps + \Delta \in \BM(\eps)$, we obtain
\[
\inf_{\widehat{\eps} \in \GM} \sup_{\eps_0 \in \BM(\eps)} 
\EB|\widehat{\eps} - \eps_0| \ge \frac{\sigma_n^\star}{\sqrt{2n}}.
\]
This completes the proof of the first part.

\paragraph{(b) General functions.}
Let $\GM$ denote the set of all general measurable functions of the $n$-dimensional random vector $(Y_1, \ldots, Y_n)$. We now establish the lower bound for any $\widehat{\eps} \in \GM$ under this broader class of functions.
Given that the $Y_t$'s are generated independently, the joint distribution $\rho_i$ is the product of $n$ independent mixture distributions:
\[
\rho_i = [(1-\eps_i) F_0 +\eps_i \bar{F}_{\bP}]^{\otimes n}.
\]
We similarly begin with \eqref{eq:lower-bound-1} and aim to upper bound the total variation distance $\TV(\rho_1, \rho_2)$. Since the proof closely resembles that of \eqref{eq:bound-TV}, we focus only on the key differences below.
\begin{align}
\TV(\rho_1, \rho_2)
&\le \sqrt{\frac{1}{2} \KL(\rho_1, \rho_2)} \nonumber \\
&= \sqrt{\frac{n}{2} \KL((1-\eps_1) F_0 +\eps_1 \bar{F}_{\bP}, (1-\eps_2) F_0 +\eps_2 \bar{F}_{\bP}) } \nonumber\\
&\overset{(a)}{\le} \sqrt{\frac{n}{2} \Chi((1-\eps_1) F_0 +\eps_1 \bar{F}_{\bP}, (1-\eps_2) F_0 +\eps_2 \bar{F}_{\bP}) } \nonumber\\
&\overset{(b)}{\le} \sqrt{\frac{|\eps_1-\eps_2|^2}{2} n \cdot \int \frac{(\rd F_0 - \rd \bar{F}_{\bP})^2}{(1-\eps)\rd F_0 + \eps \rd \bar{F}_{\bP}} } \nonumber\\
&=\Delta \sqrt{\frac{n}{2}} \cdot \sqrt{\int \frac{(\rd F_0 - \rd \bar{F}_{\bP})^2}{(1-\eps)\rd F_0 + \eps \rd \bar{F}_{\bP}} } \nonumber\\
&\overset{(c)}{=}  \sqrt{\frac{n}{2}} \frac{\Delta}{\tau_n^\star},
\label{eq:bound-TV-2}
\end{align}
where
\begin{enumerate}[label=$(\alph*)$]
    \item uses the fact that the Kullback-Leibler (KL) divergence is always upper bounded by its Chi-square divergence:
    \[
    \KL(P,Q) = \int \log \frac{\rd P}{\rd Q} \rd P \le  \int \frac{(\rd P - \rd Q)^2}{ \rd Q} = \Chi(P, Q).
    \]
    \item uses the fact that
    \begin{equation}
    \label{eq:value-chi2}
        \Chi(
(1-\eps_1) F_0 +\eps_1 \bar{F}_{\bP}, (1-\eps_2) F_0 +\eps_2 \bar{F}_{\bP}
) = |\eps_1-\eps_2|^2 \cdot \int \frac{(\rd F_0 - \rd \bar{F}_{\bP})^2}{(1-\eps)\rd F_0 + \eps \rd \bar{F}_{\bP}}
    \end{equation}
    \item uses the definition of $\tau_n^\star$,, which is given by:  
    \[
    [\tau_n^\star]^2 
    = \left[ \int \frac{(\rd F_0 - \rd \bar{F}_{\bP})^2}{(1-\eps)\rd F_0 + \eps \rd \bar{F}_{\bP}}  \right]^{-1}
    = \left[ \int \frac{(1-g(x))^2}{(1-\eps)+\eps g(x)} \rd F_0(x) \right]^{-1}
    ~~\text{where}~~
    g(x) = \frac{\rd \bar{F}_{\bP}(x)}{\rd F_0(x)}.
    \]
\end{enumerate}
The remaining argument follows the same reasoning as in part (a).
In this way, we will have
\[
\inf_{\widehat{\eps} \in \GM} \sup_{\eps_0 \in \BM(\eps)} 
\EB|\widehat{\eps} - \eps_0| \ge \frac{\tau_n^\star}{\sqrt{2n}}.
\]

\paragraph{(c) Final inequality.}
In the end, we prove the inequality that $\sigma_n^\star \geq \tau_n^\star$. The data processing inequality \citep{thomas2006elements} states that any post-processing of two distributions reduces the value of the $f$-divergence between them, including the Chi-square divergence.  
Therefore, using the same choices of $\eps_1$, $\eps_2$ as before, we have:  
\begin{align}
\label{eq:help1}
\Chi&(\mathrm{Ber}(p(\eps_1)), \mathrm{Ber}(p(\eps_2)))\nonumber \\
&=\Chi(
\mathrm{Ber}((1-\eps_1) F_0(\delta) +\eps_1 \bar{F}_{\bP}(\delta)), 
\mathrm{Ber}((1-\eps_2) F_0(\delta) +\eps_2 \bar{F}_{\bP}(\delta))
) \nonumber \\
& \le
\Chi((1-\eps_1) F_0 +\eps_1 \bar{F}_{\bP}, (1-\eps_2) F_0 +\eps_2 \bar{F}_{\bP}).
\end{align}
Note that, by definition, it follows that:
\begin{gather}
\Chi(\mathrm{Ber}(p(\eps_1)), \mathrm{Ber}(p(\eps_2))) = \frac{[p(\eps_1)-p(\eps_2)]^2}{p(\eps_2)(1-p(\eps_2))}
= |\eps_1-\eps_2|^2\cdot \frac{[F_0(\delta) -\bar{F}_{\bP}(\delta)]^2 }{\bar{F}(\delta)(1-\bar{F}(\delta))}, \label{eq:help2} \\
\Chi((1-\eps_1) F_0 +\eps_1 \bar{F}_{\bP}, (1-\eps_2) F_0 +\eps_2\bar{F}_{\bP}) = |\eps_1-\eps_2|^2  \cdot 
\int \frac{(\rd F_0 - \rd \bar{F}_{\bP})^2}{(1-\eps)\rd F_0 + \eps \rd \bar{F}_{\bP}}. \label{eq:help3}
\end{gather}
By combining \eqref{eq:help1}, \eqref{eq:help2}, and \eqref{eq:help3}, we have that
\[
\frac{1}{[\tau_n^\star]^2} = \int \frac{(\rd F_0 - \rd \bar{F}_{\bP})^2}{(1-\eps)\rd F_0 + \eps \rd \bar{F}_{\bP}} \ge 
\frac{[F_0(\delta) -\bar{F}_{\bP}(\delta)]^2 }{\bar{F}(\delta)(1-\bar{F}(\delta))} = \frac{1}{[\sigma_n^\star]^2}
~~\implies~~ \sigma_n^\star \ge \tau_n^\star.
\]
\end{proof}

\subsection{Proof of Theorem \ref{thm:failure-inherent-bias}}

\begin{proof}[Proof of Theorem \ref{thm:failure-inherent-bias}]

For clarity, let $(\widehat{\eps}_{\lambda}, \widehat{\mu}_{\lambda})$ denote the solution to the $L_2$-regularized maximum likelihood estimation (MLE) with regularization parameter $\lambda > 0$. 

By definition, the $L_2$-regularized MLE is given by:
\[
(\widehat{\eps}_{\lambda}, \widehat{\mu}_{\lambda}) = \arg\min_{\eps, \mu} \left\{ -\log L(\eps, \mu) + \lambda (\mu^2 + \eps^2) \right\},
\]
where
\[
L(\eps, \mu) = [(1 - \eps) \gamma + \eps \mu]^{\widehat{e}} \left[1 - (1 - \eps) \gamma - \eps \mu\right]^{1 - \widehat{e}}.
\]

The solution $(\widehat{\eps}_{\lambda}, \widehat{\mu}_{\lambda})$ has several key properties:
\begin{enumerate}[label=$(\alph*)$]
\item \textbf{Existence and uniqueness}: The solution exists and is unique due to the strong convexity introduced by the $L_2$-regularization term.

\item \textbf{Domain restriction}:  The condition $(1-\widehat{\eps}_{\lambda}) \gamma + \widehat{\eps}_{\lambda} \widehat{\mu}_{\lambda} \in (0,1)$ must hold; otherwise, the optimization problem is ill-defined.

\item \textbf{First-order conditions}: The solution $(\widehat{\eps}_{\lambda}, \widehat{\mu}_{\lambda})$ satisfies the following first-order conditions: 
\begin{subequations}
\begin{align}
2\lambda \eps &= (\mu - \gamma) \left[ \frac{\widehat{e}}{(1 - \eps) \gamma + \eps \mu} - \frac{1 - \widehat{e}}{1 - (1 - \eps) \gamma - \eps \mu} \right], \label{eq:first-order-cond1} \\
2\lambda \mu &= \eps \left[ \frac{\widehat{e}}{(1 - \eps) \gamma + \eps \mu} - \frac{1 - \widehat{e}}{1 - (1 - \eps) \gamma - \eps \mu} \right]. \label{eq:first-order-cond2}
\end{align}
\end{subequations}

\item \textbf{Bound on the mixture}: We claim that $(1-\widehat{\eps}_{\lambda}) \gamma + \widehat{\eps}_{\lambda} \widehat{\mu}_{\lambda} < \widehat{e}$. To see this, suppose the equality $(1-\widehat{\eps}_{\lambda}) \gamma + \widehat{\eps}_{\lambda} \widehat{\mu}_{\lambda} = \widehat{e}$ holds. Then the first-order conditions imply $\widehat{\eps}_{\lambda} = \widehat{\mu}_{\lambda} = 0$, which would force $\widehat{e} = \gamma$. This is impossible when $n$ is sufficiently large because $\widehat{e}$ converges almost surely to $(1-\eps) \gamma + \eps \mu > \gamma$, as $\mu > \gamma$ and $1 > \eps > 0$. Furthermore, by \eqref{eq:first-order-cond2}, the condition $(1-\widehat{\eps}_{\lambda}) \gamma + \widehat{\eps}_{\lambda} \widehat{\mu}_{\lambda} > \widehat{e}$ would imply $\widehat{\mu}_{\lambda} \widehat{\eps}_{\lambda} < 0$, which is a contradiction.

\item \textbf{Limits as $\lambda \to 0$}:
The following limits exist:
\begin{equation}
\label{eq:limit-exist}
\widehat{\eps} = \lim_{\lambda \to 0} \widehat{\eps}_{\lambda}, \quad \widehat{\mu} = \lim_{\lambda \to 0} \widehat{\mu}_{\lambda}.
\end{equation}
This will be proven at the end of the proof.
\end{enumerate}

By dividing \eqref{eq:first-order-cond1} by \eqref{eq:first-order-cond2}, we obtain
\[
\frac{\widehat{\eps}_{\lambda}}{\widehat{\mu}_{\lambda}} = \frac{\widehat{\mu}_{\lambda} - \gamma}{\widehat{\eps}_{\lambda}},
\]
which simplifies to $\widehat{\eps}_{\lambda}^2 = \widehat{\mu}_{\lambda} (\widehat{\mu}_{\lambda} - \gamma)$. Taking the limit as $\lambda \to 0$ and using \eqref{eq:limit-exist}, we have
\begin{equation}
\label{eq:first-equation}
\widehat{\eps}^2 = \widehat{\mu} (\widehat{\mu} - \gamma).
\end{equation}

We now claim that in the limit $\lambda \to 0$,
\begin{equation}
\label{eq:second-equation}
(1 - \widehat{\eps}) \gamma + \widehat{\eps} \widehat{\mu} = \widehat{e}.
\end{equation}
This follows from the fact that $\widehat{e} = \arg\max_{x \in (0, 1)} x^{\widehat{e}} (1-x)^{1-\widehat{e}}$. Thus, any $(\eps, \mu)$ maximizing $L(\eps, \mu)$ when $\lambda = 0$ must satisfy \eqref{eq:second-equation}.

Finally, we argue that $\widehat{\eps} > 0$ and $\widehat{\mu} > \gamma$ as long as $\eps > 0$. If this were not true, then for sufficiently large $n$, the strong law of large numbers implies $\widehat{e} \overset{a.s.}{\to} (1-\eps) \gamma + \eps \mu$, which would contradict the inequality that $\widehat{e} > \gamma$ from \eqref{eq:second-equation} when $\widehat{\eps} = 0$.

Combining \eqref{eq:first-equation} and \eqref{eq:second-equation}, and letting $x = \sqrt{\widehat{\mu} - \gamma}$ (recall that $\widehat{\mu} > \gamma$ for $\eps > 0$), we find:
\[
\widehat{\eps} = x \sqrt{x^2 + \gamma}, \quad \widehat{\mu} = x^2 + \gamma,
\]
where $x > 0$ satisfies
\[
x^3 \sqrt{x^2 + \gamma} = \widehat{e} - \gamma.
\]
\end{proof}

\begin{proof}[Proof of the existence of limits in \eqref{eq:limit-exist}.]
Rearranging \eqref{eq:first-order-cond1}, we obtain:
\begin{equation}
\label{eq:help}
(\widehat{\mu}_{\lambda} - \gamma)
[\widehat{e} - (1 - \widehat{\eps}_{\lambda}) \gamma - \widehat{\eps}_{\lambda} \widehat{\mu}_{\lambda}] 
= 2\lambda \widehat{\eps}_{\lambda} \cdot [(1 - \widehat{\eps}_{\lambda}) \gamma + \widehat{\eps}_{\lambda} \widehat{\mu}_{\lambda}] [1 - (1 - \widehat{\eps}_{\lambda}) \gamma - \widehat{\eps}_{\lambda} \widehat{\mu}_{\lambda}].
\end{equation}

Combining \eqref{eq:first-equation} and \eqref{eq:help}, and defining $x_{\lambda} = \sqrt{\widehat{\mu}_{\lambda} - \gamma}$, we have:
\[
\widehat{\eps}_{\lambda} = x_{\lambda} \sqrt{x_{\lambda}^2 + \gamma}, \quad \widehat{\mu}_{\lambda} = x_{\lambda}^2 + \gamma,
\]
and
\[
x_{\lambda}^5 \sqrt{x_{\lambda}^2 + \gamma} \leq (\widehat{e} - \gamma) \cdot x_{\lambda}^2.
\]

The inequality above implies that $x_{\lambda} > 0$ is uniformly bounded for any $\lambda$. Consequently, as a function of $x_{\lambda}$, both $\widehat{\eps}_{\lambda}$ and $\widehat{\mu}_{\lambda}$ are uniformly bounded for all $\lambda$. 
As $\lambda \to 0$, we can extract a subsequence for which $\widehat{\eps}_{\lambda}$ and $\widehat{\mu}_{\lambda}$ converge to well-defined limits. For this particular subsequence, taking the limit $\lambda \to 0$ in \eqref{eq:help} yields:
\[
\lim_{\lambda \to 0} x_{\lambda}^5 \sqrt{x_{\lambda}^2 + \gamma} - (\widehat{e} - \gamma) \cdot x_{\lambda}^2 = 0.
\]

Since $x_{\lambda} > 0$, the above equation has only one nonzero solution. Therefore, the subsequential limit of $x_{\lambda}$ is unique. Consequently, all subsequences of $\widehat{\eps}_{\lambda}$ and $\widehat{\mu}_{\lambda}$ converge to the same limits. 

\end{proof}

\section{Simulation Study: Details and Results}
\label{appen:simulation}
\subsection{Further Details}

\paragraph{Evaluation of the fixed-point equation.}
Recall that we need to solve the following fixed-point equation
\begin{equation}
\tag{\ref{eq:emprical-operator}}
\eps = \left[  \frac{\int \hvopt(\eps, x) \left[\rd F_0(x) - \rd \widehat{F}(x)\right]}{\int \hvopt(\eps, x) \left[\rd F_0(x) - \rd \widehat{F}_{\bP}(x)\right]}\right]_{[\eps_{\min}, 1-\eps_{\min}]}
\end{equation}
where $\hvopt$ is the estimated optimal weight function defined by:  
\[
\hvopt(\eps, x) = \frac{1 - \widehat{g}(x)}{(1-\eps) + \eps \widehat{g}(x)}, 
~~ \text{and} ~~ 
\widehat{g}(x) = \frac{\rd \widehat{F}_{\bP}(x)}{\rd F_0(x)}.
\] 
We use the notation $[a]_{[b, c]} = \max\{\min\{a, c\}, b\}$ to denote the projection of $a$ onto the interval $[b, c]$, ensuring that the result lies within the bounds of $b$ and $c$.

In our implementation, we first sample $n = 10^6$ values, denoted as $Z_1, \ldots, Z_n$, from $F_0$ (which is $\UM(0, 1)$ due to the probability integral transform) and approximate $\int \hvopt(\eps, x) \rd F_0(x)$ using the average over these samples: $\frac{1}{n}\sum_{t=1}^n \hvopt(\eps, Z_t)$. 
We denote the empirical sum by $\int \hvopt(\eps, x) \rd \widehat{F}_0(x)$.
Similarly, we collect the same number of pivotal statistics from $\bar{F}$ or $\bar{F}_{\bP}$ and compute $\int \hvopt(\eps, x) \rd \widehat{F}(x)$ or $\int \hvopt(\eps, x) \rd \widehat{F}_{\bP}(x)$ as their respective averages. 


\paragraph{Optimization for the refined estimator.}
To obtain the refined estimator $\epsopt$, we use numerical methods to solve the following minimization problem:
\[
\epsopt = \arg\min_{\eps \in [\eps_{\min}, 1-\eps_{\min}]} \LM(\eps),
\]
where $\eps_{\min}=10^{-3}$ is a small constant to avoid numerical issue,
\begin{equation}
\label{eq:loss-for-refined-estimator}
\LM(\eps) = \left|\eps -  \widehat{\TM}_{\mathrm{used}}(\eps) \right|
~~\text{and}~~
\widehat{\TM}_{\mathrm{used}}(\eps) = 
\left[ 
\frac{\int \hvopt(\eps, x) \left[\rd \widehat{F}_0(x) - \rd \widehat{F}(x)\right]}{\int \hvopt(\eps, x) \left[\rd \widehat{F}_0(x) - \rd \widehat{F}_{\bP}(x)\right]} 
\right]_{[\eps_{\min}, 1-\eps_{\min}]}.
\end{equation}
Here, $\LM(\eps)$ represents the deviation between $\eps$ and the estimated operator $\widehat{\TM}_{\mathrm{used}}(\eps)$, which is computed using the optimal weight function $\hvopt(\eps, x)$. The terms $\widehat{F}_0(x)$, $\widehat{F}(x)$, and $\widehat{F}_{\bP}(x)$ denote the CDFs of the null, observed, and watermarked samples, respectively.

\begin{remark}
The only difference between $\widehat{\TM}_{\mathrm{used}}$ and $\widehat{\TM}$ (defined in \eqref{eq:emprical-operator}) is that the former approximates $\int \hvopt(\eps, x) \rd F_0(x)$ using finite samples, while the latter employs exact integration, primarily for theoretical analysis.
\end{remark}

To solve this problem, we use the ``scipy.optimize.minimize'' function from the Python SciPy library \citep{virtanen2020scipy}. The objective function $\LM(\eps)$ is specified in \eqref{eq:loss-for-refined-estimator}, with bounds for $\eps$ constrained to the interval $[\eps_{\min}, 1-\eps_{\min}]$. We employ the default gradient-free optimization method, ``L-BFGS-B'', to ensure robust and efficient convergence. 
To identify a suitable initial point for the optimization, we iteratively compose the operator $\widehat{\TM}_{\mathrm{used}}$, starting from $0.9$. Specifically, we compute $\widehat{\TM}_{\mathrm{used}} \circ \cdots \circ \widehat{\TM}_{\mathrm{used}}(0.9)$ with 20 compositions, using the resulting value as the starting point.

\paragraph{Construction of NTP distributions.}  
We adopt a similar approach to \citep{li2024optimal} for constructing NTP distributions, as detailed in Algorithm \ref{alg:dominate-Ps}. The function $\text{Zipf}(a, b, k)$ generates a Zipf-like probability distribution over a specified support size $k$, where the probabilities are computed as $(i + b)^{-a}$ for $i = 1, \ldots, k$ and normalized to sum to 1.  
The NTP distributions generated by Algorithm \ref{alg:dominate-Ps} are parameterized by $\Delta$, which controls the dominance of the largest probability, ensuring it remains below $1-\Delta$ uniformly. The constructed distributions balance a dominant ``head'' component $\vh$, generated using the Zipf distribution, and a uniform ``tail'' component, ensuring the entire distribution sums to 1. For all simulation experiments, we set the vocabulary size to $|\Voca| = 10^3$.

\paragraph{Failure on green-red list watermarks.}
As suggested by the original work \citep{kirchenbauer2023watermark}, we set the parameters $(\gamma, \delta) = (0.5, 2)$ for the green-red list watermarks, which strike a balance between watermark signal strength and data distortion. 
As discussed in Sections \ref{sec:accuray-init} and \ref{sec:accuray-refinement}, our estimators \textsf{INI} and \textsf{IND} are highly biased when NTP distributions $\bP_{1:n}$ are inaccessible, likely producing estimates outside the feasible domain $[0, 1]$. To ensure meaningful comparison in Figure \ref{fig:simulation-MAE-greenredlist}, we plot the errors of projected estimators $[\epsinital(\delta)]_{[0, 1]}$ and $[\epsrefine]_{[0, 1]}$, which guarantees that MAEs do not exceed 1. 
One can easily show that the accuracy analysis for MAEs in Theorem \ref{thm:upper-bound-eps-inital} and \ref{thm:accuracyRfn} still applies due to the nonexpansiveness of the projection operator.

\begin{algorithm}[t!]
   \caption{Generate NTP distributions}
   \label{alg:dominate-Ps}
\begin{algorithmic}
   \STATE {\bfseries Input:} Dominance parameter $\Delta$, vocabulary size $|\Voca|$.

   \STATE Sample parameters $a \sim \text{Uniform}(0.95, 1.5)$ and $b \sim \text{Uniform}(0.01, 0.1)$.
   \STATE Sample the support size $k \sim \text{DiscreteUniform}[5, 15]$.
   \STATE Generate a Zipf-like head vector: $\vh = \text{Zipf}(a, b, k)$, where $\vh \in \RB^k$ is a probability vector.
   \STATE Sample the random dominance parameter $\Delta_{\mathrm{used}} \sim \text{Uniform}(0, \Delta)$.
   \STATE Define the scaling factor $b = \frac{1 - \Delta_{\mathrm{used}}}{\max_{i \in [k]} h_i}$, where $\vh = (h_1, \ldots, h_k)$.
   \STATE Initialize $\bP = \mathbf{0}_{|\Voca|}$.
   \IF{$b \leq 1$} 
       \STATE Assign the first $k$ entries of $\bP$ to $b \cdot \vh$, and distribute the remaining probability uniformly:
       \[
       \bP_{1:k} = b \cdot \vh, \quad \bP_{(k+1):|\Voca|} = (1 - b) \cdot \frac{\mathbf{1}_{|\Voca| - k}}{|\Voca| - k}.
       \]
   \ELSE
       \STATE Assign the first entry of $\bP$ to $1 - \Delta_{\mathrm{used}}$, the next $k$ entries to $\frac{\Delta_{\mathrm{used}}}{2} \cdot \vh$, and distribute the remaining probability uniformly:
       \[
       P_1 = 1 - \Delta_{\mathrm{used}}, \quad P_{2:(k+1)} = \frac{\Delta_{\mathrm{used}}}{2} \cdot \vh, \quad \bP_{(k+2):|\Voca|} = \frac{\Delta_{\mathrm{used}}}{2} \cdot \frac{\mathbf{1}_{|\Voca| - k - 1}}{|\Voca| - k - 1}.
       \]
   \ENDIF
   \STATE {\bfseries Return:} A randomly permuted version of $\bP$.
\end{algorithmic}
\end{algorithm}

\subsection{Empirical CDFs of Mixture Distributions} 
\begin{figure}[htp]
\vspace{-0.1in}
\centering
\begin{subfigure}{0.495\textwidth}
\includegraphics[width=\textwidth]{figs/gumbel_Delta0.1+_cdf.pdf}
\caption{$\Delta = 0.1$}
\end{subfigure}
\hfill
\begin{subfigure}{0.495\textwidth}
\includegraphics[width=\textwidth]{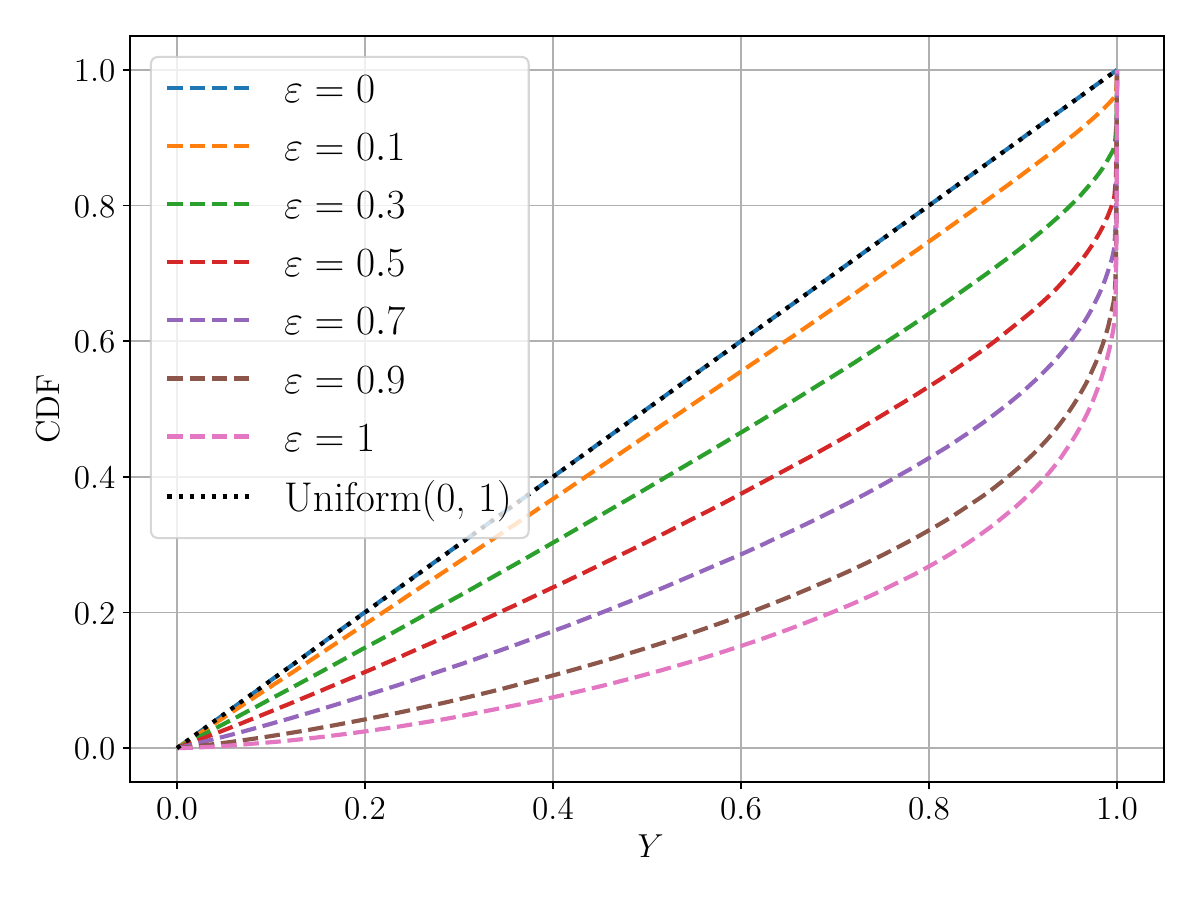}
\caption{$\Delta = 0.2$}
\end{subfigure}

\begin{subfigure}{0.495\textwidth}
\includegraphics[width=\textwidth]{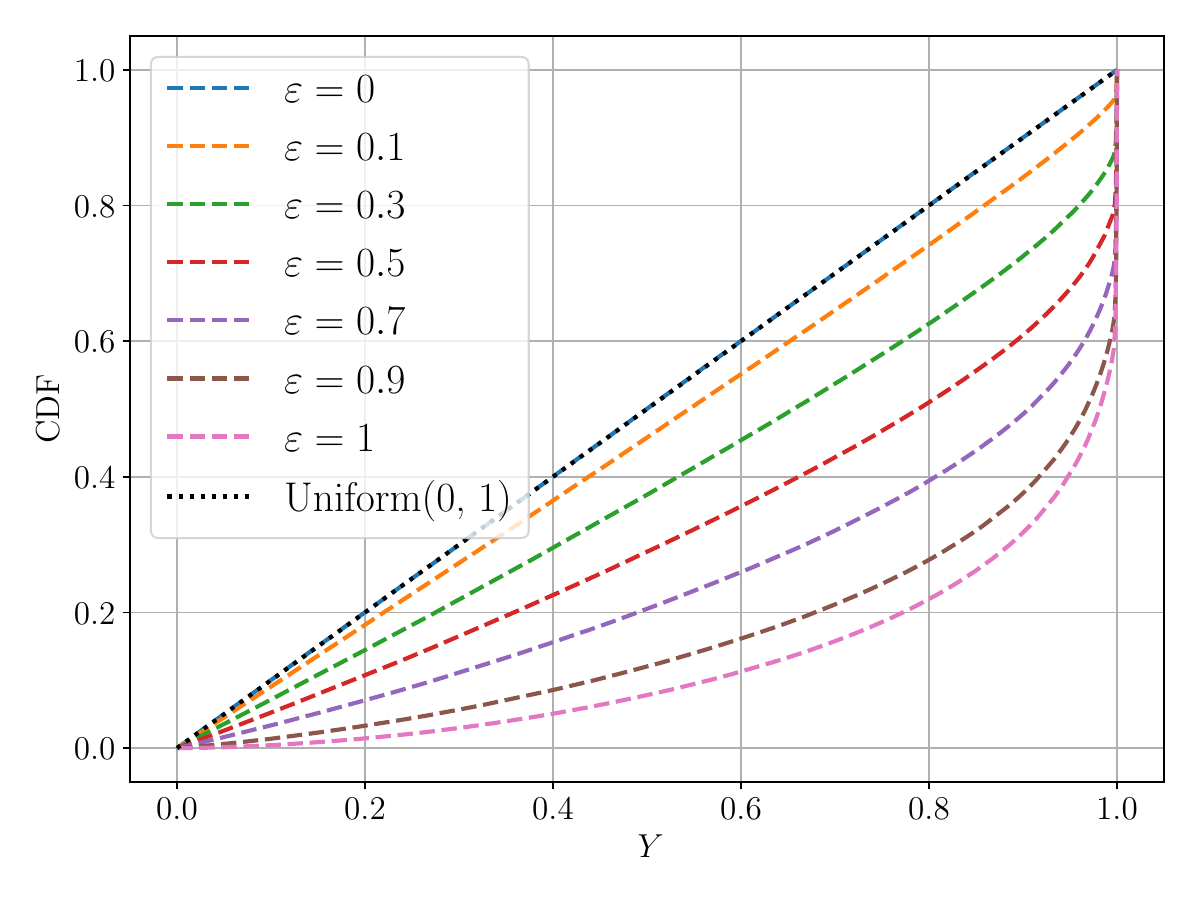}
\caption{$\Delta = 0.3$}
\end{subfigure}
\hfill
\begin{subfigure}{0.495\textwidth}
\includegraphics[width=\textwidth]{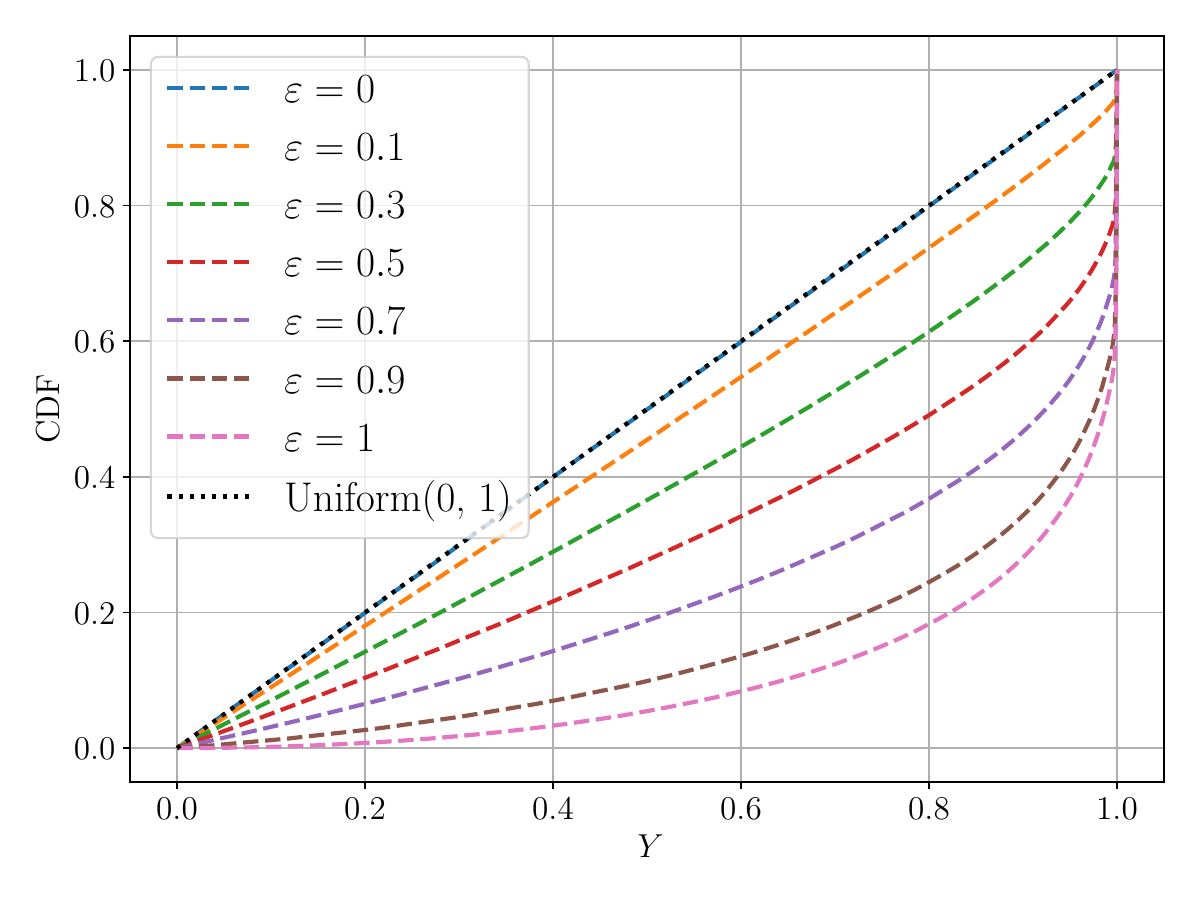}
\caption{$\Delta = 0.4$}
\end{subfigure}

\begin{subfigure}{0.495\textwidth}
\includegraphics[width=\textwidth]{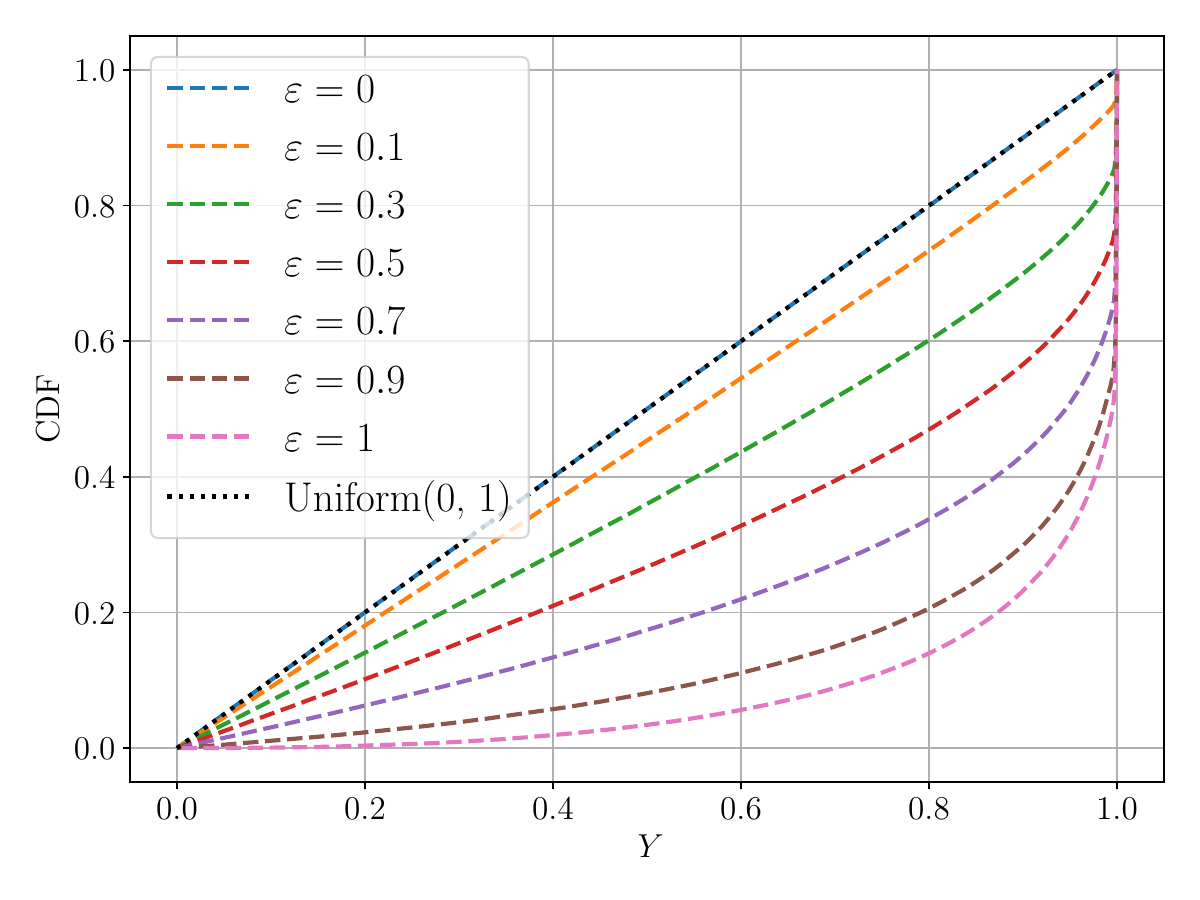}
\caption{$\Delta = 0.5$}
\end{subfigure}
\hfill
\begin{subfigure}{0.495\textwidth}
\includegraphics[width=\textwidth]{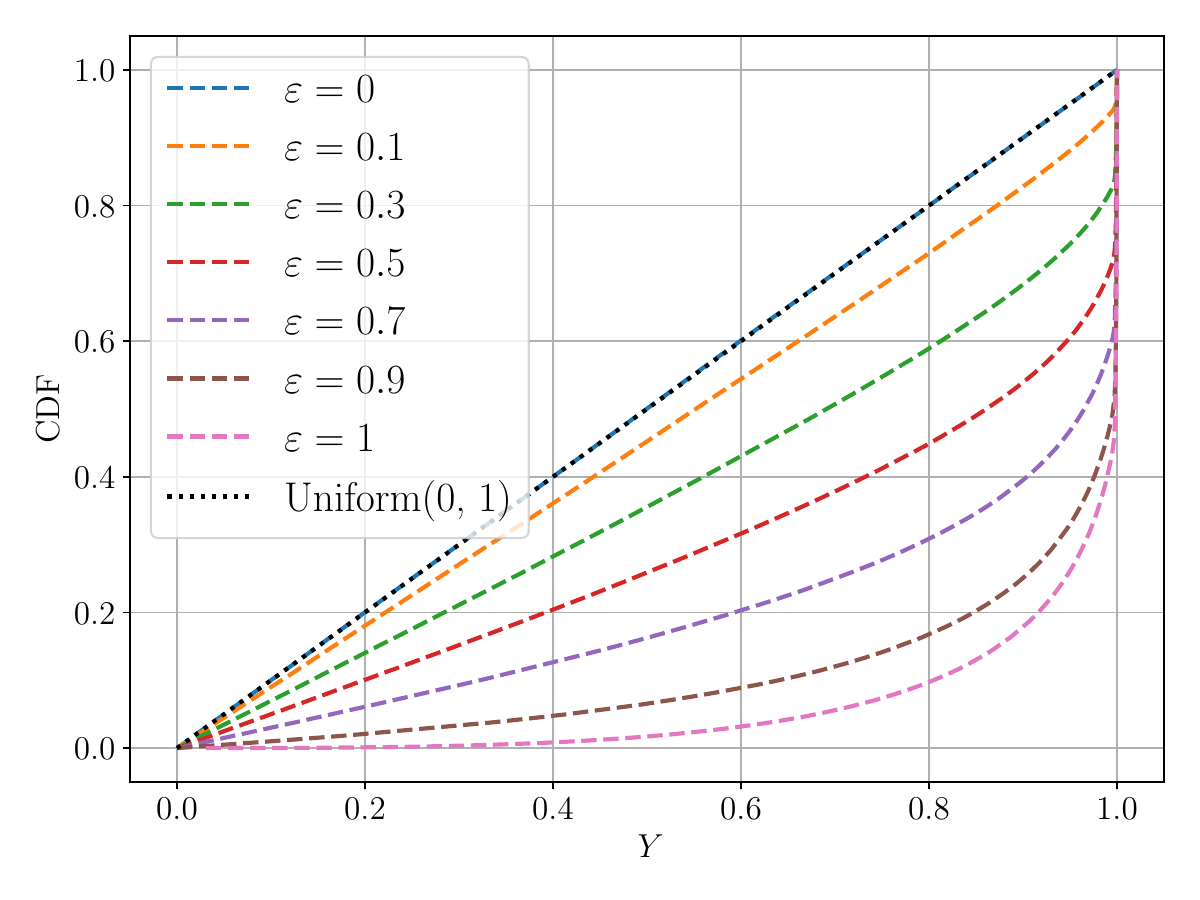}
\caption{$\Delta = 0.6$}
\end{subfigure}

\caption{CDF plots of mixture distributions for the Gumbel-max watermark.
Here, $\Delta$ represents the dominance parameter used in Algorithm \ref{alg:dominate-Ps} for generating NTP distributions, and $\eps$ denotes the true proportion.
}
\label{fig:gumbel_cdf}
\vspace{-0.1in}
\end{figure}

\begin{figure}[htp]
\vspace{-0.1in}
\centering
\begin{subfigure}{0.495\textwidth}
\includegraphics[width=\textwidth]{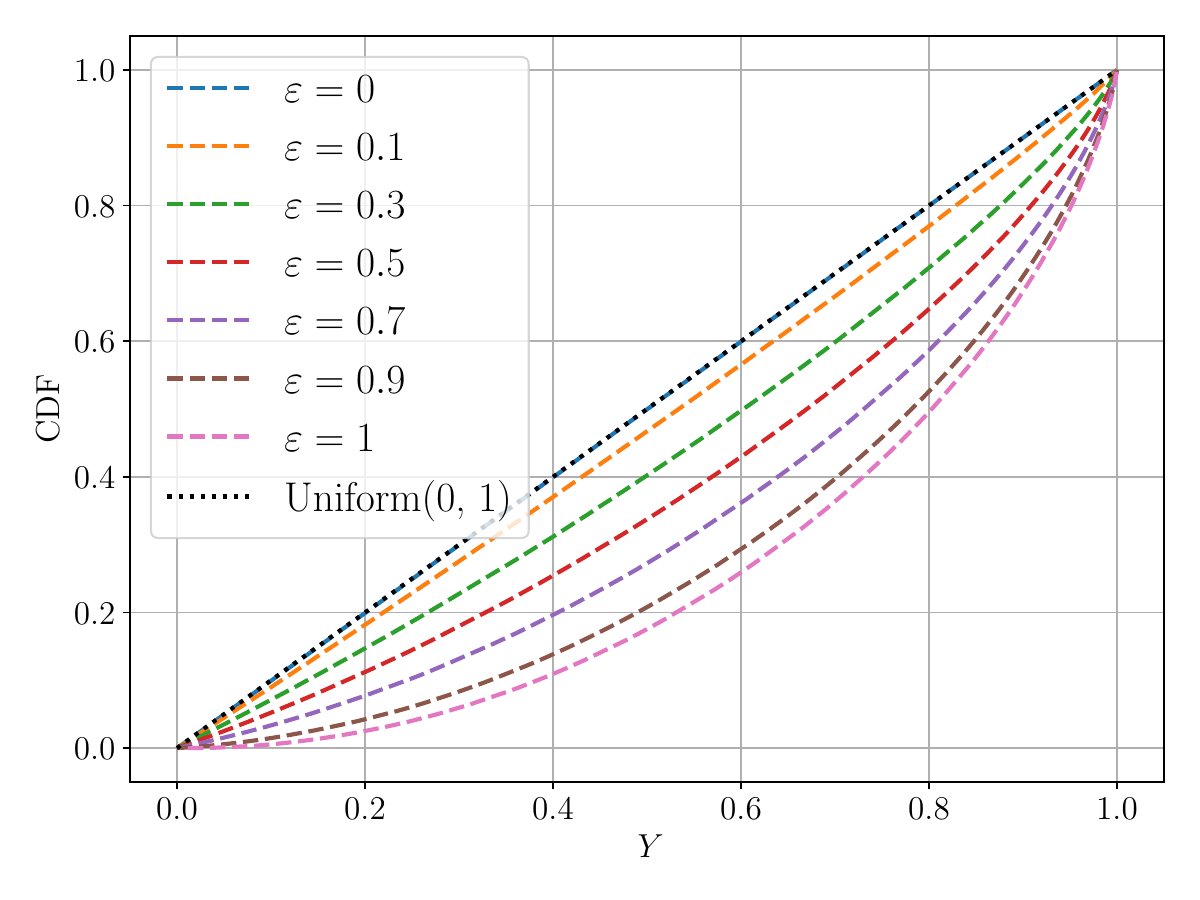}
\caption{$\Delta = 0.1$}
\end{subfigure}
\hfill
\begin{subfigure}{0.495\textwidth}
\includegraphics[width=\textwidth]{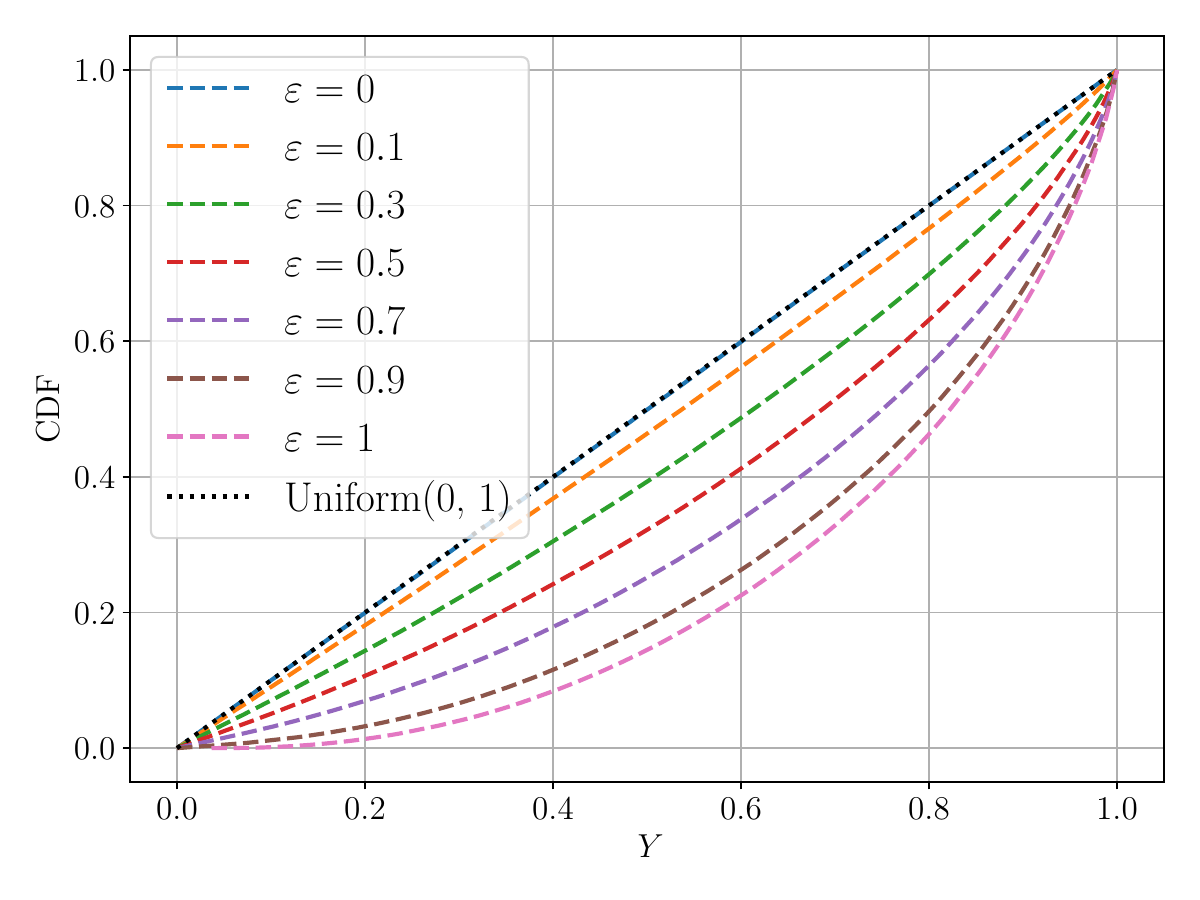}
\caption{$\Delta = 0.2$}
\end{subfigure}

\begin{subfigure}{0.495\textwidth}
\includegraphics[width=\textwidth]{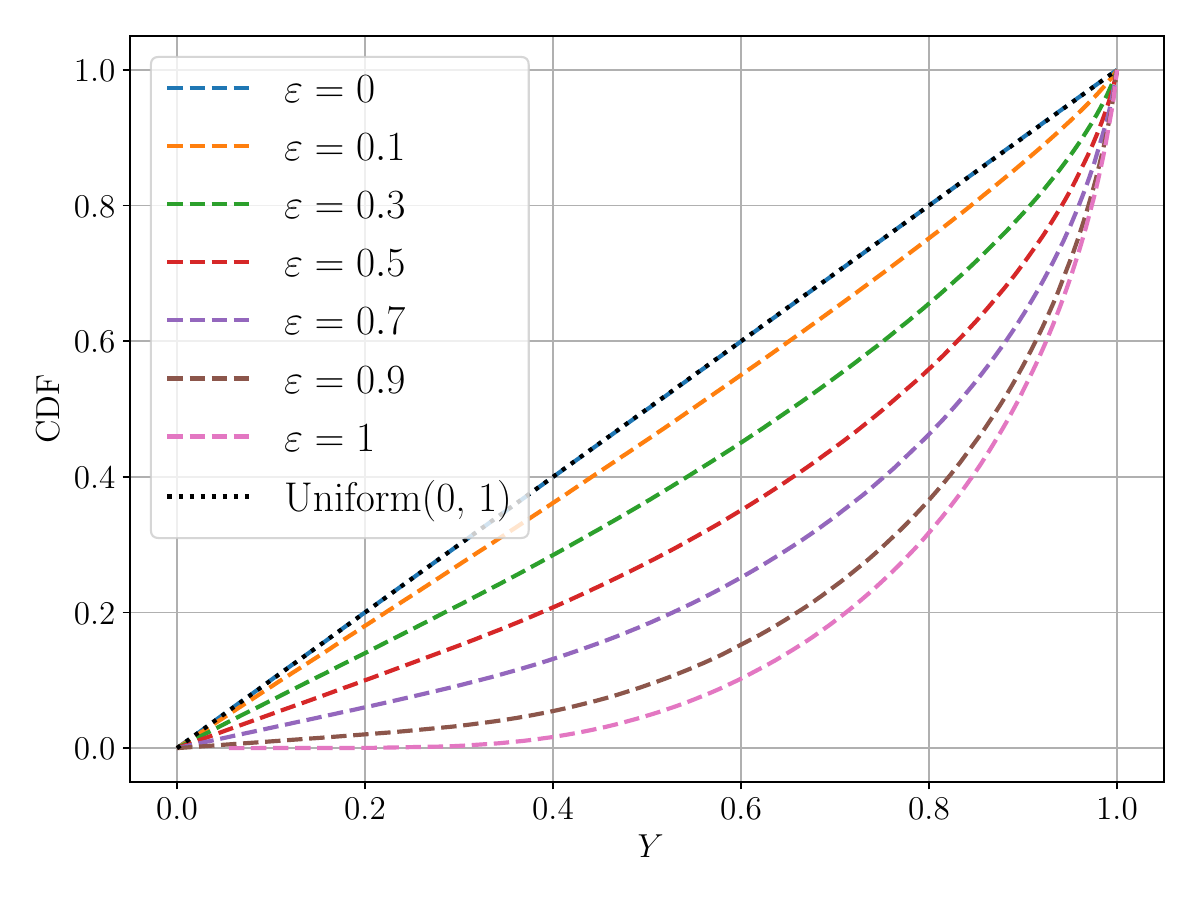}
\caption{$\Delta = 0.5$}
\end{subfigure}
\hfill
\begin{subfigure}{0.495\textwidth}
\includegraphics[width=\textwidth]{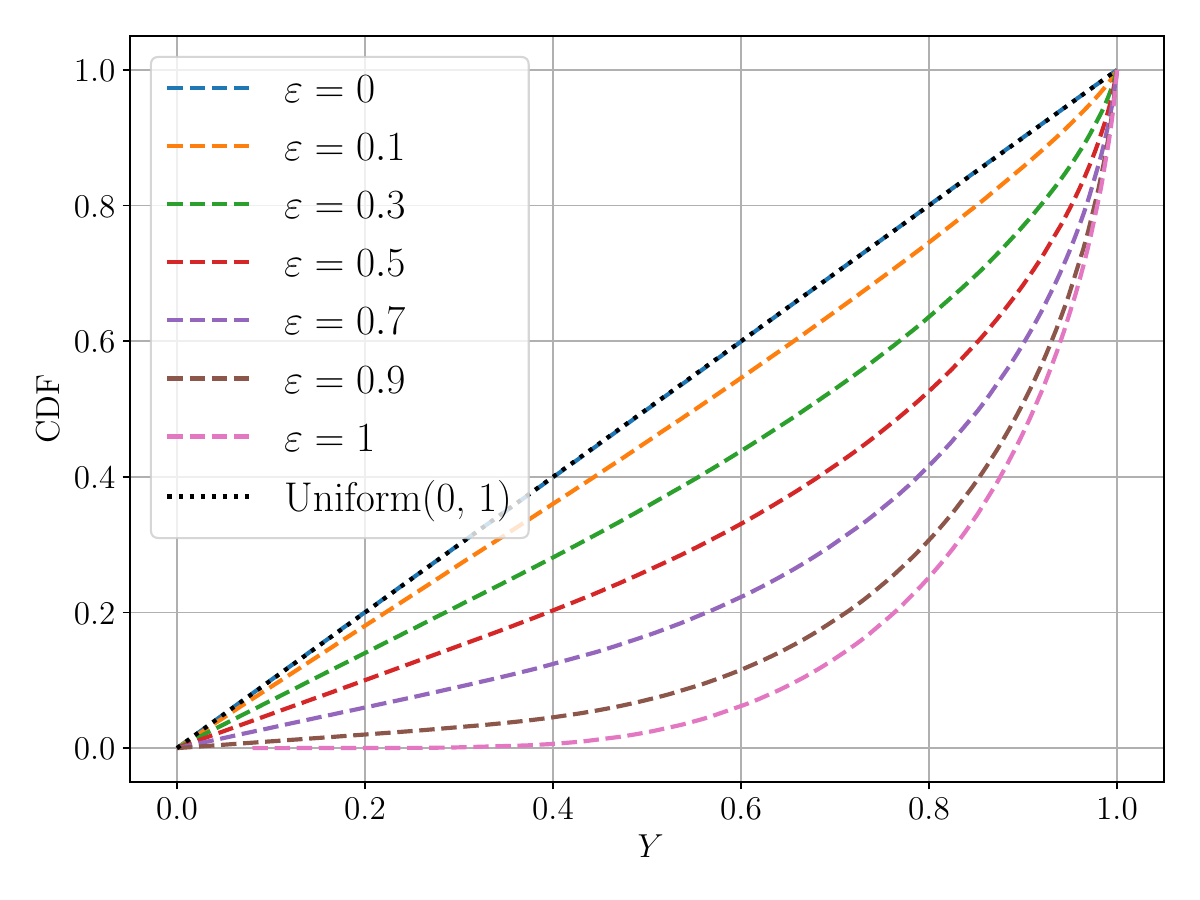}
\caption{$\Delta = 0.6$}
\end{subfigure}

\begin{subfigure}{0.495\textwidth}
\includegraphics[width=\textwidth]{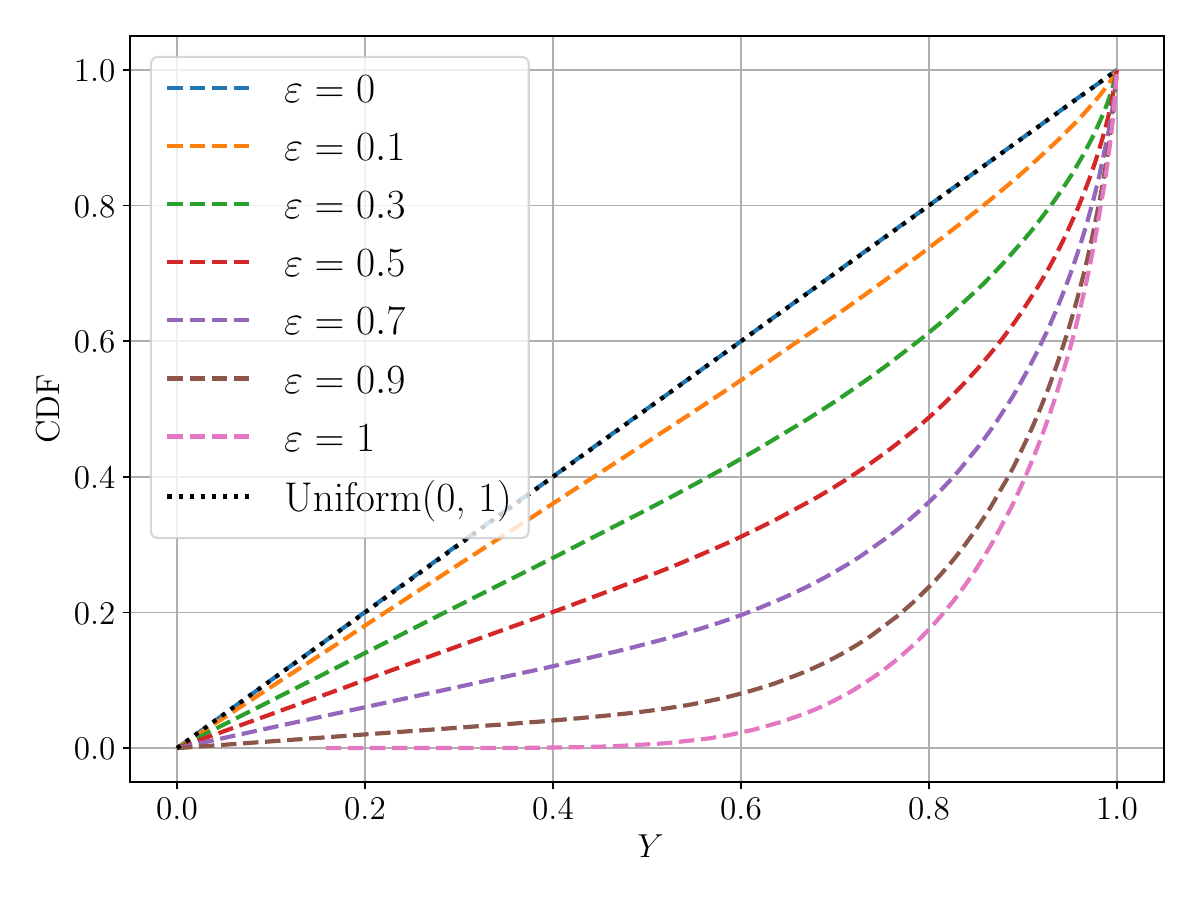}
\caption{$\Delta = 0.7$}
\end{subfigure}
\hfill
\begin{subfigure}{0.495\textwidth}
\includegraphics[width=\textwidth]{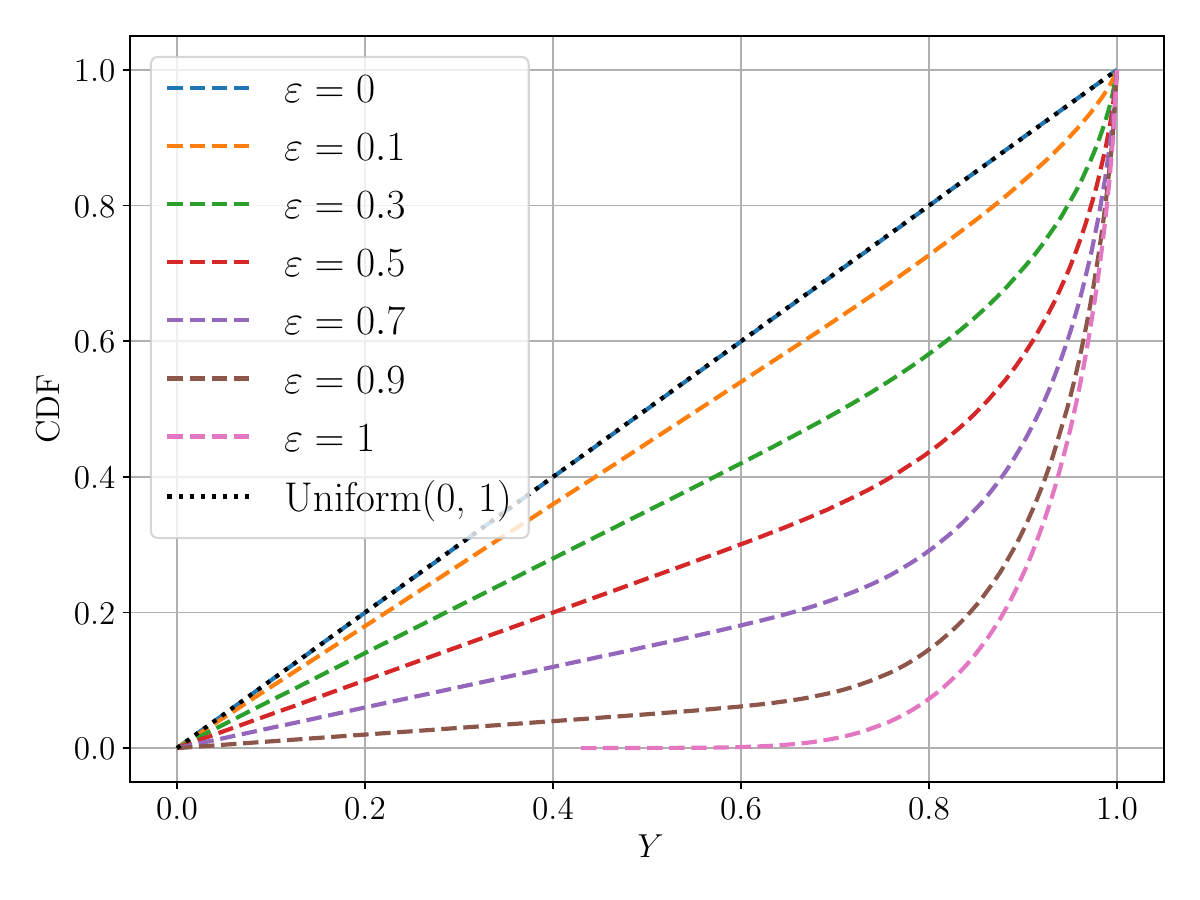}
\caption{$\Delta = 0.8$}
\end{subfigure}

\caption{CDF plots of mixture distributions for the inverse transform watermark.
Here, $\Delta$ represents the dominance parameter used in Algorithm \ref{alg:dominate-Ps} for generating NTP distributions, and $\eps$ denotes the true proportion.
}
\label{fig:inverse_cdf}
\vspace{-0.1in}
\end{figure}

Figure \ref{fig:gumbel_cdf} illustrates the CDFs of the mixture distribution $(1-\eps) F_0 + \eps \bar{F}_{\bP}$ for the Gumbel-max watermark, while Figure \ref{fig:inverse_cdf} shows the corresponding CDFs for the inverse transform watermark.
The alternative CDF, corresponding to $\eps=1$, is consistently stochastically larger than the null distribution, which corresponds to the case where $\eps=0$. As $\eps$ increases, the mixture CDF becomes progressively more right-skewed.

In constructing the NTP distributions, we ensure that each $\bP_t$ is $\Delta$-regular, meaning the largest probability in $\bP_t$ is strictly less than $1-\Delta$. Specifically, this largest probability is uniformly distributed within $(0, 1-\Delta)$. Consequently, a larger value of $\Delta$ induces a greater discrepancy between $F_0$ and $\bar{F}_{\bP}$, leading to increased right-skewness in the mixture CDF. 

\subsection{Additional Results on Estimation Accuracy}

\begin{figure}[htp]
\vspace{-0.1in}
\centering
\begin{subfigure}{0.495\textwidth}
\includegraphics[width=\textwidth]{figs/finalfig-Gumbel-K1000N2500c5key2846513T400Delta0.1+-IterFalse.pdf}
\caption{$\Delta = 0.1$}
\end{subfigure}
\hfill
\begin{subfigure}{0.495\textwidth}
\includegraphics[width=\textwidth]{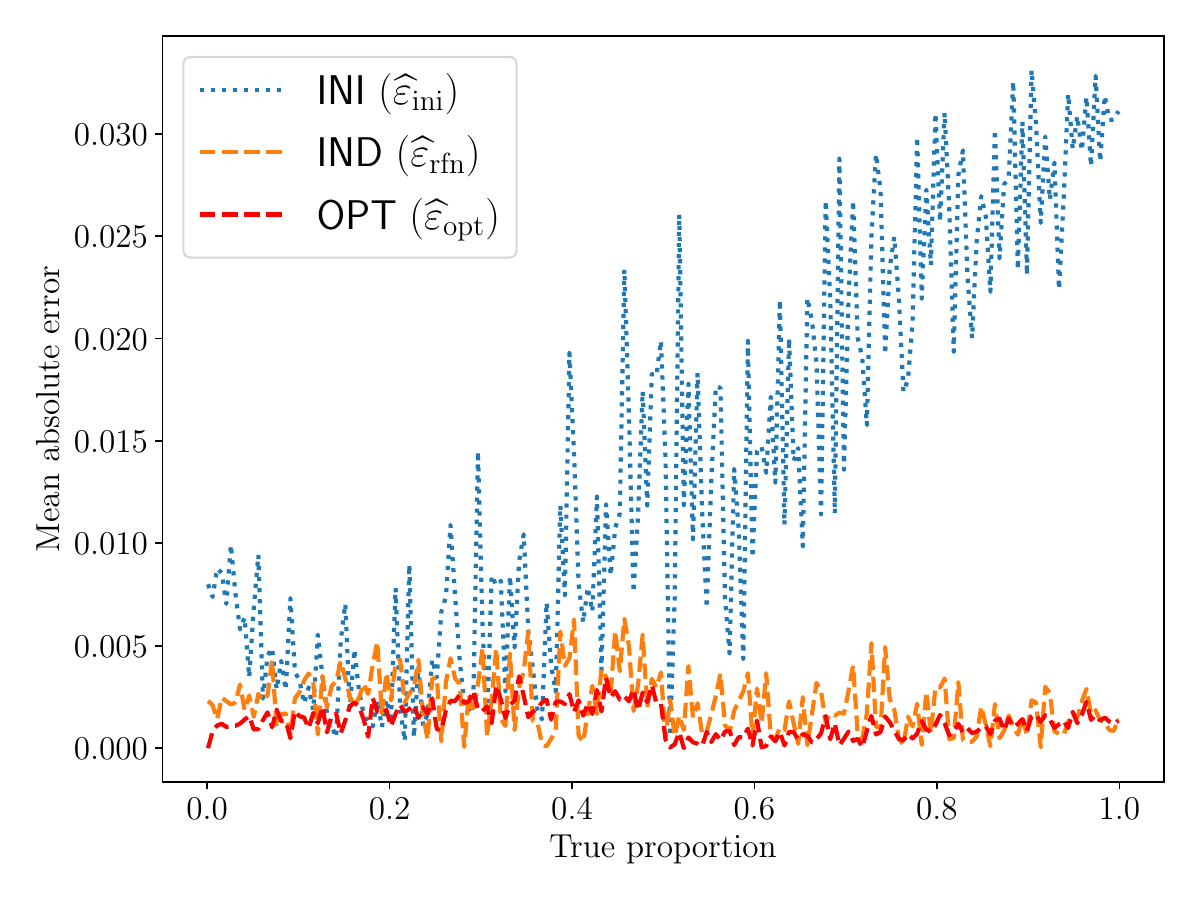}
\caption{$\Delta = 0.2$}
\end{subfigure}

\begin{subfigure}{0.495\textwidth}
\includegraphics[width=\textwidth]{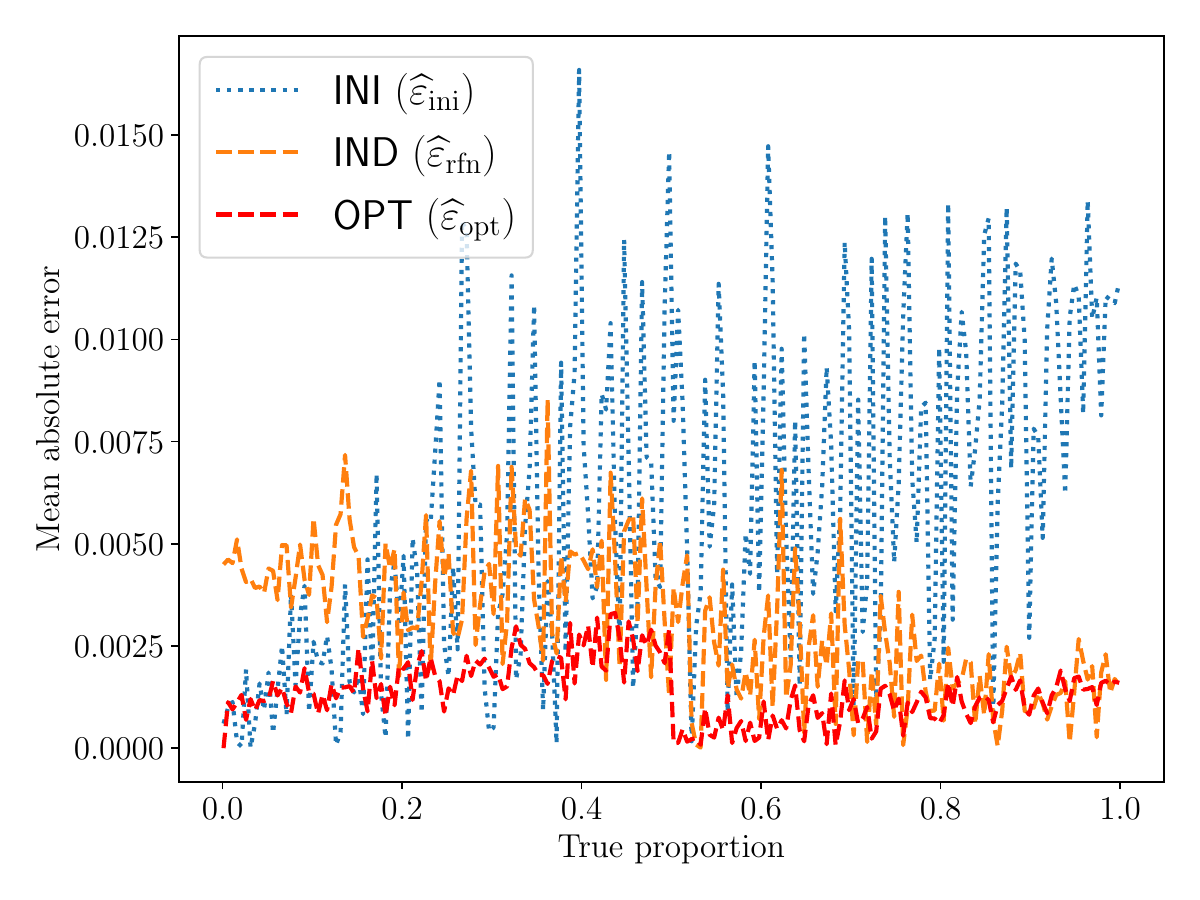}
\caption{$\Delta = 0.3$}
\end{subfigure}
\hfill
\begin{subfigure}{0.495\textwidth}
\includegraphics[width=\textwidth]{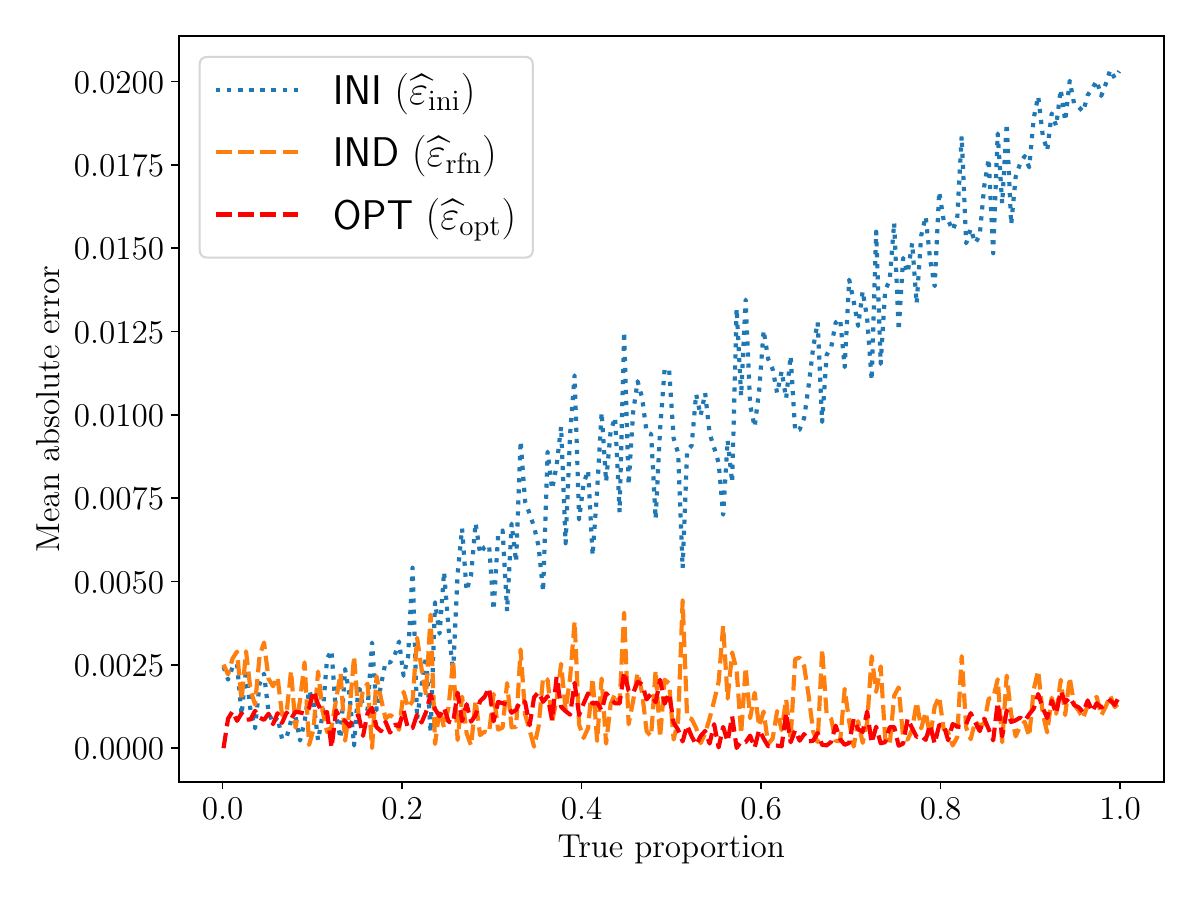}
\caption{$\Delta = 0.4$}
\end{subfigure}

\begin{subfigure}{0.495\textwidth}
\includegraphics[width=\textwidth]{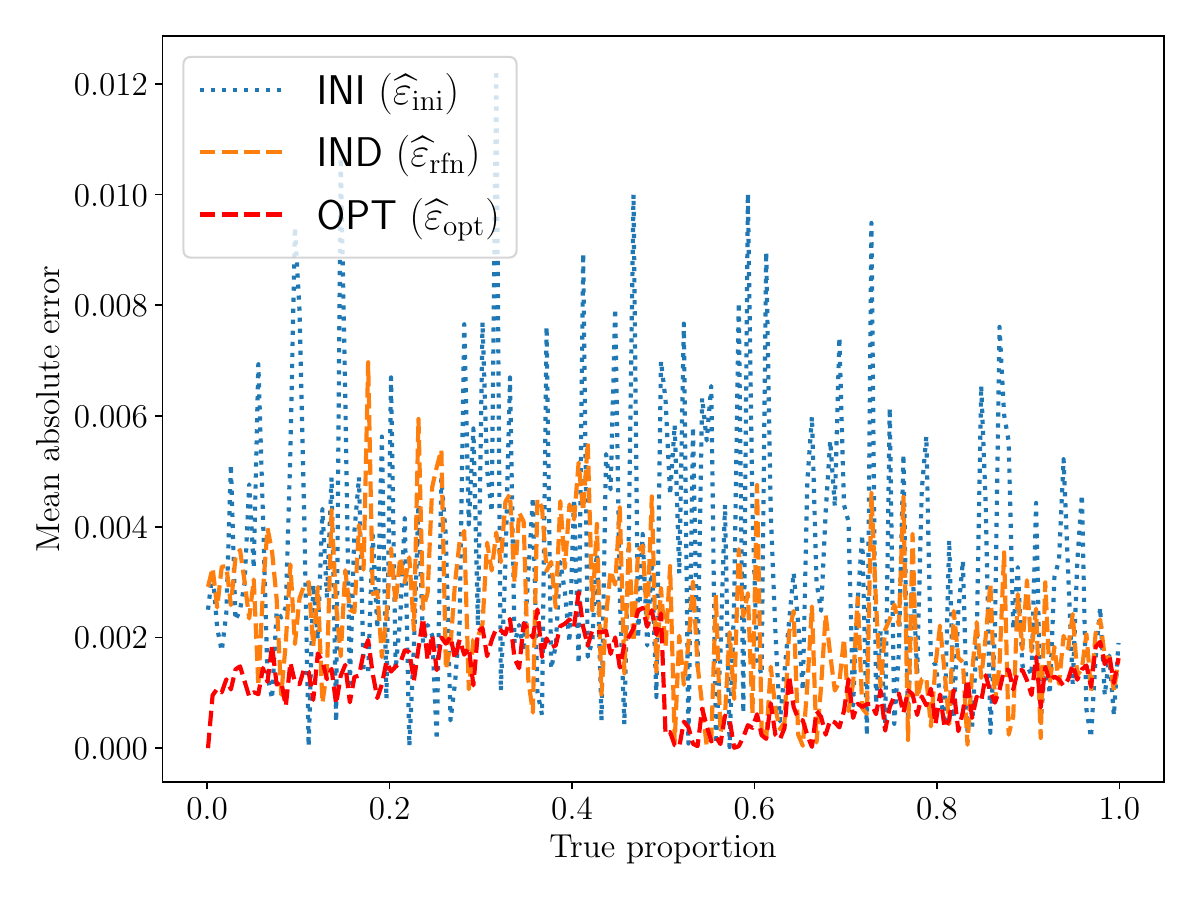}
\caption{$\Delta = 0.5$}
\end{subfigure}
\hfill
\begin{subfigure}{0.495\textwidth}
\includegraphics[width=\textwidth]{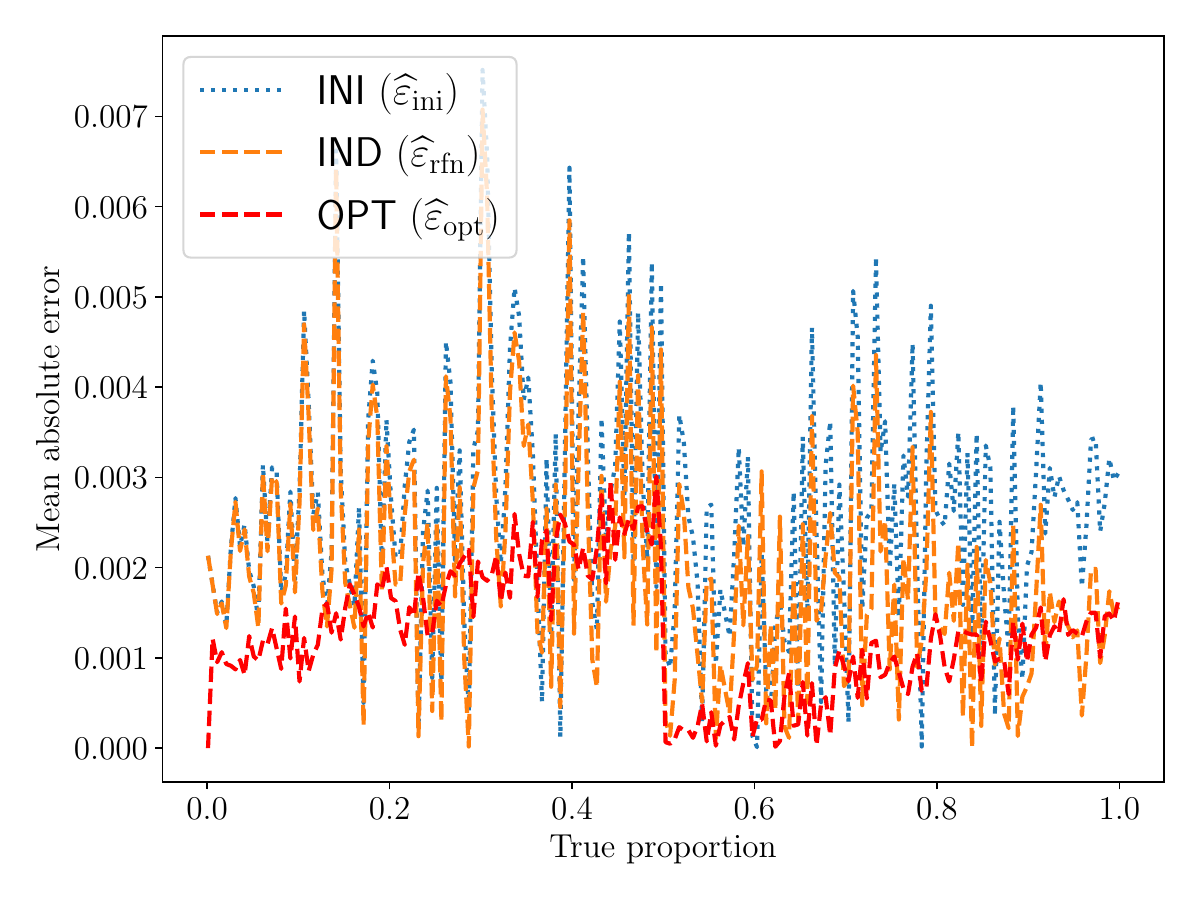}
\caption{$\Delta = 0.6$}
\vspace{-0.1in}
\end{subfigure}

\caption{Mean absolute errors (MAEs) for the Gumbel-max watermark.
	Here $\Delta$ is the dominance parameter in Algorithm \ref{alg:dominate-Ps} for the NTP distributions. The best MAEs for both \textsf{INI} $\epsinital(\delta)$ and \textsf{IND} $\epsrefine(\delta)$, are reported by searching $\delta$ over $\{10^{-1}, 10^{-2}, 10^{-3}\}$.
}
\label{fig:gumbel_MAE}
\vspace{-0.1in}
\end{figure}

\begin{figure}[htp]
\vspace{-0.1in}
\centering
\begin{subfigure}{0.495\textwidth}
\includegraphics[width=\textwidth]{figs/finalfig-Inverse-K1000N2500c5key2846513T400Delta0.1+-IterFalse.pdf}
\caption{$\Delta = 0.1$}
\end{subfigure}
\hfill
\begin{subfigure}{0.495\textwidth}
\includegraphics[width=\textwidth]{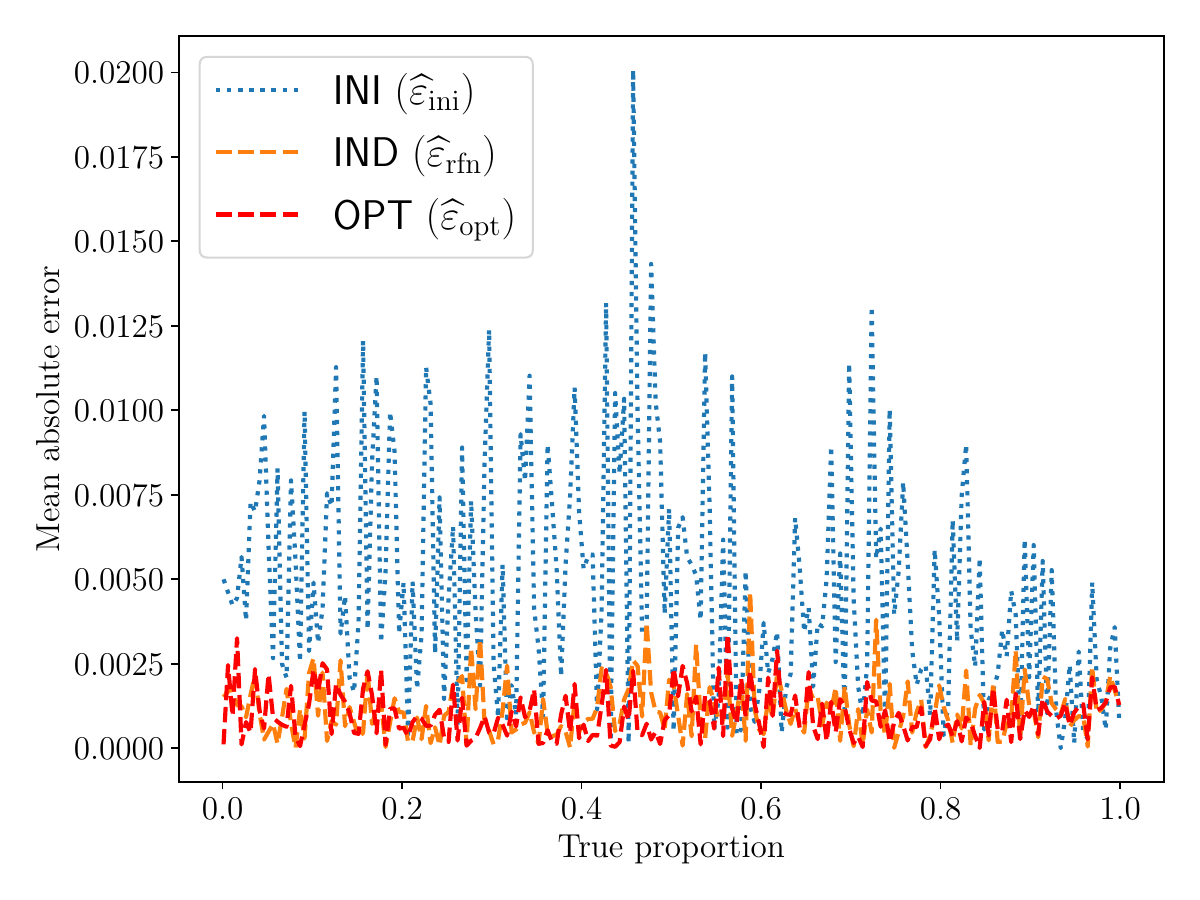}
\caption{$\Delta = 0.2$}
\end{subfigure}

\begin{subfigure}{0.495\textwidth}
\includegraphics[width=\textwidth]{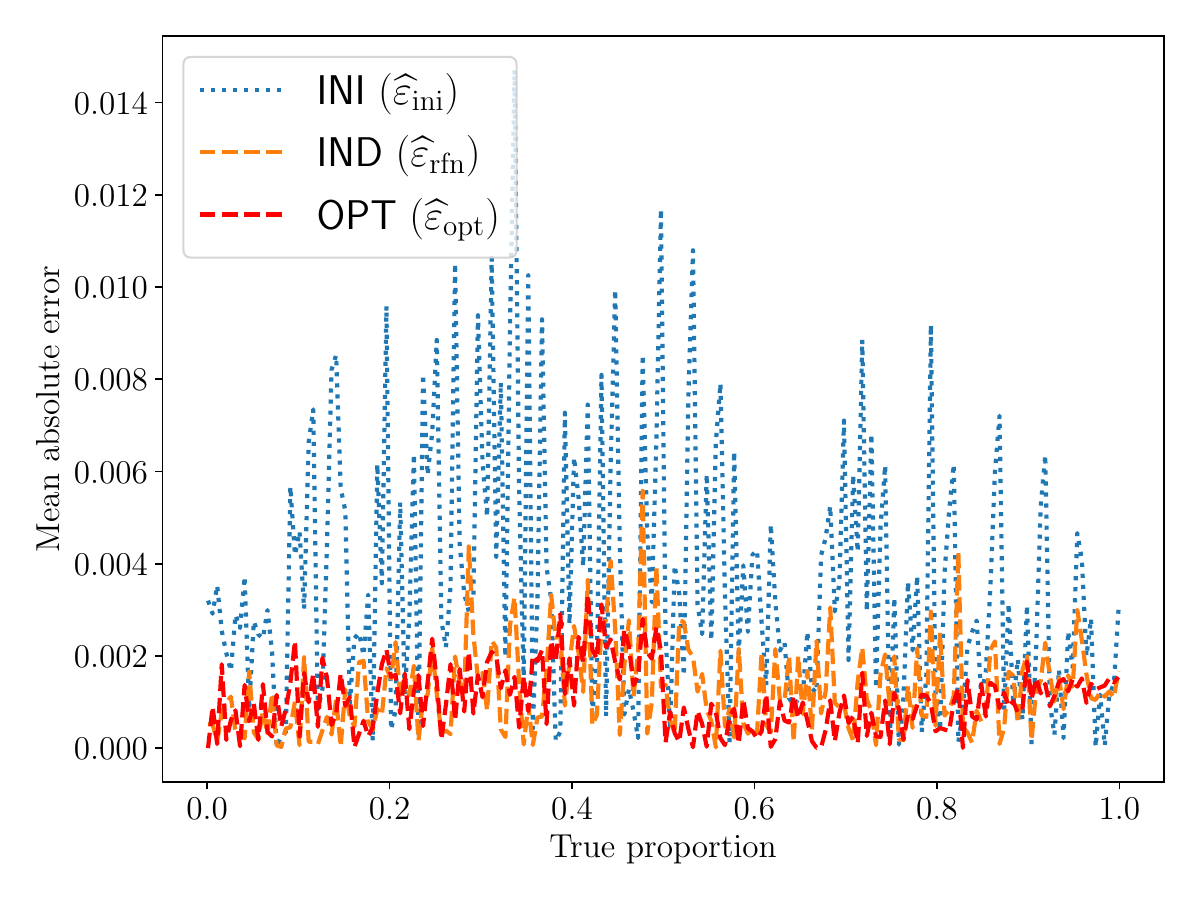}
\caption{$\Delta = 0.5$}
\end{subfigure}
\hfill
\begin{subfigure}{0.495\textwidth}
\includegraphics[width=\textwidth]{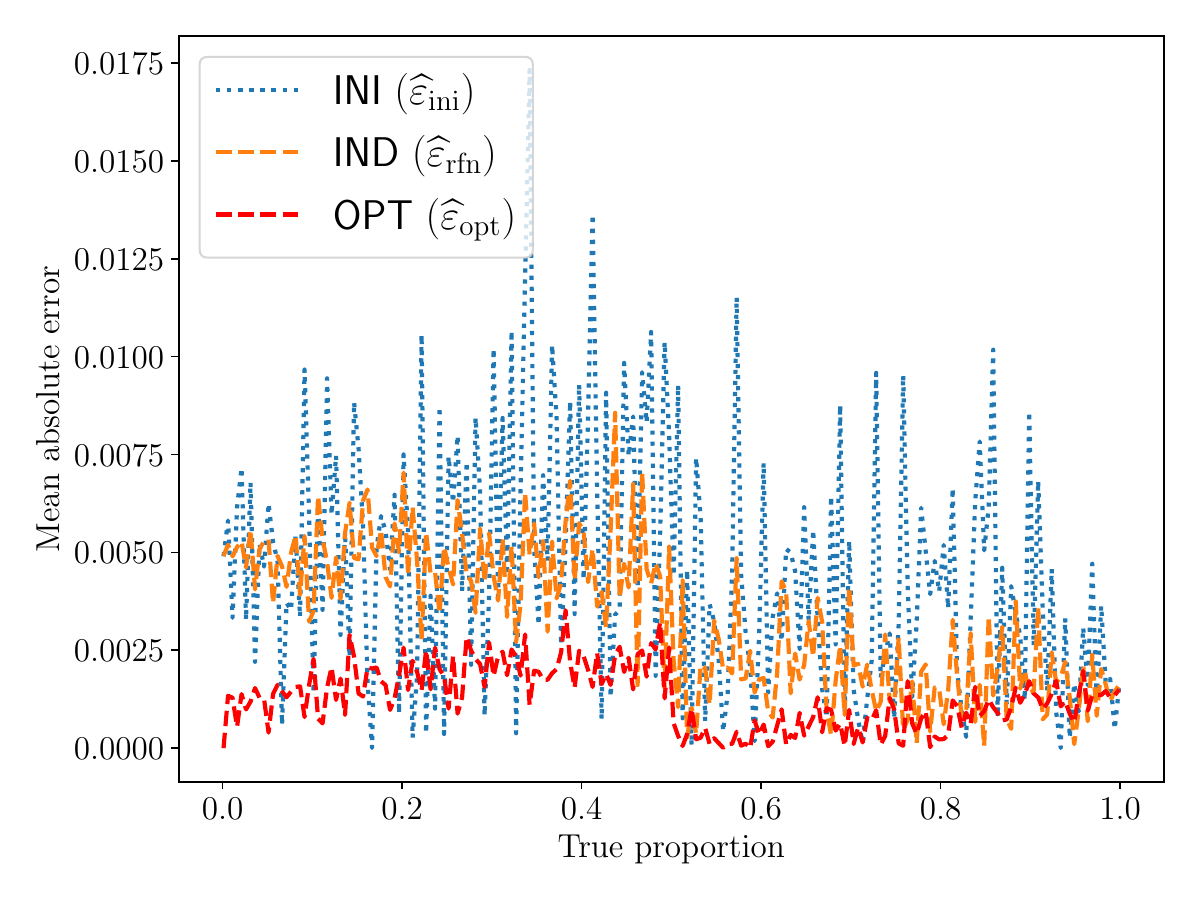}
\caption{$\Delta = 0.6$}
\end{subfigure}

\begin{subfigure}{0.495\textwidth}
\includegraphics[width=\textwidth]{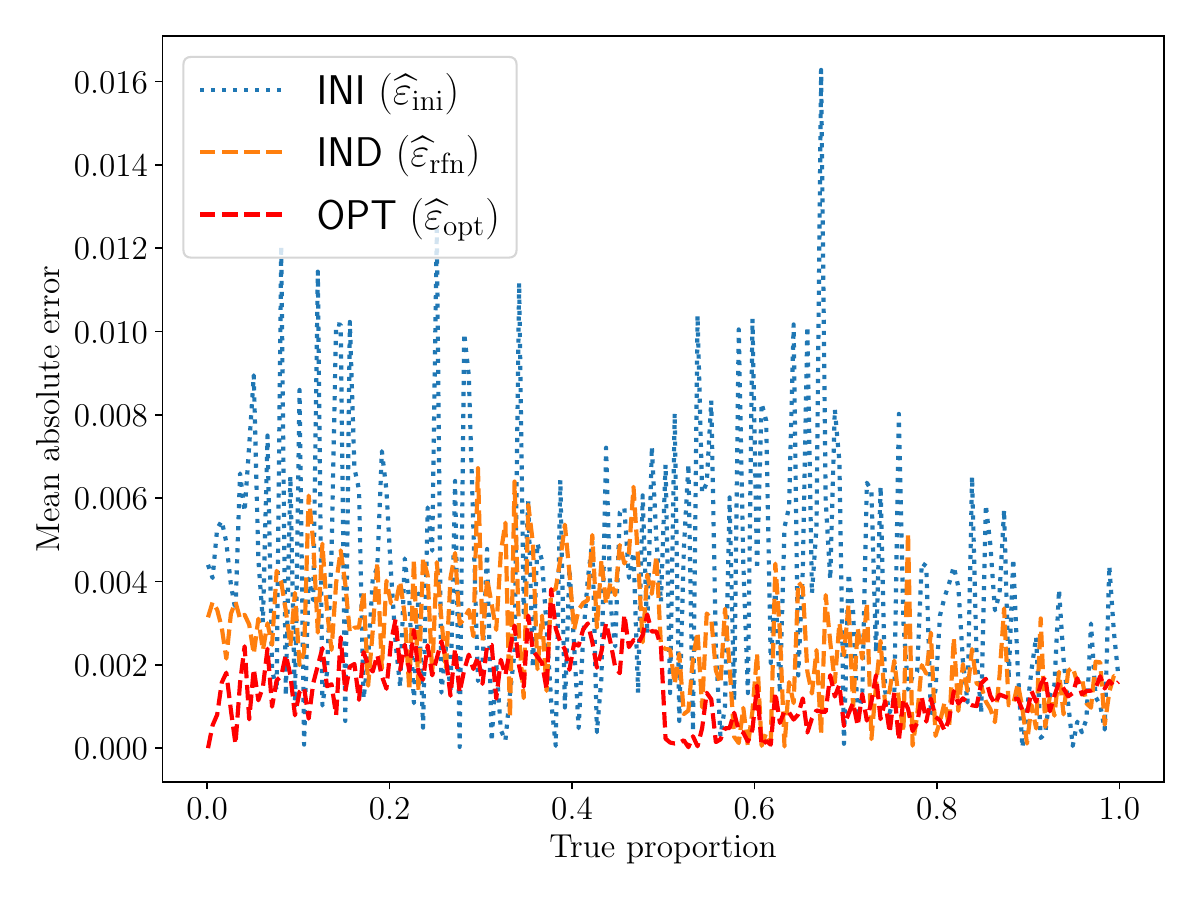}
\caption{$\Delta = 0.7$}
\end{subfigure}
\hfill
\begin{subfigure}{0.495\textwidth}
\includegraphics[width=\textwidth]{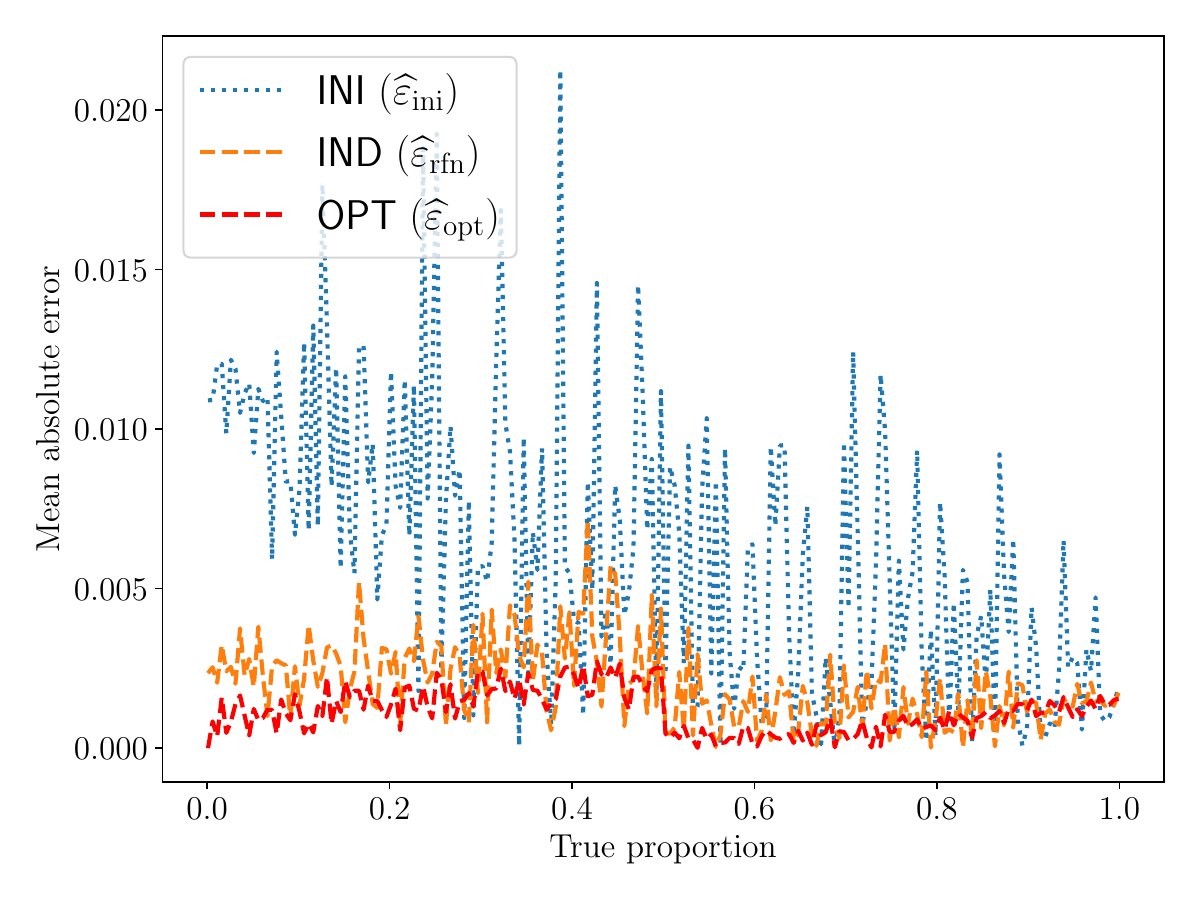}
\caption{$\Delta = 0.8$}
\vspace{-0.1in}
\end{subfigure}

\caption{Mean absolute errors (MAEs) for the inverse transform watermark.
Here $\Delta$ is the dominance parameter in Algorithm \ref{alg:dominate-Ps} for the NTP distributions. The best MAEs for both \textsf{INI} $\epsinital(\delta)$ and \textsf{IND} $\epsrefine(\delta)$, are reported by searching $\delta$ over $\{10^{-1}, 10^{-2}, 10^{-3}\}$.
}
\label{fig:inverse_MAE}
\vspace{-0.1in}
\end{figure}

Figures \ref{fig:gumbel_MAE} and \ref{fig:inverse_MAE} show the complete results of the estimation performance of various estimators on the Gumbel-max and inverse transform watermarks, respectively. In most cases, \textsf{OPT} achieves the lowest MAE.

\begin{figure}[htp]
	\vspace{-0.1in}
\centering
\begin{subfigure}{0.328\textwidth}
\includegraphics[width=\textwidth]{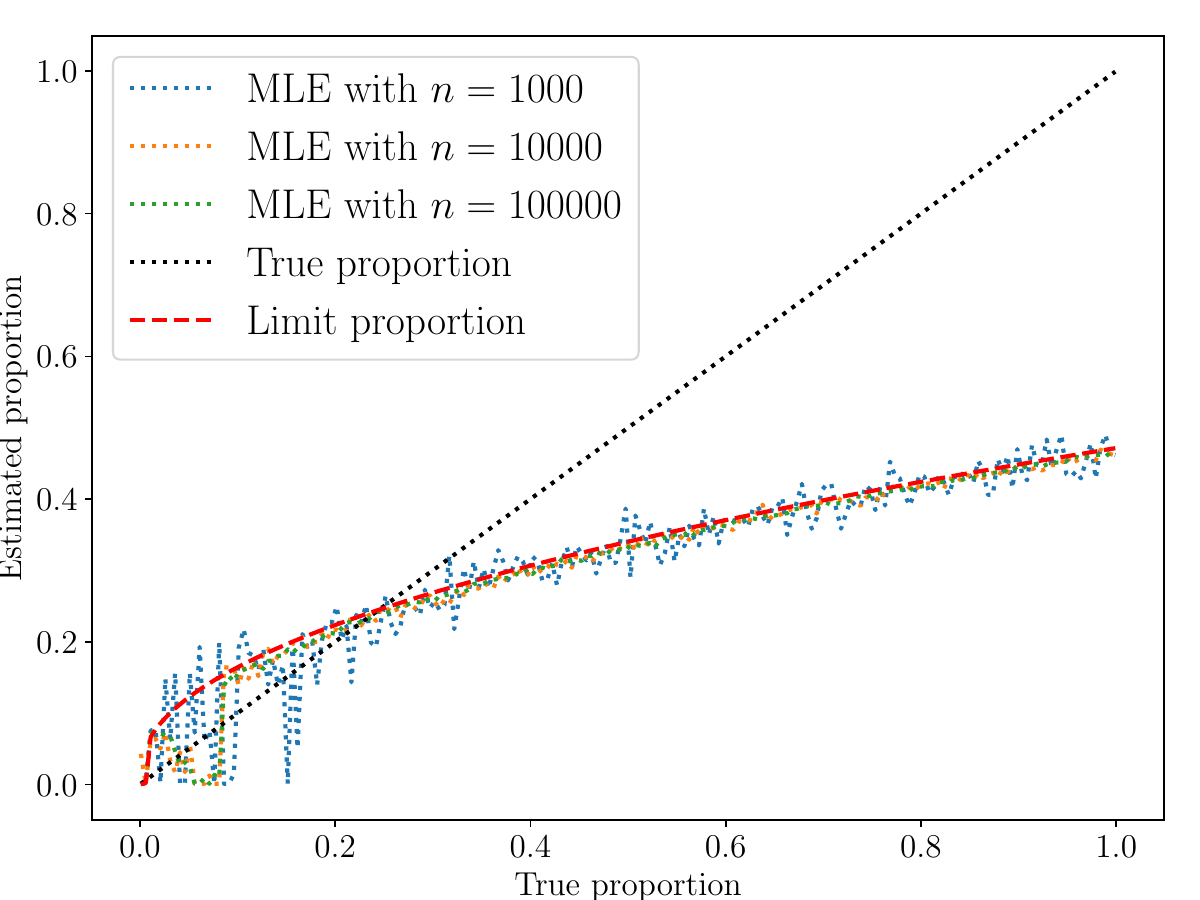}
\caption{$(\gamma, \mu)=(0.1, 0.3)$}
\end{subfigure}
\hfill
\begin{subfigure}{0.328\textwidth}
\includegraphics[width=\textwidth]{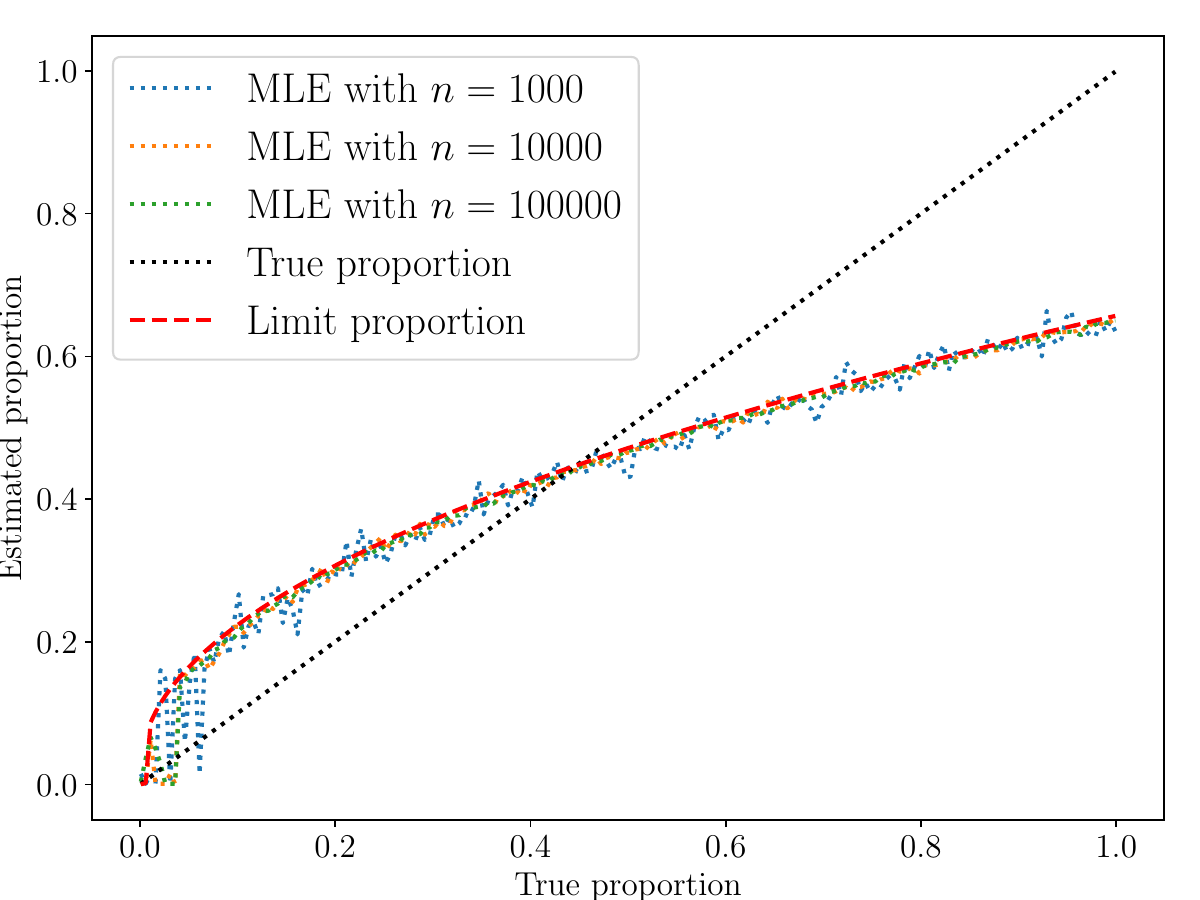}
\caption{$(\gamma, \mu)=(0.1, 0.5)$}
\end{subfigure}
\hfill
\begin{subfigure}{0.328\textwidth}
\includegraphics[width=\textwidth]{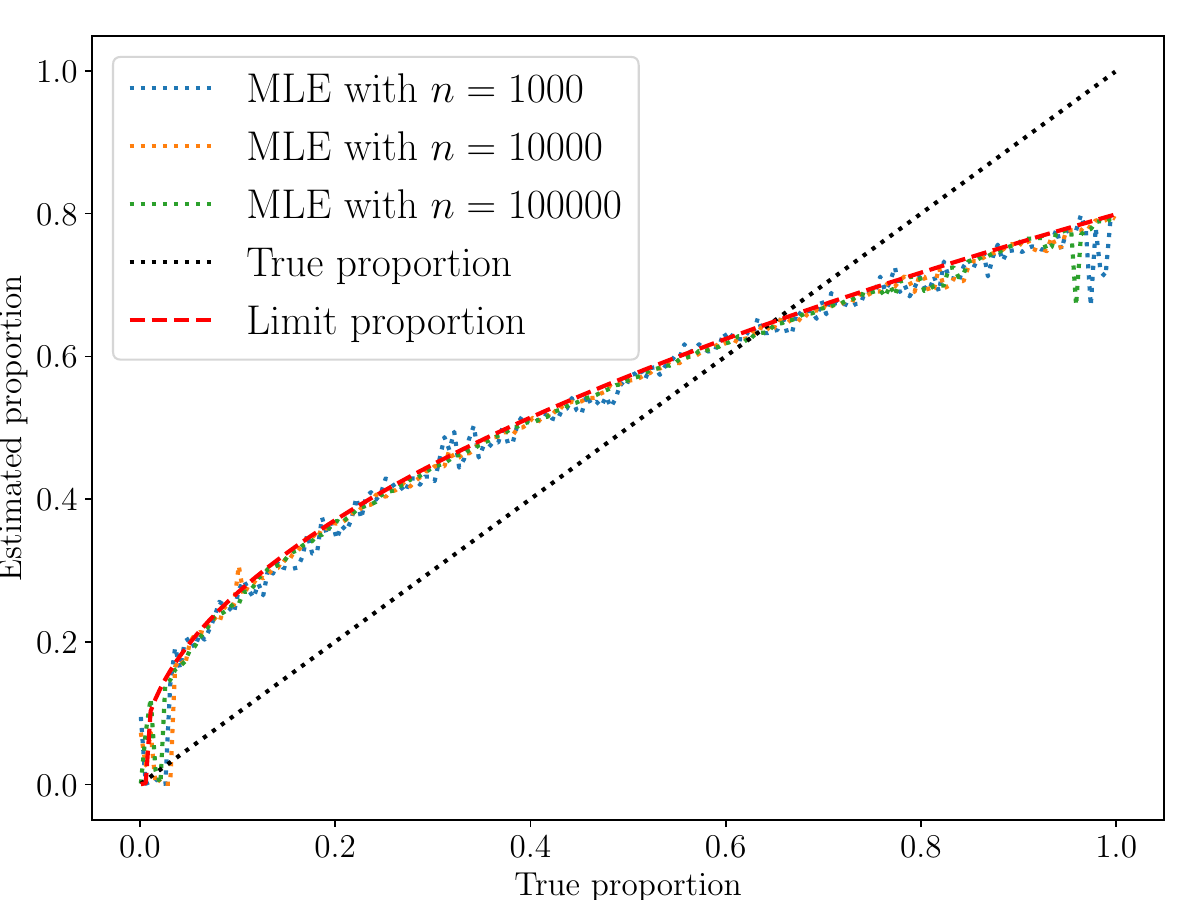}
\caption{$(\gamma, \mu)=(0.1, 0.7)$}
\end{subfigure}

\begin{subfigure}{0.328\textwidth}
\includegraphics[width=\textwidth]{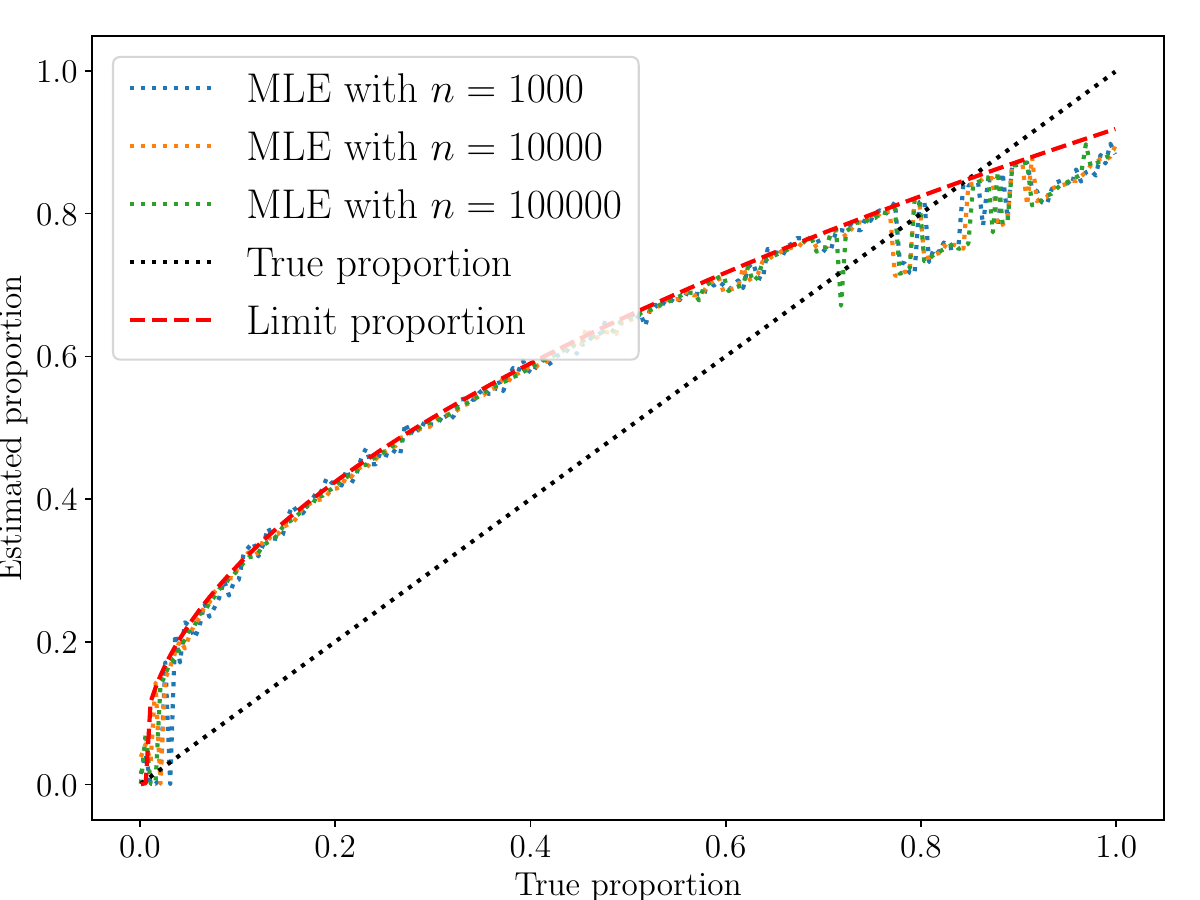}
\caption{$(\gamma, \mu)=(0.1, 0.9)$}
\end{subfigure}
\hfill
\begin{subfigure}{0.328\textwidth}
\includegraphics[width=\textwidth]{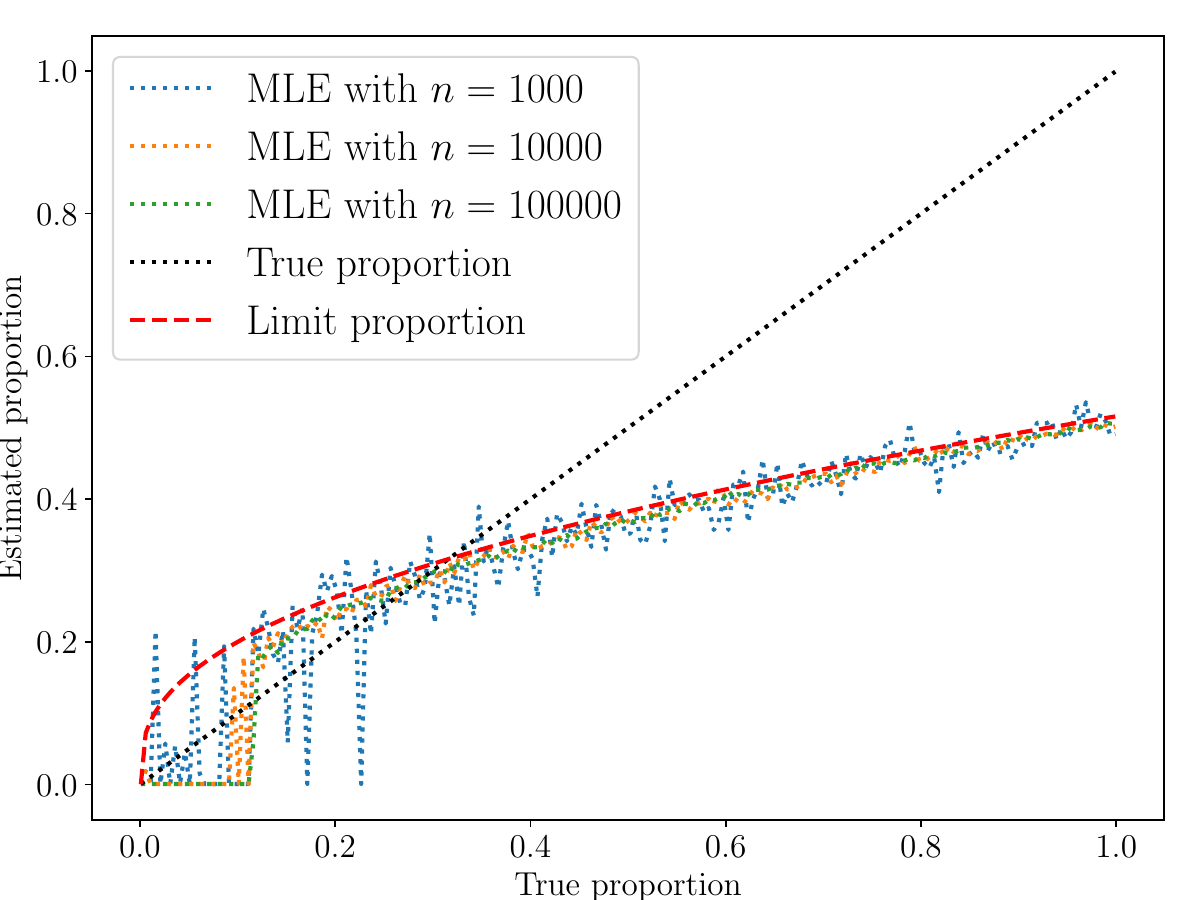}
\caption{$(\gamma, \mu)=(0.3, 0.5)$}
\end{subfigure}
\hfill
\begin{subfigure}{0.328\textwidth}
\includegraphics[width=\textwidth]{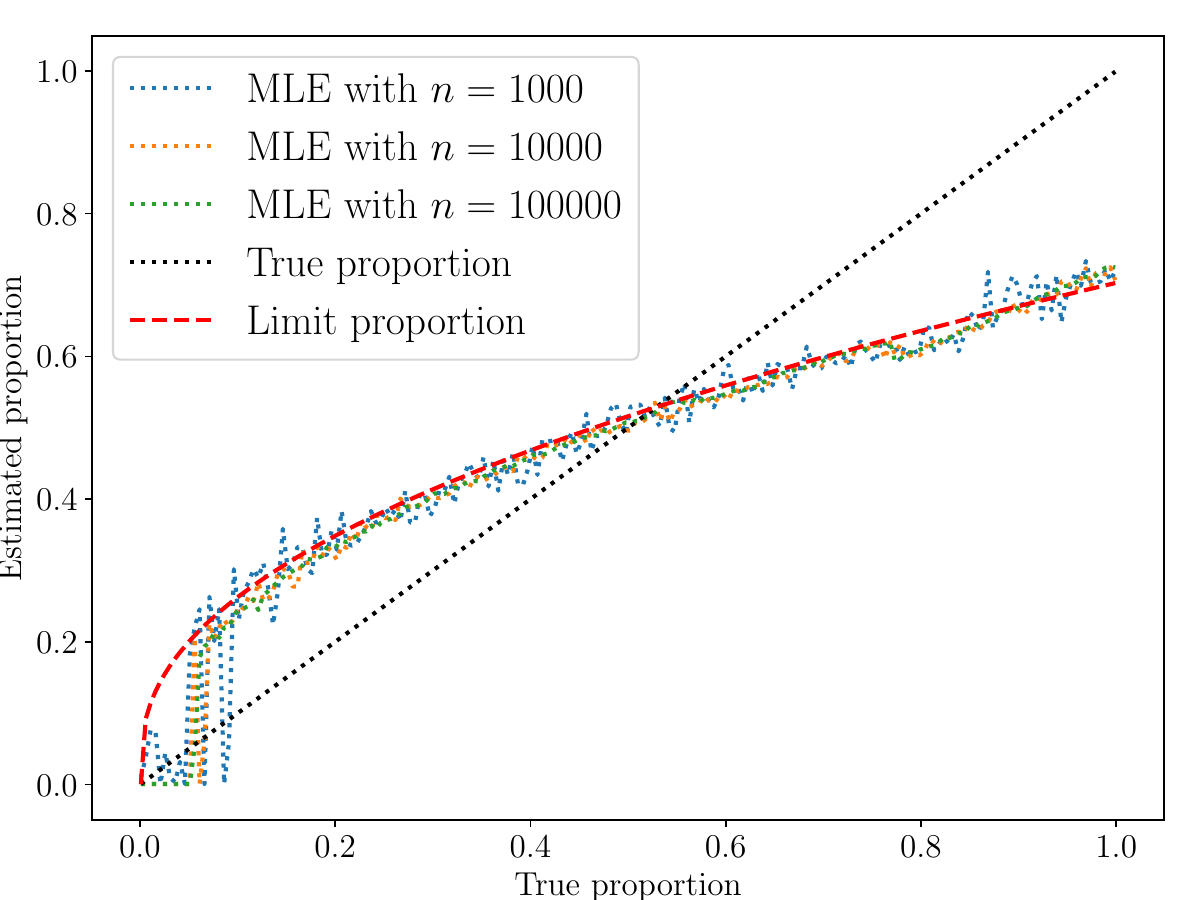}
\caption{$(\gamma, \mu)=(0.3, 0.7)$}
\end{subfigure}

\begin{subfigure}{0.328\textwidth}
\includegraphics[width=\textwidth]{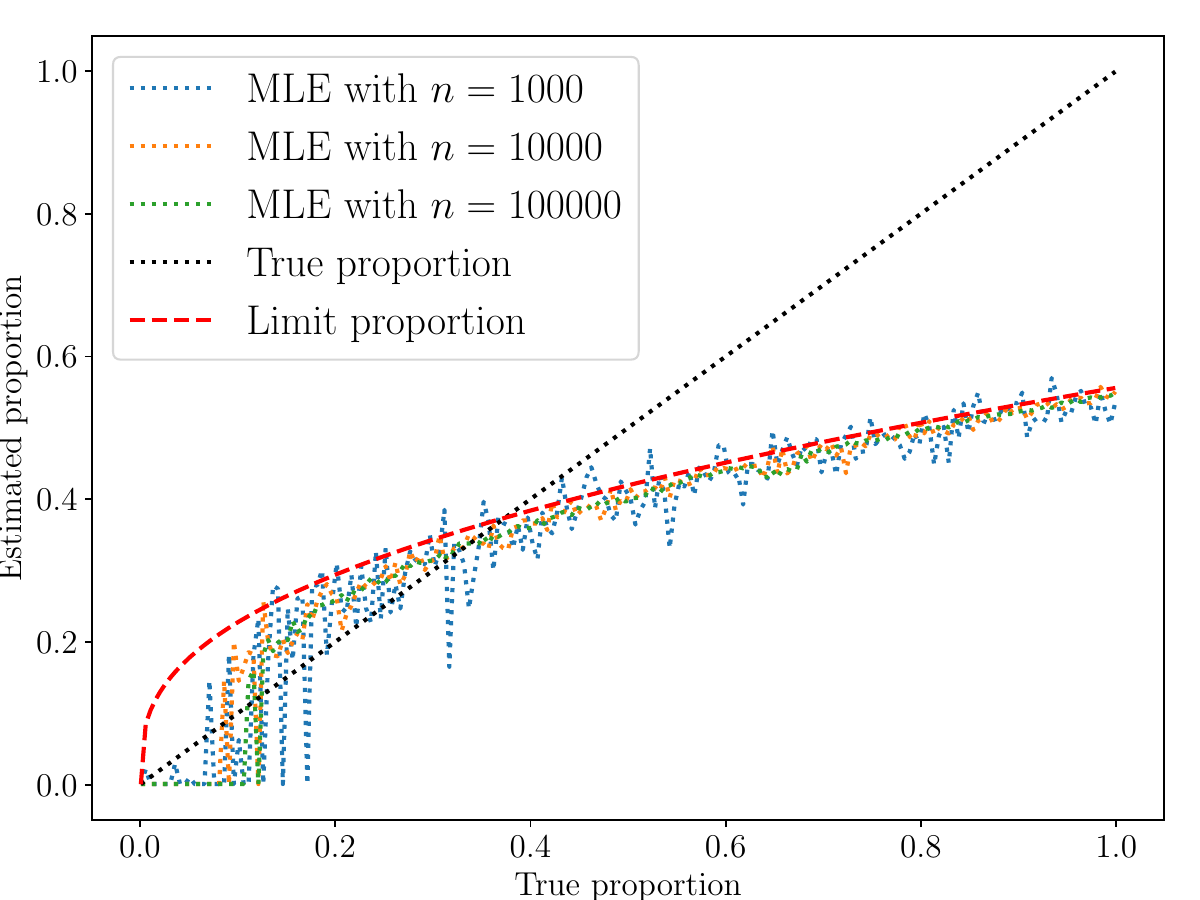}
\caption{$(\gamma, \mu)=(0.5, 0.7)$}
\end{subfigure}
\hfill
\begin{subfigure}{0.328\textwidth}
\includegraphics[width=\textwidth]{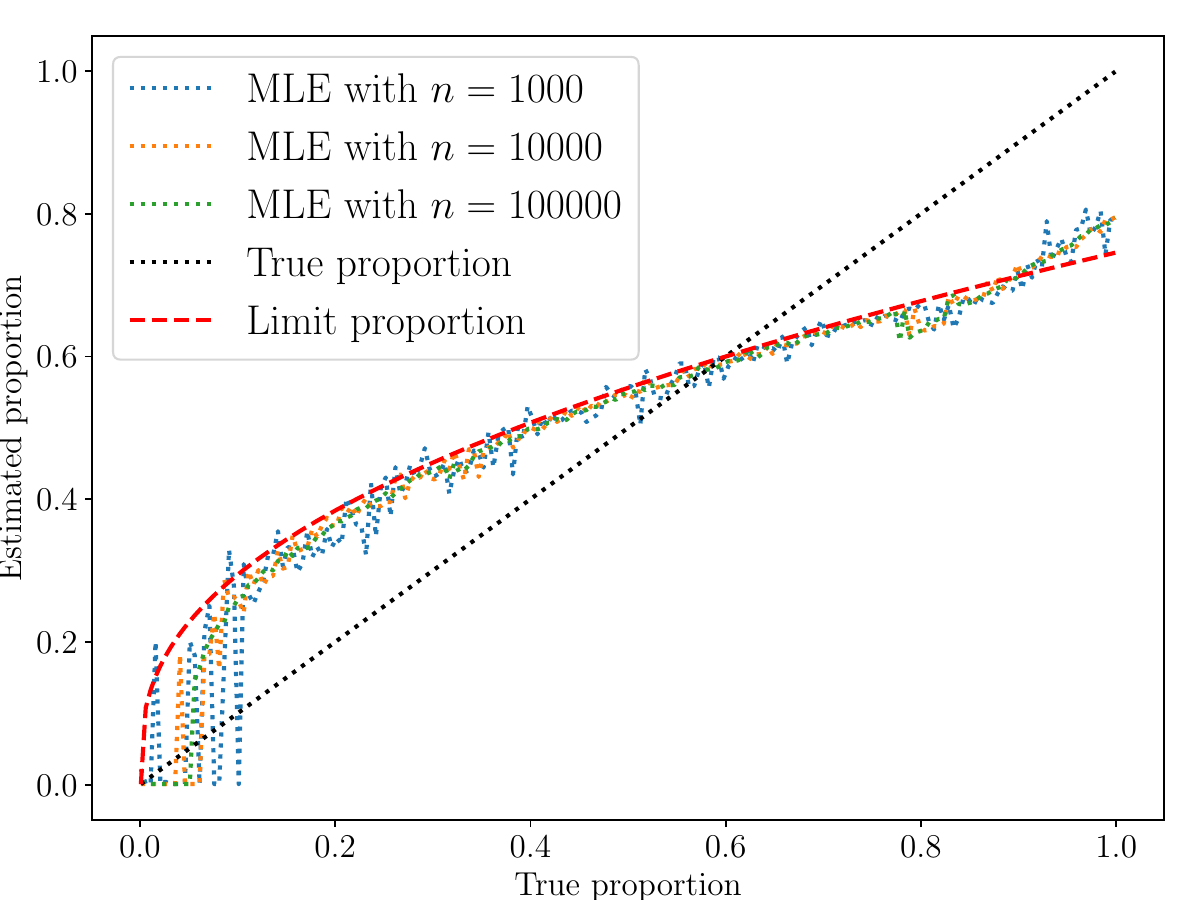}
\caption{$(\gamma, \mu)=(0.5, 0.9)$}
\end{subfigure}
\hfill
\begin{subfigure}{0.328\textwidth}
\includegraphics[width=\textwidth]{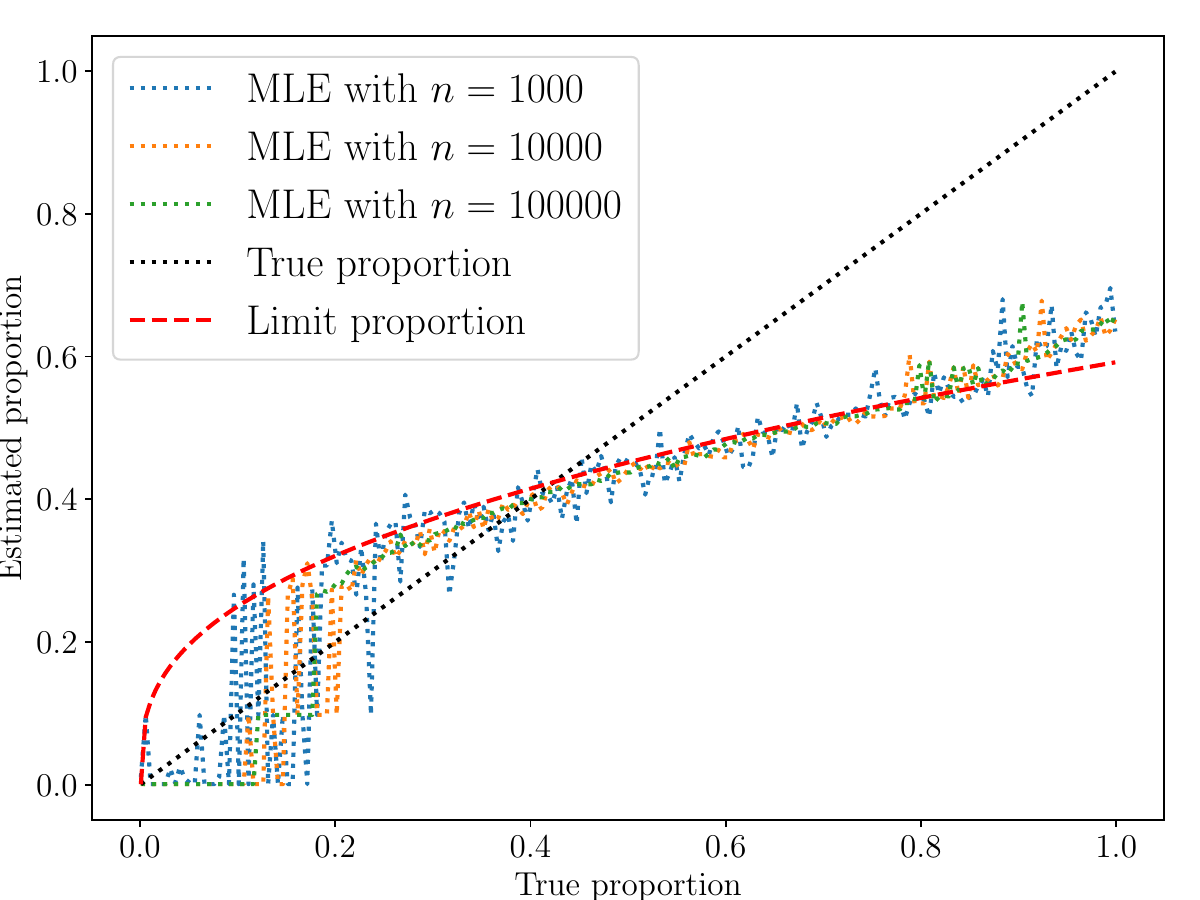}
\caption{$(\gamma, \mu)=(0.7, 0.9)$}
\end{subfigure}

\caption{More results illustrating the inherent bias caused by non-identifiability. We vary the values of $(\gamma, \mu)$. The red curves represent the limit proportion from Theorem \ref{thm:failure-inherent-bias}, while the other colored curves show the MLE estimates for different sample sizes $n$.
}
\label{fig:inherent-bias}
\end{figure}

\subsection{Additional Results on Inherent Bias}

We vary the values of $(\gamma, \mu)$ and plot the estimators obtained by solving the $L_2$-regularized MLEs described in Theorem \ref{thm:failure-inherent-bias} in Figure \ref{fig:inherent-bias}. In each subplot, we fix the $L_2$ regularization coefficient to $\lambda = 10^{-2}$ and consider three sample sizes: $n \in \{10^3, 10^4, 10^5\}$. 
The red curves represent the theoretical limit proportion $\widehat{\eps}$ from Theorem \ref{thm:failure-inherent-bias}, while the other colored curves depict the MLE estimates $\widehat{\eps}_{\lambda}$ for the different sample sizes. Across all combinations of $(\gamma, \mu)$, the $L_2$-regularized MLEs align closely with the theoretical limit proportion. This alignment validates our theory and provides strong support for the existence of inherent bias.

\section{Open-source LLM Experiments: Details and Results}
\label{appen:LLM-experiment}

\subsection{Details on Mixture Data Setting}

\paragraph{Watermark generation.}  
We use a context window of size $m=4$, where the random variable $\xi_t$ is derived as $\xi_t = \AM(s_{(t-m):(t-1)}, \Key)$, based on the preceding $m$ tokens. The randomness is generated using the hash function $\AM$, as introduced in \citep{zhao2024permute}. 
To avoid repetitive outputs, a watermark is applied only when the current text window is unique within the generation history, following the strategies outlined in \citep{hu2023unbiased, wu2023dipmark, dathathri2024scalable}. This technique is known as 1-sequence repeated context masking in \citep{dathathri2024scalable}. If no watermark is applied, tokens are sampled directly from the NTP distribution using multinomial sampling.  
When constructing a dataset with the true proportion $\eps$, we aim for approximately $\eps$-proportion of the samples (i.e., observed pivotal statistics) to be watermarked and $(1-\eps)$-proportion to be null samples.
So we independently watermark each token with probability $\eps$. However, applying 1-sequence repeated context masking can reduce the effective watermark proportion, as it may prevent some tokens that should be watermarked from being marked as such. 
As a result, in our experiments, we calculate the average number of watermarked tokens to determine the ground truth $\eps$.

\paragraph{Prompt generation.}  
Our prompt generation approach largely follows the methodology described in Appendix C.1 of \citep{li2024optimal}. Prompts are created by sampling documents from the news-focused section of the C4 dataset \citep{raffel2020exploring}. Each document is required to have at least 50 tokens, and shorter documents are excluded. Since retokenization may not perfectly align with the original token counts, we ensure that the verifier always processes at least $n$ tokens by adding special padding tokens, which are model-dependent. Additionally, to reduce the need for padding, we generate extra tokens---referred to as buffer tokens---beyond $n$. In all experiments, 20 buffer tokens are generated, which generally eliminates the necessity for padding.  

\paragraph{Details for \textsf{IND} and  \textsf{OPT}.}
The computation for the refined estimators (namely \textsf{IND} and \textsf{OPT}) follows the procedure outlined in Appendix \ref{appen:simulation}. To evaluate MAEs across various proportions $\eps$, we generate a mixed-source dataset for each trial, ensuring that approximately $\eps$-proportion of the samples are watermarked. Importantly, these datasets consist solely of pivotal statistics, not tokens.
After generating datasets for different values of $\eps$, we aggregate all the watermarked pivotal statistics into a fully watermarked dataset. This combined dataset is then used to estimate $\widehat{g}$ and $\widehat{F}_{\bP}$. As a result, some pivotal statistics are entirely relevant to the current estimation task.

\paragraph{Details for the baseline \textsf{WPL}.}

\textsf{WPL}, proposed by \citet{zhao2024efficiently}, uses an existing online learning algorithm to compute a watermark score for each token, classifying it as watermarked if the score exceeds a specified threshold. For the Gumbel-max watermark, the score is defined as $-\log(1 - Y)$, while for the inverse transform watermark, it is simply $Y$. The corresponding pivotal statistics $Y$ are summarized in Table~\ref{tab:comparison}. To ensure a fair and competitive comparison, we tune the threshold over a predefined set and report the best result that achieves the lowest MAE across both figures and tables. Specifically, the threshold set is $\{1.3, 1.5, 1.7\}$ for Gumbel-max and $\{0.73, 0.75, 0.77\}$ for the inverse transform watermark.




\subsection{Additional Results on Estimation Accuracy}

\begin{figure*}[tp]
\vspace{-0.1in}
\centering
\begin{subfigure}{0.495\textwidth}
\includegraphics[width=\textwidth]{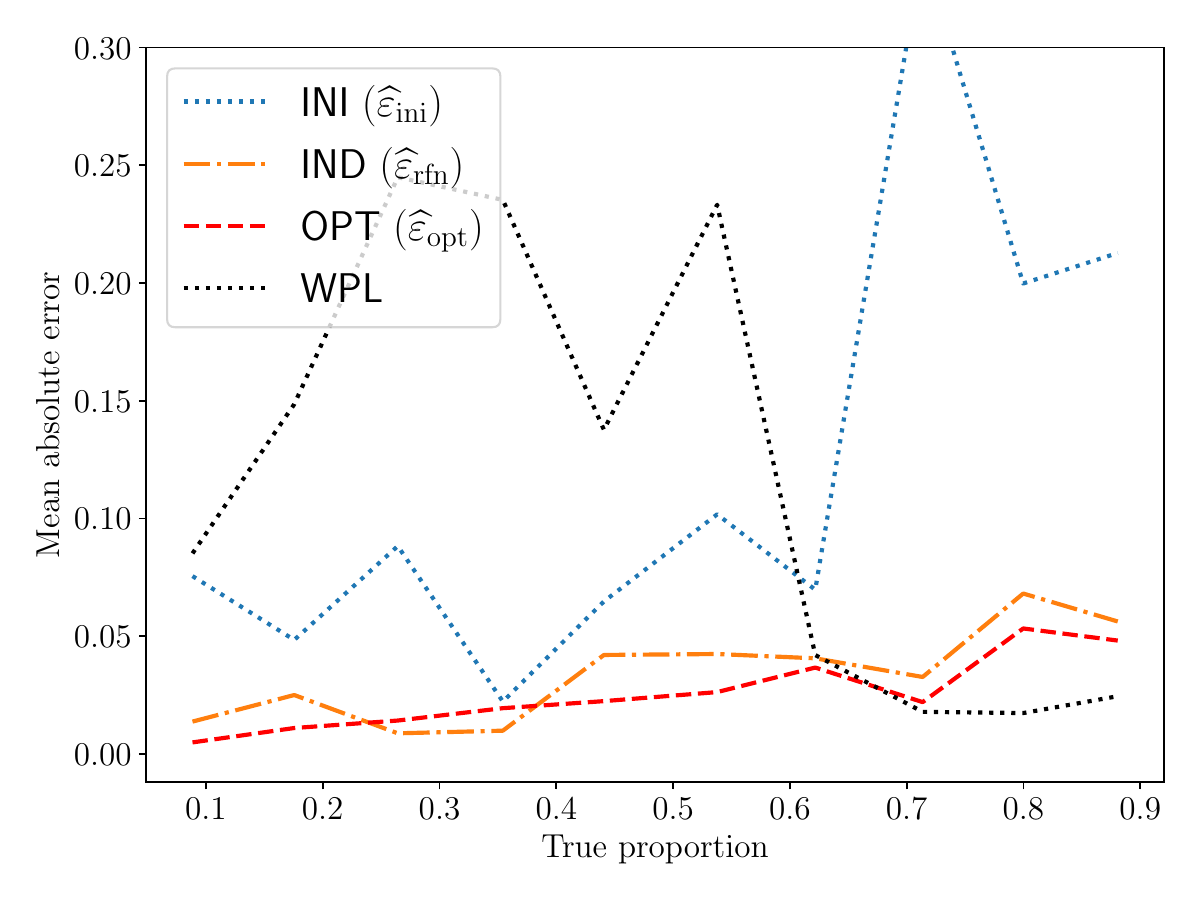}
\caption{Gumbel-max with $\mathrm{T}=0.7$}
\label{fig:13B-gumbel-0.7}
\end{subfigure}
\hfill
\begin{subfigure}{0.495\textwidth}
\includegraphics[width=\textwidth]{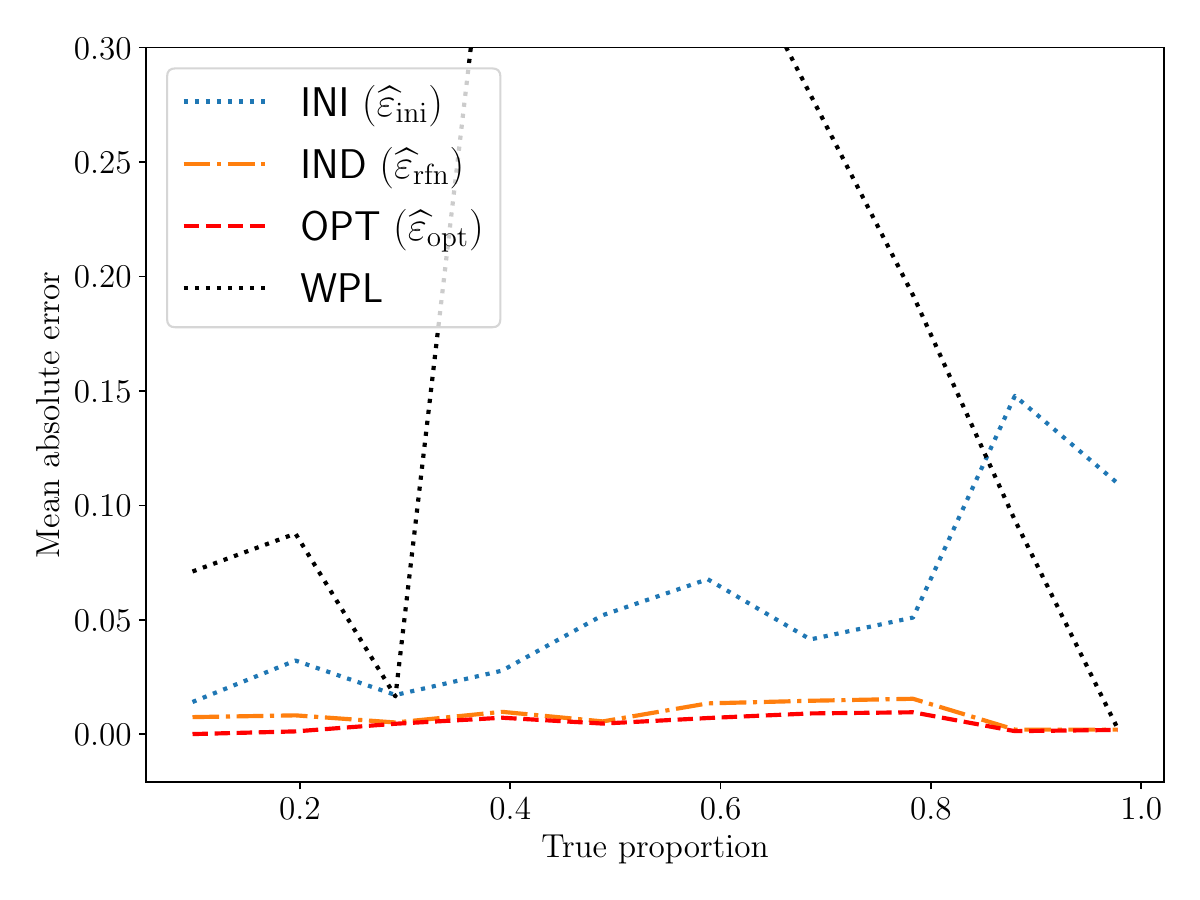}
\caption{Gumbel-max with $\mathrm{T}=1$}
\label{fig:13B-gumbel-1}
\end{subfigure} \\
\begin{subfigure}{0.495\textwidth}
\includegraphics[width=\textwidth]{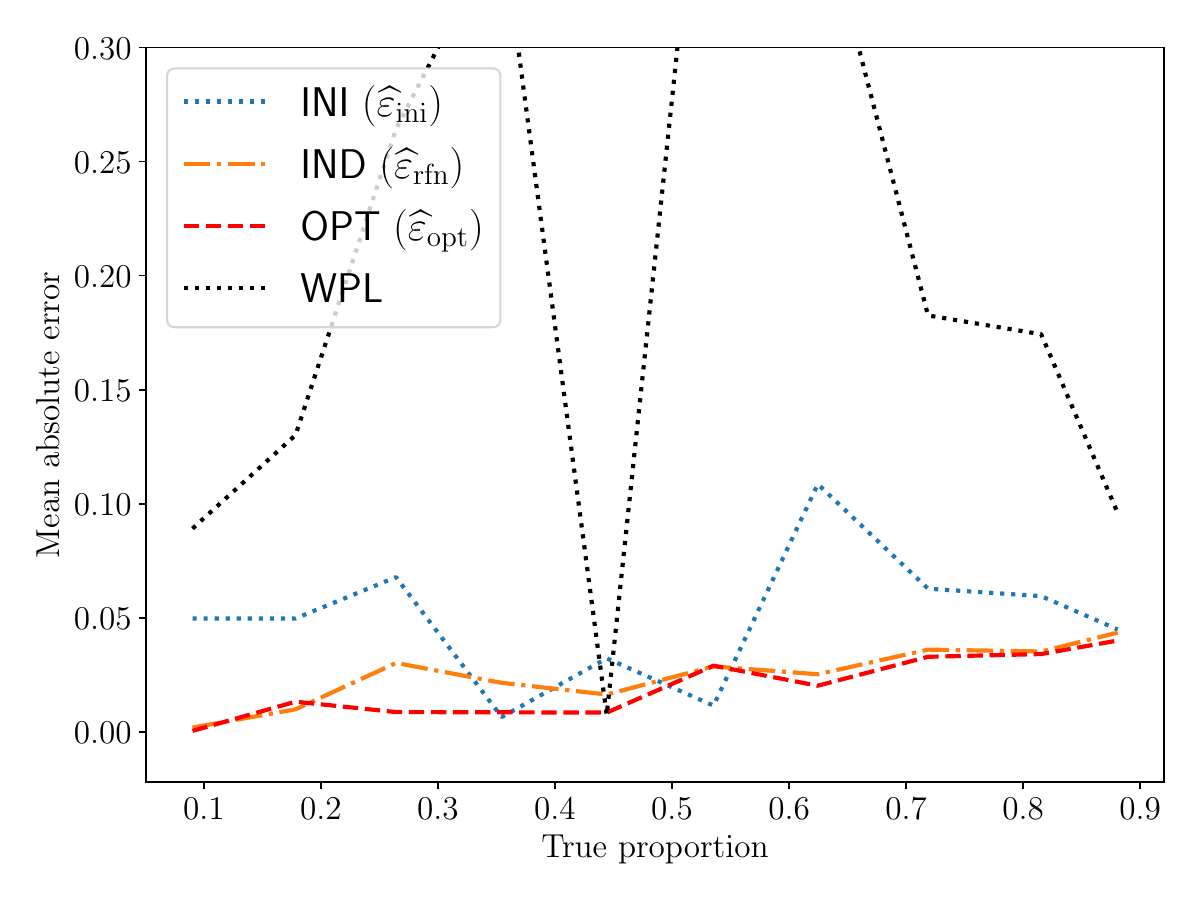}
\caption{Inverse with $\mathrm{T}=0.7$}
\label{fig:13B-inverse-0.7}
\end{subfigure}
\hfill
\begin{subfigure}{0.495\textwidth}
\includegraphics[width=\textwidth]{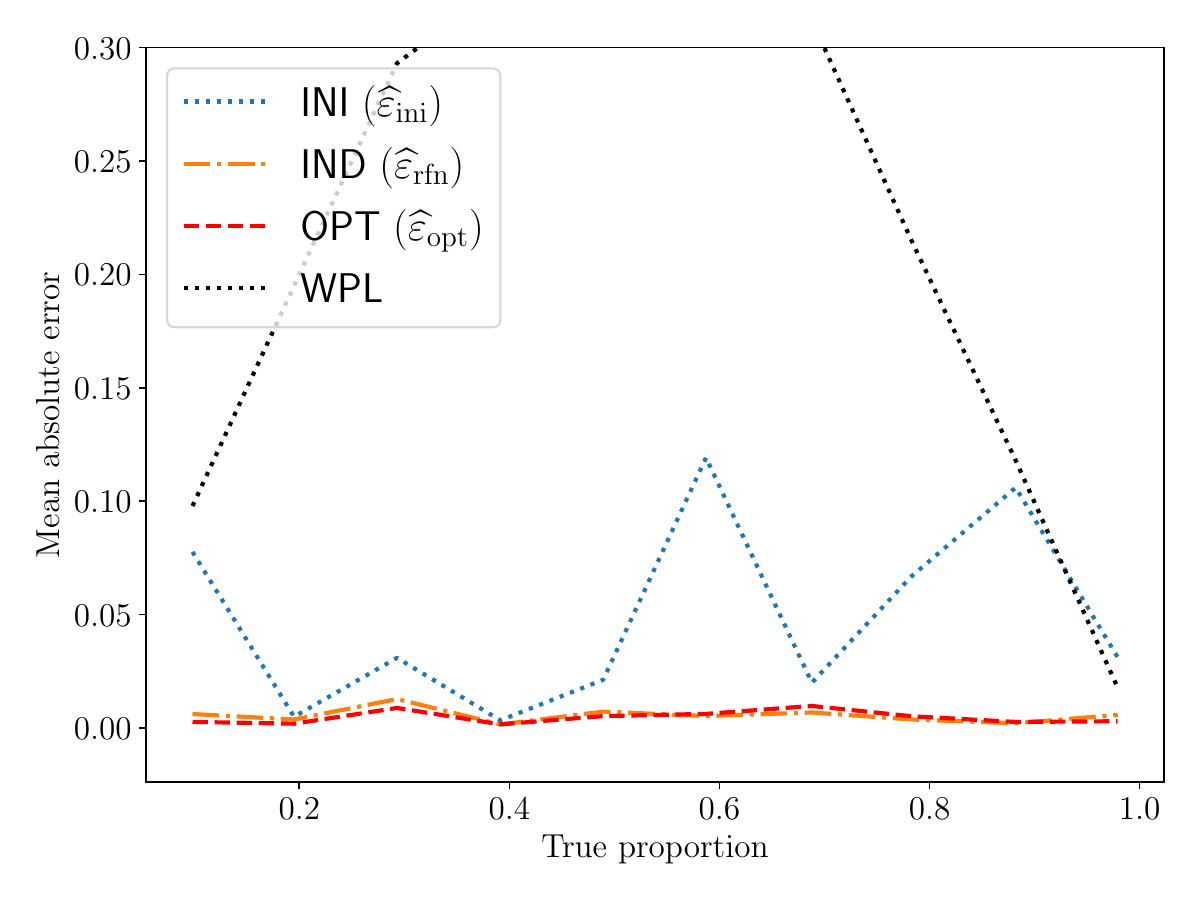}
\caption{Inverse with $\mathrm{T}=1$}
\label{fig:13B-inverse-1}
\end{subfigure}
\caption{
Mean absolute errors of estimators on the C4 dataset, evaluated across various true proportions $\eps$ and temperature parameters for the OPT-13B model.
}
\label{fig:13B-LLM}
\end{figure*}

\begin{figure*}[tp]
\vspace{-0.1in}
\centering
\begin{subfigure}{0.495\textwidth}
\includegraphics[width=\textwidth]{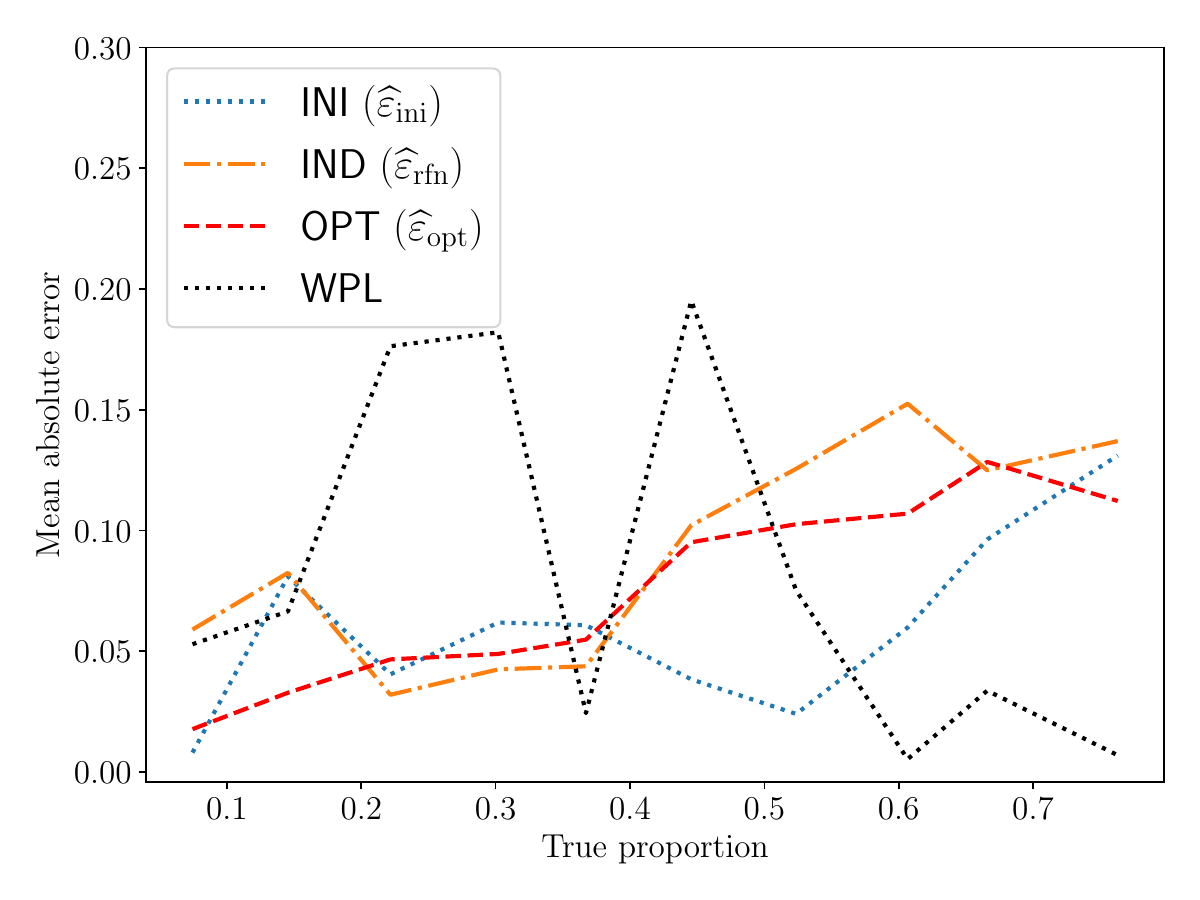}
\caption{Gumbel-max with $\mathrm{T}=0.7$}
\label{fig:8B-gumbel-0.7}
\end{subfigure}
\hfill
\begin{subfigure}{0.495\textwidth}
\includegraphics[width=\textwidth]{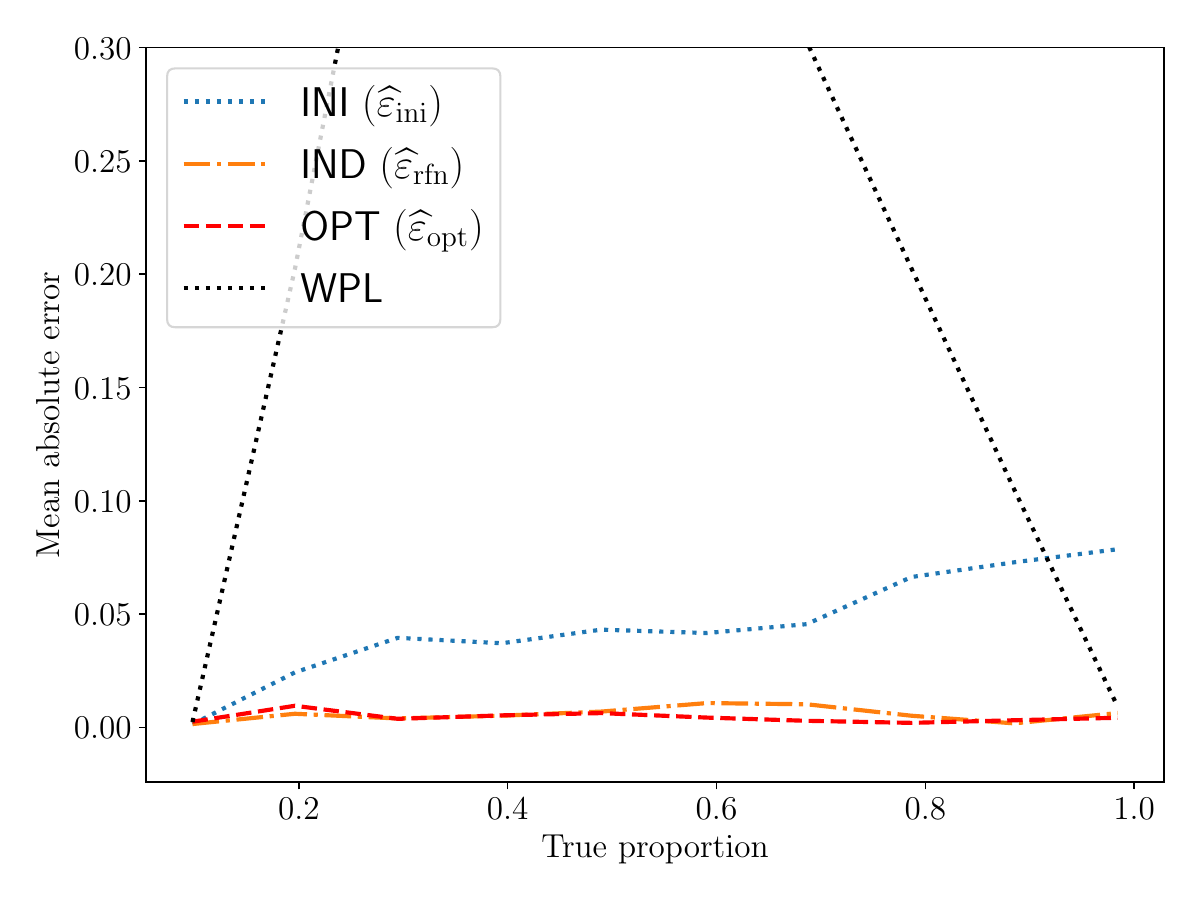}
\caption{Gumbel-max with $\mathrm{T}=1$}
\label{fig:8B-gumbel-1}
\end{subfigure} \\
\begin{subfigure}{0.495\textwidth}
\includegraphics[width=\textwidth]{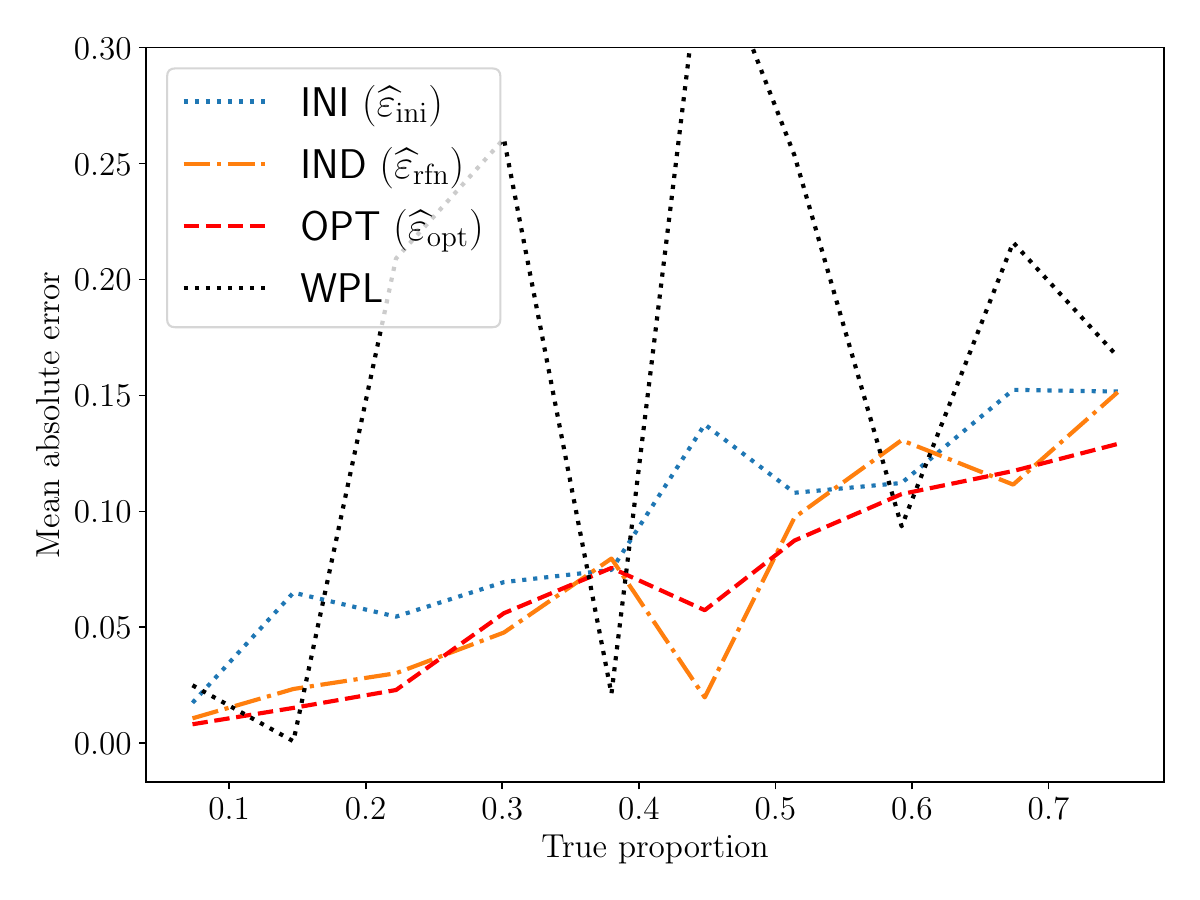}
\caption{Inverse with $\mathrm{T}=0.7$}
\label{fig:8B-inverse-0.7}
\end{subfigure}
\hfill
\begin{subfigure}{0.495\textwidth}
\includegraphics[width=\textwidth]{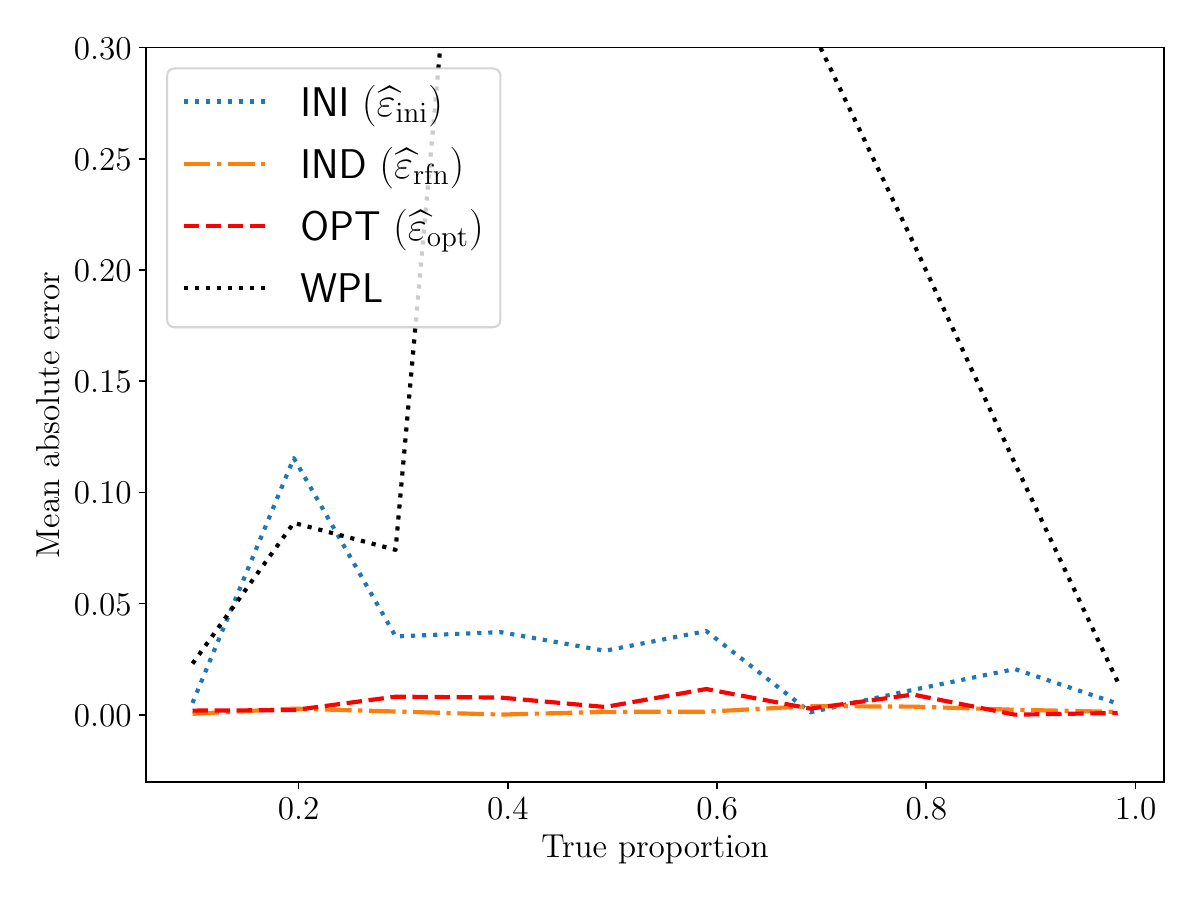}
\caption{Inverse with $\mathrm{T}=1$}
\label{fig:8B-inverse-1}
\end{subfigure}
\caption{
Mean absolute errors of estimators on the C4 dataset, evaluated across various true proportions $\eps$ and temperature parameters for the LLaMA3.1-8B model.
}
\label{fig:8B-LLM}
\end{figure*}

\begin{figure*}[tp]
\vspace{-0.1in}
\centering
\begin{subfigure}{0.495\textwidth}
\includegraphics[width=\textwidth]{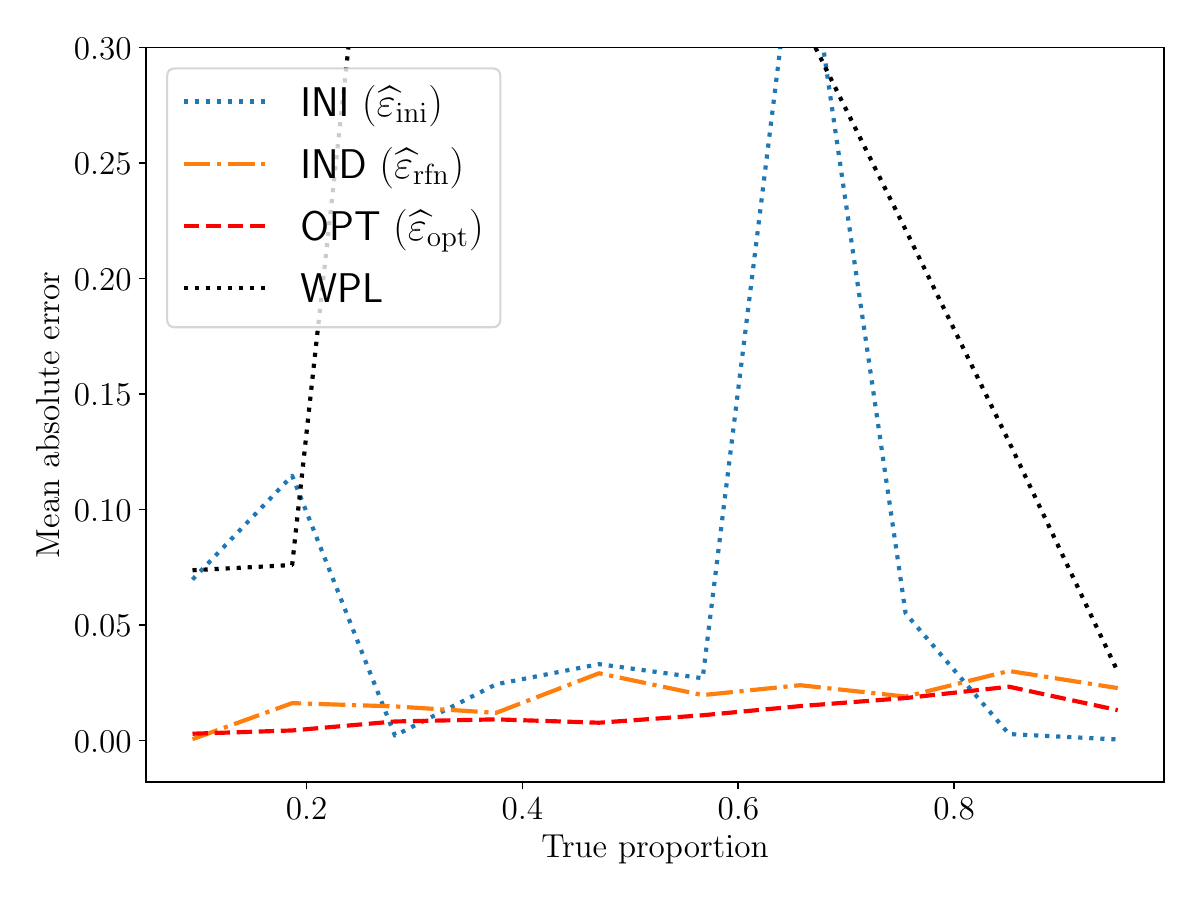}
\caption{Gumbel-max on OPT-1.3B}
\end{subfigure}
\hfill
\begin{subfigure}{0.495\textwidth}
\includegraphics[width=\textwidth]{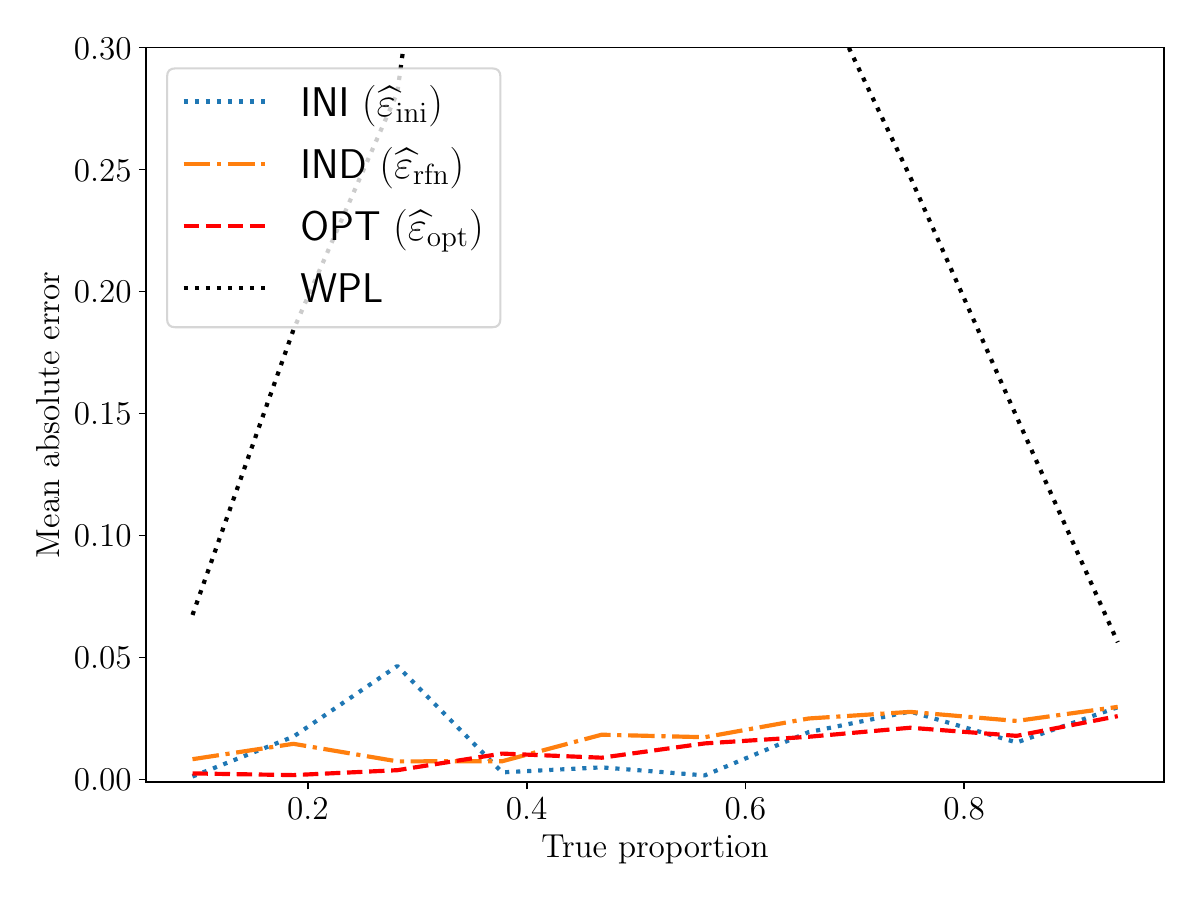}
\caption{Inverse on OPT-1.3B}
\end{subfigure}\\
\begin{subfigure}{0.495\textwidth}
\includegraphics[width=\textwidth]{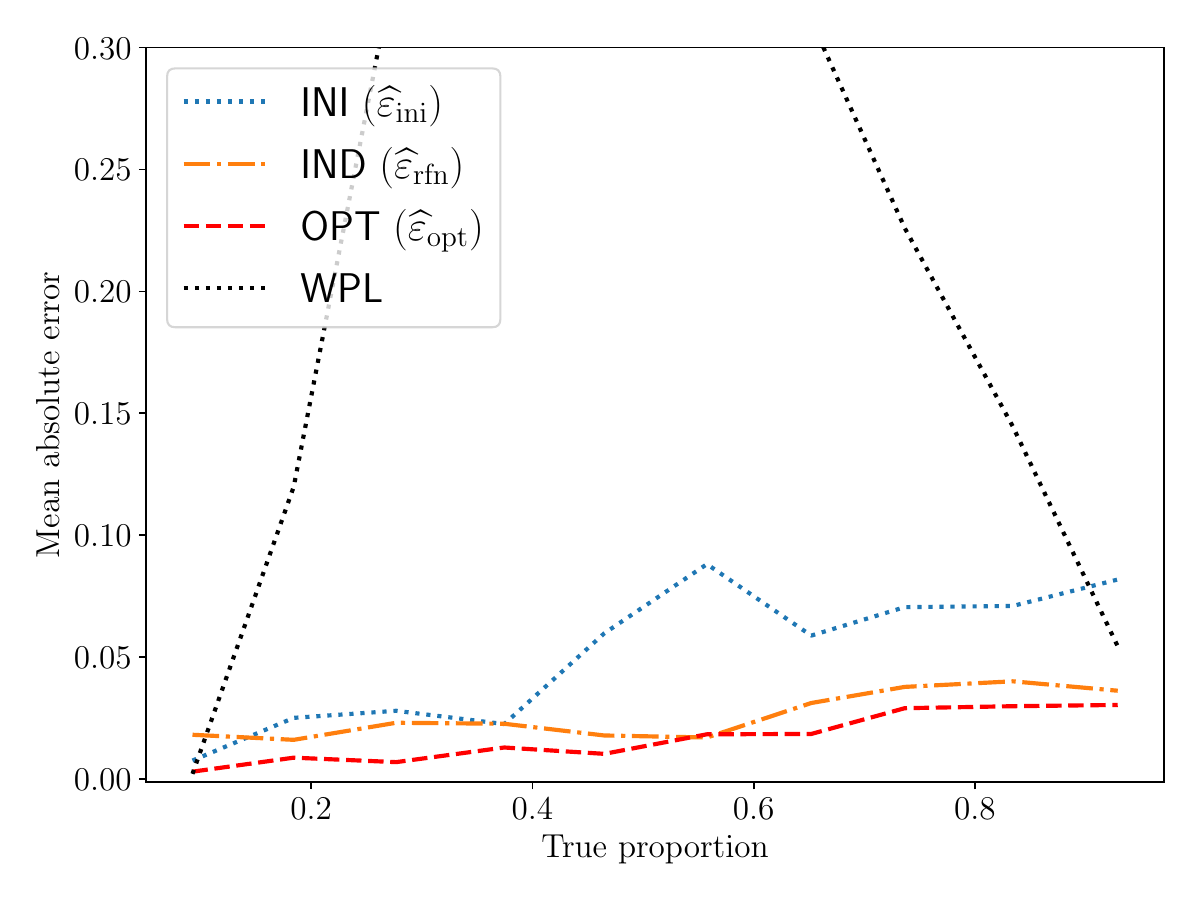}
\caption{Gumbel-max on OPT-13B}
\end{subfigure}
\hfill
\begin{subfigure}{0.495\textwidth}
\includegraphics[width=\textwidth]{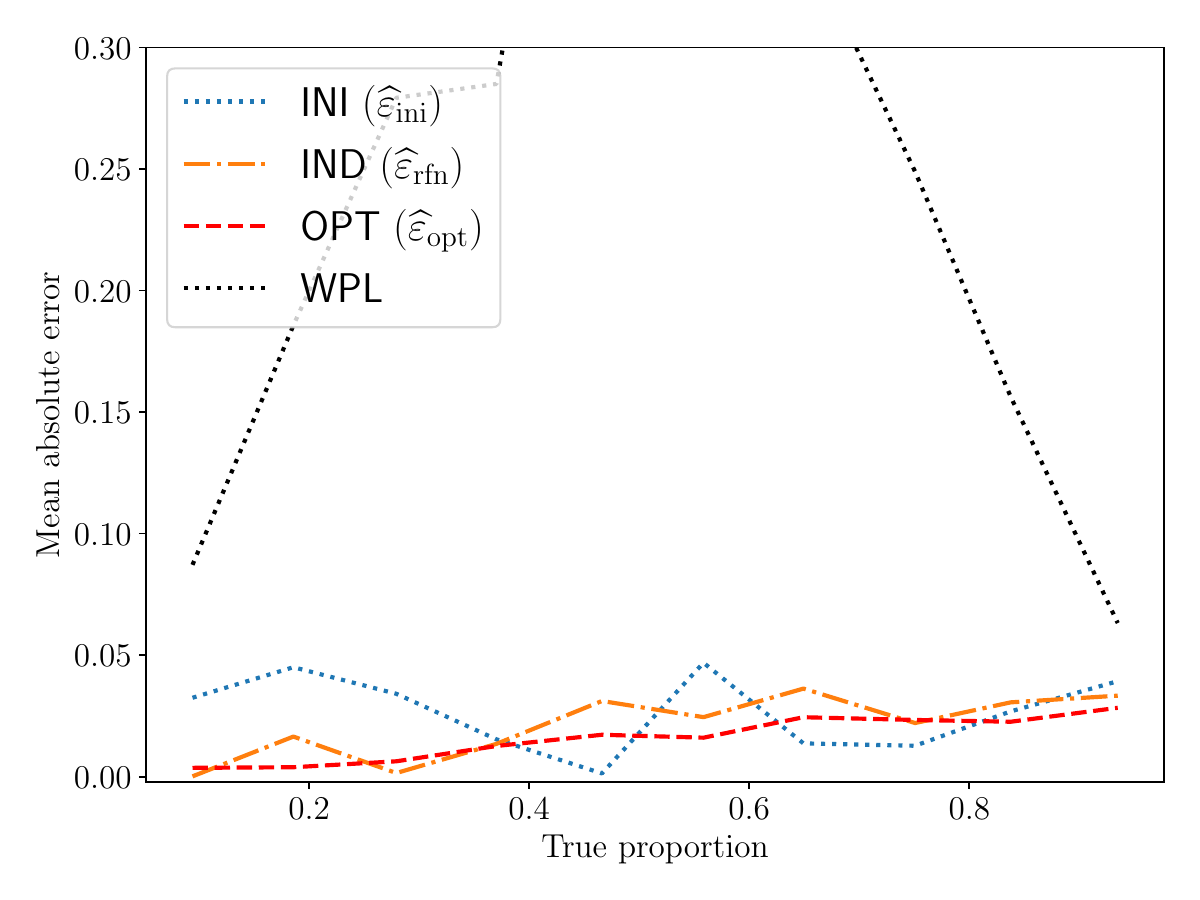}
\caption{Inverse on OPT-13B}
\end{subfigure} \\
\begin{subfigure}{0.495\textwidth}
\includegraphics[width=\textwidth]{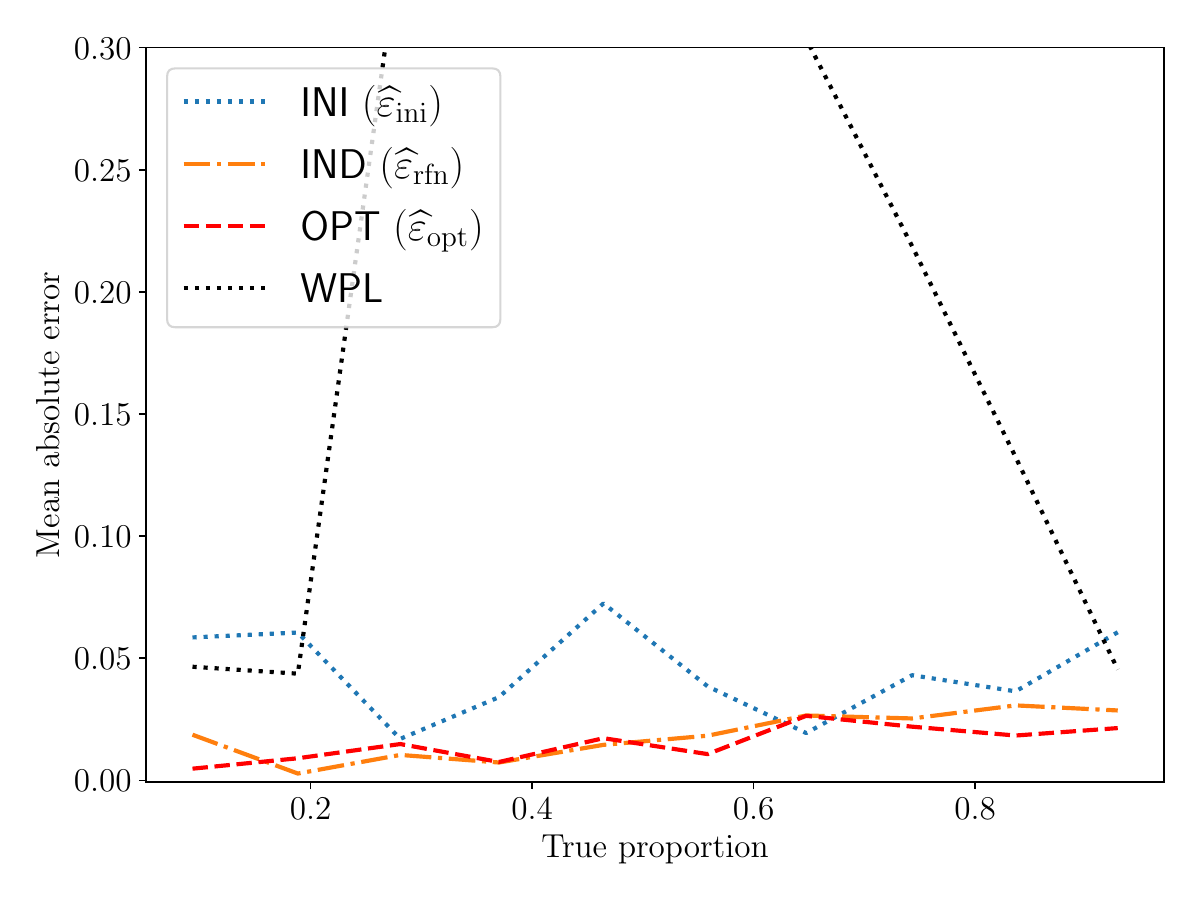}
\caption{Gumbel-max on LLaMA3.1-8B}
\end{subfigure}
\hfill
\begin{subfigure}{0.495\textwidth}
\includegraphics[width=\textwidth]{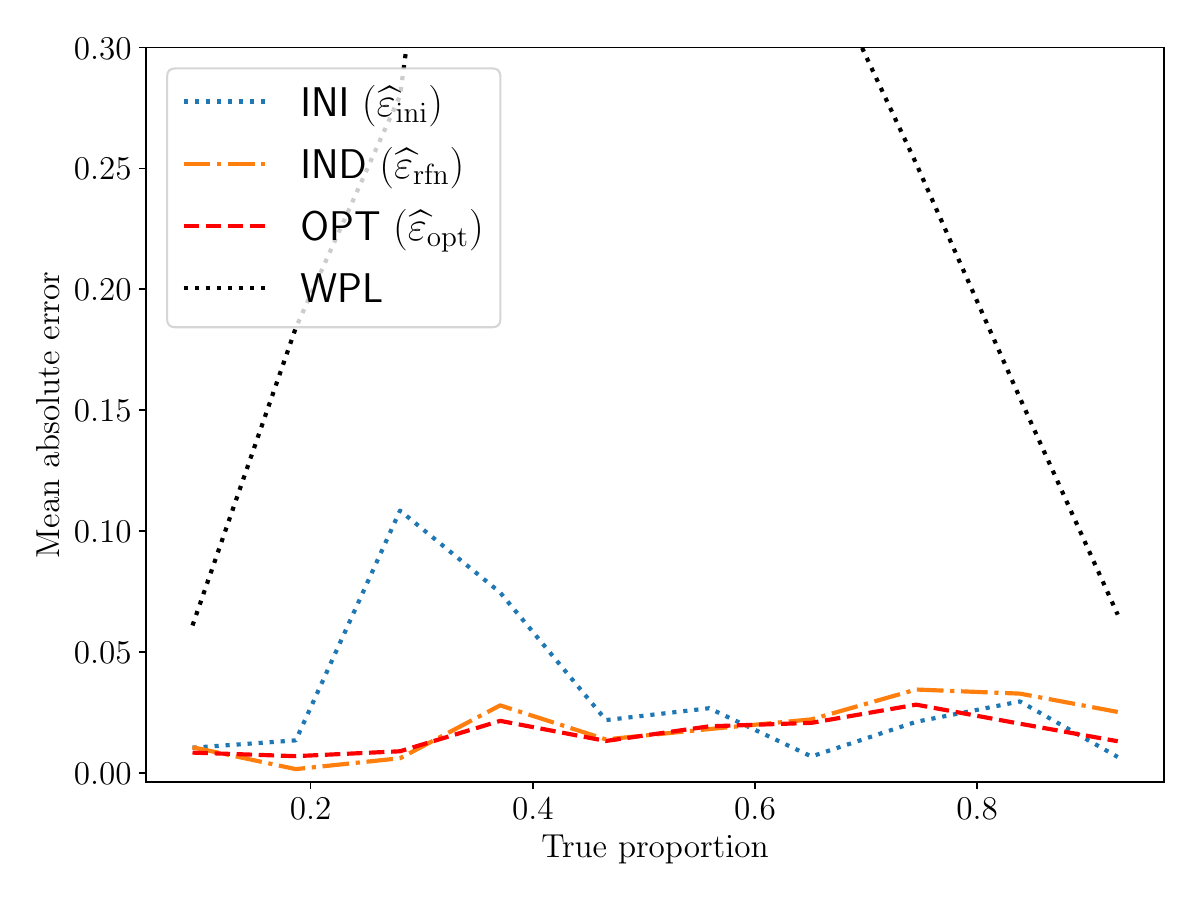}
\caption{Inverse on LLaMA3.1-8B}
\end{subfigure} 
\caption{
Mean absolute errors of estimators on the arXiv dataset, evaluated across various true proportions $\eps$ for the three different models. The temperature parameter is set to $1$. 
}
\label{fig:LLM-arxiv}
\end{figure*}

Figures \ref{fig:13B-LLM} and \ref{fig:8B-LLM} illustrate the estimation performance of various estimators on the OPT-13B and LLaMA3.1-8B models using the C4 dataset, respectively.  
Figure \ref{fig:LLM-arxiv} presents the performance on the arXiv dataset.  
In most scenarios, \textsf{OPT} achieves the lowest MAE, while \textsf{WPL} exhibits poor accuracy when the true proportion is around 0.5.

\subsection{Details on Post-processing Modification}
The post-processing modification setup closely follows the earlier mixture data setting. We use a context window of size $m=4$ and fix the temperature to 1. The same three LLMs are evaluated on both the C4 and arXiv datasets. Since the high temperature introduces sufficient randomness, we didn't apply 1-sequence repeated context masking \citep{dathathri2024scalable}. For each run, we sample prompts using the last 50 tokens of randomly selected documents and generate 500 continuation tokens. A specified fraction of tokens is then modified via random substitution, deletion, or insertion, and we apply various methods to estimate the watermark proportion.

The ground-truth proportion is defined as the fraction of pivotal statistics following the alternative distribution. Specifically, for any segment $w_{(t-m):t}$ that appears in the original modified text, the corresponding statistic $Y_t = Y(w_t, \zeta_t) = Y(w_t, \AM(w_{(t-m):(t-1)}, \Key))$ is treated as watermarked. This is equivalent to assuming $Y_t | \bP_t \sim \mu_{1, \bP_t}$. We compute the proportion of such pivotal statistics as the true $\eps$.

\subsection{Additional Results on Post-processing Modification}

Table~\ref{tab:LLM-MAE-arxiv} reports the average MAEs under the three post-processing modifications on the arXiv dataset, serving as the counterpart to Table~\ref{tab:LLM-MAE-robust}. Once again, our estimator \textsf{OPT} achieves the lowest MAE and variance in most cases, though \textsf{INI} or \textsf{IND} may occasionally perform better.

\begin{table}[H]
\vspace{-0.1in}
\caption{
Averaged MAEs under common modification calculated over 11 ground truth $\eps$ values on open-sources model experiments. Standard deviations are provided in parentheses, and all values are reported in units of $10^{-3}$. Bold numbers denote the best performance.
}
\label{tab:LLM-MAE-arxiv}
\begin{center}
\vspace{-0.1in}
\begin{small}
\resizebox{\textwidth}{!}{%
\begin{tabular}{cc|c|ccc|c|ccc}
\toprule
\multirow{2}{*}{Models} & \multirow{2}{*}{Edit types} & \multicolumn{4}{c|}{Gumbel-max} & \multicolumn{4}{c}{Inverse transform} \\
\cmidrule{3-10}
& & 
\textsf{WPL} & \textsf{INI} & \textsf{IND} & \textsf{OPT} & 
\textsf{WPL} & \textsf{INI} & \textsf{IND} & \textsf{OPT} \\
\midrule
\multirow{3}{*}{OPT-1.3B} 
  & Substitution
  & 125(82) & 29(28) & 5(3) &\textbf{1(1)}
  & 266(113) & 37(22) & 4(3) &\textbf{2(1)} \\
 & Insertion 
 & 139(80) & 41(39) & \textbf{7(4)} &{9}(5)
 & 263(104) & 34(22) & \textbf{9}(7) &10\textbf{(5)} \\
 & Deletion 
 & 224(91) & \textbf{38}(41) & 74(20) &73\textbf{(19)} 
 & 280(101) & \textbf{32}(22) & 72(19) &74\textbf{(18)} \\
\midrule
\multirow{3}{*}{OPT-13B} 
& Substitution 
&   90(53) & 77(34) & 2(2) &\textbf{1(1)}
&  251(101) & 56(34) & 5(3) &\textbf{1(1)} \\
&Insertion
& 95(58) & 53(42) & 12\textbf{(5)} &\textbf{10(5)} 
&  243(82) & 46(29) & 10(7) &\textbf{9(5)} \\
& Deletion
& 160(82) & \textbf{44}(41) & 75(19) &73\textbf{(17)} 
& 247(102) & \textbf{61}(51) & 75(18) &77\textbf{(19)} \\
\midrule
\multirow{3}{*}{LLaMA-8B}
&Substitution
&  63(42) & 79(41) & 3(3) &\textbf{1(1)}
&  239(78) & 58(38) & 5(3) &\textbf{1(1)} \\
&Insertion
&  67(43) & 83(58) & 9(7) &\textbf{3(3)} 
& 227(85) & 47(31) & 7(6) &\textbf{3(2)} \\
&Deletion
& 125(63) & \textbf{42}(35) & 61\textbf{(21)} &{48(23)} 
& 239(86) & \textbf{47}(40) & 62\textbf{(20)} &\textbf{47}(23) \\
\bottomrule
\end{tabular}}
\end{small}
\end{center}
\vspace{-0.2in}
\end{table}

\end{appendix}

\end{document}